\documentclass{article}

 \usepackage[preprint]{neurips_2026}

\usepackage[utf8]{inputenc} 
\usepackage[T1]{fontenc}    
\usepackage{hyperref}       
\usepackage{url}            
\usepackage{booktabs}       
\usepackage{amsfonts}       
\usepackage{nicefrac}       
\usepackage{microtype}      
\usepackage{xcolor}         

\usepackage{graphicx}
\usepackage{subcaption}
\usepackage{amsmath}
\usepackage{amssymb}
\usepackage{mathtools}
\usepackage{amsthm}

\newcommand{\nn}{\mathbf{n}}

\newcommand{\z}{\mathbf{z}}
\newcommand{\y}{\mathbf{y}}
\newcommand{\x}{\mathbf{x}}
\newcommand{\bb}{\mathbf{b}}

\newcommand{\R}{\mathbb{R}}

\newcommand{\G}{\mathbf{G}}
\newcommand{\M}{\mathbf{M}}

\newcommand{\I}{\mathbf{I}}
\newcommand{\D}{\mathbf{D}}
\newcommand{\V}{\mathbf{V}}
\newcommand{\bH}{\mathbf{H}}
\newcommand{\bA}{\mathbf{A}}
\newcommand{\bM}{\mathbf{M}}

\newcommand{\bLambdah}{\boldsymbol{\Lambda}_h}
\newcommand{\bLambdahcon}{\boldsymbol{\Lambda}_{h^*}}
\newcommand{\bLambdahtran}{\boldsymbol{\Lambda}_{h^Th}}
\newcommand{\bLambdahtranop}{\boldsymbol{\Lambda}_{hh^T}}

\newcommand{\0}{\boldsymbol{0}}
\newcommand{\F}{\mathbf{F}}
\newcommand{\N}{\mathcal{N}}
\newcommand{\Q}{\mathbf{Q}}
\newcommand{\bepsilon}{\boldsymbol{\epsilon}}
\newcommand{\bmu}{\boldsymbol{\mu}_0}
\newcommand{\bSigma}{\boldsymbol{\Sigma}}
\newcommand{\bSigmaZ}{\boldsymbol{\Sigma}_0}
\newcommand{\bLambda}{\boldsymbol{\Lambda}_0}
\newcommand{\balpha}{\bar{\alpha}}
\newcommand{\bbalpha}{\bar{\boldsymbol{\alpha}}}
\newcommand{\bzeta}{\boldsymbol{\zeta}}
\newcommand{\br}{\boldsymbol{r}}

\usepackage[dvipsnames]{xcolor}
\definecolor{darkgreen}{RGB}{0,100,0} 

\usepackage{empheq}
\usepackage{graphicx}
\usepackage{caption}
\usepackage{subcaption}
\usepackage{hyperref}
\usepackage{url}
\usepackage{wrapfig}
\usepackage{amsmath}
\usepackage{amssymb}
\usepackage{mathtools}
\usepackage{amsthm}
\usepackage{float}
\usepackage{booktabs}
\usepackage{multirow}

\theoremstyle{plain}
\newtheorem{theorem}{Theorem}[section]

\newtheorem{lemma}[theorem]{Lemma}

\theoremstyle{definition}

\theoremstyle{remark}

\title{Analyzing and Guiding Zero-Shot Posterior Sampling in Diffusion Models}

\author{%
Roi Benita \\
Department of Electrical and Computer Engineering\\
Technion, Haifa, Israel \\
roibenita@campus.technion.ac.il
\And
Michael Elad \\
Department of Computer Science \\
Technion, Haifa, Israel \\
elad@cs.technion.ac.il
\And
Joseph Keshet \\
Department of Electrical and Computer Engineering\\
Technion, Haifa, Israel \\
jkeshet@technion.ac.il
}

\begin{document}

\maketitle

\begin{abstract}
Recovering a signal from its degraded measurements is a long standing challenge in science and engineering. Recently, zero-shot diffusion based methods have been proposed for such inverse problems, offering a posterior sampling based solution that leverages prior knowledge. Such algorithms incorporate the observations through inference, often leaning on manual tuning and heuristics. In this work we propose a rigorous analysis of these approximate posterior samplers, relying on a Gaussianity assumption of the prior. Under this regime, we show that both the ideal posterior sampler and diffusion-based reconstruction algorithms can be expressed in closed-form, enabling their thorough analysis and comparisons in the spectral domain. Building on these representations, we introduce a principled framework for parameter design, replacing heuristic selection strategies used to date. The proposed approach is method-agnostic and yields tailored parameter choices that jointly account for the characteristics of the prior, the degraded signal, and the diffusion dynamics. We show that our spectral recommendations differ structurally from standard heuristics and vary with the diffusion step size, resulting in a consistent balance between perceptual quality and signal fidelity.
\end{abstract}


\section{Introduction}
Linear inverse problems, a common task across many scientific fields, refer to the recovery of an unknown signal from its noisy, degraded measurements. Traditionally, for a given degradation model, one can learn a direct mapping from observations to clean signals using large paired datasets \cite{dong2015image, zhang2016colorful, saharia2022palette}. However, real-world applications, such as medical imaging, computational photography, and audio processing, often require more adaptable approaches for handling a potentially wide range of degradations \cite{hershey2016deep, weiss2019pilot, yaman2021zero, zhang2021plug, chen2022zero}. Recently, diffusion-based zero-shot posterior samplers have been proposed to bridge this gap, avoiding task-specific training \cite{daras2024survey}.



In this unsupervised setting, sampling is based on a pretrained diffusion model learned on the clean signal manifold, 
with different strategies for incorporating observations during inference.
One perspective steers the synthesis process by explicitly projecting intermediate samples onto geometric constraints induced by the measurements \cite{choi2021ilvr, wang2022zero, kawar2022denoising, zhu2023denoising}. 
Another common approach adopts a Bayesian formulation, where posterior estimation involves a likelihood component \cite{bansal2023universal, xu2024provably, chung2022diffusion, song2023pseudoinverse, moufad2024variational}. This guidance term is method-dependent and reflects the underlying degradation model, directing the denoising process toward consistency with the measurements.\footnote{Throughout this work, we adopt a unified prior-likelihood terminology \cite{daras2024survey}}


 
Alongside their remarkable success, a fundamental question within these methods is how to determine the relative weight between both forces, the likelihood and the prior information. One strategy is using small guidance step sizes, allowing samples to be gradually influenced by the observations \cite{chung2022diffusion, yu2023freedom, bansal2023universal}. However, this typically requires a large number of diffusion steps, slows inference and may result in poor measurement consistency \cite{yang2024guidance, zhang2025improving}. Alternatively, large guidance weights can induce unstable or unpredictable behavior, potentially driving samples away from the posterior \cite{he2023manifold}.

In the absence of a principled solution, existing methods choose these weighting parameters manually or heuristically \cite{daras2024survey}. One approach relies on empirically chosen, dataset-dependent scaling factors adjusted based on the discrepancy with the observations \cite{chung2022diffusion}. Another approach instead uses globally defined parameters determined by the diffusion noise levels \cite{song2023pseudoinverse, bansal2023universal, zhu2023denoising}. Notably, current design choices do not explicitly account for the interaction between prior characteristics, sampling dynamics, and the observed information. This can affect performance, reflected in the balance between sample naturalness and measurement fidelity \cite{blau2018perception}, and may lead to longer sampling trajectories.

In this work, we propose a structured analysis of zero-shot posterior samplers, inspired by the spectral perspective introduced in \cite{benita2025spectral}. Under a Gaussian prior assumption, we derive closed-form expressions for the ideal posterior sampler and diffusion-based training-free reconstruction methods. These analytical representations provide a sound basis for posterior-level comparison across existing samplers. In addition, we introduce a principled way of designing the guidance weighting parameters while jointly accounting for the prior characteristics, the degraded observations, and the diffusion dynamics. Finally, we validate our framework on real-world datasets, including FFHQ \cite{karras2019stylegan}, ImageNet \cite{deng2009imagenet}, and fastMRI \cite{zbontar2018fastmri, knoll2020fastmri}, comparing our spectral recommendations with existing heuristics while evaluating image naturalness and measurement fidelity across varied diffusion step sizes.

In summary, our contributions are the following: (i) We introduce a structured spectral analysis of zero-shot posterior sampling methods, providing a unified view under a common analytical framework. (ii) Under a Gaussian prior assumption, we derive closed-form expressions for the output distributions of the ideal posterior sampler and diffusion-based training-free reconstruction methods, enabling their direct posterior-level comparison and analysis. (iii) We formulate a principled, method-agnostic optimization problem for selecting guidance weighting parameters which jointly accounts for the prior characteristics, the degraded observations, and the diffusion dynamics. We demonstrate how it can be instantiated across different posterior sampling algorithms and solved either for individual realizations or efficiently for a general degradation system. (iv) We evaluate our spectral recommendations relative to existing heuristics, considering both averaged and fixed-observation settings, and assessing sample naturalness, measurement fidelity, and dependence on the number of diffusion steps.

\section{Background} 
\paragraph{Linear inverse problems.}\label{sec:Linear_Inverse_Problem}
A general linear inverse problem is defined as: 
\begin{equation}
\y = \bH\x+\nn,
\end{equation}
where the goal is to recover an unknown signal $\x\in\mathbb{R}^d$ from a degraded observation $\y\in\mathbb{R}^m$. Here, $\bH\in\mathbb{R}^{m\times d}$ is a known linear degradation matrix, and $\nn\!\sim\!\mathcal{N}(\0,\sigma^2_y\I)$ denotes i.i.d. additive Gaussian noise. 
A common property of these problems is the loss of information induced by the degradation operator $\bH$, making perfect recovery generally impossible and rendering the problem ill-posed.
\paragraph{Denoising diffusion probabilistic models.}
Diffusion models form a generative framework that transforms a simple normal distribution into a target data distribution $p(\x_0)$, thereby enabling the synthesis of samples $\x_0\!\in\!\mathbb{R}^d$. As originally formulated in DDPM \cite{ho2020denoising}, the diffusion process consists of two Markovian paths, the \emph{forward} and \emph{reverse} processes, each comprising a sequence of $T$ latent variables. The \emph{forward} process gradually corrupts a clean signal $\x_0$ through additive Gaussian noise:
\begin{equation}
\x_t = \sqrt{\alpha_t} \x_{t-1} + \sqrt{1 - \alpha_t} \bepsilon_t,
\end{equation}
where $\left\{ {\alpha}_t \right\}_{t=1}^T$ denotes a monotonically decreasing noise schedule and $\boldsymbol{\epsilon}_t\sim \mathcal{N}(\0,\I)$. A key property of the forward process is the marginal distribution of $\x_t$ given $\x_0$:
\begin{equation}\label{eq:marginal_dist}
    \mathbf{x}_t = \sqrt{\balpha_t} \mathbf{x}_0 + \sqrt{1-\balpha_t} \boldsymbol{\epsilon} \quad  \boldsymbol{\epsilon} \sim \N(\mathbf{0},\I) ~,
\end{equation}
with $\balpha_t = \prod_{i=1}^{t} \alpha_i$, and for sufficiently large $T$, the terminal state approaches $\x_T\!\sim\! \mathcal{N}(\0,\I)$.

The \emph{reverse} process aims to reconstruct $\x_0$ from $\x_T$ by iteratively denoising. Extending the stochastic DDPM formulation \cite{ho2020denoising}, DDIM \cite{song2020denoising} relaxes the Markovian assumption to enable deterministic and accelerated sampling. As a result, the reverse trajectory is defined by:
\begin{align}
\label{eq:DDIM_sampling_procedure}
    \x_{s-1} =  \sqrt{\bar{\alpha}_{s-1}} \left( 
    \frac{\x_{s} - \sqrt{1 - \bar{\alpha}_{s}} \cdot \bepsilon_\theta(\x_{s},s)}{\sqrt{\bar{\alpha}_{s}}}\right)
     + \sqrt{1 - \bar{\alpha}_{s-1}-\sigma^2_s} \cdot \bepsilon_\theta(\x_{s},s) + \sigma_s\epsilon_s,
\end{align}
where $s\in\{0, 1, \dots, S\}$ indexes a subsequence of the original $T$ diffusion steps,
$\sigma_s=\eta \sqrt{({1-\alpha_{s-1}})/({1-\alpha_s}})\sqrt{1-{\alpha_s}/{\alpha_{s-1}}}$, and $\bepsilon_\theta(\x_{s},s)$ denotes a neural network parameterized by $\theta$, trained to estimate $\bepsilon$ from a given $\x_s$ at time $s$. When $\eta=1$, the generative process reduces to DDPM,  whereas $\eta=0$ yields the deterministic DDIM sampler. In this work, we denote the cumulative noise schedule by $\bbalpha =\left\{ {{\balpha}_t} \right\}_{t=1}^T$ for DDPM and by $\bbalpha =\left\{ {{\balpha}_s} \right\}_{s=1}^S$ for DDIM sampling.


\paragraph{Zero-shot diffusion-based methods for inverse problems}\label{sec:background_training_free_methods}
Training-free methods address inverse problems by adapting a pretrained diffusion prior at inference time, thereby avoiding task-specific retraining \cite{daras2024survey}. One class of methods relies on projection-based updates, in which the sampling process is modified to enforce data consistency at each time step \cite{wang2022zero, kawar2022denoising, choi2021ilvr, zhu2023denoising}. In contrast, Bayesian methods \cite{bansal2023universal, xu2024provably, chung2022diffusion, song2023pseudoinverse} formulate inverse problems through posterior inference by combining a pretrained diffusion prior with a measurement likelihood, leading to the following score decomposition:
\begin{align}
\nabla_{\x_t} \!\log p_t(\x_t | \y) \!=\! \nabla_{\x_t} \log p_t(\x_t)\! +\! \nabla_{\x_t} \log p_t(\y | \x_t).
\end{align} We detail the Bayesian formulations here, with projection-based methods described in Appendix~\ref{sec:appendix_DiffPIR_algorithm}.
Based on the Tweedie’s formula \cite{efron2011tweedie}, the prior score is obtained from the learned noise estimator as:
\begin{equation}
\nabla_{\x_t}\log p_t(\x_t)\approx -({1}/\sqrt{1-\balpha_t})\bepsilon_\theta(\x_t,t),
\end{equation}
and the likelihood term is approximated in a method-dependent manner to guide sampling toward consistency with the measurements. A common Bayesian approach is DPS \cite{chung2022diffusion}, which marginalizes over the clean signal and estimates the likelihood using the denoiser output, yielding $p(\y|\x_t)\simeq p(\y|\hat{\x}_0) \,\,\text{where}\,\, \hat{\x}_0:= \mathbb{E}[\x_0 | \x_t] $.
Incorporating this approximation within the deterministic DDIM formulation \eqref{eq:DDIM_sampling_procedure} leads to the following update rule:
\begin{align}\label{eq:DDIM_with_DPS}
    \x_{s-1}  = \sqrt{\bar\alpha_{s-1}}\hat\x_0 + \sqrt{1-\bar{\alpha}_{s-1}}\bepsilon_\theta(\x_s,s) - \zeta_s \nabla_{\x_s} \|\y - \bH \hat{\x}_0\|_2^2, 
\end{align}
where $\zeta_s={\zeta^{'}} / |\y-\bH\hat{\x}_0(\x_s)|$ is a heuristic guidance step size  which scales inversely with the measurement error norm, and $\zeta^{'}$ is a user-defined constant.
Alternatively, $\Pi$GDM \cite{song2023pseudoinverse} models $p(\x_0| \x_t)$ as a Gaussian distribution with mean $\hat{\x}_0$ and variance $r_t^2$, governing the uncertainty at diffusion step $t$.
This yields the likelihood approximation  $
p_t(\y | \x_t) \approx \mathcal{N}(\bH \hat{\x}_0, r_t^2 \bH \bH^T + \sigma_\y^2 \I)
$, where the variance parameter $r_t$ serves as a weighting term in the deterministic DDIM update:
\begin{align}\label{eq:PIGDM_inference_equation}
\x_{s-1} \!= \!\sqrt{\bar\alpha_{s-1}}\hat\x_0 
+ \sqrt{1 \!-\! \bar{\alpha}_{s-1}} \, \bepsilon_\theta(\x_s, s)  
+ r^2_s \nabla_{\x_s}\!(\hat\x_0)^T \bH^T ((r_s^2 \bH \bH^T \!\!+\! \sigma_\y^2 \I)^{-1})^T\!(\y \!-\! \bH\hat\x_0).  
\end{align}
Specifically, $r_s$ is heuristically defined according to the diffusion noise schedule as $r_s = \sqrt{1-\balpha_s}$. Throughout the paper, we denote by $\bzeta$ and $\br$ the heuristic parameter sets $\{\zeta_s\}_{s=1}^{S}$ and $\{r_s\}_{s=1}^{S}$, respectively.



\section{Guided zero-shot posterior samplers}\label{sec:guided_zero_shot_posterior}
In this section, we bring diffusion-based zero-shot posterior sampling methods together under a unified spectral perspective. Our framework is method-agnostic and enables simplified analysis and principled design of the underlying heuristics.
We use DPS \cite{chung2022diffusion} as a representative method and further demonstrate the framework on $\Pi \text{GDM}$ \cite{song2023pseudoinverse} and DiffPIR~\cite{zhu2023denoising}, a projection-based approach, in Appendices~\ref{sec:appendix_PIGDM_algorithm}  and \ref{sec:appendix_DiffPIR_algorithm}.



\subsection{The optimal denoiser for a gaussian prior}\label{sec:The_Optimal_Denoiser_for_a_Gaussian_Prior}
As reviewed in Section \ref{sec:background_training_free_methods}, training-free posterior samplers leverage a pretrained denoiser to incorporate prior information during the inference phase. We assume a Gaussian prior on the clean signal, $\mathbf{x}_0 \sim \mathcal{N}(\bmu, \boldsymbol{\Sigma}_0)$, where $\bmu \in \mathbb{R}^{d}$ and $\boldsymbol{\Sigma}_0 \in \mathbb{R}^{d \times d}$. Under this modeling choice, the corresponding optimal denoiser for a noisy signal $\mathbf{x}_t$ at diffusion step $t$ and noise schedule $\bbalpha$ admits the following closed-form expression \cite{benita2025spectral}:
\begin{align}\label{eq:optimal_denoiser_prior}
\x_0^* = (\balpha_t \bSigmaZ + (1-\balpha_t) \I )^{-1} (\sqrt{\balpha_t} \bSigmaZ \x_t + (1-\balpha_t) \bmu )
\end{align}

\subsection{Reverse process formulation}
We next describe the DPS \cite{chung2022diffusion} sampling procedure under the Gaussian setting. For brevity, we highlight the key steps using the DDIM \cite{song2020denoising} formulation, which facilitates an accelerated sampling framework, while the full derivation is provided in Appendix~\ref{sec:appendix_DPS_algorithm}.
Based on the marginal relation in~\eqref{eq:marginal_dist}, the deterministic update from~\eqref{eq:DDIM_with_DPS} can be expressed in terms of the optimal denoiser as follows: 
\begin{align}\label{eq:DPS_time_main_part_1}
\x_{s-1} = a_s\x_s + b_s\x_0^* - \zeta_s \nabla_{\x_s} \|\y - \bH \x_0^*\|_2^2 
\end{align}
where $\y=\bH\x_0+\nn$ denotes the linear degradation model, $a_s$ and $b_s$ are determined by the noise schedule $\bbalpha$, and $\zeta_s$ is the corresponding heuristic guidance weight.




\subsection{Spectral form of the update rule}\label{sec:prior_migrating_to_the_spectral_domain}
We adopt a spectral perspective which allows a simplified analysis of the diffusion dynamics \cite{benita2025spectral}.
In particular, we consider the case where the covariance matrix $\boldsymbol{\Sigma}_0$ is shift-invariant, a reasonable assumption for natural images \cite{unser1984approximation, freirich2021theory}, and the degradation operator $\mathbf{H}$ shares this property. This structure induces diagonalization in the Fourier basis and naturally aligns with common linear inverse problems such as deblurring, denoising, and accelerated acquisition settings (see Appendix~\ref{sec:fastMRI_DPS}).



Under this setting, the prior distribution remains Gaussian in the spectral domain, 
$\mathbf{x}_0^{\mathcal{F}} \sim \mathcal{N}(\boldsymbol{\mu}^{\mathcal{F}}_0, \mathbf{\Lambda}_0)$, where $\boldsymbol{\mu}^{\mathcal{F}}_0$ is the transformed mean vector, and $ \mathbf{\Lambda}_0\in\mathbb{R}^{d\times d}$ is a positive semidefinite diagonal matrix containing the eigenvalues of $\bSigma_0$, denoted by $\left\{\lambda_i \right\}_{i=1}^d$. 
Furthermore, $\mathbf{H} \in \mathbb{R}^{d \times d}$ admits a diagonal spectral representation $\boldsymbol{\Lambda}_h$, whose diagonal entries are the eigenvalues of $\mathbf{H}$, with $\boldsymbol{\Lambda}_{h^*}$ denoting its complex conjugate.
Accordingly, the posterior {$p(\x_0^{\mathcal{F}}\! \mid \! \y^{\mathcal{F}})$} remains Gaussian and admits a closed-form expression in the spectral domain, where $\y^{\mathcal{F}}$ denotes the  spectral domain measurements.  A detailed derivation is provided in Appendix~\ref{sec:appendix_gaussian_posterior_distribution}.

By projecting Equation \eqref{eq:DPS_time_main_part_1} onto the unitary Fourier basis $\F\in\mathbb{C}^{d \times d}$, we obtain the following relation between consecutive diffusion steps in the spectral domain. where the superscript $\mathcal{F}$ denotes the Fourier-domain representation.
\begin{align}\label{eq:DPS_freq_main_part_1}
\x^{\mathcal{F}}_{s-1} & = [a_s + b_s c_s - 2\zeta_s c_s \bLambdahcon\bLambdah c_s]\x^{\mathcal{F}}_s + [2\zeta_s c_s \bLambdahcon]\y^{\mathcal{F}} + [ b_sd_s - 2\zeta_s  c_s \bLambdahcon \bLambdah d_s]\bmu^{\mathcal{F}}
 \end{align}
Here, $a_s$ and $b_s$ are diagonal matrices determined by the noise schedule $\bbalpha$, and $c_s$, $d_s$ further reflect the spectral properties encoded in $\bLambda$. The full definitions are provided in Appendix~\ref{sec:DPS_migrating_spectral}.

Using the diagonalization property of Equation \eqref{eq:DPS_freq_main_part_1}, the update rule decouples into 
$d$ independent scalar equations, which can be recursively substituted. This yields a closed-form expression for the estimated output signal over $S$ diffusion steps.
\begin{equation}\label{eq:DDIM_X_0_X_T_main}
\mathbf{{\hat{x}}}_{0,\text{DPS}}^{\mathcal{F}} = \D_1 \mathbf{x}_S^{\mathcal{F}} + \D_2 \y^{\mathcal{F}} + \D_3\bmu^{\mathcal{F}}
\end{equation}
Where $\mathbf{D}_1$, $\mathbf{D}_2$, and $\mathbf{D}_3$ are diagonal matrices determined by the noise schedule $\bbalpha$, the number of diffusion steps $S$, the spectral characteristics of the prior $\bLambda$ and the degradation operator $\boldsymbol{\Lambda}_h$. Interestingly, Equation \eqref{eq:DDIM_X_0_X_T_main} can be interpreted as a set of $d$ scalar \emph{transfer functions} mapping the Gaussian input noise $\x_S\sim\mathcal{N}(\0,\I)$ and the degraded measurement $\y^{\mathcal{F}}$ to the
clean output estimate.

Finally, from a probabilistic perspective, the resulting posterior distribution is
\begin{align}\label{eq:DPS_spectral_reconstructed_distribution}
\mathbf{\hat{x}}_{0,\text{DPS}}^{\mathcal{F}} \,\big|\,  \y^{\mathcal{F}} \sim \N\!\!\left( \D_2 \y^{\mathcal{F}} + \D_3\bmu^{\mathcal{F}}  ~,~  \D_1^2  \right).
\end{align}
This reconstruction captures information from the prior structure, the diffusion dynamics and the degraded measurements $\y^\mathcal{F}$, with the latter incorporated through the DPS update, contributing to the posterior via its first-order statistics. 
Additional details and definitions are provided in Appendix~\ref{sec:DPS_migrating_spectral}.

\subsection{Optimal spectral guidance weights}\label{sec:Optimal_Posterior_Sampling_Weights}
By leveraging the closed-form expressions for the reconstructed distributions such as \eqref{eq:DPS_spectral_reconstructed_distribution},
we further analyze the dynamics of diffusion-based posterior sampling methods from a spectral perspective. A key aspect relates to the design choices of the prior-likelihood weighting heuristics. We consider a setting with a prior covariance matrix $\bSigma_0$, characterized by eigenvalues $\left\{\lambda_i \right\}_{i=1}^d$, a fixed linear degradation matrix $\bH$, an observation $\y$, and a diffusion process with $S$ diffusion steps and noise schedule $\bbalpha$. Our objective is to bring the estimated probability density induced by such posterior sampling methods as close as possible to the true posterior distribution. Accordingly, we seek a set of weights $\bzeta$ that minimizes a discrepancy measure $\mathcal{D}$ between the estimated reconstructed distribution, $p(\mathbf{\hat{x}}_{0,\text{DPS}}^{\mathcal{F}}\! \mid \! \mathbf{y}^{\mathcal{F}};\bzeta, \bbalpha)$, and the true posterior distribution $p(\x_0^{\mathcal{F}}\! \mid \! \y^{\mathcal{F}})$,  leading to the following optimization problem.\footnote{While $\bbalpha$ could also be treated as an optimized argument, we focus on the current formulation.}
\begin{align}
\label{eq:optimization_problem}
\bzeta^* = &\arg\min_{\bzeta}  \mathcal{D}\left(p(\mathbf{\hat{x}}_{0,\text{DPS}}^{\mathcal{F}}\! \mid \! \mathbf{y}^{\mathcal{F}}; \bzeta, \bbalpha),p(\x_0^{\mathcal{F}}\! \mid \! \y^{\mathcal{F}})\right) \end{align}
Choosing the discrepancy measure $\mathcal{D}$ is a key design choice, as it governs how measurement fidelity is balanced with prior information and ultimately shapes the trade-off between perceptual quality and distortion \cite{blau2018perception}.
In principle, for a given observation $\y$, one could define various discrepancy measures 
$\mathcal{D}$ between the corresponding posterior distributions, such as the \emph{Wasserstein-2} \cite{arjovsky2017wasserstein} or \emph{KL divergence} \cite{kullback1951information} (see Appendix~\ref{sec:dkl_derivative} for further details). In this work, we propose a more general formulation based on the \emph{Wasserstein-2} distance, referred to as the \emph{averaged Wasserstein distance} and denoted by:
\begin{align} \left[\mathcal{D}_{W_2}^2\right]_\text{Avg} &\!\!=\!\! \frac{1}{K}\!\! \textstyle\sum_{k=1}^K\! \mathcal{D}_{W_2}^2 \!\big({p}({\hat{\x}}_{0}^{\mathcal{F}} \!\,\big|\, \! \y_k^{\mathcal{F}}; \! \bzeta, \bbalpha ), p(\x_{0}^{\mathcal{F}} \!\,\big|\,\!  \y_k^{\mathcal{F}})\big).   \end{align}
This formulation accommodates two regimes. For $K=1$, the optimization is performed for a single realization $\y$,  reducing to the standard \emph{Wasserstein-2} distance:
\begin{align}\label{eq:optimization_problem_specific_realization}
\left[\mathcal{D}_{W_2}^2\right] &_\text{Avg} =
\textstyle\sum_{i=1}^{d} \left( \sqrt{\lambda_i} - 
 [\D_1]_i  \right)^2  + \textstyle\sum_{i=1}^{d} \left([\D_2 \!- \!\bA ]_i[\y^{\mathcal{F}} ]_i+[\D_3\!-\!\I \!+\! \bA\boldsymbol{\Lambda}_{h} ]_i[\bmu^{\mathcal{F}} ]_i\right)^2
\end{align}
where  $\bA = \bLambda\boldsymbol{\Lambda}_{h^*} (\boldsymbol{\Lambda}_h\bLambda\boldsymbol{\Lambda}_{h^*}\!+\! \sigma_{n}^2 \I)^{-1}$.

Alternatively, when multiple realizations are considered, the weight parameters can be adjusted to capture the expected behavior of the posterior  $p(\x_0^{\mathcal{F}}\! \mid \! \y^{\mathcal{F}})$ across $\y^{\mathcal{F}}$, thereby eliminating the need to solve the optimization for each observation. In this setting, the optimal solution admits a closed-form expression in terms of the prior characteristics:
\begin{align}\label{eq:optimization_problem_many_realizations}
[\mathcal{D}&_{W_2}^2]_\text{Avg}= \textstyle\sum_{i=1}^{d} \left( \sqrt{\lambda_i} - 
 [\D_1]_i  \right)^2  \!\!\! + \operatorname{Tr} \left(\M^T \M \left(\bLambdah \, \bLambda \, \bLambdahcon + \sigma_n^2 \, \I\right)\right) + \| \M \bLambdah \, \boldsymbol{\mu}_0^\mathcal{F} + \bb \|_2^2 
\end{align}
where $\bM=\D_2-\bA$ and  $\bb = \left(\D_3-\I + \bA\boldsymbol{\Lambda}_h\right)\bmu^{\mathcal{F}}$,
with $\D_2$ and $\D_3$ defined in \eqref{eq:DDIM_X_0_X_T_main}.
The full derivations are provided in Appendix~\ref{sec:appendix_Averaged_Wasserstein_loss}.


\section{Optimal gaussian posterior sampling}\label{sec:optimal_gaussian_posterior_sampling}
In this section, we formulate a closed-form expression for the posterior optimal denoiser under the Gaussian setting and analyze the output induced by the diffusion process. The derived sampler provides a principled reference for evaluating training-free posterior sampling methods and for assessing the effectiveness of our proposed recommendations under identical diffusion dynamics.  Such an ideal denoiser is tailored to a specific measurement operator $\mathbf{H}$ and observation $\mathbf{y}$, and generally does not admit an analytic form beyond the Gaussian setting, hence necessitating approximations in practice.
\subsection{The posterior optimal denoiser}
We adopt the notation introduced in Section~\ref{sec:The_Optimal_Denoiser_for_a_Gaussian_Prior}. The clean signal $\x_0 \in \mathbb{R}^d$ is modeled by a Gaussian prior,
$\x_0 \sim \mathcal{N}\!\left(\bmu , \bSigma_0 \right)$. Observations $\mathbf{y} \in \mathbb{R}^m$ follow
$\mathbf{y} = \mathbf{H} x_0 + \mathbf{n}$ with a general linear operator
$\mathbf{H} \in \mathbb{R}^{m \times d}$ and
$\mathbf{n} \sim \mathcal{N}(\mathbf{0}, \sigma_y^2 \mathbf{I})$.
Under this modeling framework, the true posterior distribution is also Gaussian,
$p(\mathbf{x}_0 | \mathbf{y}) \sim \mathcal{N}(\boldsymbol{\mu}_{\x_0|\y}, \boldsymbol{\Sigma}_{\x_0|\y})$,
with the resulting expressions for the posterior mean and covariance provided in Appendix~\ref{sec:appendix_gaussian_posterior_distribution}.

We begin by deriving the exact optimal denoiser for the posterior. Under the Gaussian model described above, this denoiser coincides with both the posterior mean and the MAP estimator, and reduces to the classical Wiener filter in closed form. A detailed proof is provided in Appendix \ref{sec:appendix_Map_est}. 
\begin{theorem}
Let $\x_0 \sim \N(\bmu, \bSigmaZ)$ and let $\x_t$
denote the noisy signal obtained from the forward diffusion process, as defined in \eqref{eq:marginal_dist}. Given the linear measurement model $\y=\bH\x_0+\nn$, and letting $\bbalpha$ denote the noise schedule parameters, the denoised signal obtained from the MAP estimator is given by:
\begin{align}\mathbf{\tilde{x}}_0  = \arg\max_{\x_0}  \log p(\mathbf{x}_0 | \mathbf{x}_t, \y)\end{align}
and admits the following closed-form:
\begin{align}\label{eq:posterior_optimal_denoiser}
\mathbf{\tilde{x}}_0  =& \nonumber
\left((1\!-\!\balpha_t){\boldsymbol{\Sigma}}_0\mathbf{H}^T\mathbf{H} \!+ \! \sigma_{y}^2{\balpha_t}{\boldsymbol{\Sigma}}_0\!+\!\sigma_{y}^2 (1\!-\!\balpha_t)\I\right)^{-1}\\ 
& \left((1\!-\!\balpha_t){\boldsymbol{\Sigma}}_0\mathbf{H}^T\y\!+\!\sigma_{y}^2\sqrt{\balpha_t}{\boldsymbol{\Sigma}}_0\x_t\!+\!\sigma_{y}^2 (1\!-\!\balpha_t)\boldsymbol{\mu}_0\right)
\end{align}
\end{theorem}
\subsection{Spectral form of the update rule}
We now describe the discrete sampling procedure, as introduced in \cite{ho2020denoising, song2020denoising} instantiated using optimal denoiser derived in \eqref{eq:posterior_optimal_denoiser}. In particular, we focus on the DDIM formulation \cite{song2020denoising}, as it facilitates an accelerated sampling framework. For brevity, we only present the final reconstructed distribution of the output $\mathbf{\hat{x}}_{0,\text{opt}}^{\mathcal{F}}$, where the superscript $\mathcal{F}$ denotes the Fourier-domain representation. 
A full derivation is provided in Appendices~\ref{sec:appendix_inference_Time} and~\ref{sec:appendix_inference_spectral}.

\begin{lemma}\label{proposition:optimal_denoising_spectral_distribution}
Assume $\mathbf{x}_0 \sim \mathcal{N}(\boldsymbol{\mu}_0, \boldsymbol{\Sigma}_0)$, where the covariance matrix $\boldsymbol{\Sigma}_0$ is shift-invariant, and let $\y=\bH\x_0+\nn$ denote a linear measurement model with a shift-invariant operator. The frequency domain reconstruction obtained using DDIM with the optimal denoiser in \eqref{eq:posterior_optimal_denoiser} follows a Gaussian distribution and is given by:
\begin{align}\label{eq:posterior_optimal_sampler}
p(\mathbf{\hat{x}}_{0,\text{opt}}^{\mathcal{F}}|\y^{\mathcal{F}}) \sim \mathcal{N}(\mathbf{V}_2 \mathbf{y}^{\mathcal{F}} + \mathbf{V}_3 \bmu^{\mathcal{F}},\mathbf{V}_1^2)
\end{align}
The matrices $\mathbf{V}_1$, $\mathbf{V}_2$, and $\mathbf{V}_3$ are diagonal and depend on the noise schedule parameters $\bbalpha$, the degradation operator $\boldsymbol{\Lambda}_h$, and the Fourier-domain prior, $\mathbf{x}_0^{\mathcal{F}} \sim \mathcal{N}(\boldsymbol{\mu}^{\mathcal{F}}_0, \mathbf{\Lambda}_0)$.
\end{lemma}
Consequently, under the Gaussianity assumption, the ideal posterior sampler admits a closed-form expression that decouples into $d$ independent scalar \emph{transfer functions}. This formulation provides a principled reference for analyzing and interpreting existing posterior sampling methods. 

\section{Related work} 
Diffusion models have been widely applied to inverse problems in a zero-shot setting \cite{daras2024survey}. 
One line of work incorporates measurements by enforcing data consistency through projection-based geometric operations, such as null-space decompositions and pseudoinverse operators \cite{choi2021ilvr, wang2022zero, kawar2022denoising}. Among these methods, DiffPIR \cite{zhu2023denoising} builds on the Half-Quadratic-Splitting (HQS) algorithm \cite{geman1995nonlinear} to decouple the data and the prior terms during inference. Another line adopts a Bayesian perspective, integrating measurement through likelihood-based guidance \cite{bansal2023universal, xu2024provably, moufad2024variational}. Within this framework, DPS \cite{chung2022diffusion} employs a point-estimate approximation of the clean signal motivated by Jensen’s inequality, while $\Pi\text{GDM}$ \cite{song2023pseudoinverse} further introduces time-dependent variances. 

Alongside their effectiveness, these approaches face practical challenges.  A key aspect is their reliance on manual hyperparameter tuning, which affect both stability and performance. While stability can be improved by using small guidance step sizes, this comes at the cost of increased sampling steps and slower inference \cite{he2023manifold}. Recently, DSG \cite{yang2024guidance} introduced the notion of manifold deviation during sampling and proposed constraints that enable larger guidance step sizes. Similarly, DAPS \cite{zhang2025improving} employs an annealing strategy that decouples consecutive steps in the diffusion trajectory, improving stability at larger step sizes. Related works further explore how inference-time dynamics, such as manifold constraints or noise modulation, affect the tradeoff between perceptual quality and data fidelity in diffusion models \cite{chung2022improving, he2023manifold, wang2025traversing}. Nevertheless, most training-free posterior sampling methods still rely on manual tuning to control these trade-offs.






Recently, the design of heuristics for diffusion models has received significant attention \cite{watson2021learning, xia2023towards, wang2023learning, tong2024learning, chen2024trajectory, xue2024accelerating, williams2024score}. In particular, the authors of  \cite{benita2025spectral} proposed a spectral approach for determining the noise schedule in prior sampling, expressing the diffusion process as a transfer function and formulating an associated optimization problem. In the context of posterior sampling, the authors of \cite{bellchambers2025exploiting} addressed the purely denoising task using the prior score, and considered their solution transferable to related inverse problems. While their approach provided heuristic recommendations for DPS, it involves per-instance optimization during inference, increasing synthesis time.



\section{Experiments} 
\label{sec:experiments}
We turn to validate our approach through a series of experiments under multiple degradation settings, examining our spectral recommendations compared to existing heuristics, and analyzing current sampling methods relative to the derived optimal posterior denoiser.

\subsection{Synthetic Gaussian distribution}\label{sec:Synthetic_Gaussian_distribution}

In this section, we consider a Gaussian prior distribution $\x_0\sim\mathcal{N}(\bmu,\bSigma_0)$, where $\x_0 \in \mathbb{R}^d$ and $\bSigma_0\in\mathbb{R}^{d \times d}$ is a circulant covariance matrix. The measurements $\y\in\mathbb{R}^d$ are generated by the linear degradation model $\y= \bH\x_0+\nn$, where the operator $\bH$ acts as a low-pass filter (\emph{LPF}), preserving a fraction $\mathcal{V} \in (0,1]$ of the frequency components, and $\nn\sim\mathcal{N}(\0,\sigma^2_y\I)$ is additive Gaussian noise. Further details regarding the experimental configuration are provided in Appendix~\ref{sec:Additional_results}.



We begin by solving the optimization problem in~\eqref{eq:optimization_problem_specific_realization} for DPS~\cite{chung2022diffusion} using varying numbers of diffusion steps. Specifically, we repeat the procedure independently for $N\!=\!5$ observations $\{\y^{(i)}\}_{i=1}^{N}$, where each realization corresponds to an independently sampled signal $\x_0^{(i)}$ and noise realization $\nn^{(i)}$. We set $\mathcal{V}\!=\!0.5$, $\sigma_y\! =\! 0.1$, and consider a prior with mean $\bmu\!=\!\0$ and a covariance dimension $d\!=\!50$.



Figure~(\ref{fig:Comparison_DPS_SLSQP_differnt_steps}) in Appendix \ref{sec:Additional_results_DPS}, shows the resulting spectral recommendations for the weighting coefficients $\bzeta$, along with the mean and standard deviation across realizations. While the recommendations vary with the number of diffusion steps, their overall structure is preserved. Notably, larger weights are assigned at the early stages of the diffusion process and gradually decrease toward later steps, with greater variability across realizations observed at the beginning of the reverse process.

\begin{figure}[t]
    \centering
    \begin{subfigure}[b]{0.49\textwidth}
        \centering
        \includegraphics[width=\textwidth]{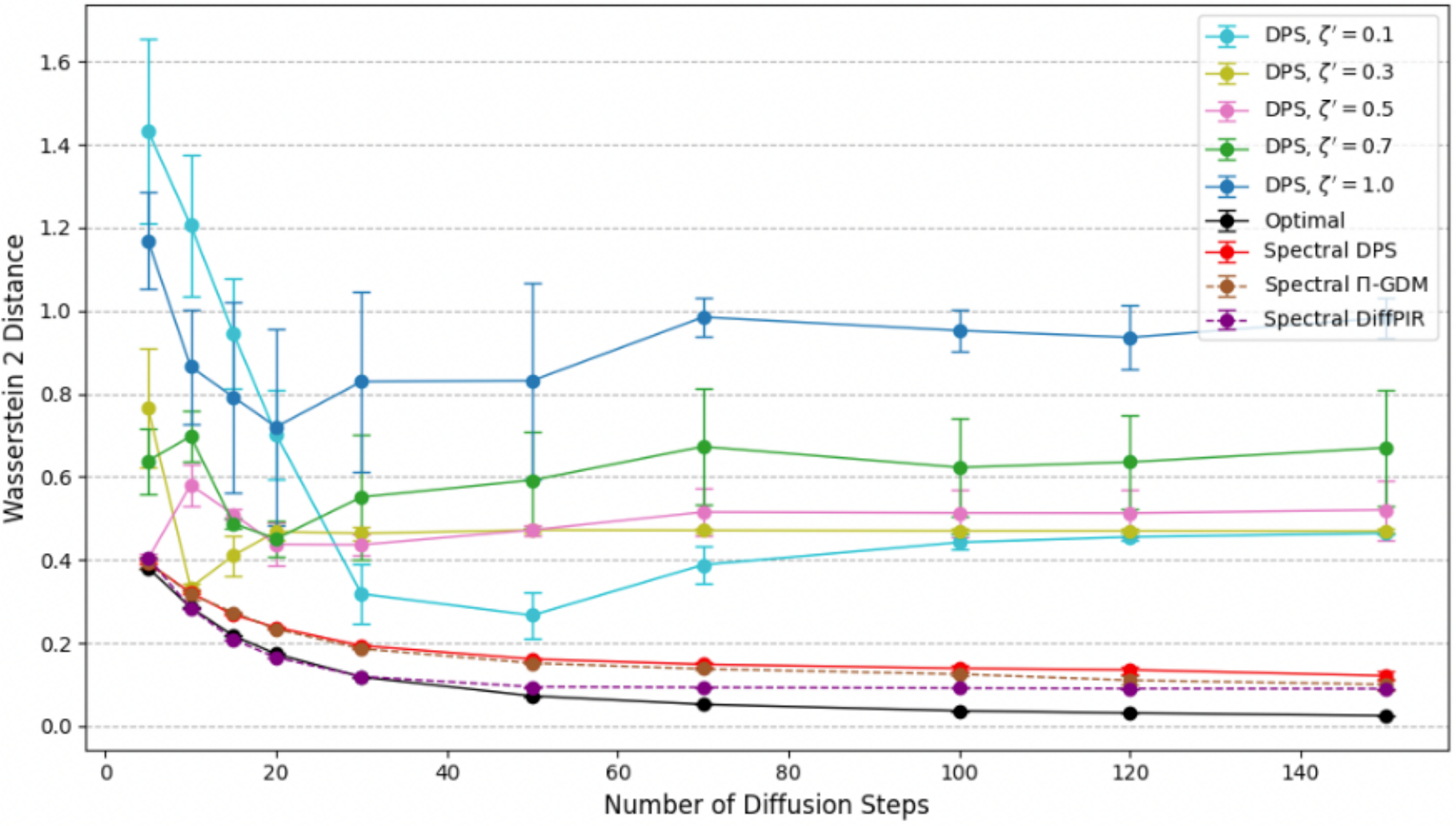}
        \caption{}
        \label{fig:Comparison_wasserstein_DPS_pigdm_spectral_optimal}
    \end{subfigure}
    \hfill
    \begin{subfigure}[b]{0.49\textwidth}
        \centering
        \includegraphics[width=\textwidth]{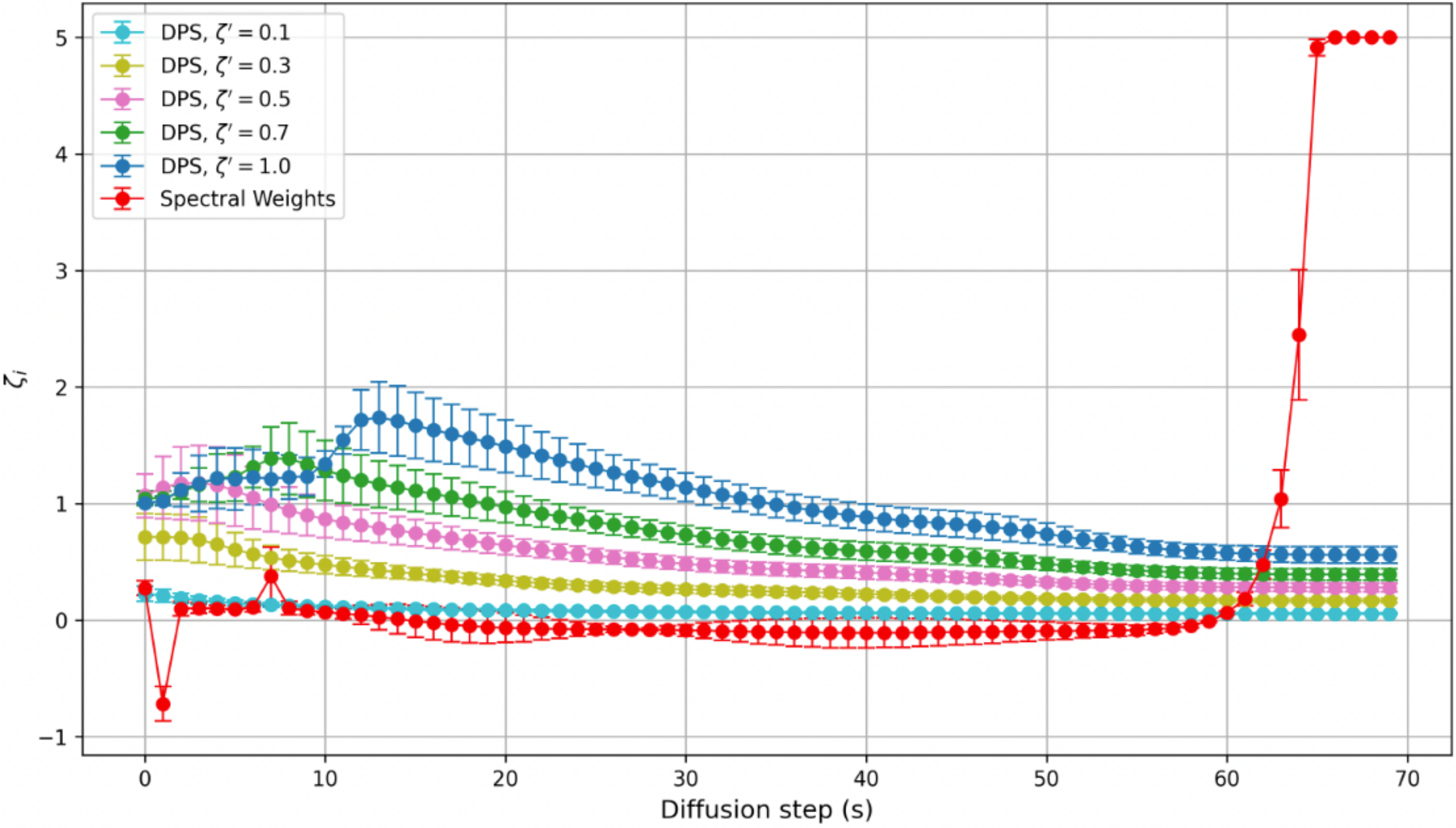}
        \caption{}
        \label{fig:Comparison_Weights_DPS_spectral}
    \end{subfigure}
    \caption{(a) Comparison of the Wasserstein-2 distance for DPS heuristic values $\zeta' \in \{0.1, 0.3, 0.5, 0.7, 1.0\}$, the spectral recommendations applied to DPS (red) and to $\Pi\text{GDM}$ (brown, dotted), and the analytically derived ideal posterior sampler (black), across different numbers of diffusion steps $S \in \{5, 10, 15, 20, 30, 50, 70, 100, 120, 150\}$. (b) Comparison of spectral recommendations (red) and DPS guidance coefficients for $70$ diffusion steps, evaluated across heuristic values.}
    \vspace{-0.35cm}
    \label{fig:combined}
\end{figure}

We compare our spectral recommendations with the guidance weights used in DPS under the same experimental setting. As these weights are selected heuristically, we evaluate several values of $\zeta'$ to assess the effect of different choices. Figure~\ref{fig:Comparison_Weights_DPS_spectral} illustrates the comparison for $70$ diffusion steps. Interestingly, due to their inverse dependence on the measurement error norm, the DPS weights exhibit an opposing trend, increasing toward the end of the diffusion process.

While the true posterior is often intractable for general priors, as only a single sample $\x_0^{(i)}$ is available for a given observation $\y^{(i)}$, the Gaussian setting admits a closed-form posterior, enabling direct distributional evaluation. Leveraging this property, Figure~\ref{fig:Comparison_wasserstein_DPS_pigdm_spectral_optimal} compares the \emph{Wasserstein-2} distance between the reconstructed and the true posterior distributions for several heuristic choices and the proposed spectral recommendation across diffusion steps and realizations. As performance varies across heuristic choices, the spectral recommendation consistently surpasses all hand-crafted alternatives and exhibits greater stability across realizations.

Figure~\ref{fig:Comparison_wasserstein_DPS_pigdm_spectral_optimal} also includes the distance obtained using the analytically derived posterior-optimal denoiser from Equation~\eqref{eq:posterior_optimal_sampler}. This solution serves as a reference for evaluating the different posterior samplers and is generally unavailable beyond the Gaussian setting.
The figure further compares the spectral recommendations for DPS~\cite{chung2022diffusion}, $\Pi\text{GDM}$~\cite{song2023pseudoinverse}, and DiffPIR~\cite{zhu2023denoising}, highlighting the method-agnostic nature of our approach. While $\Pi\text{GDM}$ achieves marginally lower Wasserstein-2 values than DPS, likely due to incorporating an additional uncertainty term, the DiffPIR recommendations outperform all other methods and, for a small number of diffusion steps, slightly surpass the exact posterior denoiser.
For brevity, Appendices~\ref{sec:Aditional_results_pigdm} and~\ref{sec:appendix_DiffPIR_algorithm_spectral_reccomendations} provide the spectral recommendations for $\Pi\text{GDM}$ and DiffPIR,  and include further comparison with their heuristic counterparts and the optimal denoiser.

\subsection{Empirical distribution}
\label{sec:Empirical_Distribution}
We turn to evaluate our method on real-world datasets, including FFHQ $256 \!\times\! 256$ \cite{karras2019stylegan}, ImageNet $256\! \times\! 256$ \cite{deng2009imagenet} and fastMRI \cite{zbontar2018fastmri, knoll2020fastmri}, a single-coil knee magnetic resonance imaging (MRI) dataset centered and cropped to a spatial size of $320\!\times\!320$. In order to derive the spectral recommendations, we fit a Gaussian distribution to each dataset by estimating the corresponding mean and shift-invariant covariance matrix (see Appendix~\ref{sec:appendix_estimating_covarince}).

\begin{wrapfigure}{r}{0.52\textwidth}
    \centering
    \vspace{-0.2cm}
    \includegraphics[width=0.5\textwidth]{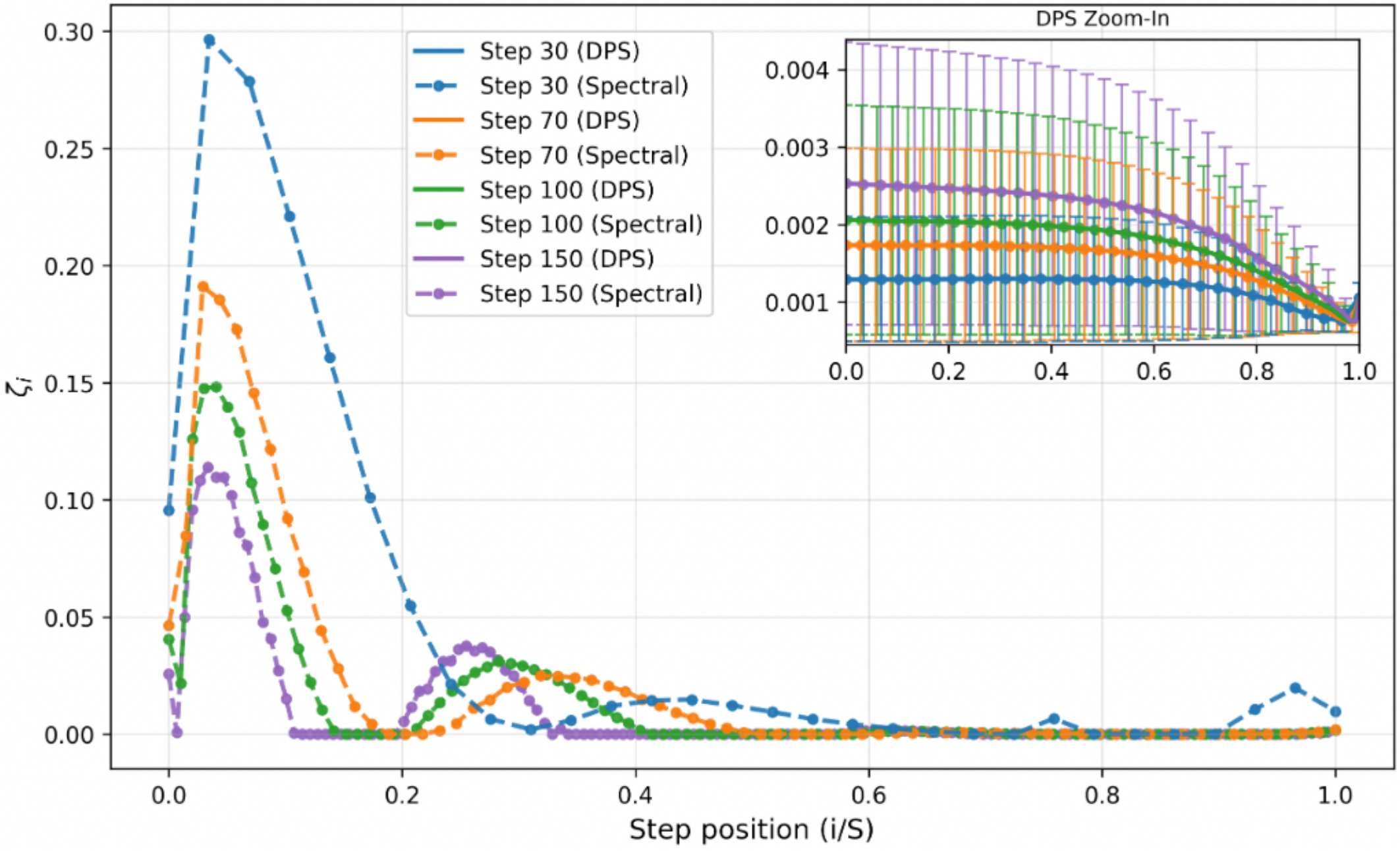}
    \caption{Comparison of spectral recommendations on the FFHQ dataset, with a zoomed-in view of the DPS heuristics. Results are shown for selected diffusion steps $S\in\{30,70,100,150\}$.}
    \label{fig:Comparison_DPS_and_Spectral_steps}
    \vspace{-0.4cm}
\end{wrapfigure}

We begin by solving the optimization problem in \eqref{eq:optimization_problem_many_realizations} within the DPS framework. 
Figure \ref{fig:Comparison_DPS_and_Spectral_steps} shows the spectral weights obtained for the FFHQ dataset under LPF degradation with $\mathcal{V}\!=\!0.1$ and $\sigma_y=0.1$, across different diffusion step counts. For comparison, we also include the DPS heuristic weights, averaged over $100$ syntheses, with variability indicated. Interestingly, the optimization reveals a markedly different weighting structure, with guidance values increasing over the reverse diffusion process. While the heuristic weights remain consistently small, the spectral recommendations place greater emphasis on the observations, suggesting a different balance between prior information and the likelihood term. Spectral recommendations and analogous analyses for the projection-based DiffPIR sampler and $\Pi\text{GDM}$, along with additional ImageNet results, are provided in Appendix~\ref{sec:Prior_comparison_results}.


As an additional degradation setting, we consider accelerated MRI acquisition for the fastMRI dataset, using the horizontal frequency subsampling protocol introduced in~\cite{jalal2021robust}. Since obtaining full Fourier measurements in MRI is time-consuming and expensive, this setting constitutes a highly relevant and challenging inverse problem that has attracted considerable research attention~\cite{weiss2019pilot, wang2022one, han2019k}. Additional details are provided in Appendix~\ref{sec:fastMRI_DPS}.

We further examine the spectral recommendations during inference, evaluating their generalization to real-world datasets beyond the Gaussian setting. Since the optimal prior denoiser~\eqref{eq:optimal_denoiser_prior} is unavailable in these settings, neural denoisers are employed instead, introducing approximation error.  For the FFHQ and ImageNet datasets, we use the pretrained models from~\cite{chung2022diffusion}, while for fastMRI we train a similar U-Net \cite{ronneberger2015u} architecture with adaptations to the MRI setting (see Appendix~\ref{sec:fastMRI_DPS}).

In our first experiment, we assess prior characteristics using $1,000$ randomly selected validation images, each independently degraded and reconstructed once with DPS using both the spectral  and the heuristic weights. To maintain consistency with the denoiser training, we employ a linear noise schedule. For evaluation, we report PSNR, SSIM~\cite{wang2004image}, LPIPS~\cite{zhang2018unreasonable}, and FID~\cite{heusel2017gans}, all computed using the piq library~\cite{kastryulin2022pytorch}, with images normalized to the $[0,\!1]$ range.

 



Table~\ref{tab:lpf_results_ffhq_H_10_sigma_1_with_fmri} reports the results on the FFHQ dataset. Overall, the spectral recommendations achieve a more balanced tradeoff between measurement fidelity and perceptual quality compared to the DPS heuristic. At relatively smaller numbers of diffusion steps, DPS heuristic attains closer FID values; however, qualitative results in Appendix~\ref{sec:Prior_comparison_results}, suggest that this behavior is associated with its small likelihood weights, leading the samples to be dominated by the pretrained prior. While this can enhance perceptual appearance, the resulting images may deviate substantially from the observed measurements, which is undesirable in inverse problems.

Table~\ref{tab:fmri_resutls_R_and_lpf} presents PSNR results on the fastMRI dataset under both LPF degradation with $\mathcal{V}\!=\!0.1$
and accelerated MRI acquisition using horizontal frequency subsampling~\cite{jalal2021robust}. Here,
$R$ denotes the acceleration factor, retaining the central $120/R$ frequencies together with an additional uniformly sampled set of $200/R$ frequencies. In both settings, additive Gaussian noise with $\sigma_y=0.1$ is applied.
Notably, the spectral recommendations consistently outperform the heuristic weighting strategies, indicating improved adherence to the underlying signal, a valuable property in MRI reconstruction. Appendix~\ref{sec:Prior_comparison_results} provides more details on the degradation operator, additional perceptual metrics, broader comparisons across $\mathcal{V}$ and $\sigma_y$ values, a range of heuristic configurations, and ImageNet results. 

While our recommendation are analytically derived for shift-invariance degradation models, we evaluate their effectiveness on settings where this property is relaxed, including random inpainting with $70\%$ pixel removal and box inpainting using a $128\times128$ mask, both with $\sigma_y = 0.1$.  We compare the results with DPS  and recent heuristic-based methods, including DAPS~\cite{zhang2025improving} and DiffPIR~\cite{zhu2023denoising}. Figures\ref{fig:inpainting_random_comparison} and \ref{fig:inpainting_box_comparison} show consistent improvements over DPS heuristics across both degradations along with frequent gains over DAPS and DiffPIR. For further discussion and images refer to appendix \ref{sec:inpainting_comparison}.

\begin{table}[t]
  \centering

  \begin{minipage}{0.48\textwidth}
    \centering
    \caption{Quantitative evaluation on FFHQ, comparing spectral recommendations with DPS heuristics ($\mathcal{V}\!=\!0.1$ and $\sigma_y\!=\!0.1$).}

    \label{tab:lpf_results_ffhq_H_10_sigma_1_with_fmri}
    \begin{small}
      \setlength{\tabcolsep}{4pt}
      \begin{tabular}{c c cccc}
        \toprule
        Method & Steps & PSNR$\uparrow$ & SSIM$\uparrow$ & LPIPS$\downarrow$ & FID$\downarrow$ \\
        \midrule
        \multirow{4}{*}{DPS}
        & $50$ & $16.00$ & $0.35$ & $0.57$ & ${69.13}$ \\
        & $100$ & $18.31$ & $0.43$ & $0.51$ & ${62.68}$ \\
        & $150$ & $19.82$ & $0.49$ & $0.45$ & $58.12$ \\
        & $200$ & $20.76$ & $0.53$ & $0.43$ & $55.88$ \\
        \midrule
        \multirow{4}{*}{Spectral}
        & $50$ & $\mathbf{25.55}$ & $\mathbf{0.69}$ & $\mathbf{0.32}$ & $\mathbf{50.86}$ \\
        & $100$ & $\mathbf{27.82}$ & $\mathbf{0.80}$ & $\mathbf{0.23}$ & $\mathbf{41.31}$ \\
        & $150$ & $\mathbf{28.10}$ & $\mathbf{0.79}$ & $\mathbf{0.23}$ & $\mathbf{37.51}$ \\
        & $200$ & $\mathbf{28.31}$ & $\mathbf{0.80}$ & $\mathbf{0.22}$ & $\mathbf{35.27}$ \\
        \bottomrule
      \end{tabular}
    \end{small}
  \end{minipage}
  \hfill
  \begin{minipage}{0.48\textwidth}
    \centering
    \caption{PSNR$\uparrow$ results on the fastMRI dataset under LPF degradation and accelerated MRI acquisition.}
    \label{tab:fmri_resutls_R_and_lpf}
    \begin{small}
      \setlength{\tabcolsep}{4pt}
      \begin{tabular}{c c cccc}
        \toprule
        Method & Steps & LPF & R=$4$& R=$8$ & R=$12$ \\
       \midrule
        \multirow{4}{*}{DPS}
        & $50$ & $17.98$ & $21.66$ & $17.62$ & $17.10$\\
        & $100$ & $19.56$ & $23.44$ & $19.06$ &  $18.26$\\
        & $150$ & $20.39$ & $24.08$ & $19.79$  & $18.96$\\
        & $200$ & $20.82$ & $24.44$ & $19.99$ & $19.27$ \\
        \midrule
        \multirow{4}{*}{Spectral}
        & $50$ & $\mathbf{23.20}$ & $\mathbf{21.66}$ & $\mathbf{19.21}$& $\mathbf{17.57}$  \\
        & $100$ & $\mathbf{24.62}$ & $\mathbf{23.44}$ & $\mathbf{20.61}$  &  $\mathbf{18.82}$ \\
        & $150$ & $\mathbf{24.90}$ & $\mathbf{24.08}$ & $\mathbf{21.26}$ & $\mathbf{19.31}$ \\
        & $200$ & $\mathbf{24.95}$ & $\mathbf{24.44}$ & $\mathbf{21.76}$  & $\mathbf{19.61}$\\
        \bottomrule
      \end{tabular}
    \end{small}
  \end{minipage}

\end{table}



We proceed with an analysis of the posterior behavior under fixed observations by selecting three representative images and applying the same LPF degradation setting. We consider both optimization formulations introduced in \eqref{eq:optimization_problem_specific_realization} and \eqref{eq:optimization_problem_many_realizations} . The first is solved independently for each observation and yields realization-specific recommendations, whereas the second captures the expected posterior behavior across degraded measurements and is solved once for a fixed degradation operator. For each image, we generate $1{,}000$ reconstructions across different numbers of diffusion steps and compute the corresponding performance metrics. 

Unlike the prior analysis, this experiment fixes the observation and probes the posterior through multiple reconstructions per image. The results, summarized in Table~\ref{tab:ffhq_lpf_10_sigma_01_1}, indicate that both spectral optimization strategies generally offer a balance between measurement fidelity and sample naturalness, with the realization-specific approach providing a modest but consistent improvement by adapting to individual observations. This benefit, however, comes at the cost of solving a separate optimization problem for each realization. Further qualitative results and details on acceleration optimization strategies, are provided in Appendices~\ref{sec:Posterior_comparison_results} and ~\ref{sec:appendix_optimization_time_analysis}.

\paragraph{Limitations} \label{sec:limitations}
Our analytical derivation assumes a Gaussian prior and relies on shift-invariance of both the covariance matrix and the degradation operator, enabling a common diagonalizing basis. While one could alternatively work in the eigenbasis of a single matrix and represent the other accordingly, this may lead to poor approximation quality, particularly under spatially varying degradations. Nevertheless, we demonstrate that the proposed recommendations remain effective beyond this setting, including on real-world datasets and under non-shift-invariant operators, as well as on practical scenarios such as accelerated MRI, which naturally align with the required conditions. Extending the framework to richer priors, such as Gaussian mixture models, or to representations that preserve tractability without joint diagonalization remains an interesting direction for future work.

\section{Conclusion}
In this work, we propose a spectral approach for analyzing zero-shot diffusion posterior sampling methods. Under a Gaussian assumption, we derive the output distributions induced by these samplers as well as that of the ideal posterior sampler, which provides a theoretical reference for analysis. We demonstrate how our approach enables a principled design of the heuristic components used by these schemes, while jointly accounting for properties of the prior distribution, the degraded observations, and the dynamics of the diffusion process. Finally, we examine our recommendation on existing methods across various real-world scenarios and characterize their behavior in terms of sample naturalness and measurement fidelity. We hope this work offers a useful perspective to inform future research on developing and analyzing zero-shot posterior sampling methods.

\bibliographystyle{plain}
\bibliography{references}

\newpage
\appendix
\section{The Optimal Denoiser for a Gaussian Posterior}
\label{sec:appendix_Map_est}

Let $\x_0 \sim \N(\bmu, \bSigmaZ)$  represent the distribution of the original dataset, where $\x_0 \in \mathbb{R}^{d}$. 
A general linear inverse problem is posed as: $\y = \bH
\x_0+\nn$ with $\y \in \mathbb{R}^{m}$, $\bH \in \mathbb{R}^{m \times d}$ a known linear degradation matrix, and $\nn \sim \mathcal{N}(\mathbf{0}, \sigma_y^2 \mathbf{\I})$ i.i.d. additive Gaussian noise with known variance.

Under the Gaussian prior and linear measurement model, the posterior distribution of $\mathbf{x}_0$ given $\mathbf{y}$ is Gaussian; consequently, the MAP estimator coincides with the posterior mean and takes the form of a Wiener filter, which is optimal under mean-squared error.

For the Maximum A Posteriori (MAP) estimation, we seek to maximize the posterior distribution: 

\begin{equation}\nonumber
\underset{x_0}{\max} \log p(\mathbf{x}_0 | \mathbf{x}_t, \y)
\end{equation}

Using Bayes' rule, this can be written as:
\begin{equation}\nonumber
\underset{x_0}{\min} -\log \left[ \frac{p(\y | \mathbf{x}_0, \mathbf{x}_t) p(\mathbf{x}_0| \x_t)}{p(\y|\mathbf{x}_t)} \right]
\end{equation}
\begin{equation}\label{eq:map_bayes}
\underset{x_0}{\min} -\log p(\y | \mathbf{x}_0, \mathbf{x}_t) -\log p(\mathbf{x}_0| \x_t)
\end{equation}


Since $\y$ is conditionally independent of $\x_t$ given $\x_0$, The conditional log-likelihood $ \log p(\y | \mathbf{x}_0, \mathbf{x}_t)$ is given by:

\begin{equation}\nonumber
p(\y | \mathbf{x}_0, \mathbf{x}_t) = \mathcal{N}(\y; \bH \x_0, \sigma_y^2 \I) = \frac{1}{\sqrt{\left(2\pi\sigma_{y}^2\right)^d}} \exp \left\{-\frac{1}{2\sigma_{y}^2}\left(\y-\mathbf{H} \x_0\right)^T\left(\y-\mathbf{H} \x_0\right)\right\}
\end{equation}

\begin{equation}\label{eq:y_x_0_x_t}
\log{p(\y | \mathbf{x}_0, \mathbf{x}_t)} = -\frac{1}{2}\log{\left(2\pi\sigma_y^2\right)^d } -\frac{1}{2\sigma_{y}^2}\left(\y-\mathbf{H} \x_0\right)^T\left(\y-\mathbf{H} \x_0\right)
\end{equation}

Using Bayes' rule, $ \log{p(\x_0 | \x_t)} \propto \log{p(\x_t | \x_0)} +  \log{p(\x_0)}$

Through the diffusion process, the signal undergoes noise contamination, leading to the following marginal expression for $\x_t$:
\begin{equation}\label{eq:marginal_diffusion}
    \mathbf{x}_t = \sqrt{\bar{\alpha}_t} \mathbf{x}_0 + \sqrt{1-\bar{\alpha}_t} \boldsymbol{\epsilon} ~~~~~  \boldsymbol{\epsilon} \sim \mathcal{N}(\mathbf{0},\boldsymbol{I})
\end{equation}

where $\alpha_t $ for $t\in[1,T]$ is referred to as the incremental noise schedule and $\balpha_t = \prod_{i=1}^{t} \alpha_i$.

Hence, the conditional log-likelihood $ \log{p(\x_t | \x_0)} $ is given by:
\begin{equation}\nonumber\label{eq:x_t_x_0}
p(\x_t|\x_0) =  \mathcal{N}(\mathbf{x}_t; \sqrt{\bar{\alpha}_t}\, \mathbf{x}_0, (1-\bar{\alpha}_t)\mathbf{I}) = \frac{1}{\sqrt{\left(2\pi(1-\balpha_t)\right)^d)}} \exp \left\{-\frac{1}{2(1-\balpha_t)}\left(\x_t-\sqrt{\balpha_t}\x_0\right)^T\left(\x_t-\sqrt{\balpha_t}\x_0\right)\right\}
\end{equation}

\begin{equation}
\log \, p(\x_t|\x_0) = -\frac{1}{2}\log{\left(2\pi(1-\balpha_t)\right)^d} - \frac{1}{2(1-\balpha_t)}\left(\x_t-\sqrt{\balpha_t} \x_0\right)^T\left(\x_t-\sqrt{\balpha_t} \x_0\right)
\end{equation}


The log-likelihood $ \log{p(\x_0)} $ is given by:
\begin{equation}\label{eq:x_0}
p(\x_0) =  \mathcal{N}(\mathbf{x}_0; \boldsymbol{\mu}, \boldsymbol{\Sigma}_0) = \frac{1}{\sqrt{\left(2\pi\right)^d|\bSigmaZ|}} \exp \left\{-\frac{1}{2}\left(\x_0-\bmu\right)^T\bSigmaZ^{-1}\left(\x_0-\bmu\right)\right\}
\end{equation}

\begin{equation}
\log{p(\x_0)} = -\frac{1}{2}\log{\left(2\pi\right)^d|\bSigmaZ|} - \frac{1}{2}\left(\x_0-\bmu\right)^T\bSigmaZ^{-1}\left(\x_0-\bmu\right)
\end{equation}

By substituting the expressions from Equations~\ref{eq:y_x_0_x_t}, \ref{eq:x_t_x_0}, and \ref{eq:x_0} into Equation~\ref{eq:map_bayes} and differentiating with respect to $\mathbf{x}_0$, we obtain:

\begin{equation}\nonumber
-\frac{2\mathbf{H}^T\left(\y - \mathbf{H} \x_0\right)}{2{\sigma_{y}}^2}
-\frac{2\sqrt{\balpha_t}\left(\mathbf{x}_t - \sqrt{\balpha_t}\mathbf{x}_0\right)}{2(1-\balpha_t)} + \frac{2{\boldsymbol{\Sigma}}_0^{-1} \left(\mathbf{x}_0 - \boldsymbol{\mu}_0\right)}{2}  = 0
\end{equation}

This simplifies to:
\begin{equation}\nonumber
-\frac{\mathbf{H}^T\left(\y - \mathbf{H} \x_0\right)}{{\sigma_{y}}^2} -\frac{\sqrt{\balpha_t}\left(\mathbf{x}_t - \sqrt{\balpha_t}\mathbf{x}_0\right)}{(1-\balpha_t)} + {\boldsymbol{\Sigma}}_0^{-1} \left({\mathbf{x}}_0 - \boldsymbol{\mu}_0\right) = 0
\end{equation}

Resulting in:
\begin{equation}\nonumber
-(1-\balpha_t){\boldsymbol{\Sigma}}_0\mathbf{H}^T\left(\y- \mathbf{H} \x_0\right) -  \sigma_{y}^2\sqrt{\balpha_t}{\boldsymbol{\Sigma}}_0\left( \x_t - \sqrt{\balpha_t}\x_0\right)
+ \sigma_{y}^2 (1-\balpha_t)\left(\x_0 - \boldsymbol{\mu}_0\right) = 0
\end{equation}

Thus:
\begin{equation}\nonumber
\left((1-\balpha_t){\boldsymbol{\Sigma}}_0\mathbf{H}^T\mathbf{H} +  \sigma_{y}^2{\balpha_t}{\boldsymbol{\Sigma}}_0+I\sigma_{y}^2 (1-\balpha_t)\right)\x_0 = (1-\balpha_t){\boldsymbol{\Sigma}}_0\mathbf{H}^T\y+\sigma_{y}^2\sqrt{\balpha_t}{\boldsymbol{\Sigma}}_0\x_t+\sigma_{y}^2 (1-\balpha_t)\boldsymbol{\mu}_0
\end{equation}

Finally:

\begin{align}\label{eq:map_estimator_posterior}
\mathbf{x}_0^{*}  &= 
\left((1-\balpha_t){\boldsymbol{\Sigma}}_0\mathbf{H}^T\mathbf{H} +  \sigma_{y}^2{\balpha_t}{\boldsymbol{\Sigma}}_0+I\sigma_{y}^2 (1-\balpha_t)\right)^{-1}\\ \nonumber
& \left((1-\balpha_t){\boldsymbol{\Sigma}}_0\mathbf{H}^T\y+\sigma_{y}^2\sqrt{\balpha_t}{\boldsymbol{\Sigma}}_0\x_t+\sigma_{y}^2 (1-\balpha_t)\boldsymbol{\mu}_0\right)
\end{align}

$$
$$

\section{The Reverse Process in the Time Domain}
\label{sec:appendix_inference_Time}
Here, we present the reverse process in the time domain for the DDIM \cite{song2020denoising}.
Let $\x_0$ follow the distribution:
$$ 
\x_0 \sim \mathcal{N}(\bmu, \bSigmaZ), \quad \x_0 \in \R^{d}
$$
Using the procedure outline in \cite{song2020denoising}, the diffusion process begins with $ \x_S\sim \mathcal{N}(\boldsymbol{0}, \I)$, where $ \x_S \in \R^{d} $ and progresses through an iterative denoising process described as follows:\footnote{We follow here the DDIM notations that replaces $t$ with $s$, where the steps \(\left[1, \ldots, S\right]\) form a subsequence of \(\left[1, \ldots, T\right]\) and \(S = T\).}
\begin{equation}
    \x_{s-1}(\eta) = \sqrt{\balpha_{s-1}} \left( 
    \frac{\x_{s} - \sqrt{1 - \balpha_{s}} \cdot \bepsilon_\theta(\x_{s},s)}{\sqrt{\balpha_{s}}}\right)
    + \sqrt{1 - \balpha_{s-1} - \sigma_{s}^2(\eta)} \cdot \bepsilon_\theta(\x_s,s) + \sigma_{s}(\eta)\mathbf{z}_s
\end{equation}

where
\begin{equation}\label{eq:sgma_eta}
    \sigma_{s}(\eta) = \eta \sqrt{\frac{1 - \balpha_{{s-1}}}{1 - \balpha_s}} \sqrt{1 - \frac{\balpha_{s}}{\balpha_{{s-1}}}}
\end{equation}  
Substituting the marginal property from \eqref{eq:marginal_dist}:
\begin{equation}
\boldsymbol{\epsilon}_\theta(\x_s, s) = \frac{\x_s-\sqrt{{\balpha}_s}\hat{\x}_0}{\sqrt{1-\balpha_s}} \quad \quad \hat{\x}_0 = \frac{\x_{s} - \sqrt{1 - \balpha_{s}} \cdot \bepsilon_\theta(\x_{s},s)}{\sqrt{\balpha_{s}}}
\end{equation}  

\begin{equation}
    \x_{s-1}(\eta) = \sqrt{\balpha_{s-1}} \hat{\x}_0
    + \sqrt{1 - \balpha_{s-1} - \sigma_{s}^2(\eta)} \left(\frac{\x_s-\sqrt{{\balpha}_s}\hat{\x}_0}{\sqrt{1-\balpha_s}}\right) + \sigma_{s}(\eta)\mathbf{z}_s
\end{equation}
For the deterministic scenario, we choose \(\eta = 0\) in \eqref{eq:sgma_eta} and obtain $ \sigma_{s}(\eta=0) = 0$. Therefore:
\begin{align}\nonumber
    \x_{s-1}(\eta=0) &= \sqrt{\balpha_{s-1}} \hat{\x}_0
    + \sqrt{1 - \balpha_{s-1}} \left(\frac{\x_s-\sqrt{{\balpha}_s}\hat{\x}_0}{\sqrt{1-\balpha_s}}\right) \\
     &= \frac{\sqrt{1 - \balpha_{s-1}}}{\sqrt{1-\balpha_s}}\x_s + \left[  \sqrt{\balpha_{s-1}} - \frac{\sqrt{{\balpha}_s}\sqrt{1 - \balpha_{s-1}}}{\sqrt{1 - \balpha_{s}}}\right]\hat{\x}_0
\end{align}
We define:
$$a_s = \left[\frac{\sqrt{1 - \balpha_{s-1}}}{\sqrt{1-\balpha_s}}\right]\I \quad \quad b_s =   \left[\sqrt{\balpha_{s-1}} - \frac{\sqrt{{\balpha}_s}\sqrt{1 - \balpha_{s-1}}}{\sqrt{1 - \balpha_{s}}}\right]\I$$
Therefore, the update can be written as:
$$
\x_{s-1} = a_s\x_s +b_s\x_0^* ~.
$$

Substituting the MAP estimator from Equation~\ref{eq:map_estimator_posterior} into the DDIM update yields

\begin{align}\nonumber
\mathbf{x}_{s-1} &= a_s\mathbf{x}_s + \\ \nonumber
& b_s\left[\left((1-\balpha_s){\boldsymbol{\Sigma}}_0\mathbf{H}^T\mathbf{H} +  \sigma_{y}^2{\balpha_s}{\boldsymbol{\Sigma}}_0+\I\sigma_{y}^2 (1-\balpha_s)\right)^{-1} \left((1-\balpha_s){\boldsymbol{\Sigma}}_0\mathbf{H}^T\y+\sigma_{y}^2\sqrt{\balpha_s}{\boldsymbol{\Sigma}}_0\x_s+\sigma_{y}^2 (1-\balpha_s)\boldsymbol{\mu}_0\right)\right] 
\end{align}
\begin{align}\nonumber
\mathbf{x}_{s-1} = \left[a_s + b_s\left[ \left((1-\balpha_s){\boldsymbol{\Sigma}}_0\mathbf{H}^T\mathbf{H} +  \sigma_{y}^2\balpha_s{\boldsymbol{\Sigma}}_0+\I\sigma_{y}^2 (1-\balpha_s)\right)^{-1}\right]\sigma_{y}^2\sqrt{\balpha_s}{\boldsymbol{\Sigma}}_0\right]\x_s 
+ \\ \nonumber b_s\left((1-\balpha_s){\boldsymbol{\Sigma}}_0\mathbf{H}^T\mathbf{H} +  \sigma_{y}^2\balpha_s{\boldsymbol{\Sigma}}_0+\I\sigma_{y}^2 (1-\balpha_s)\right)^{-1} \left((1-\balpha_s){\boldsymbol{\Sigma}}_0\mathbf{H}^T\y\right)
+ \\ \nonumber b_s\left((1-\balpha_s){\boldsymbol{\Sigma}}_0\mathbf{H}^T\mathbf{H} +  \sigma_{y}^2\balpha_s{\boldsymbol{\Sigma}}_0+\I\sigma_{y}^2 (1-\balpha_s)\right)^{-1} \left(\sigma_{y}^2 (1-\balpha_s)\boldsymbol{\mu}_0\right)
\end{align}

We define:
\begin{align} \nonumber
\bar{\boldsymbol{\Sigma}}_{0,s} =(1-\balpha_s){\boldsymbol{\Sigma}}_0\mathbf{H}^T\mathbf{H} +  \sigma_{y}^2\balpha_s{\boldsymbol{\Sigma}}_0+\I\sigma_{y}^2 (1-\balpha_s)
\end{align}

Using this notation, the update becomes:
\begin{align}\nonumber
    \mathbf{x}_{s-1} = & \left(a_s+ b_s \left(\bar{\boldsymbol{\Sigma}}_{0,s} \right)^{-1} \sigma_{y}^2 \sqrt{\balpha_s}{\boldsymbol{\Sigma}}_0 \right)\mathbf{x}_s  + b_s\left(\bar{\boldsymbol{\Sigma}}_{0,s} \right)^{-1}\left((1-\balpha_s){\boldsymbol{\Sigma}}_0\mathbf{H}^T\y\right)
     \\ & + b_s\left(\bar{\boldsymbol{\Sigma}}_{0,s} \right)^{-1} \left(\sigma_{y}^2 (1-\balpha_s)\boldsymbol{\mu}_0\right)
\end{align}







\section{Spectral Form of the Update Rule}\label{sec:appendix_inference_spectral}


To migrate to the spectral domain, we apply the Discrete Fourier Transform (DFT), represented by the unitary matrix $\mathbf{F}$, to both sides of the DDIM update:

\begin{align}
\mathbf{x}^{\mathcal{F}}_{s-1} 
&= \mathbf{F} \left(a_s + b_s \bar{\boldsymbol{\Sigma}}_{0,s}^{-1} \sigma_y^2 \sqrt{\balpha_s} \boldsymbol{\Sigma}_0 \right) \mathbf{x}_s
\mathbf{F}^T \nonumber \\
& \quad + \mathbf{F} b_s \bar{\boldsymbol{\Sigma}}_{0,s}^{-1} \left( (1-\balpha_s)\boldsymbol{\Sigma}_0 \mathbf{H}^T \mathbf{y} + \sigma_y^2 (1-\balpha_s) \boldsymbol{\mu}_0 \right) \mathbf{F}^T,
\end{align}

where the superscript $\mathcal{F}$ indicates the Fourier-domain representation of a vector, e.g., $\mathbf{x}^{\mathcal{F}}_s = \mathbf{F}\mathbf{x}_s$.

\begin{align} \mathbf{x}^{\mathcal{F}}_{s-1} & =  a_s \mathbf{x}^{\mathcal{F}}_{s} + \F b_s \left(\bar{\boldsymbol{\Sigma}}_{0,s} \right)^{-1} \F^T\F\sigma_{y}^2 \sqrt{\balpha_s}{\boldsymbol{\Sigma}}_0 \F^T\F\mathbf{x}_s \nonumber \\ 
 & + b_s\F\left(\bar{\boldsymbol{\Sigma}}_{0,s} \right)^{-1}\F^T\F(1-\balpha_s){\boldsymbol{\Sigma}}_0\F^T\F\mathbf{H}^T\F^T\F\y  +  b_s\F\left(\bar{\boldsymbol{\Sigma}}_{0,s} \right)^{-1}\F^T\F\sigma_{y}^2 (1-\balpha_s)\bmu 
\end{align}

\begin{align}\label{eq:spectral_domain_middle}
    \mathbf{x}^{\mathcal{F}}_{s-1}  & = a_s \mathbf{x}^{\mathcal{F}}_{s} 
    +  b_s\F \left(\bar{\boldsymbol{\Sigma}}_{0,s} \right)^{-1} \F^T \sqrt{\balpha_s}\sigma_{y}^2\F{\boldsymbol{\Sigma}}_0 \F^T\mathbf{x}^{\mathcal{F}}_{s}  \nonumber 
    \nonumber \\ \quad & + b_s\F\left(\bar{\boldsymbol{\Sigma}}_{0,s} \right)^{-1}\F^T(1-\balpha_s)\left(\F{\boldsymbol{\Sigma}}_0\F^T\right)\left(\F\mathbf{H}^T\F^T\right)\y^{\mathcal{F}}  +  b_s\F\left(\bar{\boldsymbol{\Sigma}}_{0,s} \right)^{-1}\F^T\sigma_{y}^2 (1-\balpha_s)\bmu^{\mathcal{F}}
\end{align}

Assuming that $\mathbf{H}$ is circulant with eigenvalues $\{h_i\}_{i=1}^d$, and noting that $\boldsymbol{\Sigma}_0$ is also circulant, both matrices commute and can be diagonalized via the Discrete Fourier Transform (DFT) matrix $\F$, which greatly simplifies the matrix inversion. specifically:





\begin{itemize}

\item $\mathbf{\Lambda}_0 = \F\boldsymbol{\Sigma}_0 \F^T  $, \quad $ \boldsymbol{\Sigma}_0 = \F^T \mathbf{\Lambda}_0 \F $
    
\item $  \mathbf{H} = \F^T \mathbf{\Lambda}_{\mathbf{H}} \F , \quad   \mathbf{H^T} = \F^T \mathbf{\Lambda}_{\mathbf{H^T}} \F , \quad   \mathbf{H}^T\mathbf{H} = \F^T \mathbf{\Lambda}_{\mathbf{H}^T\mathbf{H}} \F  $

\item $a \bSigma_0 + b \I = a \F \mathbf{\Lambda}_0 \F^T + b \F \I \F^T = \F \left(a \mathbf{\Lambda}_0 + b\I\right) \F^T$
\item $ \bSigma_0^{-1} = \F \mathbf{\Lambda}_0^{-1} \F^T $  , \quad $\mathbf{\Lambda}_0^{-1} = \F^T\boldsymbol{\Sigma}_0^{-1} \F $

\end{itemize}

Therefore, we obtain:

$$ \boldsymbol{\Sigma}_0\mathbf{H}^T\mathbf{H} = \F^T \mathbf{\Lambda}_0 \F\F^T \mathbf{\Lambda}_{\mathbf{H}^T\mathbf{H}} \F =   \F^T \mathbf{\Lambda}_0 \mathbf{\Lambda}_{\mathbf{H}^T\mathbf{H}} \F $$

and:

$$
\F\left(\bar{\boldsymbol{\Sigma}}_{0,s}\right)^{-1}\F^T = \F\left[(1-\balpha_s){\boldsymbol{\Sigma}}_0\mathbf{H}^T\mathbf{H} +  \sigma_{y}^2\balpha_s{\boldsymbol{\Sigma}}_0+\sigma_{y}^2 (1-\balpha_s)\I\right]^{-1}\F^T
$$

$$
=\left((1-\balpha_s) \mathbf{\Lambda}_0 \mathbf{\Lambda}_{\mathbf{H}^T\mathbf{H}}  +  \sigma_{y}^2\balpha_s \mathbf{\Lambda}_0 +\sigma_{y}^2 (1-\balpha_s)\I\right)^{-1}
$$

$$
 = diag\left((1-\balpha_s)\lambda_ih^2_i +  \sigma_{y}^2\balpha_s\lambda_i +\sigma_{y}^2 (1-\balpha_s)\I\right)^{-1}
$$

$$
\F\left(\bar{\boldsymbol{\Sigma}}_{0,s}\right)^{-1}\F^T = diag\left(\frac{1}{(1-\balpha_s)\lambda_ih^2_i +  \sigma_{y}^2\balpha_s\lambda_i +\sigma_{y}^2 (1-\balpha_s)}\right)
$$

We denote the expression as: 
\begin{equation}\label{eq:experssion_FsigF}
    \F\left(\bar{\boldsymbol{\Sigma}}_{0,s}\right)^{-1}\F^T = diag\left(\frac{1}{\lambda_{sum}}\right) = \mathbf{\Lambda}_{sum}^{-1}
\end{equation}\\
Substitute \ref{eq:experssion_FsigF} into \ref{eq:spectral_domain_middle}


$$ \mathbf{x}^{\mathcal{F}}_{s-1} = a_s \mathbf{x}^{\mathcal{F}}_{s} +  b_s\mathbf{\Lambda}_{sum}^{-1} \sqrt{\balpha_s}\sigma_{y}^2\mathbf{\Lambda}_0\mathbf{x}^{\mathcal{F}}_{s}  + b_s\mathbf{\Lambda}_{sum}^{-1}(1-\balpha_s)\mathbf{\Lambda}_0\mathbf{\Lambda}_{\mathbf{H}^T}\y^{\mathcal{F}} +  b_s\mathbf{\Lambda}_{sum}^{-1}\sigma_{y}^2 (1-\balpha_s)\bmu^{\mathcal{F}}  $$

$$ \mathbf{x}^{\mathcal{F}}_{s-1} = \left[a_s  +  b_s\mathbf{\Lambda}_{sum}^{-1} \sqrt{\balpha_s}\sigma_{y}^2\mathbf{\Lambda}_0\right]\mathbf{x}^{\mathcal{F}}_{s}  + b_s\mathbf{\Lambda}_{sum}^{-1}(1-\balpha_s)\mathbf{\Lambda}_0\mathbf{\Lambda}_{\mathbf{H}^T}\y^{\mathcal{F}} +  b_s\sigma_{y}^2 (1-\balpha_s)\mathbf{\Lambda}_{sum}^{-1}\bmu^{\mathcal{F}} $$\\

We can then recursively obtain
\(\mathbf{x}_m^{\mathcal{F}}\) for a general \(m\):
$$
\mathbf{x}_{m}^{\mathcal{F}} = \prod_{s=m+1}^{S} \left[a_s  +  b_s\mathbf{\Lambda}_{sum}^{-1} \sqrt{\balpha_s}\sigma_{y}^2\mathbf{\Lambda}_0\right]\mathbf{x}_S^{\mathcal{F}}+\left[\sum_{i=m+1}^{S}\left[b_i\mathbf{\Lambda}_{sum}^{-1}(1-\balpha_s)\mathbf{\Lambda}_0\mathbf{\Lambda}_{\mathbf{H}^T}\right]\prod_{j=m+1}^{i-1} \left[a_j  +  b_j\mathbf{\Lambda}_{sum}^{-1} \sqrt{\balpha_s}\sigma_{y}^2\mathbf{\Lambda}_0\right]\right]\y^{\mathcal{F}}$$
$$+\left[\sum_{i=m+1}^{S}\left[b_i\sigma_{y}^2 (1-\balpha_s)\mathbf{\Lambda}_{sum}^{-1}\right]\prod_{j=m+1}^{i-1} \left[a_j  +  b_j\mathbf{\Lambda}_{sum}^{-1} \sqrt{\balpha_s}\sigma_{y}^2\mathbf{\Lambda}_0\right]\right]\bmu^{\mathcal{F}}$$

specifically for m=0:

$$
\mathbf{x}_{0}^{\mathcal{F}} = \prod_{s=1}^{S} \left[a_s  +  b_s\mathbf{\Lambda}_{sum}^{-1} \sqrt{\balpha_s}\sigma_{y}^2\mathbf{\Lambda}_0\right]\mathbf{x}_S^{\mathcal{F}}+\left[\sum_{i=1}^{S}\left[b_i\mathbf{\Lambda}_{sum}^{-1}(1-\balpha_s)\mathbf{\Lambda}_0\mathbf{\Lambda}_{\mathbf{H}^T}\right]\prod_{j=1}^{i-1} \left[a_j  +  b_j\mathbf{\Lambda}_{sum}^{-1} \sqrt{\balpha_s}\sigma_{y}^2\mathbf{\Lambda}_0\right]\right]\y^{\mathcal{F}}$$
$$
+\left[\sum_{i=1}^{S}\left[b_i\sigma_{y}^2 (1-\balpha_s)\mathbf{\Lambda}_{sum}^{-1}\right]\prod_{j=1}^{i-1} \left[a_j  +  b_j\mathbf{\Lambda}_{sum}^{-1} \sqrt{\balpha_s}\sigma_{y}^2\mathbf{\Lambda}_0\right]\right]\bmu^{\mathcal{F}}
$$

we will denote the following:

\begin{equation}\nonumber
    \V_1 = \prod_{s=1}^{S} \left[a_s  +  b_s\mathbf{\Lambda}_{sum}^{-1} \sqrt{\balpha_s}\sigma_{y}^2\mathbf{\Lambda}_0\right]\mathbf{x}_S^{\mathcal{F}}  
\end{equation}

\begin{equation}\nonumber
      \V_2 = \left[\sum_{i=1}^{S}\left[b_i\mathbf{\Lambda}_{sum}^{-1}(1-\balpha_s)\mathbf{\Lambda}_0\mathbf{\Lambda}_{\mathbf{H}^T}\right]\prod_{j=1}^{i-1} \left[a_j  +  b_j\mathbf{\Lambda}_{sum}^{-1} \sqrt{\balpha_s}\sigma_{y}^2\mathbf{\Lambda}_0\right]\right]
\end{equation}

\begin{equation}\nonumber
      \V_3 = \left[\sum_{i=1}^{S}\left[b_i\sigma_{y}^2 (1-\balpha_s)\mathbf{\Lambda}_{sum}^{-1}\right]\prod_{j=1}^{i-1} \left[a_j  +  b_j\mathbf{\Lambda}_{sum}^{-1} \sqrt{\balpha_s}\sigma_{y}^2\mathbf{\Lambda}_0\right]\right]
\end{equation}

\begin{equation}\label{eq:closed_form_optimal_case}
    \mathbf{x}_{0}^{\mathcal{F}} = \V_1\mathbf{x}_S^{\mathcal{F}}+ \V_2\y^{\mathcal{F}} + \V_3\bmu^{\mathcal{F}}
\end{equation}\\ \\ 

Given the closed-form expression for $\x_{0}^{\mathcal{F}}$ in Equation~\ref{eq:closed_form_optimal_case}, we derive its mean and covariance as functions of the problem settings.

\textbf{Mean:}
\begin{align}\nonumber
\boldsymbol{\mu}_{opt} = \mathbb{E}\left[\V_1\mathbf{x}_S^{\mathcal{F}}+ \V_2\y^{\mathcal{F}} + \V_3\bmu^{\mathcal{F}}\mid \mathbf{y}\right]
   &= \V_1 \F \mathbb{E}\left[\mathbf{x}_S \mid \mathbf{y}\right] +  \V_2 \F \mathbb{E}\left[\mathbf{y} \mid \mathbf{y}\right] +  \V_3 \F \mathbb{E}\left[\bmu \mid \mathbf{y}\right]    \\ \nonumber \quad &= \V_1 \F \mathbb{E}\left[\mathbf{x}_S\right] +  \V_2 \F \mathbf{y} + \V_3 \F \bmu \\ \nonumber  \quad & = \V_2 \mathbf{y}^{\mathcal{F}} +  \V_3 \bmu^{\mathcal{F}}
\end{align}

\begin{empheq}[box=\fbox]{align}    \boldsymbol{\mu}_{opt} = \mathbb{E}\left[ \V_1\mathbf{x}_S^{\mathcal{F}}+ \V_2\y^{\mathcal{F}} + \V_3\bmu^{\mathcal{F}}\right] =  \V_2 \mathbf{y}^{\mathcal{F}} + \V_3 \bmu^{\mathcal{F}} \end{empheq}

\textbf{Covariance:}
\begin{align}\nonumber
\mathbf{\Lambda}_{opt} = & \mathbb{E}[\left(\V_1 \mathbf{x}_S^{\mathcal{F}} + \V_2 \mathbf{y}^{\mathcal{F}} + \V_3\bmu^{\mathcal{F}} - \mathbb{E}\left[\V_1 \mathbf{x}_S^{\mathcal{F}} + \V_2 \mathbf{y}^{\mathcal{F}} + \V_3\bmu^{\mathcal{F}}|\y\right]\right) \\ \nonumber \quad &\left(\V_1 \mathbf{x}_S^{\mathcal{F}} + \V_2 \mathbf{y}^{\mathcal{F}} + \V_3\bmu^{\mathcal{F}} - \mathbb{E}\left[\V_1 \mathbf{x}_S^{\mathcal{F}} + \V_2 \mathbf{y}^{\mathcal{F}}+ \V_3\bmu^{\mathcal{F}}|\y\right]\right)^T \mid \mathbf{y}]
 \\ \nonumber \quad &=
 \mathbb{E}\left[\left(\V_1 \mathbf{x}_S^{\mathcal{F}} + \V_2 \mathbf{y}^{\mathcal{F}} +\V_3\bmu^{\mathcal{F}}  - \V_2 \mathbf{y}^{\mathcal{F}}- \V_3\bmu^{\mathcal{F}} \right)\left(\V_1 \mathbf{x}_S^{\mathcal{F}} + \V_2 \mathbf{y}^{\mathcal{F}}+\V_3\bmu^{\mathcal{F}}  - \V_2 \mathbf{y}^{\mathcal{F}}- \V_3\bmu^{\mathcal{F}} \right)^T \mid \mathbf{y}\right]
 \\ \nonumber \quad &= \mathbb{E}\left[\left(\V_1 \mathbf{x}_S^{\mathcal{F}}\right)\left(\V_1 \mathbf{x}_S^{\mathcal{F}}\right)^T \mid \mathbf{y}\right]
 \\ \nonumber \quad &= \V_1 \mathbb{E}\left[\mathbf{x}_S^{\mathcal{F}} \left(\mathbf{x}_S^{\mathcal{F}}\right)^T\right] {\V_1}^T
   \\ \nonumber \quad &= \V_1 {\V_1}^T
\end{align}

\begin{empheq}[box=\fbox]{align}     \mathbf{\Lambda}_{opt} =\V_1^2 \end{empheq}

\section{Gaussian Posterior Distribution}
\label{sec:appendix_gaussian_posterior_distribution}

Here, we derive a closed-form expression for $p(\x \mid \y)$ under a linear degradation model. 
Specifically, we assume that the clean signal $\x \sim \mathcal{N}(\bmu, \bSigma_0)$ and the noise $\nn \sim \mathcal{N}(\0, \sigma_n^2 \I)$ are independent Gaussian vectors, and that the observation is given by a linear transformation $\y = \bH \x + \nn$, where $\bH$ is a deterministic matrix. 
Under these assumptions, the joint distribution $p(\x, \y)$ is Gaussian.

Define the concatenated vector $\z = \begin{pmatrix} \x \\ \y \end{pmatrix} = \begin{pmatrix} \x \\ \bH \x + \nn \end{pmatrix}$.  
Then $\z$ is jointly Gaussian with mean and covariance:
$$ \mathbb{E}[\z] = \boldsymbol{\mu}_{(x, y)} =
\begin{pmatrix} \boldsymbol{\mu}_x \\ \boldsymbol{\mu}_y \end{pmatrix}, 
\quad
\bSigma_{\z} = \bSigma_{(x, y)} =
\begin{pmatrix} \bSigma_x & \bSigma_{xy} \\ \bSigma_{yx} & \bSigma_y \end{pmatrix} $$

where:

$$ \boldsymbol{\mu}_x = \bmu $$
$$ \boldsymbol{\mu}_y =\mathbb{E}[\y] = \mathbb{E}[\bH\x + \nn] = \bH\mathbb{E}[\x] + \mathbb{E}[\nn] =  \bH\bmu $$
The covariance of $\y$ is then simply
$$
\bSigma_y = \mathrm{Cov}[\y] = \bH \bSigma_x \bH^T + \sigma_n^2 \I,
$$




Hence:

\begin{empheq}[box=\fbox]{align}\label{eq:true_y_distribution}
  p(\y) = \mathcal{N}(\y; \bH\bmu, \bH\bSigma_x\bH^T  + \sigma_n^2\I)
\end{empheq}

The cross-covariance is then:
$$
\bSigma_{xy} = \mathbb{E}[(\x -  \boldsymbol{\mu}_x)(\y -  \boldsymbol{\mu}_y)^T] $$

$$
\bSigma_{xy} = \mathrm{Cov}[\x, \y] = \bSigma_x \bH^T, \quad
\bSigma_{yx} = \mathrm{Cov}[\y, \x] = \bH \bSigma_x.
$$







Therefore, we can derive the Joint Mean and Covariance of $p\left(\x,\y\right)$:

$$ \boldsymbol{\mu}_{(x, y)} = \begin{pmatrix} \bmu \\ \bH\boldsymbol{\mu}_x
\end{pmatrix} 
\quad \quad ,
 \bSigma_{(x, y)} = \begin{pmatrix} \bSigma_x & \bSigma_0 \bH^T \\ \bH^T \bSigma_0&\bH\Sigma_x\bH^T + \sigma_n^2\I  \end{pmatrix} $$




To compute $p(\x \mid \y)$, we use the standard result for joint Gaussian distributions: the conditional distribution is also Gaussian with mean and covariance:

\textbf{Mean of} $p(\x \mid \y)$:
\begin{align}
 \boldsymbol{\mu}_{x \mid y} &= \boldsymbol{\mu}_x + \bSigma_{xy} \bSigma_y^{-1} (\y-\boldsymbol{\mu}_y) \\ \nonumber \quad &=
 \bmu + \bSigma_0 \bH^T (\bH\Sigma_0\bH^T + \sigma_n^2\I)^{-1} (\y - \bH\boldsymbol{\mu}_0) 
\end{align}
\textbf{Covariance of} $p(\x \mid \y)$: 
\begin{align}
 \bSigma_{x \mid y} &=  \bSigma_x - \bSigma_{xy} \bSigma_y^{-1} \bSigma_{yx} \\ \nonumber \quad &=
 \bSigma_0 - \bSigma_0 \bH^T (\bH\Sigma_0\bH^T  + \sigma_n^2\I)^{-1} \bH \bSigma_0 
\end{align}

Therefore, the conditional probability distribution $p(x \mid y)$ is:

$$
\boxed{
\begin{aligned}
p(\x \mid \y) &= \mathcal{N}\!\big(\x;\, 
\bmu + \bSigma_0 \bH^T (\bH\Sigma_0\bH^T + \sigma_n^2\I)^{-1} (\y - \bH\boldsymbol{\mu}_0),\\
&\quad \bSigma_0 - \bSigma_0 \bH^T (\bH\Sigma_0\bH^T  + \sigma_n^2\I)^{-1} \bH \bSigma_0 
\big)
\end{aligned}
}
$$

\textbf{Migrating to the spectral domain:}\label{sec:true_posterior_distribuion}

Here we describe the posterior $p(\x \mid \y)$ in the spectral domain. Let $\F$ denote the unitary Discrete Fourier Transform (DFT) matrix and the superscript $\mathcal{F}$ denotes the Fourier-domain representation. Then
\begin{align}\nonumber
\boldsymbol{\mu}_{\mathbf{x}_0 \mid \mathbf{y}}^{\mathcal{F}} = &\F \mathbb{E}\left[\mathbf{x}_{0} \mid \mathbf{y}\right]   \\ \quad = & \nonumber \F \left[ \bmu + \bSigma_0 \bH^T (\bH\Sigma_0\bH^T + \sigma_n^2\I)^{-1} (\y- \bH\bmu ) \right] \\ \quad = & \nonumber  \F \bmu +  \F \bSigma_0  \F^T\F \bH^T\F^T\F (\bH \Sigma_{\x_0} \bH^T   + \sigma_{n}^2 \I)^{-1} \F^T\F(\y-\bH\bmu ) 
\\ \quad = & \nonumber  \bmu^{\mathcal{F}} +  \bLambda\boldsymbol{\Lambda}_{H^T} (\boldsymbol{\Lambda}_H\bLambda\boldsymbol{\Lambda}_{H^T} -\nonumber \boldsymbol{\Lambda}_H \bmu^{\mathcal{F}}\{\bmu^{\mathcal{F}}\}^T\boldsymbol{\Lambda}_{H^T} + \sigma_{n}^2 \I)^{-1} (\y^{\mathcal{F}}-\boldsymbol{\Lambda}_H\bmu^{\mathcal{F}} ) 
\end{align}

By defining:
\begin{align}\nonumber
\mathbf{H} = \F^T \mathbf{\Lambda}_{\mathbf{H}} \F,\quad
 \mathbf{H^T} = \F^T \mathbf{\Lambda}_{\mathbf{H^T}} \F,\quad
 \mathbf{H}^T\mathbf{H} = \F^T \mathbf{\Lambda}_{\mathbf{H}^T\mathbf{H}} \F 
\end{align}
and: 
\begin{align}\nonumber
\F \left[ \bH \bSigma_{0} \bH^T \right] \F^T = \F  \bH \F^T\F \bSigma_{0} \F^T\F \bH^T \F^T = \boldsymbol{\Lambda}_H\bLambda\boldsymbol{\Lambda}_{H^T}
\end{align}

We get:

\begin{empheq}[box=\fbox]{align}\label{eq:mean_real_distribution_x_y_spectral}    \boldsymbol{\mu}_{\mathbf{x}_0 \mid \mathbf{y}}^{\mathcal{F}}  =
\bmu^{\mathcal{F}} +  \bLambda\boldsymbol{\Lambda}_{H^T} (\boldsymbol{\Lambda}_H\bLambda\boldsymbol{\Lambda}_{H^T}   + \sigma_{n}^2 \I)^{-1} (\y^{\mathcal{F}}-\boldsymbol{\Lambda}_H\bmu^{\mathcal{F}} ) \end{empheq}

\begin{align}\nonumber
\boldsymbol{\Lambda}_{x_0|y} =& \F\Sigma_{\x_0|\y} \F^T  
\\ \quad = & \nonumber \F\left[\bSigma_0 - \bSigma_0 \bH^T (\bH\Sigma_0\bH^T  + \sigma_n^2\I)^{-1} \bH \bSigma_0\right]\F^T
\\ \quad = & \nonumber \F\bSigma_0\F^T - \F\bSigma_0 \F^T\F \bH^T\F^T\F (\bH \bSigma_0 \bH^T   + \sigma_n^2\I)^{-1}\F^T\F \bH\F^T\F \bSigma_0\F^T 
\\ \quad = & \nonumber   \bLambda -  \bLambda  \boldsymbol{\Lambda}_{H^T} (\boldsymbol{\Lambda}_H\bLambda\boldsymbol{\Lambda}_{H^T} + \sigma_{n}^2 \I)^{-1}\boldsymbol{\Lambda}_H\bLambda
\end{align}

Therefore:


\begin{empheq}[box=\fbox]{align}    \boldsymbol{\Lambda}_{x_0|y} =         \bLambda -  \bLambda  \boldsymbol{\Lambda}_{H^T} (\boldsymbol{\Lambda}_H\bLambda\boldsymbol{\Lambda}_{H^T} + \sigma_{n}^2 \I)^{-1}\boldsymbol{\Lambda}_H\bLambda\end{empheq}

\section{DPS algorithm}
\label{sec:appendix_DPS_algorithm}
In this section, we present the DPS algorithm \cite{chung2022diffusion} from a spectral perspective and show how it can be integrated into our method.

As discussed in Section~\ref{sec:background_training_free_methods}, to avoid retraining a model for a specific degradation matrix $\bH$, DPS \cite{chung2022diffusion} uses a Bayesian formulation, in which the posterior score gradient is expressed as
\begin{equation}
\nabla_{\x_t} \log p_t(\x_t | \y) = \nabla_{\x_t} \log p_t(\x_t) + \nabla_{\x_t} \log p_t(\y | \x_t)
\end{equation}

While the first term can be obtained directly from a pretrained prior model, the second term is expressed by marginalizing over $\x_0$:
\begin{equation}\label{eq:marginilization_DPS}
p(\y | \x_t) = \int_{\x_0} p(\x_0 | \x_t) p(\y | \x_0) \, d\x_0
\end{equation}
Here, the conditional probability $p(\y |\x_0)$ is determined by the linear observation model $\y = \bH \x_0 + \nn$, which implies a Gaussian distribution:
$$
p(\y | \x_0) = \mathcal{N}(\bH \x_0, \sigma_y^2\I)
$$
$$
p(\y|\x_0) = \frac{1}{\sqrt{(2\pi)^d\sigma_y^2}}\exp\left[-\frac{||\y-\bH\x_0||^2_2}{2\sigma^2}\right]
$$  
A key approximation in DPS is modeling the distribution $p(\x_0 | \x_t)$ as a deterministic Gaussian centered at the estimated mean:
$$
p(\x_0 | \x_t) \approx \mathcal{N}(\hat{\x}_0, \0)
$$
Applying this deterministic model in \eqref{eq:marginilization_DPS} gives a closed-form expression for the conditional likelihood:
$$
p_t(\y | \x_t) \approx \mathcal{N}(\bH \hat{\x}_0, \sigma_\y^2 \I)
$$  
which leads to the gradient:
$$
\nabla_{\x_t} \log p_t(\y | \x_t) \simeq - \frac{1}{\sigma_y^2} \nabla_{\x_t} \|\y - \bH \hat{\x}_0\|_2^2
$$  
The authors of \cite{chung2022diffusion} reweight this term using the heuristic
$\zeta_i = \zeta' / \lvert \y - \bH \hat{\x}_0(\x_s) \rvert$,
which scales the step size inversely with the norm of the measurement error, where $\zeta'$ is a user-defined constant.
$$
\nabla_{\x_t} \log p_t(\y | \x_t) \simeq - \zeta_i \nabla_{\x_t} \|\y - \bH \hat{\x}_0\|_2^2
$$

\subsection{The Reverse Process in the Time Domain}\label{sec:DPS_time_domain}

The original formulation of the DPS algorithm is based on DDPM \cite{ho2020denoising}. 
Here, we present it \cite{chung2022diffusion} within the DDIM  \cite{song2020denoising} framework under the Variance-Preserving (VP) parameterization \cite{song2020score}.
\begin{equation}\label{eq:DDIM_DPS_eq}
\x_{s-1} = \sqrt{\bar\alpha_{t-1}}\hat\x_0 
+ \sqrt{1 - \bar{\alpha}_{t-1} - \sigma_s^2(\eta)} \, \bepsilon_\theta(\x_s, s)  
- \zeta_i \nabla_{\x_t} \|\y - \bH \hat{\x}_0\|_2^2
+ \sigma_s(\eta) \z_s.
\end{equation}

where:
$$
\sigma_s(\eta) = \eta \sqrt{\frac{1 - \bar\alpha_{s-1}}{1 - \bar\alpha_s}} \sqrt{1 - \frac{\bar\alpha_s}{\bar\alpha_{s-1}}}.
$$
We consider the deterministic setting, corresponding to $\eta = 0$.
$$\x_{s-1} = \sqrt{\bar\alpha_{s-1}}\hat\x_0 + \sqrt{1-\bar{\alpha}_{s-1}}\bepsilon_\theta(\x_s,s)  - \zeta_i \nabla_{\x_t} \|\y - \bH \hat{\x}_0\|_2^2 $$
Using the marginal property,
\begin{equation}
\boldsymbol{\epsilon}_\theta(\x_s, s) = \frac{\x_s - \sqrt{\balpha_s}\hat{\x}0}{\sqrt{1-\balpha_s}} .
\end{equation}
which, substituted into Equation \ref{eq:DDIM_DPS_eq}, gives:
$$\x_{s-1} = \sqrt{\bar\alpha_{s-1}}\hat\x_0 + \sqrt{1-\bar{\alpha}_{s-1}}\left[ \frac{\x_s-\sqrt{{\balpha}_s}\hat{\x}_0}{\sqrt{1-\balpha_s}} \right] - \zeta_i \nabla_{\x_t} \|\y - \bH \hat{\x}_0\|_2^2 $$
$$\x_{s-1} = \frac{\sqrt{1 - \balpha_{s-1}}}{\sqrt{1-\balpha_s}}\x_s + \left[  \sqrt{\balpha_{s-1}} - \frac{\sqrt{{\balpha}_s}\sqrt{1 - \balpha_{s-1}}}{\sqrt{1 - \balpha_{s}}}\right]\hat{\x}_0 - \zeta_i \nabla_{\x_t} \|\y - \bH \hat{\x}_0\|_2^2 $$
Introducing:
\begin{equation}\nonumber
a_s = \left[\frac{\sqrt{1 - \balpha_{s-1}}}{\sqrt{1-\balpha_s}}\right]\I
\end{equation}
\begin{equation}\nonumber
b_s = \left[\sqrt{\balpha_{s-1}} - \frac{\sqrt{{\balpha}_s}\sqrt{1 - \balpha_{s-1}}}{\sqrt{1 - \balpha_{s}}}\right]\I
\end{equation}
we obtain:
$$\x_{s-1} =a_s\x_s + b_s\hat{\x}_0- \zeta_i \nabla_{\x_t} \|\y - \bH \hat{\x}_0\|_2^2$$
Where $\hat{\x}_0$ denotes the MMSE denoiser, corresponds to the denoised signal produced by a pretrained network. Assuming a Gaussian prior, the optimal solution $\x_0^*$ is given by \cite{benita2025spectral}:
$$\x_0^* = \left(\balpha_s \bSigmaZ + (1-\balpha_s) \I \right)^{-1} \left(\sqrt{\balpha_s} \bSigmaZ \x_s + (1-\balpha_s) \bmu \right)$$
Therefore we get:
\begin{empheq}[box=\fbox]{align}\label{eq:DPS_time_part_1}
\x_{s-1} = a_s\x_s + b_s\x_0^* - \zeta_i \nabla_{\x_t} \|\y - \bH \x_0^*\|_2^2 
\end{empheq}
Since $\x_0^*$ depends on $\x_s$ we denote it as $\x_0^*(\x_s)$:
$$\zeta_i\nabla_{x_s}||\y-\bH\x_0^*||_2^2 = \zeta_i\nabla_{x_s}||\y-\bH\x_0^*(\x_s)||_2^2$$
Hence, the gradient can be written as:
$$\zeta_i\nabla_{x_t}||\y-\bH\x_0^*||_2^2 = \zeta_i\left[-2\left[\nabla_{x_s}\x_0^*(\x_s)\right]^T\bH^T\left(\y-\bH\x_0^*(\x_s)\right)\right]$$
Where:
$$\nabla_{x_s}\x_0^*(\x_s) = \nabla_{x_s}\left[ \left(\balpha_s \bSigmaZ + (1-\balpha_s) \I \right)^{-1} \left(\sqrt{\balpha_s} \bSigmaZ \x_s + (1-\balpha_s) \bmu \right) \right]$$
\begin{align}\label{eq:gradient_optimal_denoiser}
\nabla_{x_s}\x_0^*(\x_s) = \left(\balpha_s \bSigmaZ + (1-\balpha_s) \I \right)^{-1} \sqrt{\balpha_s} \bSigmaZ  
\end{align}
This yields:

$$\zeta_i\nabla_{x_s}||\y-\bH\x_0^*||_2^2 = \zeta_i\left[-2 \sqrt{\balpha_s} \bSigmaZ \left(\balpha_s \bSigmaZ + (1-\balpha_s) \I \right)^{-1} \bH^T\left(\y-\bH\x_0^*(\x_s)\right)\right]$$
Substituting into \eqref{eq:DPS_time_part_1} gives:
\begin{empheq}[box=\fbox]{align}\label{eq:DPS_time_sampling_part_1}
 \x_{s-1} = a_s\x_s + b_s\x_0^* +2 \zeta_i\left[ \sqrt{\balpha_s} \bSigmaZ \left(\balpha_s \bSigmaZ + (1-\balpha_s) \I \right)^{-1} \bH^T\left(\y-\bH\x_0^*\right)\right]
\end{empheq}


\subsection{Spectral form of the update rule}\label{sec:DPS_migrating_spectral}
Next, we apply the Discrete Fourier Transform (DFT), denoted by $\mathcal{F}$, to Equation \eqref{eq:DPS_time_sampling_part_1}. Assuming that $\boldsymbol{\Sigma}_0$ and $\mathbf{H}$ are shift-invariant, both matrices are diagonalized by the DFT. Specifically we denote:
\begin{align}\nonumber
\mathbf{F} \mathbf{H} \mathbf{F}^T &= \boldsymbol{\Lambda}_h = \mathrm{diag}(h_1, h_2, \dots, h_d), \\\nonumber
\mathbf{F} \mathbf{H}^T \mathbf{F}^T &= \boldsymbol{\Lambda}_h^* = \mathrm{diag}(\bar{h}_1, \bar{h}_2, \dots, \bar{h}_d), \\\nonumber
\mathbf{F} \mathbf{H}^T \mathbf{H} \mathbf{F}^T &= \boldsymbol{\Lambda}_{|h|^2} = \mathrm{diag}(|h_1|^2, |h_2|^2, \dots, |h_d|^2).
\end{align}
Applying the unitary Fourier operator $\mathbf{F}$ transforms Equation \eqref{eq:DPS_time_sampling_part_1} into the frequency domain.
$$
\F\x_{s-1} = \F\left[a_s\x_s + b_s\x_0^* +2 \zeta_i\left[ \sqrt{\balpha_s} \bSigmaZ \left(\balpha_s \bSigmaZ + (1-\balpha_s) \I \right)^{-1} \bH^T\left(\y-\bH\x_0^*\right)\right]\right]
$$
\begin{align}\label{eq:DFT_DPS_part_1}
\x^{\mathcal{F}}_{s-1} = a_s\x^{\mathcal{F}}_s + b_s\F\x_0^* +2 \zeta_i\F\left[ \sqrt{\balpha_s} \bSigmaZ \left(\balpha_s \bSigmaZ + (1-\balpha_s) \I \right)^{-1} \bH^T\left(\y-\bH\x_0^*\right)\right]
\end{align}
Where:
\begin{align}\nonumber
\F\x_0^* &= \F \left[ \left(\balpha_s \bSigmaZ + (1-\balpha_s) \I \right)^{-1} \left(\sqrt{\balpha_s} \bSigmaZ \x_s + (1-\balpha_s) \bmu \right)\right]
\\ \nonumber
 &= \F  \left(\balpha_s \bSigmaZ + (1-\balpha_s) \I \right)^{-1} \F^T\F \left(\sqrt{\balpha_s} \bSigmaZ \x_s + (1-\balpha_s) \bmu \right)
\\ \nonumber
 &= \left[\balpha_s \mathbf{\Lambda}_0 + (1-\balpha_s)\I\right]^{-1} \left(\sqrt{\balpha_s} \F \bSigmaZ \F^T\F \x_s + (1-\balpha_s) \F \bmu \right)
\end{align}
\begin{align}
\label{eq:DFT_optimal_solution}
\F\x_0^* = \left[\balpha_s \mathbf{\Lambda}_0 + (1-\balpha_s)\I\right]^{-1} \left(\sqrt{\balpha_s}\bLambda \x^{\mathcal{F}}_s + (1-\balpha_s)  \bmu^{\mathcal{F}} \right)
\end{align}
and:
\begin{align}\nonumber\label{eq:DFT_part_1_DPS}
 &\F\left[\sqrt{\balpha_s} \bSigmaZ \left(\balpha_s \bSigmaZ +(1-\balpha_s) \I \right)^{-1} \bH^T\left(\y-\bH\x_0^*\right)\right]
\\ \nonumber
  &= \nonumber \left[\sqrt{\balpha_s} \F\bSigmaZ \F^T\F \left(\balpha_s \bSigmaZ + (1-\balpha_s) \I \right)^{-1}\F^T\F \bH^T\F^T\F\left(\y-\bH\x_0^*\right)\right]
\\ \nonumber
  &=  \left[\sqrt{\balpha_s} \bLambda\left[\balpha_s \mathbf{\Lambda}_0 + (1-\balpha_s)\I\right]^{-1} \bLambdahcon\left(\F\y-\F\bH\F^T\F\x_0^*\right)\right]
\\ \nonumber
  &= \left[\sqrt{\balpha_s} \bLambda\left[\balpha_s \mathbf{\Lambda}_0 + (1-\balpha_s)\I\right]^{-1} \bLambdahcon\left(\F\y-\F\bH\F^T\F\x_0^*\right)\right]
\\ 
  &=  \left[\sqrt{\balpha_s} \bLambda\left[\balpha_s \mathbf{\Lambda}_0 + (1-\balpha_s)\I\right]^{-1} \bLambdahcon\left(\y^{\mathcal{F}}-\bLambdah\F\x_0^*\right)\right].
\end{align}

By defining: 
\begin{equation}\nonumber
c_s = \sqrt{\balpha_s}\bLambda \left[\balpha_s \mathbf{\Lambda}_0 + (1-\balpha_s)\I\right]^{-1}
\end{equation}
\begin{equation}\nonumber
d_s = (1-\balpha_s) \left[\balpha_s \mathbf{\Lambda}_0 + (1-\balpha_s)\I\right]^{-1}
\end{equation}
Equation \ref{eq:DFT_optimal_solution} can be rewritten as:
\begin{equation}\label{eq:DFT_part_1_DPS_signed}
\F\x_0^* = c_s \x^{\mathcal{F}}_s + d_s  \bmu^{\mathcal{F}} 
\end{equation}
Substitute it into Equation \ref{eq:DFT_part_1_DPS}, we get:
$$
\F\left[\sqrt{\balpha_s} \bSigmaZ \left(\balpha_s \bSigmaZ +(1-\balpha_s) \I \right)^{-1} \bH^T\left(\y-\bH\x_0^*\right)\right]
 =  \left[c_s \bLambdahcon\left(\y^{\mathcal{F}}-\bLambdah(c_s \x^{\mathcal{F}}_s + d_s  \bmu^{\mathcal{F}} )\right)\right]
$$
Finally, Equation \ref{eq:DFT_DPS_part_1} can be rewritten as:
$$\x^{\mathcal{F}}_{t-1} = a_s\x^{\mathcal{F}}_s + b_s [c_s \x^{\mathcal{F}}_s + d_s  \bmu^{\mathcal{F}} ] + 2\zeta_i  \left[c_s \bLambdahcon\left(\y^{\mathcal{F}}-\bLambdah(c_s \x^{\mathcal{F}}_s + d_s  \bmu^{\mathcal{F}} )\right)\right]
$$
$$\x^{\mathcal{F}}_{t-1} = [a_s + b_s c_s - 2\zeta_i c_s \bLambdahcon\bLambdah c_s]\x^{\mathcal{F}}_s  + [2\zeta_i c_s \bLambdahcon]\y^{\mathcal{F}} + [ b_sd_s - 2\zeta_i  c_s \bLambdahcon \bLambdah d_s]\bmu^{\mathcal{F}}
$$
Denoting:
$$\G(s) = [a_s + b_s c_s - 2\zeta_i c_s \bLambdahcon\bLambdah c_s] $$
$$\Q(s) = [ 2\zeta_i c_s \bLambdahcon] $$
$$\M(s) = [ b_s d_s - 2\zeta_i  c_s \bLambdahcon \bLambdah d_s] $$
\begin{align}\label{eq:spectral_domain_dps_one_step}
\x^{\mathcal{F}}_{s-1} = \G(s)\x^{\mathcal{F}}_s+\Q(s)\y^{\mathcal{F}}   + \M(s)\bmu^{\mathcal{F}}
\end{align}

By recursively applying Equation \eqref{eq:spectral_domain_dps_one_step}, we obtain a frequency-domain representation for each diffusion step $\mathbf{x}_l^{\mathcal{F}}$:
$$
\mathbf{x}_{l}^{\mathcal{F}} = \left[ \prod_{j=l+1}^{S}\G(j)\right] \mathbf{x}_S^{\mathcal{F}} + \left[\sum_{i=l+1}^{S}\left(\prod_{j=l+1}^{i-1}\G(j)\right)\Q(i)\right]\boldsymbol{\y}^{\mathcal{F}} + \left[\sum_{i=l+1}^{S}\left(\prod_{j=l+1}^{i-1}\G(j)\right)\M(i)\right]\boldsymbol{\mu_0}^{\mathcal{F}}  
$$
Specifically for $l=0$:
$$
\mathbf{x}_{0}^{\mathcal{F}} = \left[ \prod_{j=1}^{S}\G(j)\right] \mathbf{x}_S^{\mathcal{F}}  + \left[\sum_{i=1}^{S}\left(\prod_{j=1}^{i-1}\G(j)\right)\Q(i)\right]\boldsymbol{\y}^{\mathcal{F}} + \left[\sum_{i=1}^{S}\left(\prod_{j=1}^{i-1}\G(j)\right)\M(i)\right]\boldsymbol{\mu_0}^{\mathcal{F}} 
$$
Denoting:
\begin{equation}\nonumber
 \D_1 =    \prod_{j=l+1}^{S}\G(j)
\end{equation}
\begin{equation}\nonumber
 \D_2 = \sum_{i=l+1}^{S}\left(\prod_{j=l+1}^{i-1}\G(j)\right)\Q(i)
\end{equation}
\begin{equation}\nonumber
 \D_3 =\sum_{i=l+1}^{S}\left(\prod_{j=l+1}^{i-1}\G(j)\right)\M(i)
\end{equation}
\begin{equation}\label{eq:appen_DDPM_X_0_X_T}
\mathbf{{x}}_{0}^{\mathcal{F}} = \D_1 \mathbf{x}_S^{\mathcal{F}} + \D_2 \y^{\mathcal{F}} + \D_3\boldsymbol{\mu_0}^{\mathcal{F}} 
\end{equation}
Equation \ref{eq:appen_DDPM_X_0_X_T} can be interpreted as a transfer function, describing the relationship between the input Gaussian noise $\x_S^{\mathcal{F}} \sim \mathcal{N}(\mathbf{0}, \mathbf{I})$ and the estimated output signal $\mathbf{x}_{0,\text{DPS}}^{\mathcal{F}}$. The output signal is also Gaussian, with the following distributional properties:

\textbf{Mean:}
\begin{equation}\label{eq:mean_DPS_spectral}\nonumber
\mathbb{E}  \left[\mathbf{x}_{0}^{\mathcal{F}} \right]=  \D_2 \y^{\mathcal{F}} + \D_3\bmu^{\mathcal{F}} 
\end{equation}
\textbf{Covariance:}
\begin{equation} \nonumber
\bSigma_{\mathbf{{x}}_{0}^{\mathcal{F}}} =\mathbb{E}  \left[ \left[\mathbf{{x}}_{0}^{\mathcal{F}}-\mathbb{E}  \left[\mathbf{{x}}_{0}^{\mathcal{F}} \right]\right]\left[\mathbf{{x}}_{0}^{\mathcal{F}}-\mathbb{E}  \left[\mathbf{{x}}_{0}^{\mathcal{F}} \right]\right]^T\right] = \left[ \D_1\right]^2
\end{equation}



\begin{empheq}[box=\fbox]{align}\label{eq:DPS_freq_final}
\mathbf{\hat{x}}_{0}^{\mathcal{F}} \sim \N\!\!\left( \D_2 \y^{\mathcal{F}} + \D_3\bmu^{\mathcal{F}}  ~,~  \D_1^2  \right)
\end{empheq}
$$
$$

\section{Averaged Wasserstein Distance}
\label{sec:appendix_Averaged_Wasserstein_loss}

To bridge the analytical spectral recommendations with practical implementation, we formulate the optimization in terms of an average loss over the distribution of observations $\y$, modeled as $\y \sim  \mathcal{N}(\boldsymbol{\mu}_y, \bSigma_y)$ with $\y = \bH \x + \boldsymbol{n}$. We propose the Average Wasserstein distance between the original and estimated posterior distributions as the loss function.

This formulation naturally accommodates two regimes. For a single observation ($K=1$), the optimization reduces to the standard Wasserstein-2 loss associated with that realization. Alternatively, when multiple realizations are considered, the weight parameters are optimized to capture the expected behavior of the posterior $p(\x \mid \y)$ across $\y$, eliminating the need to solve the optimization individually for each sample. In this average setting, the optimal solution can be expressed directly in terms of the statistical properties of $p(\y)$ rather than for a specific observation.
In what follows, we derive the resulting average loss formulation for both the single-sample case ($K\! = \!1$) and the regime of large $K$.

The Wasserstein-2 distance between two Gaussian distributions with means  $\mu_1$ and $\mu_2$, and covariance matrices $\Sigma_1$ and $\Sigma_2$,  and the corresponding eigenvalues $\{\lambda_1^{(i)}\}_{i=1}^d$ and $\{\lambda_2^{(i)}\}_{i=1}^d$  is given by:

\begin{equation}
W_2(\mathcal{N}_1, \mathcal{N}_2) = \sqrt{(\boldsymbol{\mu}_1 - \boldsymbol{\mu}_2)^T (\boldsymbol{\mu}_1 - \boldsymbol{\mu}_2) + \sum_i \left( \sqrt{\lambda_1^{(i)}} - \sqrt{\lambda_2^{(i)}} \right)^2}
\end{equation}
In addition, We define the average Wasserstein distance as: 
\begin{align}
\left[\mathcal{L}_{W_2}^2\right]_\text{Avg} &=\frac{1}{K} \sum_{k=1}^KW^2_2\left( p(\x_0^{\mathcal{F}}|\y), p(\mathbf{\hat{x}}_{0}^{\mathcal{F}} \,\big|\,  \y^{\mathcal{F}})\right)
\end{align}

Using the true posterior distribution derived in Section \ref{sec:true_posterior_distribuion} and, as a representative example, the DPS posterior from Equation \eqref{eq:DPS_freq_final}, we have:
$$ \quad \mathbf{\hat{x}}_{0}^{\mathcal{F}} \,\big|\,  \y^{\mathcal{F}} \sim \N\!\!\left( \D_2 \y^{\mathcal{F}} + \D_3\bmu^{\mathcal{F}}  ~,~  \D_1^2  \right)  ~, \quad \mathbf{\hat{x}}_{0}^{\mathcal{F}}  \in \mathbb{R}^{d} \quad \text{and} \quad \x_0^{\mathcal{F}}|\y \sim \mathcal{N}( \boldsymbol{\mu}_{\mathbf{x}_0 \mid \mathbf{y}}^{\mathcal{F}},\boldsymbol{\Lambda}_{x_0|y}), \quad \x_0^{\mathcal{F}} \in \mathbb{R}^{d}\quad $$ 
Substituting these expressions into the Wasserstein formula yields:
\begin{align}\label{eq:avg_wasserstein_2_distance_simpler}
\left[\mathcal{L}_{W_2}^2\right]_\text{Avg} &= \frac{1}{K} \sum_{k=1}^K \Bigg( \sum_i \left( \sqrt{\lambda_0^{(i)}} - \D_1^{(i)} \right)^2 + \sum_i \left( \boldsymbol{\mu}_{\mathbf{x}^{\mathcal{F}}|\y,k}^{(i)} - 
\boldsymbol{\mu}_{\mathbf{\hat{x}}^{\mathcal{F}}|\y,k}^{(i)} \right)^2 \Bigg) 
\end{align}
Where $\left\{\lambda_0^{(i)}\right\}_{i=1}^{d}$ are the eigenvalues of $p(\x_0^{\mathcal{F}}|\y)$ located on the diagonal of the matrix $\boldsymbol{\Lambda}_{x_0|y}$ and $\D_1^{(i)}$ denotes the $i$th diagonal entry of the matrix $\D_1$.

We now derive the mean discrepancy between the two posteriors. The mean of the true posterior from Equation~\eqref{eq:mean_real_distribution_x_y_spectral} is:

\begin{equation}\label{eq:original_mu_posterior_freq}
 \boldsymbol{\mu}_{\mathbf{x}^{\mathcal{F}} \mid \mathbf{y}}  =
\bmu^{\mathcal{F}} +  \bA (\y^{\mathcal{F}}-\boldsymbol{\Lambda}_H\bmu^{\mathcal{F}} ) = \bA \y^{\mathcal{F}} + \left(\I-\bA\boldsymbol{\Lambda}_H\right)\bmu^{\mathcal{F}}\end{equation} 
where:
 $$\bA = \bLambda\boldsymbol{\Lambda}_{H^T} (\boldsymbol{\Lambda}_H\bLambda\boldsymbol{\Lambda}_{H^T}   + \sigma_{n}^2 \I)^{-1}$$

Similarly, the mean of the DPS posterior from Equation~\eqref{eq:mean_DPS_spectral} is:
$$   \boldsymbol{\mu}_{\mathbf{\hat{x}}^{\mathcal{F}} \mid \mathbf{y}}  =
\D_2 \y^{\mathcal{F}} + \D_3\bmu^{\mathcal{F}} $$

The mean discrepancy between the two posteriors is therefore:

\begin{equation}\label{eq:delta_mu}
\Delta\boldsymbol{\mu}(\y) = \boldsymbol{\mu}_{\mathbf{\hat{x}}^{\mathcal{F}} \mid \mathbf{y}}  -  \boldsymbol{\mu}_{\mathbf{x}^{\mathcal{F}} \mid \mathbf{y}}= \left(\D_2-\bA\right)\y^{\mathcal{F}}+\left(\D_3-\I + \bA\boldsymbol{\Lambda}_H\right)\bmu^{\mathcal{F}}
\end{equation}

As for $K=1$ the average Wasserstein distance trivially reduces to the standard Wasserstein distance, substituting $\Delta\boldsymbol{\mu}(\y)$ into Equation~\eqref{eq:avg_wasserstein_2_distance_simpler} for $K=1$ yields:

\begin{align}\label{eq:optimization_problem_specific_realization_k_1}
\left[\mathcal{D}_{W_2}^2\right] &_\text{Avg} =
\textstyle\sum_{i=1}^{d} \left( \sqrt{\lambda_i} - 
 [\D_1]_i  \right)^2  + \textstyle\sum_{i=1}^{d} \left([\D_2 \!- \!\bA ]_i[\y^{\mathcal{F}} ]_i+[\D_3\!-\!\I \!+\! \bA\boldsymbol{\Lambda}_{h} ]_i[\bmu^{\mathcal{F}} ]_i\right)^2
\end{align}

We now turn to the regime of large $K$ Introducing the shorthand:
$$ \M = \left(\D_2-\bA\right) \quad \text{and} \quad \bb=\left(\D_3-\I + \bA\boldsymbol{\Lambda}_H\right)\bmu^{\mathcal{F}}$$
the mean discrepancy simplifies to:

$$\Delta\boldsymbol{\mu}(\y) =  \M \y^{\mathcal{F}}+ \bb$$

Focusing on the mean term of the Wasserstein-2 loss:

\begin{align}\nonumber
&\frac{1}{K} \sum_{k=1}^K \| \boldsymbol{\mu}_{\mathbf{x}^{\mathcal{F}}|\y,k} - 
\boldsymbol{\mu}_{\mathbf{\hat{x}}^{\mathcal{F}}|\y,k} \|_2^2  \\\nonumber
&= \frac{1}{K} \sum_{k=1}^K \| \Delta\boldsymbol{\mu}(\y_k) \|_2^2 
= \frac{1}{K} \sum_{k=1}^K \| \M \y_k^{\mathcal{F}}+ \bb
\|_2^2  \\\nonumber
&= \frac{1}{K} \sum_{k=1}^K (\y_k^\mathcal{F})^T \M^T \M \y_k^\mathcal{F} + 2 \bb^T \M \y_k^\mathcal{F} + \bb^T \bb \\ \nonumber
&= \operatorname{Tr} \Big( \M^T \M \frac{1}{K} \sum_{k=1}^K \y_k^\mathcal{F} (\y_k^\mathcal{F})^T \Big) + 2 \bb^T \M \boldsymbol{\mu}_y^\mathcal{F} + \bb^T \bb \\ \nonumber
&= \operatorname{Tr}\Big( \M^T \M (\boldsymbol{\Lambda}_y + \boldsymbol{\mu}_y^\mathcal{F} (\boldsymbol{\mu}_y^\mathcal{F})^T) \Big) + 2 \bb^T \M \boldsymbol{\mu}_y^\mathcal{F} + \bb^T \bb \\\label{eq:mean_discrepancy_frequency} 
&= \operatorname{Tr} (\M^T \M \boldsymbol{\Lambda}_y) + \| \M \boldsymbol{\mu}_y^\mathcal{F} + \bb \|_2^2
\end{align}

where 
$$
\boldsymbol{\mu}_y^\mathcal{F} = \frac{1}{K} \sum_{k=1}^K \mathbf{y}_k^\mathcal{F}, 
\qquad
\boldsymbol{\Lambda}_y = \frac{1}{K} \sum_{k=1}^K \left( \mathbf{y}_k^\mathcal{F} - \boldsymbol{\mu}_y^\mathcal{F} \right) 
\left( \mathbf{y}_k^\mathcal{F} - \boldsymbol{\mu}_y^\mathcal{F} \right)^T
$$
denote the empirical mean and covariance of $\mathbf{y}$ in the frequency domain.
Substituting \ref{eq:mean_discrepancy_frequency} into the average Wasserstein distance \ref{eq:avg_wasserstein_2_distance_simpler} yields:

\begin{align}\label{eq:average_wasserstein_3}
\left[\mathcal{L}_{W_2}^2\right]_\text{Avg} &= \frac{1}{K} \sum_{k=1}^K \Bigg( \sum_i \left( \sqrt{\lambda_0^{(i)}} - \D_1^{(i)} \right)^2 + \sum_i \left( \boldsymbol{\mu}_{\mathbf{x}^{\mathcal{F}}|\y,k}^{(i)} - 
\boldsymbol{\mu}_{\mathbf{\hat{x}}^{\mathcal{F}}|\y,k}^{(i)} \right)^2 \Bigg) 
\\ \nonumber
 &= \sum_i \left( \sqrt{\lambda_0^{(i)}} - \D_1^{(i)} \right)^2  +  \operatorname{Tr} (\M^T \M \boldsymbol{\Lambda}_y) + \| \M \boldsymbol{\mu}_y^\mathcal{F} + \bb \|_2^2 
\end{align}

Finally, using the true distribution of $\mathbf{y}$ in terms of the prior from Equation~\eqref{eq:true_y_distribution}:

$$
\boldsymbol{\mu}_y^\mathcal{F} = \bLambdah \, \boldsymbol{\mu}_x^\mathcal{F}, 
\qquad
\boldsymbol{\Lambda}_y = \bLambdah \, \bLambda \, \bLambdahcon + \sigma_n^2 \, \I
$$

substituting into Equation~\eqref{eq:average_wasserstein_3} gives the final closed-form expression for the averaged Wasserstein-2 distance:

\begin{empheq}[box=\fbox]{align} \label{eq:final_average_wasserstein_distance}   \left[\mathcal{L}_{W_2}^2\right]_\text{Avg} &= \sum_i \left( \sqrt{\lambda_0^{(i)}} - \D_1^{(i)} \right)^2  +  \operatorname{Tr} \left(\M^T \M \left(\bLambdah \, \bLambda \, \bLambdahcon + \sigma_n^2 \, \I\right)\right) + \| \M \bLambdah \, \boldsymbol{\mu}_x^\mathcal{F} + \bb \|_2^2 \end{empheq}

\section{Kullback-Leibler divergence}
\label{sec:dkl_derivative}
The Kullback-Leibler (KL) divergence between two Gaussian distributions
with means $\boldsymbol{\mu}_1$ and $\boldsymbol{\mu}_2$,
covariance matrices $\bSigma_1$ and $\bSigma_2$,
and corresponding eigenvalues
$\{\lambda_1^{(i)}\}_{i=1}^d$
and
$\{\lambda_2^{(i)}\}_{i=1}^d$,
is given by:

\begin{equation} \nonumber
D_{\text{KL}} \left(\mathcal{N}(\boldsymbol{\mu}_1, \bSigma_1) \parallel \mathcal{N}(\boldsymbol{\mu}_2, \bSigma_2) \right) = \frac{1}{2} \left( \log \frac{|\bSigma_2|}{|\bSigma_1|} - d + \operatorname{tr}\left( \bSigma_2^{-1} \bSigma_1 \right) + (\boldsymbol{\mu}_2 - \boldsymbol{\mu}_1)^T \bSigma_2^{-1} (\boldsymbol{\mu}_2 - \boldsymbol{\mu}_1) \right)
\end{equation}

In our setting, $$ \quad \mathbf{\hat{x}}_{0}^{\mathcal{F}} \,\big|\,  \y^{\mathcal{F}} \sim \N\!\!\left( \D_2 \y^{\mathcal{F}} + \D_3\bmu^{\mathcal{F}}  ~,~  \D_1^2  \right)  ~, \quad \mathbf{\hat{x}}_{0}^{\mathcal{F}}  \in \mathbb{R}^{d} \quad \text{and} \quad \x_0^{\mathcal{F}}|\y \sim \mathcal{N}( \boldsymbol{\mu}_{\mathbf{x}_0 \mid \mathbf{y}}^{\mathcal{F}},\boldsymbol{\Lambda}_{x_0|y}), \quad \x_0^{\mathcal{F}} \in \mathbb{R}^{d}\quad $$ 


Accordingly, the KL divergence between the true posterior distribution
and the estimated posterior distribution is

\begin{equation} \nonumber
 D_{\text{KL}}\left(  \x_0^{\mathcal{F}}|\y  \parallel \mathbf{\hat{x}}_{0}^{\mathcal{F}} \,\big|\,  \y^{\mathcal{F}}    \right) = D_{\text{KL}} \left(  \mathcal{N}( \boldsymbol{\mu}_{\mathbf{x}_0 \mid \mathbf{y}}^{\mathcal{F}},\boldsymbol{\Lambda}_{x_0|y}) , \, \mathcal{N}\!\!\left( \D_2 \y^{\mathcal{F}} + \D_3\bmu^{\mathcal{F}}  ~,~  \D_1^2  \right)  \right)   
\end{equation}

Denote by
$\left\{\lambda_0^{(i)}\right\}_{i=1}^{d}$
the eigenvalues of
$p(\x_0^{\mathcal{F}}|\y)$,
corresponding to the diagonal entries of
$\boldsymbol{\Lambda}_{x_0|y}$.


We now derive each term in the KL divergence separately.


\begin{itemize}
    \item The determinant of the estimated posterior covariance is $$|\bSigma_2| = |\D_1^T\D_1| = |\D_1^2| =  \prod_{i=1}^{d} {\D_1^{(i)}}^2 
    $$
   
    \item The determinant of the true posterior covariance is $$|\bSigma_1| = \prod_{i=1}^{d} \lambda_0^{(i)}$$

    \item The trace term becomes: $$\operatorname{tr}\left(\bSigma_2^{-1} \bSigma_1\right) = \sum_{i=1}^{d}\frac{\lambda_0^{(i)}}{    {\D_1^{(i)}}^2} $$

    \item The quadratic mean term is given by: $$ (\boldsymbol{\mu}_2 - \boldsymbol{\mu}_1)^T \bSigma_2^{-1} (\boldsymbol{\mu}_2 - \boldsymbol{\mu}_{\mathbf{x}_0 }) = (\D_2\y^\mathcal{F}+\D_3 \bmu^\mathcal{F} - \boldsymbol{\mu}_{\mathbf{x}_0 \mid \mathbf{y}}^{\mathcal{F}})^T(\D_1^2)^{-1}(\D_2\y^\mathcal{F}+\D_3 \bmu^\mathcal{F} - \boldsymbol{\mu}_{\mathbf{x}_0 \mid \mathbf{y}}^{\mathcal{F}})$$
    \\By substituting $\boldsymbol{\mu}_{\mathbf{x}^{\mathcal{F}} \mid \mathbf{y}}  = \bA \y^{\mathcal{F}} + \left(\I-\bA\boldsymbol{\Lambda}_H\right)\bmu^{\mathcal{F}}$ from Equation \eqref{eq:original_mu_posterior_freq} and assuming $\bmu=0$, we obtain: 
     $$ (\boldsymbol{\mu}_2 - \boldsymbol{\mu}_1)^T \bSigma_2^{-1} (\boldsymbol{\mu}_2 - \boldsymbol{\mu}_{1 }) = \left(\D_2\y^\mathcal{F}-\bA\y^\mathcal{F}\right)^T(\D_1^2)^{-1}\left(\D_2\y^\mathcal{F}-\bA\y^\mathcal{F}\right)$$
     $$=(\y^\mathcal{F})^T(\D_2^T-{\bA}^T)(\D_1^2)^{-1}(\D_2-\bA)(\y^\mathcal{F}).$$
Rearranging terms gives:
$$ (\boldsymbol{\mu}_2 - \boldsymbol{\mu}_1)^T \bSigma_2^{-1} (\boldsymbol{\mu}_2 - \boldsymbol{\mu}_{\mathbf{x}_0 }) = \sum_{i=1}^{d}\frac{\left(\D_2^{(i)}-\bA^{(i)}\right)^2}{{\D_1^{(i)}}^2}({\y^\mathcal{F}_{i})}^2.$$

\end{itemize}

Substituting these terms into the KL divergence gives:
\begin{equation} \nonumber
 D_{\text{KL}}\left(  \x_0^{\mathcal{F}}|\y  \parallel \mathbf{\hat{x}}_{0}^{\mathcal{F}} \,\big|\,  \y^{\mathcal{F}}    \right) = D_{\text{KL}} \left(  \mathcal{N}( \boldsymbol{\mu}_{\mathbf{x}_0 \mid \mathbf{y}}^{\mathcal{F}},\boldsymbol{\Lambda}_{x_0|y}) , \, \mathcal{N}\!\!\left( \D_2 \y^{\mathcal{F}} + \D_3\bmu^{\mathcal{F}}  ~,~  \D_1^2  \right)  \right)   
\end{equation}

\begin{equation}\nonumber
= \frac{1}{2} \left[ \sum_{i=1}^{d} \log{\D_1^{(i)}}^2 -  \sum_{i=1}^{d} \log{\lambda_0^{(i)}} \ - d +\sum_{i=1}^{d}\frac{ \lambda_0^{(i)}}{    {\D_1^{(i)}}^2 }  + \sum_{i=1}^{d}\frac{\left(\D_2^{(i)}-\bA^{(i)}\right)^2}{{\D_1^{(i)}}^2}({\y^\mathcal{F}_{i})}^2  \right]
\end{equation}

\begin{align}
& D_{\text{KL}}\left(  \x_0^{\mathcal{F}}|\y  \parallel \mathbf{\hat{x}}_{0}^{\mathcal{F}} \,\big|\,  \y^{\mathcal{F}}    \right) = \nonumber  \\ & \frac{1}{2} \left[ \sum_{i=1}^{d} 2\log{\D_1^{(i)}} -  \sum_{i=1}^{d} \log{\lambda_0^{(i)}} \ - d +  \sum_{i=1}^{d}\frac{ \lambda_0^{(i)} + \left(\D_2^{(i)}-\bA^{(i)}\right)^2({\y^\mathcal{F}_{i})}^2}{{\D_1^{(i)}}^2} \right]
\end{align}

\section{\texorpdfstring{$\Pi\text{GDM}$ Algorithm}{PiGDM algorithm}}
\label{sec:appendix_PIGDM_algorithm}
The $\Pi\text{GDM}$ method shares the same Bayesian principle as DPS, but differs in its approximation of the likelihood term. Specifically, $p(\y|\x_t)$ is modeled as a Gaussian distribution with covariance that varies across diffusion time steps $t$:

$$
p_t(\y | \x_t) \approx \mathcal{N}(\bH \hat{\x}_0, r_t^2 \bH \bH^T + \sigma_\y^2 \I)
$$

Thus, the gradient of the log-probability is given by:  

$$
\nabla_{\x_t} \log p_t(\y | \x_t) \approx \nabla_{\x_t} (\hat{\x}_0)^T \bH^T \left( (r_t^2 \bH \bH^T + \sigma_\y^2 \I)^{-1} \right)^T (\y - \bH \hat{\x}_0)
$$

\subsection{The reverse process in the time domain}

Based on the DDIM~\cite{song2020denoising} inference process and the Variance-Preserving (VP) framework, the inference procedure using $\Pi\text{GDM}$ is formulated as:
\begin{align}
\x_{s-1} &= \sqrt{\bar\alpha_{s-1}}\hat\x_0 
+ \sqrt{1 - \bar{\alpha}_{s-1} - \sigma_s^2(\eta)} \, \bepsilon_\theta(\x_s, s)  
\\&+ \nonumber \sqrt{\bar\alpha_s} \nabla_{\x_s}(\hat\x_0)^T \bH^T ((r_s^2 \bH \bH^T + \sigma_\y^2 \I)^{-1})^T (\y - \bH\hat\x_0)  
+ \sigma_s(\eta) \z_s.
\end{align}
where:
$$
\sigma_s(\eta) = \eta \sqrt{\frac{1 - \bar\alpha_{s-1}}{1 - \bar\alpha_s}} \sqrt{1 - \frac{\bar\alpha_s}{\bar\alpha_{s-1}}}.
$$
Since the VP formulation is already used for $\hat{\x}_0$, we omit the conversion factor $\sqrt{\bar\alpha_t}$ in the gradient term, leading to:
\begin{align}
\x_{s-1} &= \sqrt{\bar\alpha_{s-1}}\hat\x_0 
+ \sqrt{1 - \bar{\alpha}_{s-1} - \sigma_s^2(\eta)} \, \bepsilon_\theta(\x_s, s) 
\\&+ \nonumber \nabla_{\x_s}(\hat\x_0)^T \bH^T ((r_s^2 \bH \bH^T + \sigma_\y^2 \I)^{-1})^T (\y - \bH\hat\x_0)  
+ \sigma_s(\eta) \z_s.
\end{align}
By using the deterministic process ($\eta = 0$):
\begin{align}\label{eq:DDIM_PIGDM_eq}
\x_{s-1} &= \sqrt{\bar\alpha_{s-1}}\hat\x_0 
+ \sqrt{1 - \bar{\alpha}_{s-1}} \, \bepsilon_\theta(\x_s, s)  
\\&+ \nonumber \nabla_{\x_s}(\hat\x_0)^T \bH^T ((r_s^2 \bH \bH^T + \sigma_\y^2 \I)^{-1})^T (\y - \bH\hat\x_0)  
\end{align}

Using the marginal property,
\begin{equation}
\boldsymbol{\epsilon}_\theta(\x_s, s) = \frac{\x_s - \sqrt{\balpha_s}\hat{\x}0}{\sqrt{1-\balpha_s}} .
\end{equation}
which, substituted into Equation \ref{eq:DDIM_PIGDM_eq}, gives:
\begin{align}
\x_{s-1} &=  \nonumber \sqrt{\bar\alpha_{s-1}}\hat\x_0 + \sqrt{1-\bar{\alpha}_{s-1}}\left[ \frac{\x_s-\sqrt{{\balpha}_s}\hat{\x}_0}{\sqrt{1-\balpha_s}} \right] + \nabla_{\x_s}(\hat\x_0)^T\bH^T{((r_s^2\bH\bH^T + \sigma_\y^2\I)^{-1}})^T(\y-\bH\hat\x_0) 
\\&= \nonumber \frac{\sqrt{1 - \balpha_{s-1}}}{\sqrt{1-\balpha_s}}\x_s + \left[  \sqrt{\balpha_{s-1}} - \frac{\sqrt{{\balpha}_s}\sqrt{1 - \balpha_{s-1}}}{\sqrt{1 - \balpha_{s}}}\right]\hat{\x}_0 +\nabla_{\x_s}(\hat\x_0)^T\bH^T{((r_s^2\bH\bH^T + \sigma_\y^2\I)^{-1}})^T(\y-\bH\hat\x_0) 
\end{align}
Introducing:
\begin{equation}\nonumber
a_s = \left[\frac{\sqrt{1 - \balpha_{s-1}}}{\sqrt{1-\balpha_s}}\right]\I
\end{equation}
\begin{equation}\nonumber
b_s = \left[\sqrt{\balpha_{s-1}} - \frac{\sqrt{{\balpha}_s}\sqrt{1 - \balpha_{s-1}}}{\sqrt{1 - \balpha_{s}}}\right]\I
\end{equation}
we obtain:
\begin{empheq}[box=\fbox]{align}\label{eq:PIGDM_time_part_1}
\x_{s-1} =a_s\x_s + b_s\hat{\x}_0 +\nabla_{\x_s}(\hat\x_0)^T\bH^T{((r_s^2\bH\bH^T + \sigma_\y^2\I)^{-1}})^T(\y-\bH\hat\x_0) 
\end{empheq}

Where $\hat{\x}_0$ denotes the MMSE denoiser, corresponds to the denoised signal produced by a pretrained network. Assuming a Gaussian prior, the optimal solution $\x_0^*$ is given by \cite{benita2025spectral}:
$$\x_0^* = \left(\balpha_s \bSigmaZ + (1-\balpha_s) \I \right)^{-1} \left(\sqrt{\balpha_s} \bSigmaZ \x_s + (1-\balpha_s) \bmu \right)$$
Therefore we get:
$$\x_{s-1} = a_s\x_s + b_s\x_0^* +\nabla_{\x_s}(\x_0^*(\x_s))^T\bH^T{((r_s^2\bH\bH^T + \sigma_\y^2\I)^{-1}})^T(\y-\bH\x^*_0) $$

from Equation \ref{eq:gradient_optimal_denoiser}:
$${\nabla_{x_s}\x_0^*(\x_s) = \left(\balpha_s \bSigmaZ + (1-\balpha_s) \I \right)^{-1} \sqrt{\balpha_s} \bSigmaZ   }$$

\textcolor{black}{Using the property $(\mathbf{A}^{-1})^T = (\mathbf{A}^T)^{-1}$ for any invertible matrix $\mathbf{A}$, with 
$\mathbf{A} = \alpha_s \boldsymbol{\Sigma}_0 + (1-\alpha_s)\mathbf{I}$, it can be rewritten as:}
$$\nabla_{x_s}\x_0^*(\x_s)^T = \sqrt{\balpha_s} \bSigmaZ  \left(\balpha_s \bSigmaZ + (1-\balpha_s) \I \right)^{-1}   $$
Substitute it into Equation \ref{eq:PIGDM_time_part_1}, we get:
\begin{empheq}[box=\fbox]{align}\label{eq:pigdm_time_final}
\x_{s-1} = a_s\x_s + b_s\x_0^* +\sqrt{\balpha_s}  \bSigmaZ  \left(\balpha_t \bSigmaZ + (1-\balpha_s) \I \right)^{-1} \bH^T{(r_s^2\bH\bH^T + \sigma_\y^2\I)^{-1}}(\y-\bH\x^*_0)
\end{empheq}

\subsection{Spectral form of the update rule}

Next, we apply the Discrete Fourier Transform (DFT), denoted by $\mathcal{F}{\cdot}$, to Equation \eqref{eq:pigdm_time_final}. Assuming that $\boldsymbol{\Sigma}_0$ and $\mathbf{H}$ are shift-invariant, both matrices are diagonalized by the DFT.
$$\F\x_{s-1} = \F[a_s\x_s + b_s\x_0^* +\sqrt{\balpha_s}  \bSigmaZ  \left(\balpha_t \bSigmaZ + (1-\balpha_s) \I \right)^{-1} \bH^T{(r_s^2\bH\bH^T + \sigma_\y^2\I)^{-1}}(\y-\bH\x^*_0)] $$
\begin{align}\label{eq:freq_domain_pigdm_first}
\x^{\mathcal{F}}_{s-1} = a_s\x^{\mathcal{F}}_s + b_s\F\x_0^* +\sqrt{\balpha_s} \F[ \bSigmaZ  \left(\balpha_t \bSigmaZ + (1-\balpha_s) \I \right)^{-1} \bH^T{(r_s^2\bH\bH^T + \sigma_\y^2\I)^{-1}})(\y-\bH\x^*_0)] 
\end{align}

From Equation~\ref{eq:DFT_part_1_DPS_signed}, we obtain
\begin{equation}\label{eq:pigdm_freq_optimal_x_0}
\F\x_0^* = c_s \x^{\mathcal{F}}_s + d_s  \bmu^{\mathcal{F}} 
\end{equation}

Where:
\begin{equation}\nonumber
c_s = \sqrt{\balpha_s}\bLambda \left[\balpha_s \mathbf{\Lambda}_0 + (1-\balpha_s)\I\right]^{-1}
\end{equation}
\begin{equation}\nonumber
d_s = (1-\balpha_s) \left[\balpha_s \mathbf{\Lambda}_0 + (1-\balpha_s)\I\right]^{-1}
\end{equation}
\begin{align}
&=\sqrt{\balpha_s} \F[ \bSigmaZ  \left(\balpha_s \bSigmaZ + (1-\balpha_s) \I \right)^{-1} \bH^T{(r_s^2\bH\bH^T + \sigma_\y^2\I)^{-1}})(\y-\bH\x^*_0)]
\\&= \nonumber \sqrt{\balpha_s} [ \F\bSigmaZ \F^T\F \left(\balpha_s \bSigmaZ + (1-\balpha_s) \I \right)^{-1} \F^T\F  \bH^T\F^T\F {(r_s^2\bH\bH^T + \sigma_\y^2\I)^{-1}})\F^T\F (\y-\bH\x^*_0)]
\\&= \nonumber \sqrt{\balpha_s} [ \bLambda\left[\balpha_t \mathbf{\Lambda}_0 + (1-\balpha_s)\I\right]^{-1}\bLambdahcon\F {(r_s^2\bH\bH^T + \sigma_\y^2\I)^{-1}})\F^T (\F\y-\F\bH\F^T\F\x^*_0)]
\end{align}
Since:
$$
 \F {(r_s^2\bH\bH^T + \sigma_\y^2\I)^{-1}})\F^T = (r_s^2\bLambdahtranop +\sigma_\y^2\I )^{-1}$$
and denoting:
\begin{equation}\nonumber
e_s = (r_s^2\bLambdahtranop +\sigma_\y^2\I )^{-1} 
\end{equation}
\begin{equation}\label{eq:DFT_pigdm_part_2} \sqrt{\balpha_s} \F[ \bSigmaZ  \left(\balpha_s \bSigmaZ + (1-\balpha_s) \I \right)^{-1} \bH^T{(r_s^2\bH\bH^T + \sigma_\y^2\I)^{-1}})(\y-\bH\x^*_0)]= c_s \bLambdahcon e_s (\y^{\mathcal{F}}-\bLambdah\F\x^*_0)] \end{equation}

Substituting Equations \ref{eq:DFT_pigdm_part_2} and \ref{eq:pigdm_freq_optimal_x_0} in  \ref{eq:freq_domain_pigdm_first} yields:

$$\x^{\mathcal{F}}_{s-1} = a_s\x^{\mathcal{F}}_s + b_s(c_s \x^{\mathcal{F}}_s + d_s  \bmu^{\mathcal{F}}) +c_s \bLambdahcon e_s (\y^{\mathcal{F}}-\bLambdah(c_s\x^{\mathcal{F}}_s + d_s  \bmu^{\mathcal{F}})] $$

$$\x^{\mathcal{F}}_{s-1} = [a_s + b_s c_s - c_s \bLambdahcon e_s \bLambdah c_s]\x^{\mathcal{F}}_s + [b_s d_s  - c_s \bLambdahcon e_s \bLambdah d_s ]\bmu^{\mathcal{F}} +[c_s \bLambdahcon e_s ]\y^{\mathcal{F}} $$
We further define:
$$\G_{\Pi\text{GDM}}(s) = [a_s + b_s c_s - c_s \bLambdahcon e_s \bLambdah c_s] $$
$$\Q_{\Pi\text{GDM}}(s) = [ c_s \bLambdahcon e_s] $$
$$\M_{\Pi\text{GDM}}(s) = [ b_s d_s  - c_s \bLambdahcon e_s \bLambdah d_s ] $$
This leads to the following spectral-domain recursion:
\begin{align}\label{eq:spectral_domain_pigdm_one_step}
\x^{\mathcal{F}}_{s-1} = \G_{\Pi\text{GDM}}(s)\x^{\mathcal{F}}_s +\Q_{\Pi\text{GDM}}(s)\y^{\mathcal{F}} + \M_{\Pi\text{GDM}}(s)\bmu^{\mathcal{F}}
\end{align}

By recursively applying Equation~\eqref{eq:spectral_domain_pigdm_one_step} and following the same derivation steps and notation as in Appendix~\ref{sec:appendix_DPS_algorithm}, we arrive at the following closed-form distribution for $\mathbf{x}_{0,\Pi\text{GDM}}^{\mathcal{F}}$.

\begin{equation}
\mathbf{x}_{0,{\Pi\text{GDM}}}^{\mathcal{F}} \sim \N\!\!\left(  \D_2 \y^{\mathcal{F}} + \D_3\bmu^{\mathcal{F}} ~,~  \D_1^2  \right)
\end{equation}

\newpage
\section{DiffPIR Algorithm}
\label{sec:appendix_DiffPIR_algorithm}

In this section, we present the DiffPIR algorithm \cite{zhu2023denoising} from a spectral perspective and show how it can be integrated into our method. DiffPIR formulates posterior sampling through the following optimization problem:

$$\hat{\x} =\underset{\x}{\arg\min}\frac{1}{2\sigma^2_n}|| \y - \bH\x||^2+ \lambda \mathcal{P}(\x)$$

where the measurements follow the degradation model $\y=\bH\x+\nn$, $\bH$ is a known linear degradation operator, $\mathbf{n} \sim \mathcal{N}(\mathbf{0}, \sigma_y^2 \mathbf{I})$ denotes additive Gaussian noise, and $\mathcal{\lambda}\mathcal{P}(\x)$ represents the prior regularization term weighted by the hyperparameter  $\mathcal{\lambda}$.

The main idea behind plug-and-play inverse reconstruction methods is to decouple the data-consistency and prior terms using the Half-Quadratic Splitting (HQS) framework \cite{geman1995nonlinear}:

\begin{subequations}\label{eq:solving_measurements}
\begin{align}
\mathbf{x}_0^{(t)} 
&= \arg\min_{\mathbf{z}} 
\frac{1}{2\bar{\sigma}_t^2} \|\mathbf{z} - \mathbf{x}_t\|^2 
+  \mathcal{P}(\mathbf{z})
\label{eq:prior_eq} \\[6pt]
\hat{\mathbf{x}}_0^{(t)} 
&= \arg\min_{\mathbf{x}} 
\|\mathbf{y} - \mathbf{H}\mathbf{x}\|^2 
+ \rho_t \|\mathbf{x} - \mathbf{x}_0^{(t)}\|^2
\label{eq:measurement_eq}
\end{align}
\end{subequations}

 Equation\eqref{eq:prior_eq} promotes consistency with the prior distribution, while Equation \eqref{eq:measurement_eq} enforces adherence to the measurements. The weighting parameter $\rho_s $ is defined as $\rho_s \triangleq \lambda \sigma_n^2 / \bar{\sigma}_t^2$, where $\bar{\sigma}_t=\sqrt{\frac{1-\bar{a}_t}{\bar{\alpha}_t}}$ depends on the diffusion noise schedule $\{\bar{\alpha}\}_{t=1}^{T}$ and determines the effective noise level.

Since the prior term $\mathcal{P}(\x)$  is generally unknown, DiffPIR employs the denoising process to approximate samples from the prior distribution through $\mathbf{x}_0^{(t)}$. In addition, the optimization problem in Equation~\eqref{eq:measurement_eq} generally admits the following closed-form solution:

\begin{equation}\label{eq:solving_measurments}
\hat{\mathbf{x}}_0^{(t)} 
= \left(\mathbf{H}^\top \mathbf{H} + \rho_t \mathbf{I} \right)^{-1}
\left(\mathbf{H}^\top \mathbf{y} + \rho_t \mathbf{x}_0^{(t)} \right)
\end{equation}

\subsection{The reverse process in the time domain}
\label{sec:diffpir_time_domain}

 We now analyze the DiffPIR reverse process under the Gaussian prior assumption $\x_0\sim\mathcal{N}(\bmu,\bSigma_0)$. We begin from the sampling update introduced in \cite{zhu2023denoising}:

$$\mathbf{x}_{t-1} = \sqrt{\bar{\alpha}_{t-1}} \hat{\mathbf{x}}_0^{(t)} + \sqrt{1 - \bar{\alpha}_{t-1}} \left( \sqrt{1 - \zeta}\,\boldsymbol{\epsilon}_\theta(\x_t, t) + \sqrt{\zeta}\,\boldsymbol{\epsilon}_t \right)$$

where: $\boldsymbol{\epsilon}_\theta(\x_t, t)$ denotes the effective predicted noise and is given by:

\begin{equation}
\boldsymbol{\epsilon}_\theta(\x_t, t) = \frac{\x_t - \sqrt{\balpha_t}{\hat{\x}^{(t)}_0}}{\sqrt{1-\balpha_t}} .
\end{equation}\label{eq:predicted_noise}
Here,  $\boldsymbol{\epsilon}_t \sim \mathcal{N}(\mathbf{0}, \mathbf{I})$ and $\zeta$ controls the amount of stochastic noise injected at each diffusion step.

To enable comparison with other posterior sampling methods, we focus on the deterministic setting $\zeta=0$, corresponding to DDIM sampling \cite{song2020denoising}. Accordingly, we use the indexing $\{\dot\}_{s=0}^S $ to denote the accelerated diffusion trajectory.

\begin{equation}\label{eq:reverse_update}\mathbf{x}_{s-1} = \sqrt{\bar{\alpha}_{s-1}} \hat{\mathbf{x}}_0^{(s)} + \sqrt{1 - \bar{\alpha}_{s-1}}  \,\boldsymbol{\epsilon}_\theta(\x_s, s)   \end{equation}

Substituting the predicted noise expression from Equation \eqref{eq:predicted_noise} into the reverse update in \eqref{eq:reverse_update} yields:

$$\mathbf{x}_{s-1} = \sqrt{\bar{\alpha}_{s-1}} {\hat{\x}^{(s)}_0} + \sqrt{1 - \bar{\alpha}_{s-1}}  \left[\frac{\x_s - \sqrt{\balpha_s}{\hat{\x}^{(s)}_0}}{\sqrt{1-\balpha_s}} \right]  $$

which can be rewritten as:

\begin{equation}\label{eq:DiffPIR_time_synthesis} \x_{s-1} = \frac{\sqrt{1 - \balpha_{s-1}}}{\sqrt{1-\balpha_s}}\x_s + \left[  \sqrt{\balpha_{s-1}} - \frac{\sqrt{{\balpha}_s}\sqrt{1 - \balpha_{s-1}}}{\sqrt{1 - \balpha_{s}}}\right]\hat{\x}_0 \end{equation}

We define: 
\begin{equation}\nonumber
a_s = \left[\frac{\sqrt{1 - \balpha_{s-1}}}{\sqrt{1-\balpha_s}}\right]\I
\end{equation}
\begin{equation}\nonumber
b_s = \left[\sqrt{\balpha_{s-1}} - \frac{\sqrt{{\balpha}_s}\sqrt{1 - \balpha_{s-1}}}{\sqrt{1 - \balpha_{s}}}\right]\I.
\end{equation}
Substituting these definitions into Equation~\eqref{eq:DiffPIR_time_synthesis} gives
$$ \x_{s-1} = a_s\x_s +b_s{\hat{\x}^{(s)}_0}. $$

At this stage, DiffPIR incorporates measurement information into the diffusion process through the solution of the optimization problem in Equation
\eqref{eq:measurement_eq}. Substituting the closed-form solution for ${\hat{\x}^{(s)}_0}$ yields:

$$ \x_{s-1} = a_s\x_s +b_s \left(\bH^\top\bH + \rho_t \mathbf{I} \right)^{-1}
\left(\bH^\top \y + \rho_t \x_0^{(t)} \right) $$

In DiffPIR, the estimate $\x_0^{(t)}$ is obtained using Tweedie’s formula adapted to variance-preserving diffusion:

$$\mathbf{x}_0^{(t)} = \frac{1}{\sqrt{\bar{\alpha}_t}} \left( \mathbf{x}_t + (1 - \bar{\alpha}_t) s_\theta(\mathbf{x}_t, t) \right)$$ 

  Here, we replace the pretrained denoiser with the prior-optimal denoiser under a Gaussian prior assumption. In this setting, the MMSE estimator coincides with the Wiener filter and is given by \cite{benita2025spectral}:

$$\x_0^{(t)} = \x_0^* = \left(\balpha_s \bSigmaZ + (1-\balpha_s) \I \right)^{-1} \left(\sqrt{\balpha_s} \bSigmaZ \x_s + (1-\balpha_s) \bmu \right)$$

Substituting this expression into the reverse process yields:
\begin{empheq}[box=\fbox]{align}\label{eq:DiffPIR_time_final}
 \x_{s-1} = a_s\x_s +b_s \left(\bH^\top\bH + \rho_t \mathbf{I} \right)^{-1}
\left(\bH^\top \y + \rho_s \x_0^* \right)
\end{empheq}

\subsection{Spectral form of the update rule}

Next, we apply the Discrete Fourier Transform (DFT), denoted by  $\mathcal{F}{\cdot}$, to Equation \eqref{eq:DiffPIR_time_final}.  Assuming that $\boldsymbol{\Sigma}_0$ and $\mathbf{H}$ are shift-invariant, both matrices are diagonalized by the DFT.

$$ \F\x_{s-1} = \F[a_s\x_s +b_s \left(\bH^\top\bH + \rho_t \mathbf{I} \right)^{-1}
\left(\bH^\top \y + \rho_s \x_0^* \right)] $$

\begin{equation}\label{eq:DiffPIR_freq_first}
\x^{\mathcal{F}}_{s-1}   = a_s\x^{\mathcal{F}}_s +b_s \F\left(\bH^\top\bH + \rho_t \mathbf{I} \right)^{-1}\F^T\F
\left(\bH^\top \y + \rho_s \x_0^* \right)] 
\end{equation}

Using the diagonalization property:
$$
\F\left(\bH^\top\bH + \rho_t \mathbf{I} \right)^{-1}\F^T = (\bLambdahtran +\rho_s\I)^{-1} $$

and denoting:
\begin{equation}\nonumber
m_s = (\bLambdahtran +\rho_s\I)^{-1}
\end{equation}

together with Equation~\ref{eq:DFT_part_1_DPS_signed}, we obtain:

\begin{equation}
\F\x_0^* = c_s \x^{\mathcal{F}}_s + d_s  \bmu^{\mathcal{F}} 
\end{equation}

where
\begin{equation}\nonumber
c_s = \sqrt{\balpha_s}\bLambda \left[\balpha_s \mathbf{\Lambda}_0 + (1-\balpha_s)\I\right]^{-1}
\end{equation}
and
\begin{equation}\nonumber
d_s = (1-\balpha_s) \left[\balpha_s \mathbf{\Lambda}_0 + (1-\balpha_s)\I\right]^{-1}.
\end{equation}

Substituting these expressions into Equation Equestion\eqref{eq:DiffPIR_freq_first} yields:

$$a_s\x^{\mathcal{F}}_s  + b_s m_s \bLambdah \y^{\mathcal{F}} + b_s m_s \rho_s c_s \x^{\mathcal{F}}_s + b_s m_s \rho_s d_s  \bmu^{\mathcal{F}} $$
which can be rearranged as:
$$(a_s+b_s m_s \rho_s c_s)\x^{\mathcal{F}}_s  + b_s m_s \bLambdahcon \y^{\mathcal{F}} + b_s m_s \rho_s d_s  \bmu^{\mathcal{F}} $$
We further define

$$\G_{\text{DiffPIR}}(s) = [a_s+b_s m_s \rho_s c_s] $$
$$\Q_{\text{DiffPIR}}(s) = [ b_s m_s \bLambdahcon] $$
and
$$\M_{\text{DiffPIR}}(s) = [b_s m_s \rho_s d_s   ] $$
This leads to the following spectral-domain recursion:
\begin{align}\label{eq:spectral_domain_Diffpir_one_step}
\x^{\mathcal{F}}_{s-1} = \G_{\text{DiffPIR}}(s)\x^{\mathcal{F}}_s +\Q_{\text{DiffPIR}}(s)\y^{\mathcal{F}} + \M_{\text{DiffPIR}}(s)\bmu^{\mathcal{F}}
\end{align}
Recursively applying Equation~\eqref{eq:spectral_domain_Diffpir_one_step}, together with the derivation steps used in Appendix~\ref{sec:appendix_DPS_algorithm}, yields the following closed-form distribution for $\mathbf{x}_{0,\text{DiffPIR}}^{\mathcal{F}}$:

\begin{empheq}[box=\fbox]{align}\label{eq:DiffPIR_freq_final}
\mathbf{x}_{0,\text{DiffPIR}}^{\mathcal{F}} \sim \N\!\!\left(  \D_2 \y^{\mathcal{F}} + \D_3\bmu^{\mathcal{F}} ~,~  \D_1^2  \right)
\end{empheq}

\newpage
\section{Additional Results: Synthetic Gaussian Distribution}
\label{sec:Additional_results}

\begin{figure}[h]
    \centering
    \includegraphics[width=0.3\textwidth]{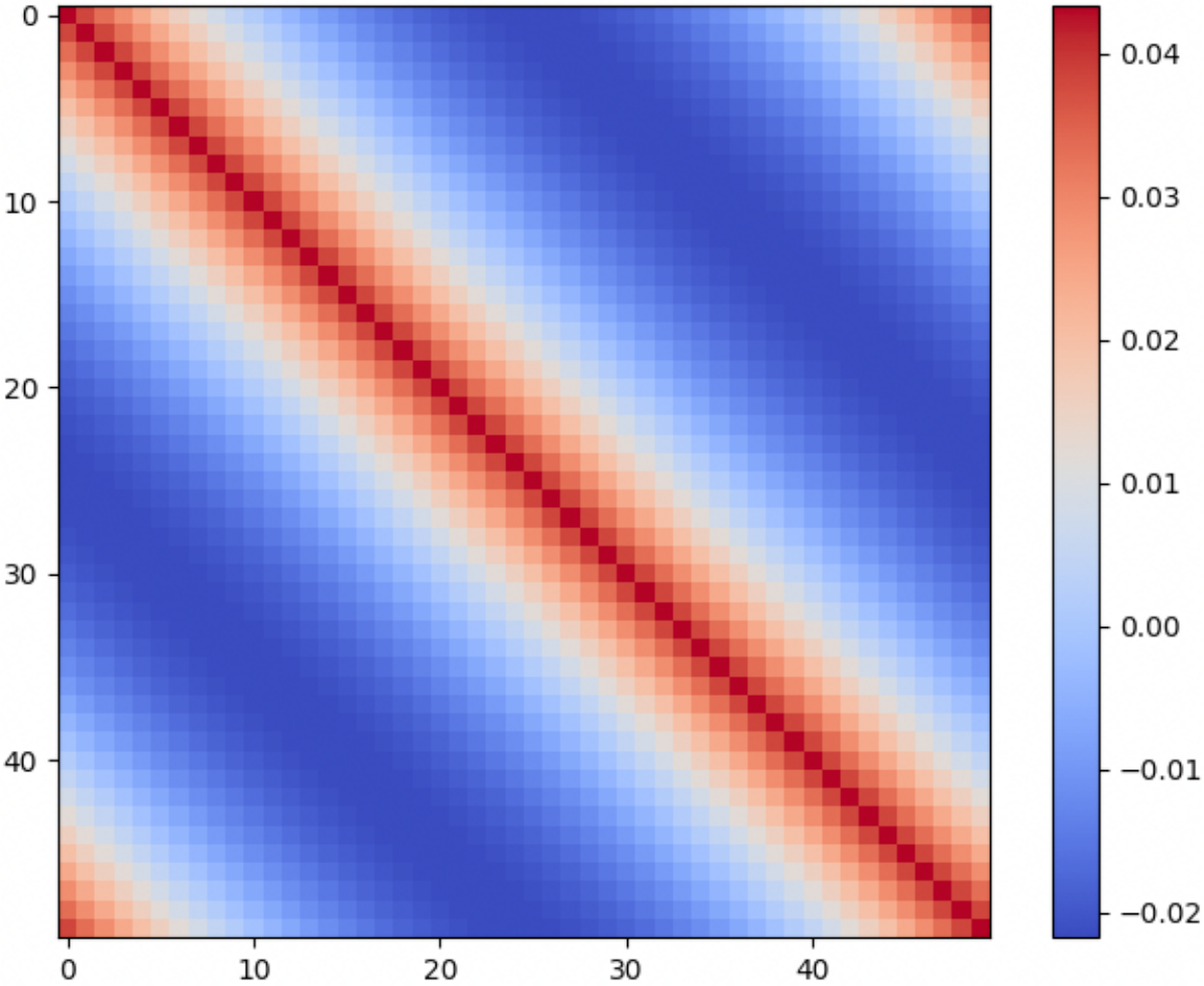}
    \caption{Covariance matrix obtained for $d=50$ and $l=0.05$.}
    \label{fig:covariance_d50_l001}
\end{figure}

Section~\ref{sec:Synthetic_Gaussian_distribution} considers a Gaussian prior $\x_0 \sim \mathcal{N}(\bmu, \bSigma_0)$, where $\x_0 \in \mathbb{R}^d$ and $\bSigma_0 \in \mathbb{R}^{d \times d}$ is a shift invariant covariance matrix. The covariance is constructed as $\bSigma_0 = A^\top A$, where $A$ is a shift invariant matrix whose first row is given by $a = [-l, -l + \tfrac{1}{d-1}, \dots, l - \tfrac{1}{d-1}, l]$. The mean vector $\bmu$ is chosen to be constant, corresponding to a stationary model. Figure~\ref{fig:covariance_d50_l001} illustrates the resulting covariance matrix for $d=50$ and $l=0.05$.

\subsection{Optimization procedure}
\label{sec:optimization_details}
Throughout the paper, the optimization problem is solved using the Sequential Least Squares Programming (SLSQP) method~\cite{kraft1988software}, as implemented in SciPy.

\subsection{DPS algorithm}
\label{sec:Additional_results_DPS}
Using the optimization problem in Equation~\eqref{eq:optimization_problem_specific_realization} under the DPS framework, together with the experimental setting described in Section~\ref{sec:Synthetic_Gaussian_distribution}, we obtain the following spectral recommendation for the hyperparameter vector $\boldsymbol{\zeta}$:
\begin{figure}[h]
    \centering
    \includegraphics[width=0.6\textwidth]{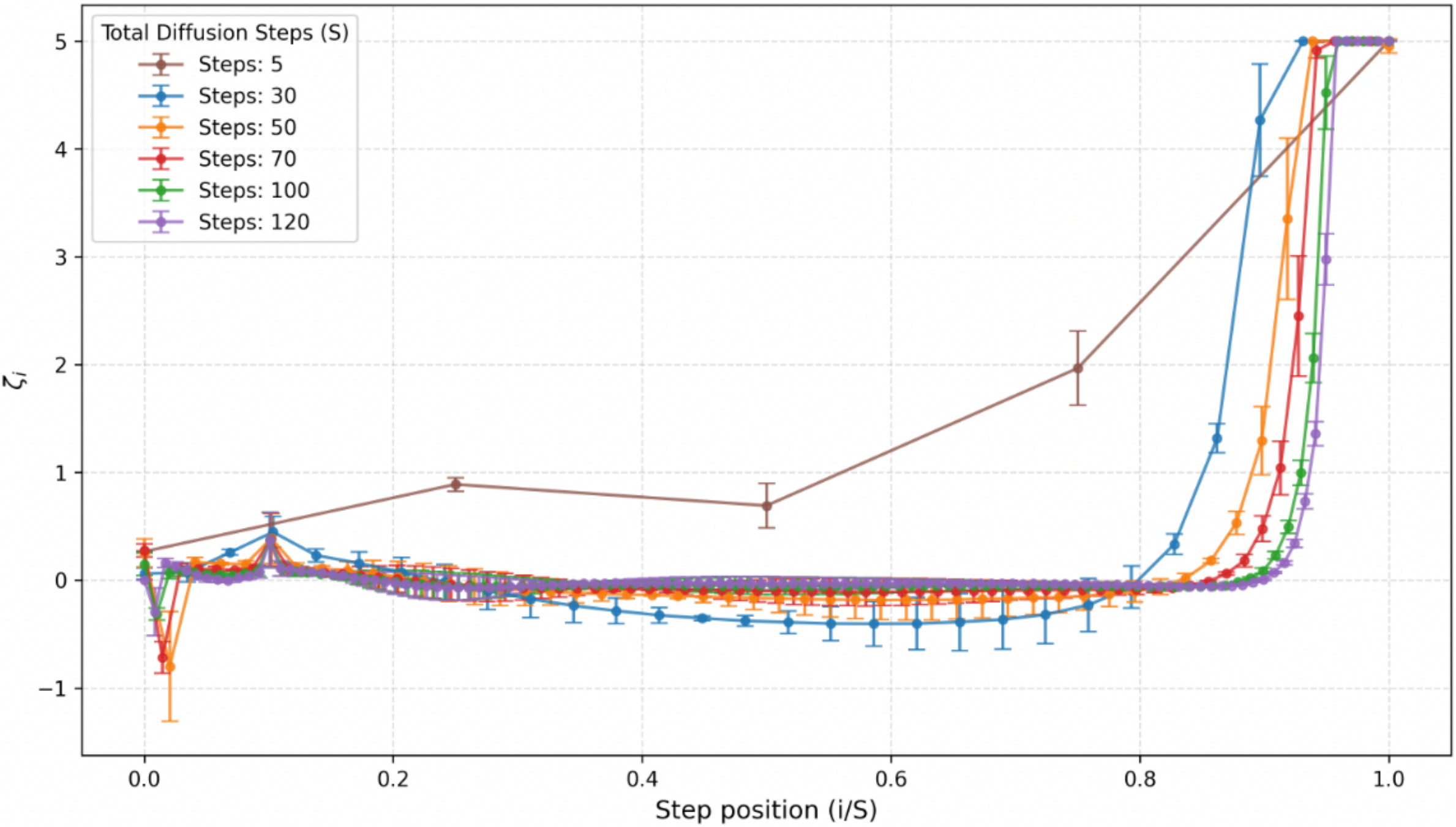}
\caption{Spectral recommendations for the weighting coefficients $\boldsymbol{\zeta}$ in the DPS framework for different numbers of diffusion steps $S\in[5, 30, 50, 70, 100, 120]$, with variability across realizations.}
    \vspace{-0.35cm}
    \label{fig:Comparison_DPS_SLSQP_differnt_steps}
\end{figure}

\subsection{\texorpdfstring{$\Pi\text{GDM}$ algorithm}{PiGDM algorithm}}
\label{sec:Aditional_results_pigdm}

Here, we derive the spectral recommendations for the $\Pi\text{GDM}$ algorithm under the setup introduced in Section~\ref{sec:Synthetic_Gaussian_distribution}. In the original formulation of $\Pi\text{GDM}$, the heuristic parameter $r_s$ plays a dual role: it both controls the uncertainty at diffusion step $s$ and serves as a weighting term for the likelihood component in the inference equation (see Eq.~\ref{eq:PIGDM_inference_equation}). These two roles are handled through a single heuristic definition based on the diffusion noise schedule,

$$r_s = \sqrt{{1-\balpha_s}}.$$ 

In our approach, we consider these roles separately, allowing for greater flexibility in balancing their respective influences. Specifically, we introduce a separate guidance parameter $g_s$ while retaining $r_s^2$ as the uncertainty-related term. 
The resulting inference update is given by:

\begin{align}\label{eq:PIGDM_inference_equation_ours}
\x_{s-1} &= \sqrt{\bar\alpha_{s-1}}\hat\x_0 
+ \sqrt{1 - \bar{\alpha}_{s-1}} \, \bepsilon_\theta(\x_s, s)  
\\&+ g_s\nonumber \nabla_{\x_s}(\hat\x_0)^T \bH^T ((r_s^2 \bH \bH^T + \sigma_\y^2 \I)^{-1})^T (\y - \bH\hat\x_0)  
\end{align}

\begin{figure}[h]
    \centering
    \begin{subfigure}[t]{0.48\textwidth}
        \centering
        \includegraphics[width=\textwidth]{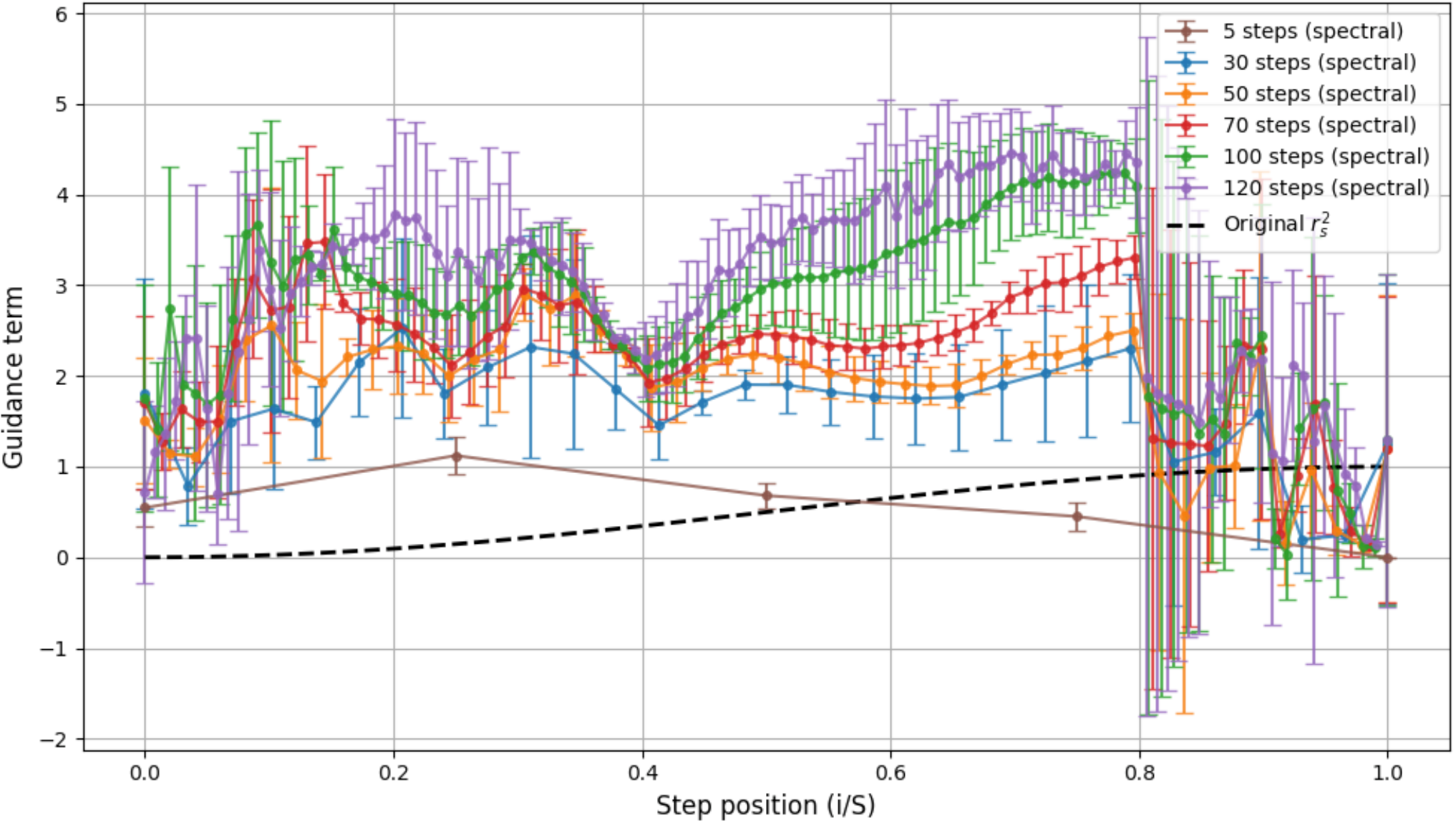}
        \caption{Guidance term $g_s$}
        \label{fig:Comparison_Weights_Guidance_PIGDM_spectral_many}
    \end{subfigure}
    \hfill
    \begin{subfigure}[t]{0.48\textwidth}
        \centering
        \includegraphics[width=\textwidth]{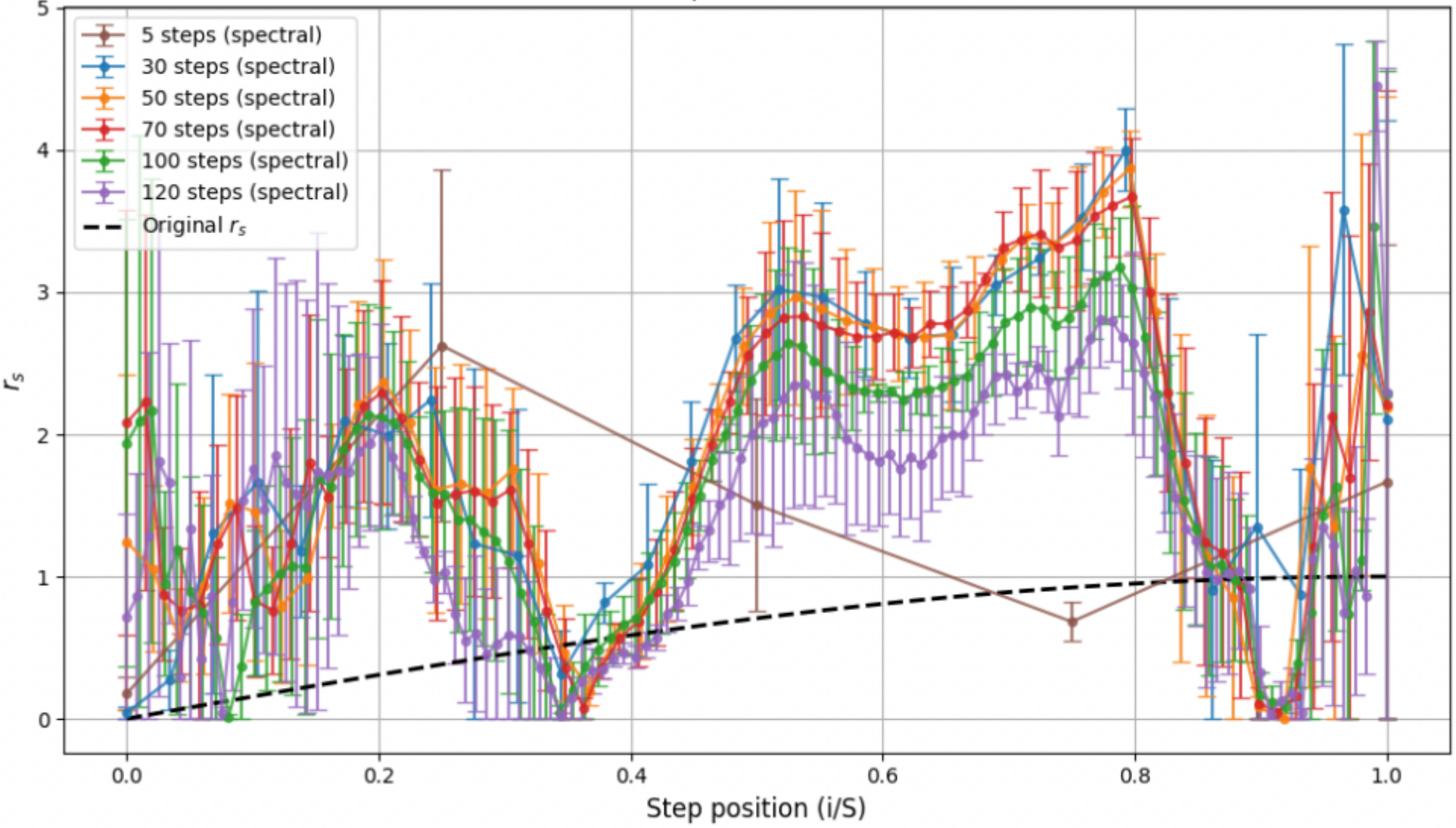}
        \caption{Uncertainty term $r_s$}
        \label{fig:Comparison_Weights_r_s_PIGDM_spectral_many}
    \end{subfigure}
    \caption{Figures~\ref{fig:Comparison_Weights_Guidance_PIGDM_spectral_many} and~\ref{fig:Comparison_Weights_r_s_PIGDM_spectral_many} show the spectral recommendations for the $\Pi$GDM weighting coefficients $g_s$ and $r_s$, respectively. Results are shown for different diffusion step counts, with the standard deviation across realizations. The original $\Pi$GDM heuristics indicated in black curves}

    \vspace{-0.35cm}
    \label{fig:Comparison_Weights_PIGDM_spectral_many}
\end{figure}

Figure~\ref{fig:Comparison_Weights_PIGDM_spectral_many} presents the resulting spectral recommendations for the weighting coefficients ${\{r_s\}}_{s=1}^d$ and ${\{g_s\}}_{s=1}^d$, together with their mean and standard deviation computed over $N=5$ realizations $\{y_i\}_{i=1}^N$, for different numbers of diffusion steps, as described in Section~\ref{sec:Synthetic_Gaussian_distribution}. For comparison, we also include the heuristic parameter values used in the original $\Pi\text{GDM}$ algorithm.

Notably, our recommendations take into account prior information, the observed measurements, and the diffusion dynamics, while the original $\Pi\text{GDM}$ heuristics define the weighting coefficients solely based on the diffusion noise schedule and decrease over diffusion steps.

\begin{figure}[h]
    \centering
    \includegraphics[width=0.5\textwidth]{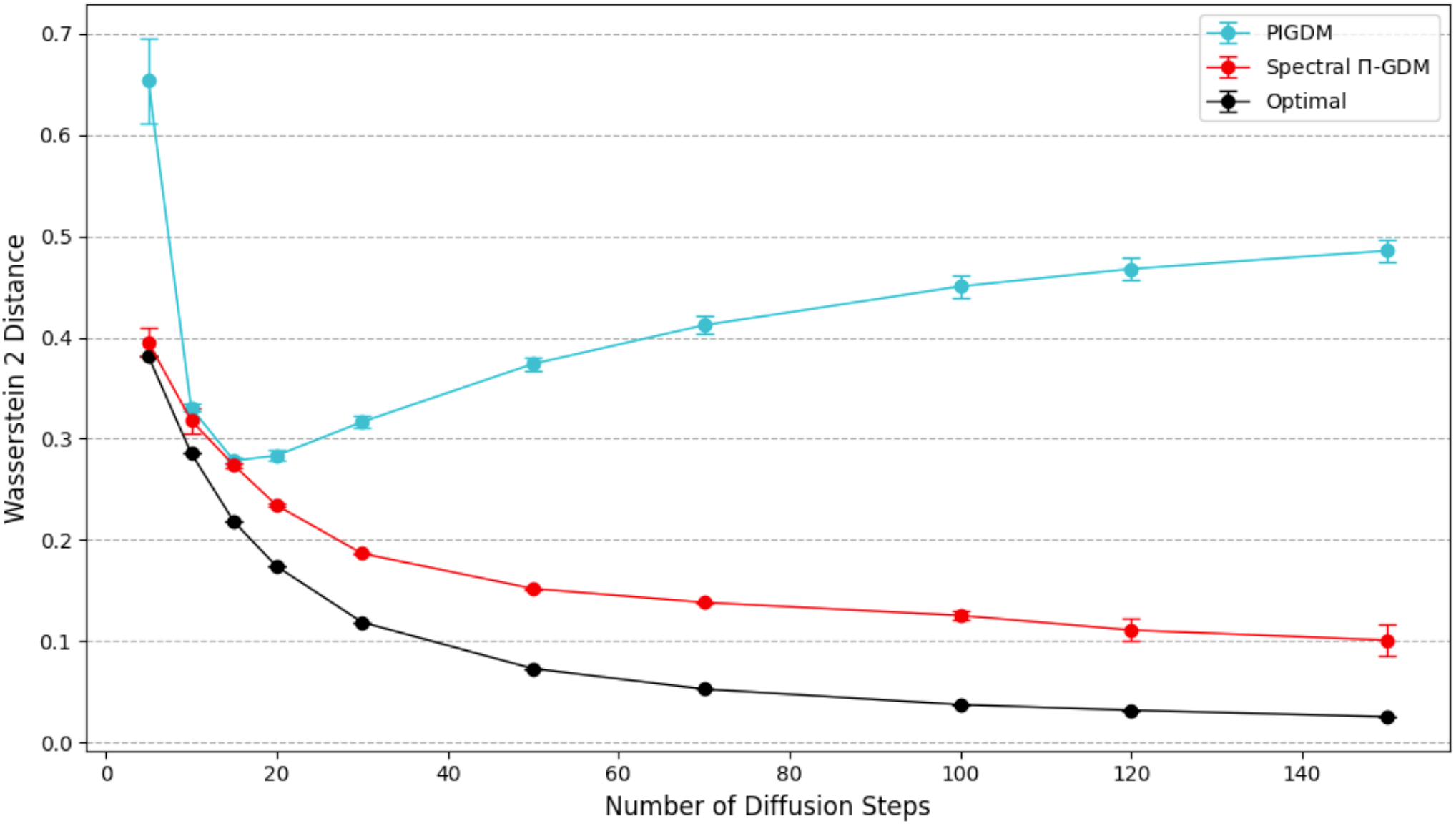}
    \caption{Comparison of the Wasserstein-2 distance for $\Pi\text{GDM}$ heuristic values, the spectral recommendations applied to $\Pi\text{GDM}$ (red), and the analytically derived ideal posterior denoiser (black), across different number of diffusion steps $S \in \{5, 10, 15, 20, 30, 50, 70, 100, 120, 150\}$.}
    \vspace{-0.35cm}
    \label{fig:Comparison_wasserstein_PIGDM_spectral_optimal}
\end{figure}

Figure~\ref{fig:Comparison_wasserstein_PIGDM_spectral_optimal} compares the \emph{Wasserstein-2} distance between the reconstructed posterior distributions and the true posterior, for both the heuristic parameter choices in $\Pi\text{GDM}$ and the proposed spectral recommendations, evaluated across multiple diffusion steps and measurement realizations ${y_i}$. We also report the distance obtained from the reconstructed posterior produced by the optimal posterior denoiser derived in \eqref{eq:posterior_optimal_denoiser}. The spectral recommendations yield consistently lower Wasserstein distances across diffusion steps and show reduced variability across realizations compared to the $\Pi\text{GDM}$ heuristics. In addition the exact solution achieves the lowest Wasserstein distance overall.

\subsection{DiffPIR algorithm}
\label{sec:appendix_DiffPIR_algorithm_spectral_reccomendations}

Here, we derive the spectral recommendations for the DiffPIR~\cite{zhu2023denoising} algorithm under the setup introduced in Section~\ref{sec:Synthetic_Gaussian_distribution}. As discussed in Section~\ref{sec:appendix_DiffPIR_algorithm}, DiffPIR introduces a single hyperparameter $\lambda$, which controls the balance between the prior and likelihood terms. The remaining weighting parameters are then determined through $\lambda$ together with the predefined noise level $\bar{\sigma}_t=\sqrt{\frac{1-\bar{\alpha}_t}{\bar{\alpha}_t}}$. In practice, the choice of $\lambda$ remains heuristic, with no principled rule currently existing for selecting it.


In our approach, we extend the scalar hyperparameter $\lambda$ to a diffusion-step-dependent vector,$\{{\lambda_s}\}_{s=1}^S$, enabling adaptive weighting across diffusion steps according to the predefined noise schedule.

\begin{figure}[h]
    \centering
    \includegraphics[width=0.6\textwidth]{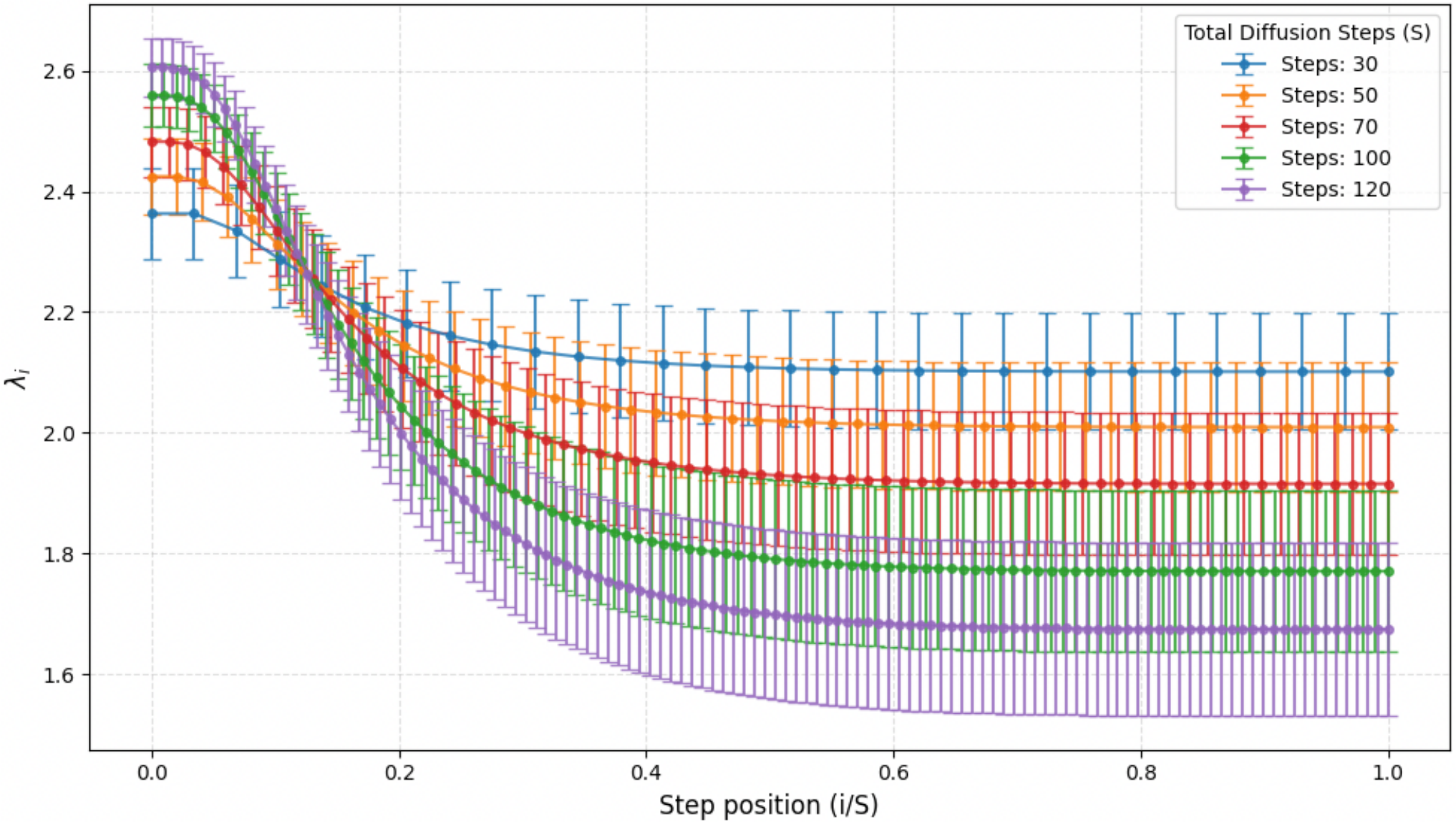}
\caption{Spectral recommendations for the DiffPIR weighting coefficients $\{\lambda_s\}_{s=1}^S$ across different numbers of diffusion steps. The curves include the corresponding standard deviation across realizations.}
    \vspace{-0.35cm}
    \label{fig:spectral_reccomendations_DiffPIR}
\end{figure}

Figure~\ref{fig:spectral_reccomendations_DiffPIR} presents the resulting spectral recommendations for the weighting coefficients $\{\lambda_s\}_{s=1}^S$, together with their mean and standard deviation computed over $N=5$ realizations $\{y_i\}_{i=1}^N$, for different numbers of diffusion steps under the setup described in Section~\ref{sec:Synthetic_Gaussian_distribution}. We note that the original DiffPIR heuristic is not included, since unlike DPS and $\Pi$GDM, DiffPIR relies on a manually predefined scalar parameter that does not explicitly depend on the diffusion schedule or the measurement consistency term. Moreover, the experimental setting considered here was not studied in the original DiffPIR paper, and therefore no corresponding heuristic recommendation is available. Notably, the proposed recommendations jointly account for the prior characteristics, the observed measurements, and the diffusion dynamics, unlike the original DiffPIR heuristics, which are manually predefined.

\begin{figure}[h]
    \centering
    \includegraphics[width=0.5\textwidth]{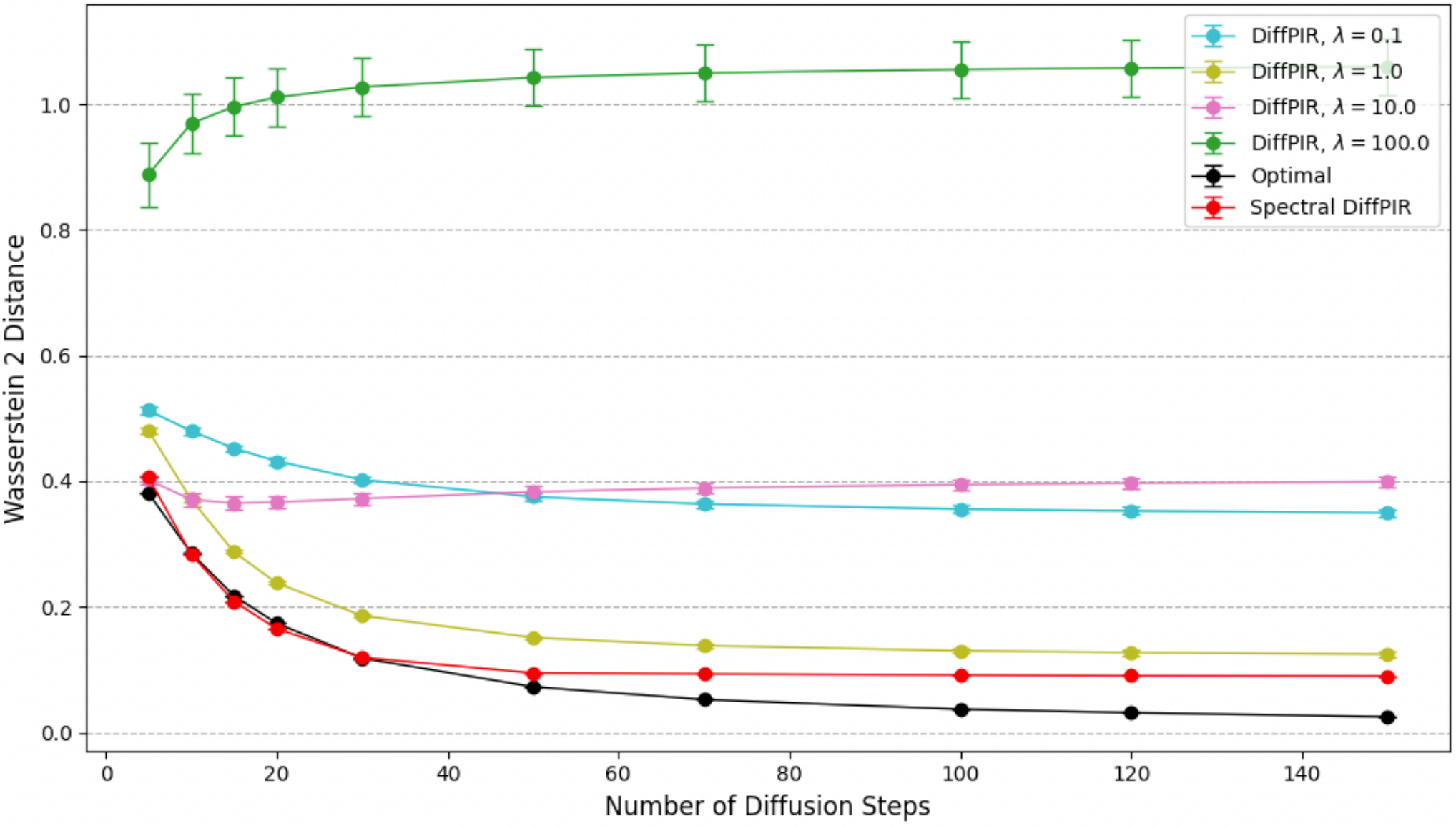}
\caption{Comparison of the Wasserstein-2 distance for the spectral recommendations applied to DiffPIR (red) and the analytically derived ideal posterior denoiser (black), across different numbers of diffusion steps $S \in \{5, 10, 15, 20, 30, 50, 70, 100, 120, 150\}.$}
    \vspace{-0.35cm}
    \label{fig:Comparison_wasserstein_DiffPIR_spectral_optimal}
\end{figure}

Figure~\ref{fig:Comparison_wasserstein_DiffPIR_spectral_optimal} compares the \emph{Wasserstein-2} distance between the reconstructed posterior distributions and the true posterior for the DiffPIR heuristic parameters with $\lambda=\{0.1,1,10,100\}$ and the proposed spectral recommendations, evaluated across multiple diffusion steps and $N$ measurement realizations $\{y_i\}_{i=1}^N$. We also report the distance obtained using the posterior-optimal denoiser derived in \eqref{eq:posterior_optimal_denoiser}. The proposed spectral recommendations consistently yield lower Wasserstein distances across diffusion steps and exhibit reduced variability across realizations compared to the original DiffPIR heuristics.

For a small number of diffusion steps, the spectral recommendations slightly outperform the posterior-optimal denoiser in terms of the Wasserstein distance, whereas for larger numbers of diffusion steps the opposite behavior is observed, with a growing gap in favor of the posterior-optimal denoiser. This behavior does not contradict the optimality of the posterior denoiser, since it is derived to minimize the MMSE at each diffusion step rather than directly minimizing the Wasserstein distance between the reconstructed and true posterior distributions.

\newpage
\section{Prior Sampling Results}
\label{sec:Prior_comparison_results}

\subsection{Comprehensive comparison}
\label{sec:prior_comprehensive_comparison}

In the following section, we provide a comprehensive comparison between the proposed spectral recommendations and the heuristic weighting strategies used in DPS~\cite{chung2022diffusion} and  DiffPIR~\cite{zhu2023denoising} under the same experimental setting described in Section~\ref{sec:Empirical_Distribution}. For DPS, we evaluate the heuristic values $\zeta^{'}=\{0.01,0.05,0.1,0.3,0.5,0.7\}$. For DiffPIR, we use the hyperparameter configuration reported in the original paper ($\lambda=7$), where \emph{DiffPIR (D)} denotes the deterministic variant and \emph{DiffPIR (S)} denotes the stochastic variant. The results are presented in terms of FID, LPIPS, SSIM and PSNR.


\begin{figure}[h]
    \centering

    \begin{subfigure}[b]{0.48\textwidth}
        \centering
        \includegraphics[width=\textwidth]{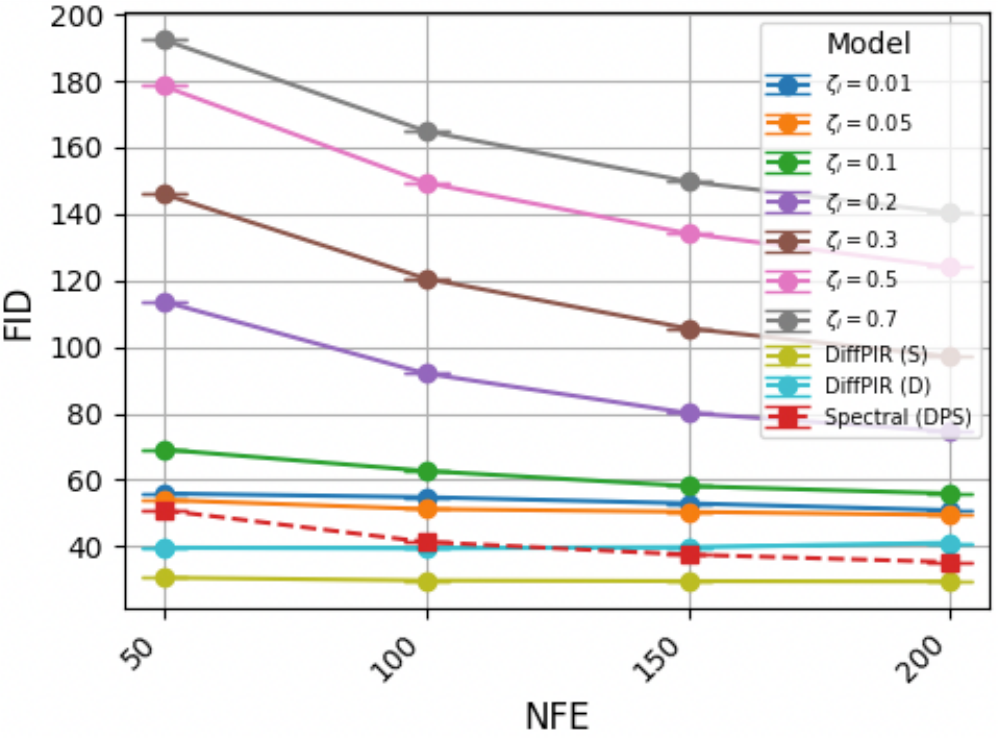}
        \caption{FID $\downarrow$}
        \label{fig:LPF_FFHQ_with_heuristics_fid}
    \end{subfigure}
    \hfill
    \begin{subfigure}[b]{0.48\textwidth}
        \centering
        \includegraphics[width=\textwidth]{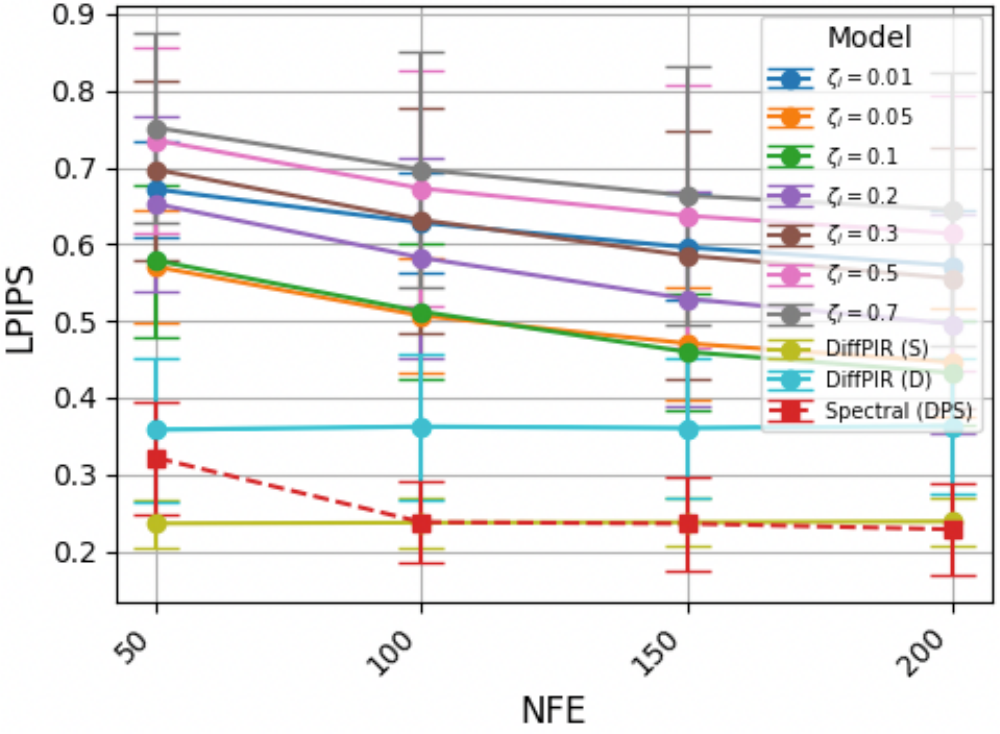}
        \caption{LPIPS $\downarrow$}
        \label{fig:LPF_FFHQ_with_heuristics_lpips}
    \end{subfigure}

    \vspace{0.5em}

    \begin{subfigure}[b]{0.48\textwidth}
        \centering
        \includegraphics[width=\textwidth]{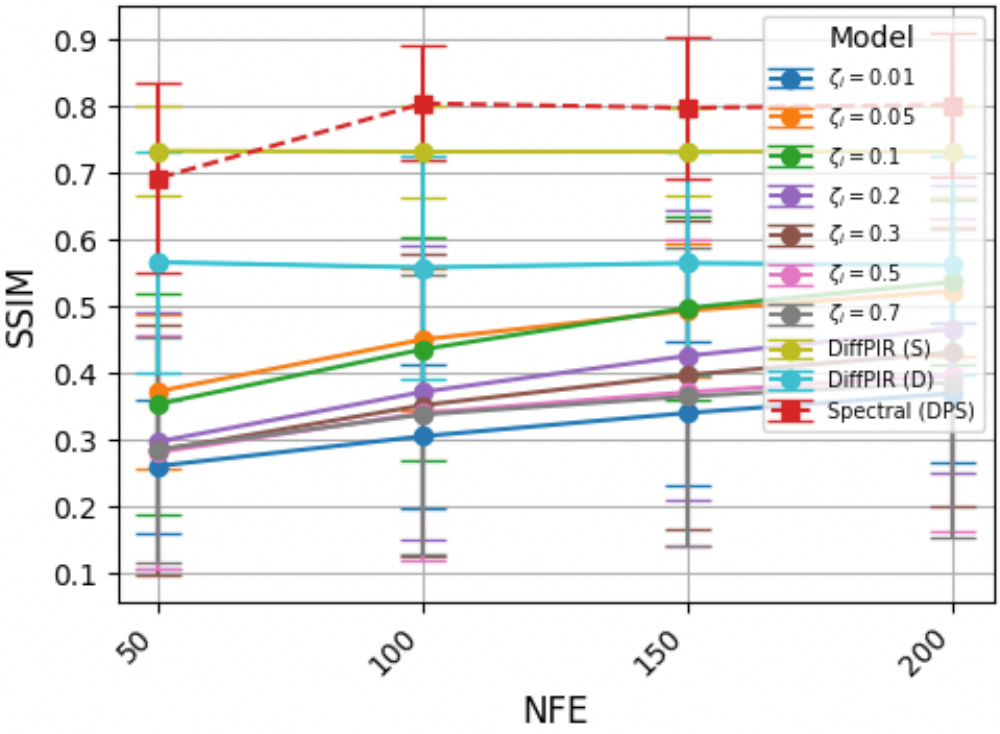}
        \caption{SSIM $\uparrow$}
        \label{fig:LPF_FFHQ_with_heuristics_ssim}
    \end{subfigure}
    \hfill
    \begin{subfigure}[b]{0.48\textwidth}
        \centering
        \includegraphics[width=\textwidth]{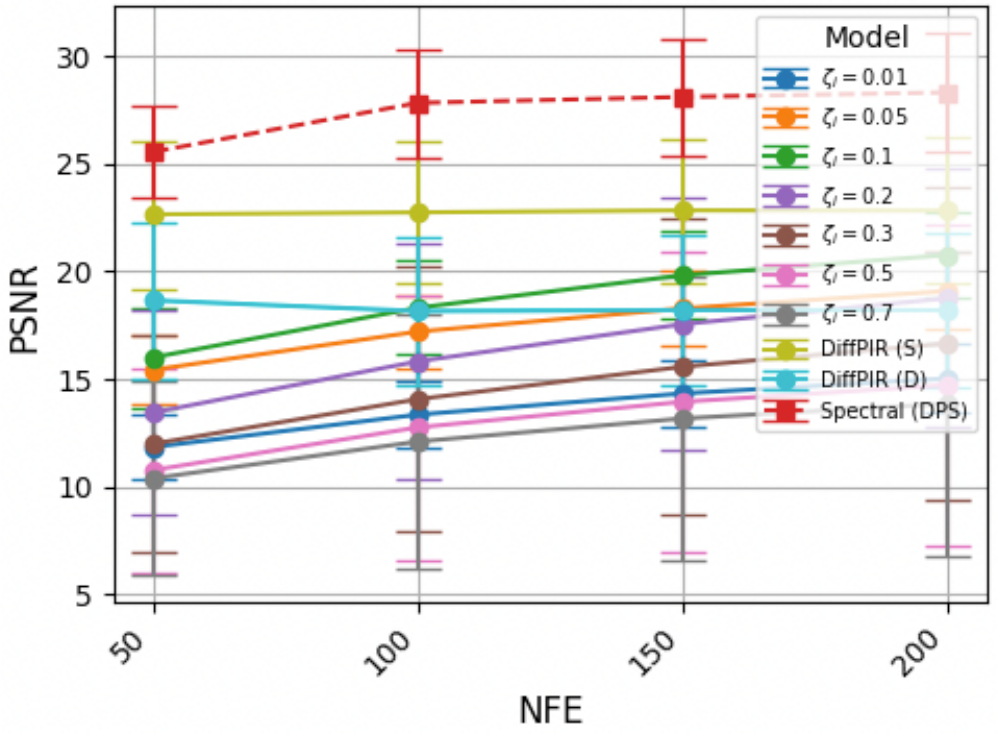}
        \caption{PSNR $\uparrow$}
        \label{fig:LPF_FFHQ_with_heuristics_psnr}
    \end{subfigure}
\caption{\textbf{LPF degradation.}
Comparison between DPS and DiffPIR heuristic configurations and the proposed DPS spectral recommendations under LPF degradation with $\mathcal{V}=0.1$ and additive Gaussian noise $\sigma_y=0.1$, evaluated across the number of function evaluations (NFE) on the FFHQ dataset.}
    \label{fig:LPF_comprehenssive_comparison}
\end{figure}


Figure~\ref{fig:LPF_comprehenssive_comparison} shows that the proposed spectral recommendations consistently outperform all DPS heuristic configurations in terms of reconstruction fidelity, achieving improved distortion metrics and stronger adherence to the underlying signal. In terms of perceptual quality, the spectral recommendations generally achieve lower FID values, although for small numbers of diffusion steps, heuristic configurations with lower guidance weights $\zeta'$ attain nearly comparable FID scores. This behavior is consistent with the observation that smaller likelihood guidance weights often favor perceptual quality by placing greater emphasis on the prior distribution, whereas increasing the guidance strength initially improves PSNR before deteriorating for larger values. As discussed in Section~\ref{sec:Empirical_Distribution}, perceptual quality alone is insufficient in inverse problems, where consistency with the measurements and accurate recovery of the underlying signal remain necessary.

Figure~\ref{fig:LPF_comprehenssive_comparison} further compares our approach with DiffPIR heuristics. The proposed spectral recommendations generally achieve higher PSNR values than both the deterministic and stochastic DiffPIR variants. In terms of perceptual metrics such as FID, however, the stochastic DiffPIR process achieves better performance, reflecting the effect of stochastic sampling on perceptual quality. This comparison is intended to position the proposed framework relative to existing posterior sampling approaches. We emphasize that since the spectral recommendations are optimized separately for each posterior sampling method, improved performance for one algorithm does not necessarily imply superior performance relative to alternative methods across all metrics and experimental settings.

\subsection{ImageNet spectral recommendation}
\label{sec:ImageNet_spectral_reccomendations}

\begin{figure}[h]
    \centering
    \includegraphics[width=0.5\textwidth]{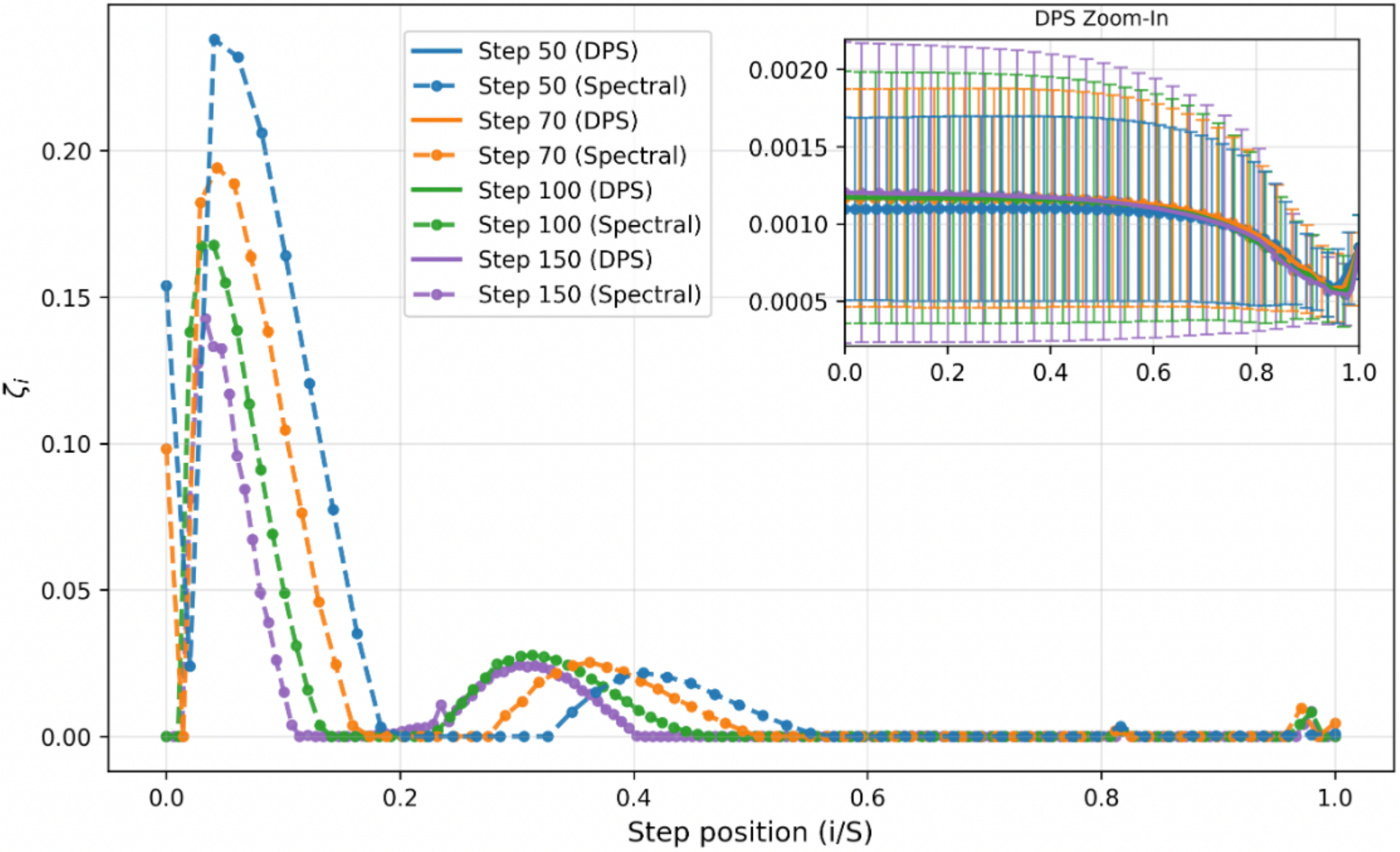}
    \caption{Comparison of spectral recommendations on the ImageNet $256$ dataset, with a zoomed-in view of the DPS heuristics for selected diffusion steps $S\in\{50, 70, 100,150\}$.}
    \vspace{-0.35cm}
    \label{fig:Comparison_DPS_and_Spectral_steps_ImageNet}
\end{figure}

Figure \ref{fig:Comparison_DPS_and_Spectral_steps_ImageNet} shows the spectral recommendations obtained for ImageNet the $256$ dataset. The overall behavior is similar to that observed for FFHQ, with guidance values increasing as the diffusion process progresses; however, the absolute magnitudes differ between the two datasets. In contrast, the DPS heuristic weights remain relatively smaller and exhibit a similar trend across both datasets.

Table~\ref{tab:lpf_results} presents a comprehensive comparison between our spectral recommendations and the original DPS heuristics. The comparison includes results on both FFHQ-$256$ and ImageNet-$256$ and considers several linear degradation settings with parameter pairs
$(\mathcal{V}, \sigma_y) \in \{(0.01, 0.1), (0.1, 0.1),  (0.1, 0.2)\}$.
For each degradation setting, we evaluate a range of diffusion step counts
$S \in \{50, 100, 200, 300, 400\}$ thereby evaluating posterior sampling performance in regimes that also permit relatively efficient synthesis. 


In this setting, selecting the DPS heuristic parameter $\zeta'$ for the FFHQ and ImageNet datasets is non-trivial.  \cite{chung2022diffusion} reports recommended values that are associated with specific high numbers of diffusion steps and particular degradation settings. Guided by the methodology described in \cite{chung2022diffusion}, we therefore evaluated several candidate values across different step counts and selected $\zeta' = 0.1$ for FFHQ (see Figure \ref{sec:prior_comprehensive_comparison})  and $\zeta' = 0.2$ for ImageNet, falling within the range of values considered in \cite{chung2022diffusion}. While smaller values resulted in insufficient measurement fidelity, larger values led to unstable behavior in the final reconstructions. This sensitivity underscores the advantage of the proposed spectral recommendations, which eliminate the need for repeated heuristic tuning.

Table~\ref{tab:lpf_results} reports results across four evaluation metrics, SSIM, PSNR, LPIPS, and FID, reflecting measurement fidelity and perceptual quality. All experiments were conducted using the implementation and pretrained models from \cite{chung2022diffusion}.  Overall, the spectral recommendations tend to achieve a more balanced tradeoff between these two aspects compared to the original DPS heuristics. While some reported values differ from those in \cite{chung2022diffusion}, this discrepancy is mainly attributable to the use of fewer diffusion steps and different degradation settings.

\begin{figure*}[h]

    \centering
    \setlength{\tabcolsep}{2pt}
    \renewcommand{\arraystretch}{1}
    \begin{tabular}{cc|ccc|ccc}
        \toprule
        Reference & Measurement
        & \multicolumn{3}{c}{DPS Heuristic}
        & \multicolumn{3}{c}{Spectral Weights} \\

        \midrule

        \includegraphics[width=0.1\textwidth]{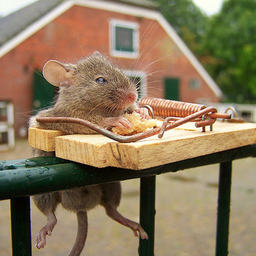}
        & \includegraphics[width=0.1\textwidth]{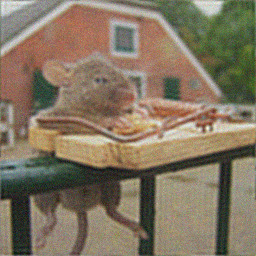}
        & \includegraphics[width=0.1\textwidth]{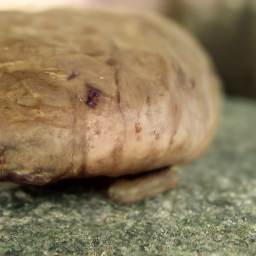}
        & \includegraphics[width=0.1\textwidth]{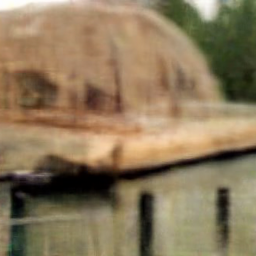}
        & \includegraphics[width=0.1\textwidth]{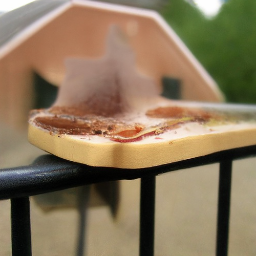}
        & \includegraphics[width=0.1\textwidth]{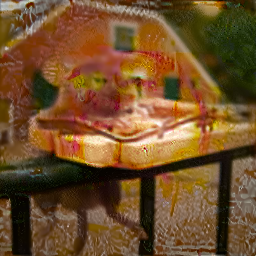}
        & \includegraphics[width=0.1\textwidth]{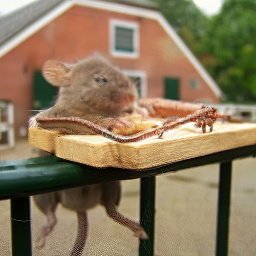}
        & \includegraphics[width=0.1\textwidth]{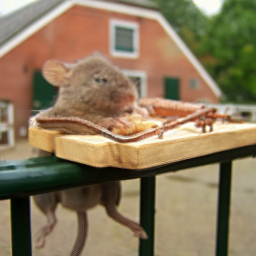} \\

        \includegraphics[width=0.1\textwidth]{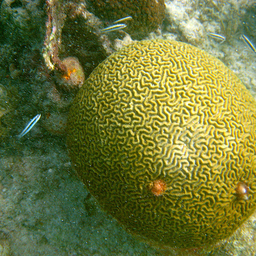}
        & \includegraphics[width=0.1\textwidth]{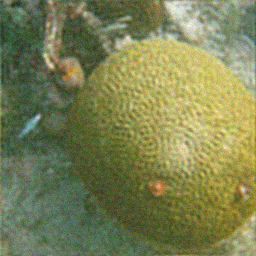}
        & \includegraphics[width=0.1\textwidth]{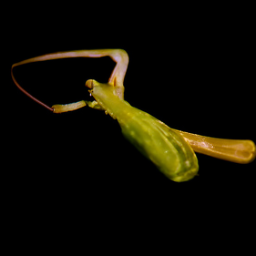}
        & \includegraphics[width=0.1\textwidth]{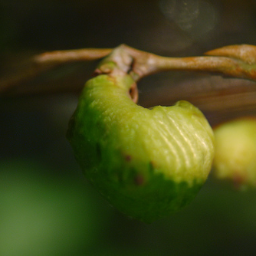}
        & \includegraphics[width=0.1\textwidth]{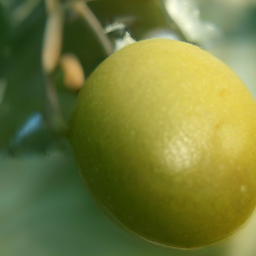}
        & \includegraphics[width=0.1\textwidth]{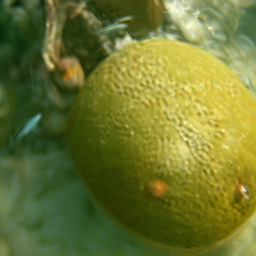}
        & \includegraphics[width=0.1\textwidth]{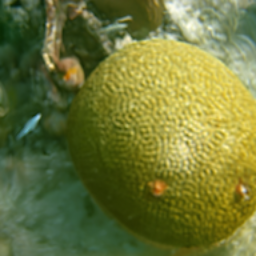}
        & \includegraphics[width=0.1\textwidth]{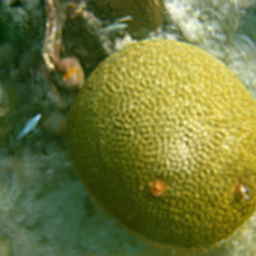} \\

        \includegraphics[width=0.1\textwidth]{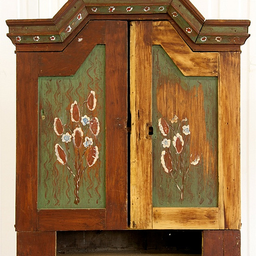}
        & \includegraphics[width=0.1\textwidth]{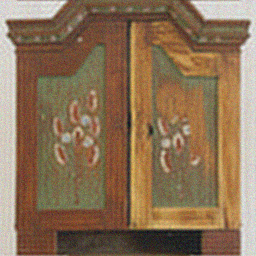}
        & \includegraphics[width=0.1\textwidth]{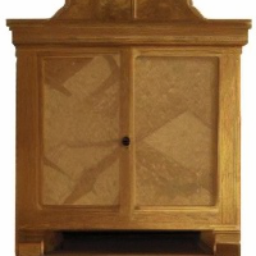}
        & \includegraphics[width=0.1\textwidth]{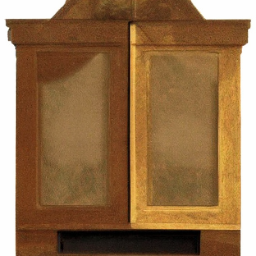}
        & \includegraphics[width=0.1\textwidth]{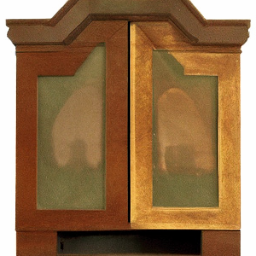}
        & \includegraphics[width=0.1\textwidth]{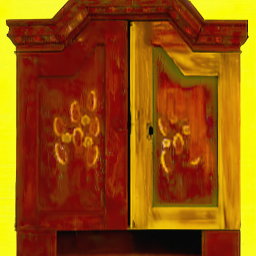}
        & \includegraphics[width=0.1\textwidth]{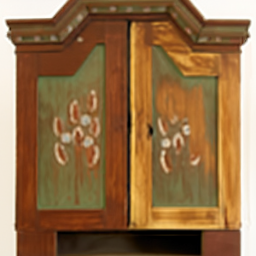}
        & \includegraphics[width=0.1\textwidth]{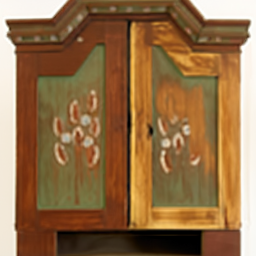} \\

    \end{tabular}

    \caption{Qualitative comparison of visual results on ImageNet. Each row shows the reference image, the degraded measurement, samples obtained using the DPS heuristic at 100, 200, and 400 diffusion steps (from left to right), and samples obtained using the proposed spectral recommendations at the same step counts.}
        \label{fig:ImageNet_phenomenon}
\end{figure*}

In some settings with a small number of diffusion steps, the DPS heuristic attains lower FID values; however, this is typically accompanied by reduced performance on measurement fidelity metrics. This behavior may be related to the small guidance step sizes commonly used by DPS, which can yield visually plausible samples that are less strongly constrained by the noisy measurements when using fewer diffusion steps. Representative examples are shown in Figure~\ref{fig:ImageNet_phenomenon}. These observations are also consistent with the findings reported in the original DPS paper~\cite{chung2022diffusion}.

Figure~\ref{fig:ImageNet_phenomenon} compares visual results obtained using the DPS heuristic and the proposed spectral recommendations under a degradation with $\mathcal{V}=0.1$ and $\sigma_y=0.1$, across different numbers of diffusion steps $S\in\{100,200,400\}$ and for the ImageNet dataset.

For a small number of diffusion steps, the DPS heuristic produces visually plausible images; however, these reconstructions often deviate from the reference images in terms of both global structure and fine details. In contrast, the spectral recommendations yield visual results that more closely align with the reference images and the measurements, although they may exhibit minor effects such as localized color variations. This behavior helps explain why the DPS heuristic can attain favorable FID values at low step counts while showing reduced measurements fidelity.

As the number of diffusion steps increases, the DPS heuristic increasingly incorporates information from the measurements, leading to visual results that more closely resemble the reference images, although deviations remain due to the limited heuristic step size. At the same time, the spectral recommendations produce visually natural outputs that remain consistent with the measurements. This convergence in behavior aligns with the quantitative trends reported in Table~\ref{fig:ImageNet_phenomenon}, where the spectral recommendations exhibit a balanced tradeoff between perceptual quality and measurement fidelity.

Figure~\ref{fig:qual_ffhq_500} shows additional visual results on the ImageNet dataset, evaluated at $500$ diffusion steps. While the DPS heuristic produces visually natural patterns, some differences with respect to the measurements remain. By comparison, the spectral recommendations yield visually natural samples that are more consistently aligned with the measurements, including at finer levels of detail.

\begin{table*}[t]
\caption{
Quantitative comparison between the proposed spectral recommendations and the DPS heuristic.
Results are reported on FFHQ-$256$ and ImageNet-$256$ across multiple degradation settings $(\mathcal{V}, \sigma_y)$ and diffusion step counts, using PSNR, SSIM, LPIPS, and FID.
}

  \label{tab:lpf_results}
  \begin{center}
    \begin{small}
      \begin{sc}
        \setlength{\tabcolsep}{4.5pt}
        \begin{tabular}{ccc c cccc cccc}
          \toprule
          \multirow{2}{*}{$\mathcal{V}$} &
          \multirow{2}{*}{$\sigma_y$} &
          \multirow{2}{*}{Method} &
          \multirow{2}{*}{Steps} &
          \multicolumn{4}{c}{FFHQ $256$} &
          \multicolumn{4}{c}{ImageNet $256$} \\
          \cmidrule(lr){5-8}\cmidrule(lr){9-12}
          & & & &
          PSNR$\uparrow$ & SSIM$\uparrow$ & LPIPS$\downarrow$ & FID$\downarrow$  &
          PSNR$\uparrow$ & SSIM$\uparrow$ &  LPIPS$\downarrow$ & FID$\downarrow$ \\
          \midrule


                 \multirow{8}{*}{0.01} &
          \multirow{8}{*}{0.1} &
          \multirow{4}{*}{Spectral} &
                      $50$  &$\mathbf{11.32}$ & $\mathbf{0.26}$& $0.71$&$197.83$ &$8.10$ &$0.12$ & $0.73$& $283.29$\\
          & & & $100$ & $\mathbf{13.24}$& $\mathbf{0.29}$& $\mathbf{0.67}$&$171.32$ & $11.85$&$\underline{0.24}$ &$0.68$ &$184.89$ \\
          & & & $200$ & $\mathbf{16.27}$ & $\mathbf{0.34}$& $\mathbf{0.60}$& $\mathbf{133.89}$&$\mathbf{16.04}$ &$\mathbf{0.30}$ &$\mathbf{0.60}$ &$174.88$\\
          & & & $300$ &$\mathbf{17.18}$ &$\mathbf{0.37}$ &$\mathbf{0.59}$ & $\mathbf{132.80}$&$\mathbf{14.01}$ &$\mathbf{0.28}$ &$\mathbf{0.64}$ &$\mathbf{137.44}$\\
                  & & & $400$ &$\mathbf{18.27}$ & $\mathbf{0.39}$& $\mathbf{0.57}$&$\mathbf{121.08}$ &$\mathbf{15.09}$ &$\mathbf{0.29}$ &$\mathbf{0.62}$ &$\mathbf{144.01}$\\
          \cmidrule(lr){3-12}
          & & \multirow{4}{*}{DPS} &
                      $50$  &$4.59$ & $0.24$&$\mathbf{0.52}$ & $ \mathbf{169.29}$ &$\mathbf{12.55}$ &$\mathbf{0.24}$ &$\mathbf{0.68}$ & $\mathbf{132.02}$\\
          & & & $100$ & $12.47$ & $0.28$& $0.68$& $\mathbf{155.47 }$&$\mathbf{12.01}$ &$\underline{0.24}$ &$\mathbf{0.67}$ & $\mathbf{142.94}$\\
          & & & $200$ & $13.53$&$0.30$ & $0.66$& $142.03 $& $11.83$& $0.24$& $0.66$& $\mathbf{147.25}$\\
          & & & $300$ & $14.56$&$0.33$ &$0.64$ &$137.69$ & $11.94$& $0.24$&$0.66$ &$144.88$ \\
                  & & & $400$ &$14.68$ & $0.33$& $0.63$& $137.13$&$11.94$ & $0.25$& $0.66$& $144.08$\\

          \midrule

             \multirow{8}{*}{0.1} &
          \multirow{8}{*}{0.1} &
          \multirow{4}{*}{{Spectral}} &

                $50$  & $\mathbf{25.55}$ & $\mathbf{0.69}$ & $\mathbf{0.32}$ & $\mathbf{50.86}$ & $\mathbf{15.99}$&$\mathbf{0.41}$ & $\mathbf{0.59}$&$262.73$ \\
          & & & $100$ & $\mathbf{27.82}$ & $\mathbf{0.80}$ & $\mathbf{0.23}$ & $\mathbf{41.31}$ &$10.22$ &$\textbf{0.40}$ & $\mathbf{0.58}$&$231.42$ \\
          & & & $200$ &$\mathbf{28.31}$ & $\mathbf{0.80}$ & $\mathbf{0.22}$ & $\mathbf{35.27}$ & $\mathbf{15.67}$& $\mathbf{0.41}$& $\mathbf{0.55}$& $\mathbf{120.55}$ \\
          & & & $300$ &$\mathbf{28.53}$ & $\mathbf{0.81}$&$\mathbf{0.21}$ &$\mathbf{31.72}$ &$\mathbf{14.56}$ &$\mathbf{0.37}$ & $\mathbf{0.58}$& $\mathbf{139.86}$ \\
            & & & $400$ &$\mathbf{28.59}$ & $\mathbf{0.82}$&$\mathbf{0.20}$ &$\mathbf{29.36}$ & $\mathbf{13.92}$& $\mathbf{0.35}$& $\mathbf{0.60}$& $\mathbf{54.22}$ \\
          \cmidrule(lr){3-12}
          & & \multirow{4}{*}{{DPS}} &
                $50$ & $16.00$ & $0.35$ & $0.57$ & ${69.13}$ &$12.73$ & $0.26$& $0.68$& $\mathbf{130.60}$\\
          & & & $100$ &$18.31$ & $0.43$ & $0.51$ & ${62.68}$
          & $\mathbf{12.41}$&$0.27$ &$0.67$ & $\mathbf{142.66}$\\
          & & & $200$ & $20.76$ & $0.53$ & $0.43$ & $55.88$ & $12.55$&$0.28$ & $0.66$&$143.66$ \\
          & & & $300$ & $22.00$&$0.57$ & $0.40$ &$54.41$ & $12.92$&$0.29$ &$0.64$ &$141.28$ \\
          & & & $400$ &$22.86$ &$0.60$ & $0.38$ &$53.78$ &$13.29$ & $0.31$&$0.63$ &$131.13$ \\


          \midrule

          \multirow{8}{*}{0.1} &
          \multirow{8}{*}{0.2} &
          \multirow{4}{*}{Spectral} &
          $50$  & $\mathbf{14.56}$ &$\mathbf{0.42}$ &$\mathbf{0.65}$ & $\mathbf{144.32}$&$1\mathbf{4.73}$ & $\mathbf{0.37}$&$\mathbf{0.64}$ & $152.59$\\
          & & & $100$ &$\mathbf{15.73}$ & $\mathbf{0.45}$&$\mathbf{0.59}$ &$\mathbf{120.04}$ & $\mathbf{14.81}$& $\mathbf{0.37}$& $\mathbf{0.63}$& $152.18$\\
          & & & $200$ & $\mathbf{18.72}$&$\mathbf{0.50}$ & $\mathbf{0.50}$&$\mathbf{85.81}$ & $\mathbf{14.83}$&$\mathbf{0.38}$ &$\mathbf{0.62}$ &$145.89$ \\
          & & & $300$ & $\mathbf{23.35}$& $\mathbf{0.58}$& $\mathbf{0.41}$&$\mathbf{57.94}$&$\mathbf{14.52}$ & $\mathbf{0.37}$& $\mathbf{0.64}$& $\mathbf{137.42}$\\
                  & & & $400$ & $\mathbf{22.69}$&$\mathbf{0.55}$ &$\mathbf{0.44}$ &$\mathbf{61.74}$ &$\mathbf{16.42}$ &$\mathbf{0.42}$ &$\mathbf{0.57}$ &$\mathbf{120.90}$ \\
          \cmidrule(lr){3-12}
          & & \multirow{4}{*}{DPS} &
          $50$  & $11.95$& $0.28$& $0.69$&$144.70$  &$12.96$ & $0.26$ &$0.68$ & $\mathbf{125.55}$\\
          & & & $100$ & $13.85$&$0.34$ &$0.63$ &$120.49$ & $12.58$&$0.27$ & $0.67$&$\mathbf{139.52}$ \\
          & & & $200$ & $17.08$& $0.43$& $0.54$&$91.55$  &$12.44$ &$0.28$ &$0.66$ & $\mathbf{141.26}$\\
          & & & $300$  &$17.63$ &$0.45$ &$0.53$    &$88.43$ & $12.90$&$0.29$&$0.65$&$137.63$ \\
                  & & & $400$  &$18.64$ & $0.49$& $0.50$&$82.39$&$13.12$ & $0.30$& $0.64$& $136.29$\\

          \bottomrule
        \end{tabular}
      \end{sc}
    \end{small}
  \end{center}
  \vskip -0.1in
\end{table*}

\begin{figure*}[h]
    \centering
    \setlength{\tabcolsep}{3pt}
    \renewcommand{\arraystretch}{1}

    \begin{tabular}{cccc}
        \toprule
        Original & Degraded & DPS Heuristic & Spectral weights \\
        \midrule

        \includegraphics[width=0.18\textwidth]{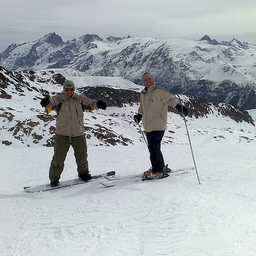} &
        \includegraphics[width=0.18\textwidth]{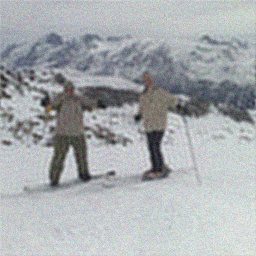} &
        \includegraphics[width=0.18\textwidth]{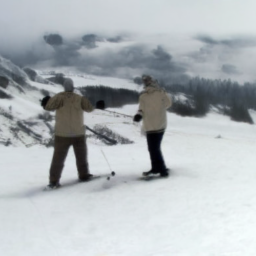} &
        \includegraphics[width=0.18\textwidth]{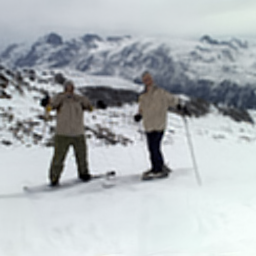} \\

        \includegraphics[width=0.18\textwidth]{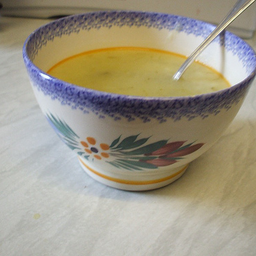} &
        \includegraphics[width=0.18\textwidth]{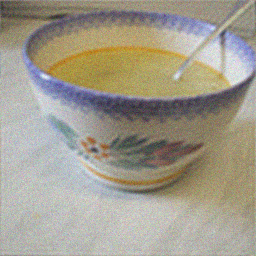} &
        \includegraphics[width=0.18\textwidth]{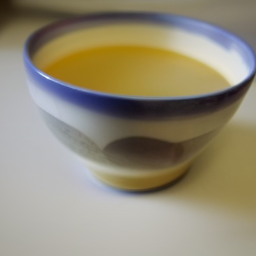} &
        \includegraphics[width=0.18\textwidth]{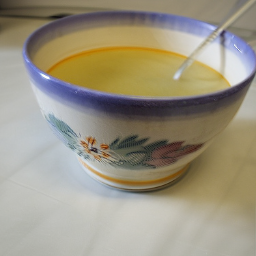} \\

        \includegraphics[width=0.18\textwidth]{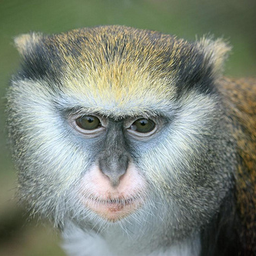} &
        \includegraphics[width=0.18\textwidth]{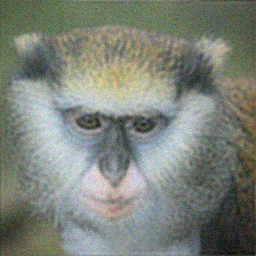} &
        \includegraphics[width=0.18\textwidth]{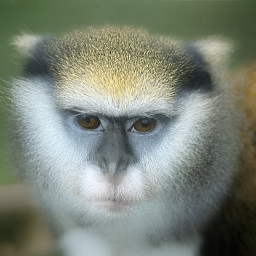} &
        \includegraphics[width=0.18\textwidth]{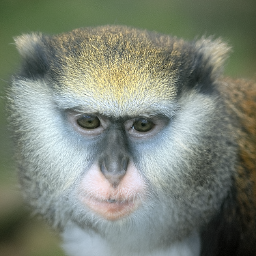} \\

        \includegraphics[width=0.18\textwidth]{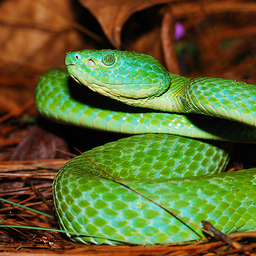} &
        \includegraphics[width=0.18\textwidth]{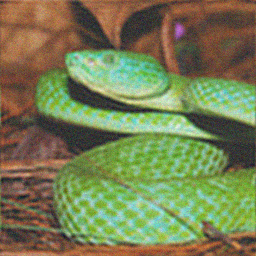} &
        \includegraphics[width=0.18\textwidth]{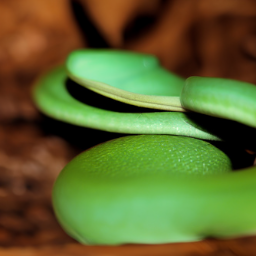} &
        \includegraphics[width=0.18\textwidth]{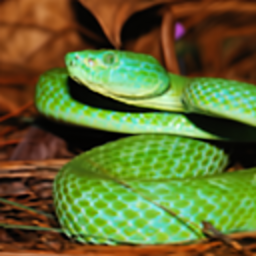} \\

            \includegraphics[width=0.18\textwidth]{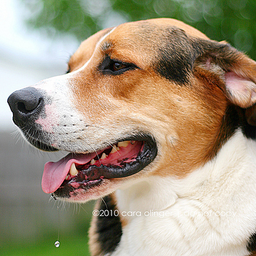} &
        \includegraphics[width=0.18\textwidth]{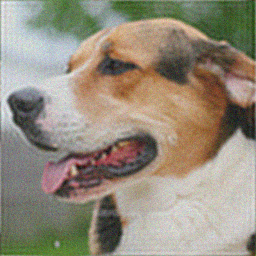} &
        \includegraphics[width=0.18\textwidth]{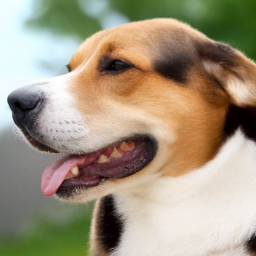} &
        \includegraphics[width=0.18\textwidth]{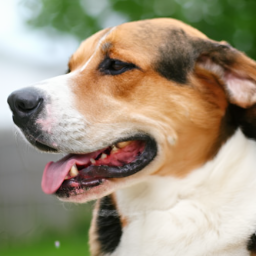} \\

                \includegraphics[width=0.18\textwidth]{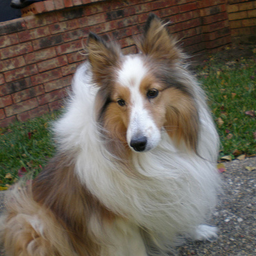} &
        \includegraphics[width=0.18\textwidth]{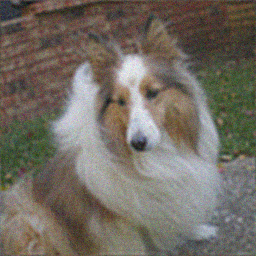} &
        \includegraphics[width=0.18\textwidth]{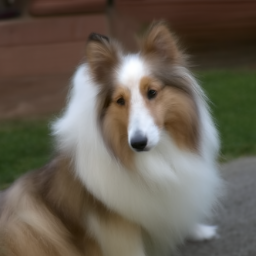} &
        \includegraphics[width=0.18\textwidth]{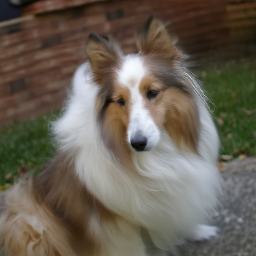} \\

        \bottomrule
    \end{tabular}

    \caption{Qualitative comparison of visual results on the ImageNet dataset.
Each row shows the reference image, the degraded observation with $\mathcal{V}=0.1$ and $\sigma_y=0.1$, and samples obtained using the DPS heuristic and the proposed spectral weighting, both evaluated at $500$ diffusion steps.}
    \label{fig:qual_ffhq_500}

\end{figure*}
\clearpage

\subsection{Comparison under inpainting degradation}
\label{sec:inpainting_comparison}

Section~\ref{sec:prior_comprehensive_comparison} demonstrated that the proposed spectral recommendations extend beyond the Gaussian setting through experiments on FFHQ and ImageNet. Here, we further examine how these recommendations generalize to degradation operators beyond the shift-invariant setting. To this end, we apply the spectral recommendations to two inpainting tasks: random masking with $70\%$ random pixel removal and a $128\times128$ box mask, both corrupted by additive Gaussian noise with standard deviation $\sigma_y=0.1$. We compare the results with DPS across a range of heuristic guidance weights $\zeta'=\{0.01,0.05,0.1,0.3,0.5,0.7,1\}$, as well as with recent posterior sampling methods, including DAPS~\cite{zhang2025improving} and DiffPIR~\cite{zhu2023denoising}, using their predefined heuristic configurations.

\begin{figure}[h]
    \centering

    \begin{subfigure}[b]{0.48\textwidth}
        \centering
        \includegraphics[width=\textwidth]{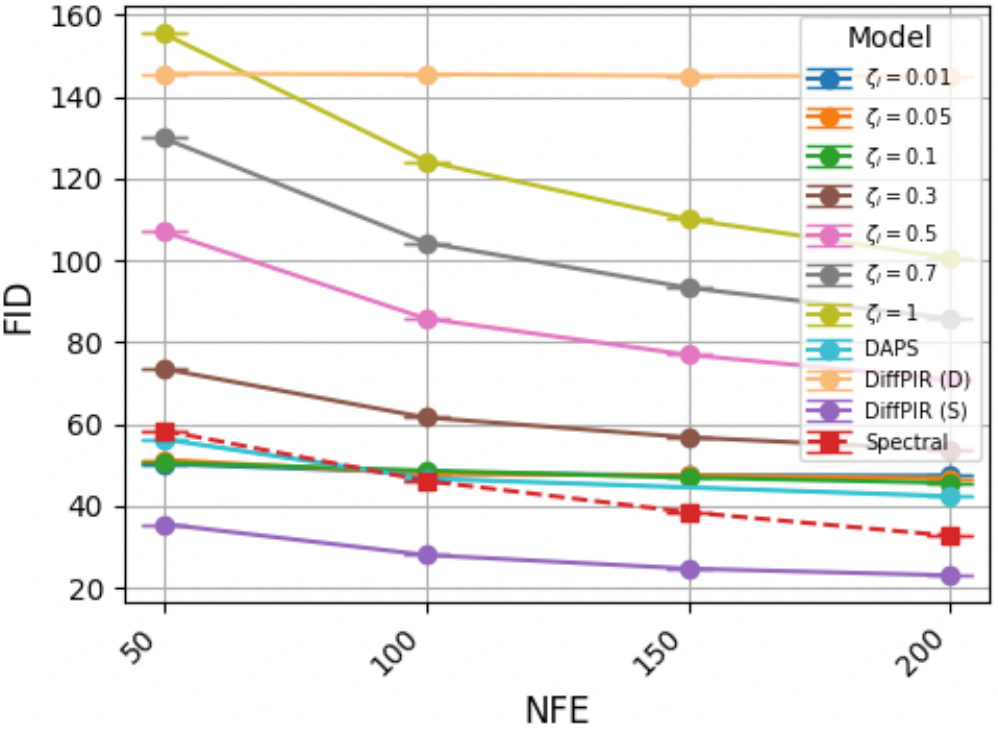}
        \caption{FID $\downarrow$}
        \label{fig:inpainting_random_fid}
    \end{subfigure}
    \hfill
    \begin{subfigure}[b]{0.48\textwidth}
        \centering
        \includegraphics[width=\textwidth]{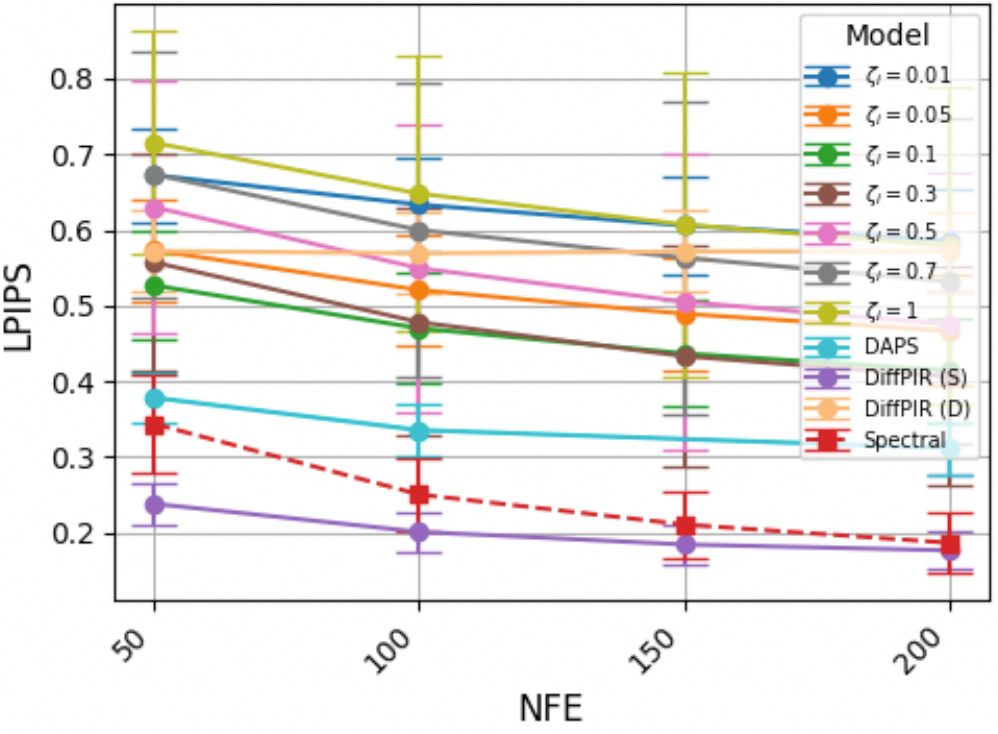}
        \caption{LPIPS $\downarrow$}
        \label{fig:inpainting_random_lpips}
    \end{subfigure}

    \vspace{0.5em}

    \begin{subfigure}[b]{0.48\textwidth}
        \centering
        \includegraphics[width=\textwidth]{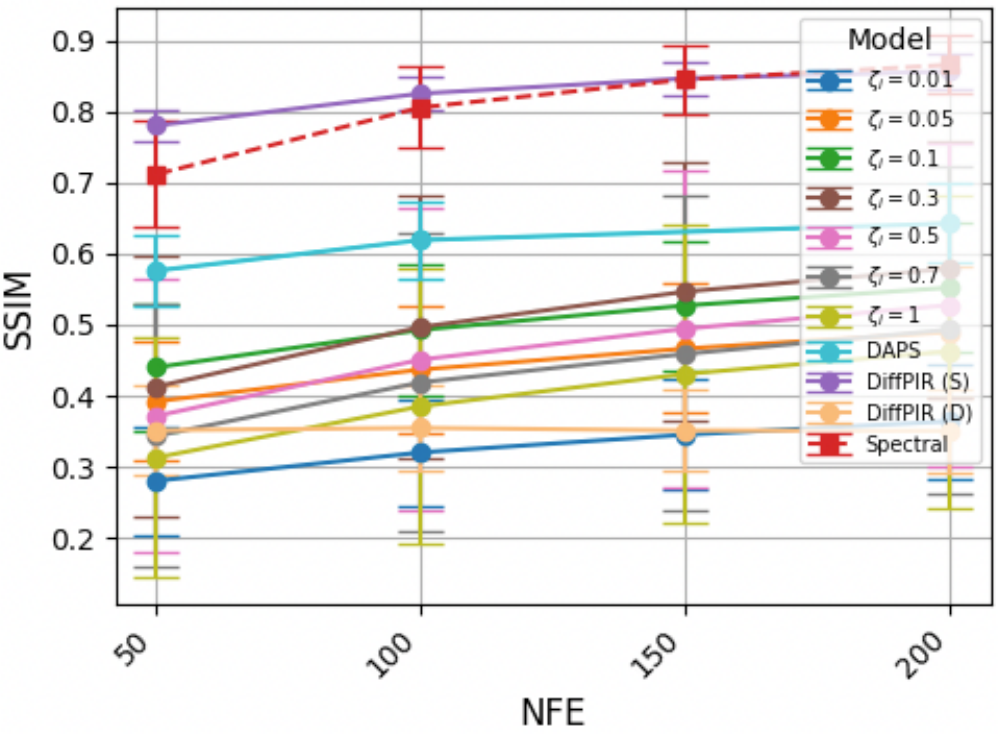}
        \caption{SSIM $\uparrow$}
        \label{fig:inpainting_random_ssim}
    \end{subfigure}
    \hfill
    \begin{subfigure}[b]{0.48\textwidth}
        \centering
        \includegraphics[width=\textwidth]{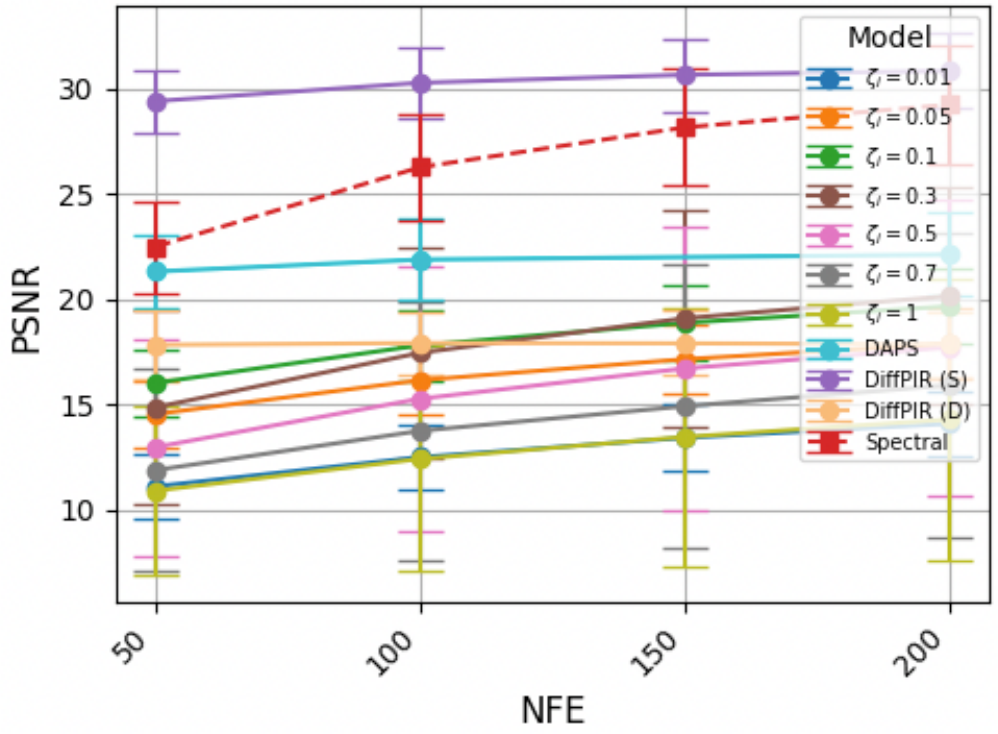}
        \caption{PSNR $\uparrow$}
        \label{fig:inpainting_random_psnr}
    \end{subfigure}
\caption{\textbf{Random inpainting degradation.}
Comparison between DPS, DiffPIR, and DAPS heuristic configurations and the proposed DPS spectral recommendations under inpainting with $70\%$ randomly removed pixels and additive Gaussian noise $\sigma_y=0.1$, evaluated across the number of function evaluations (NFE) on the FFHQ dataset.
}
    \label{fig:inpainting_random_comparison}
\end{figure}

\begin{figure}[h]
    \centering

    \begin{subfigure}[b]{0.48\textwidth}
        \centering
        \includegraphics[width=\textwidth]{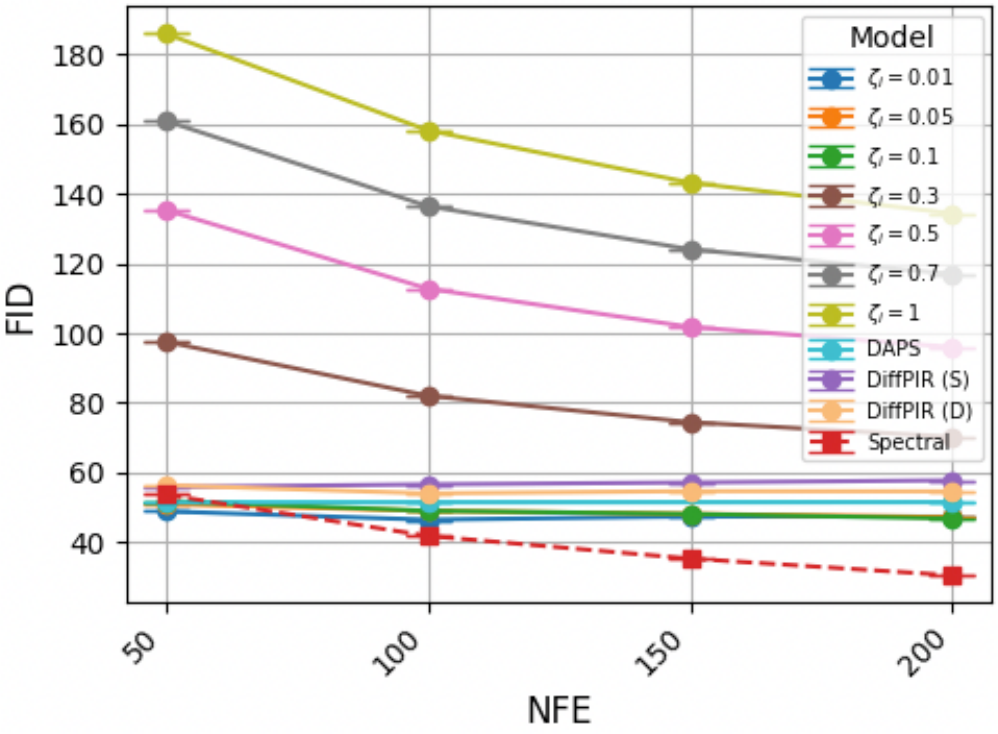}
        \caption{FID $\downarrow$}
        \label{fig:inpainting_box_fid}
    \end{subfigure}
    \hfill
    \begin{subfigure}[b]{0.48\textwidth}
        \centering
        \includegraphics[width=\textwidth]{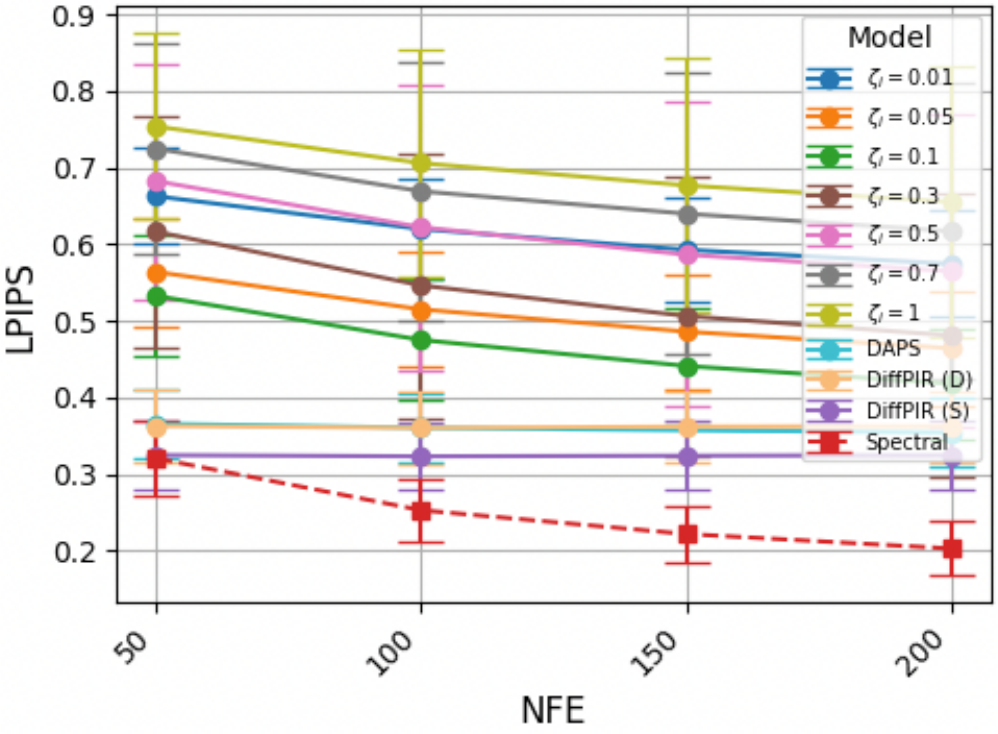}
        \caption{LPIPS $\downarrow$}
        \label{fig:inpainting_box_lpips}
    \end{subfigure}

    \vspace{0.5em}

    \begin{subfigure}[b]{0.48\textwidth}
        \centering
        \includegraphics[width=\textwidth]{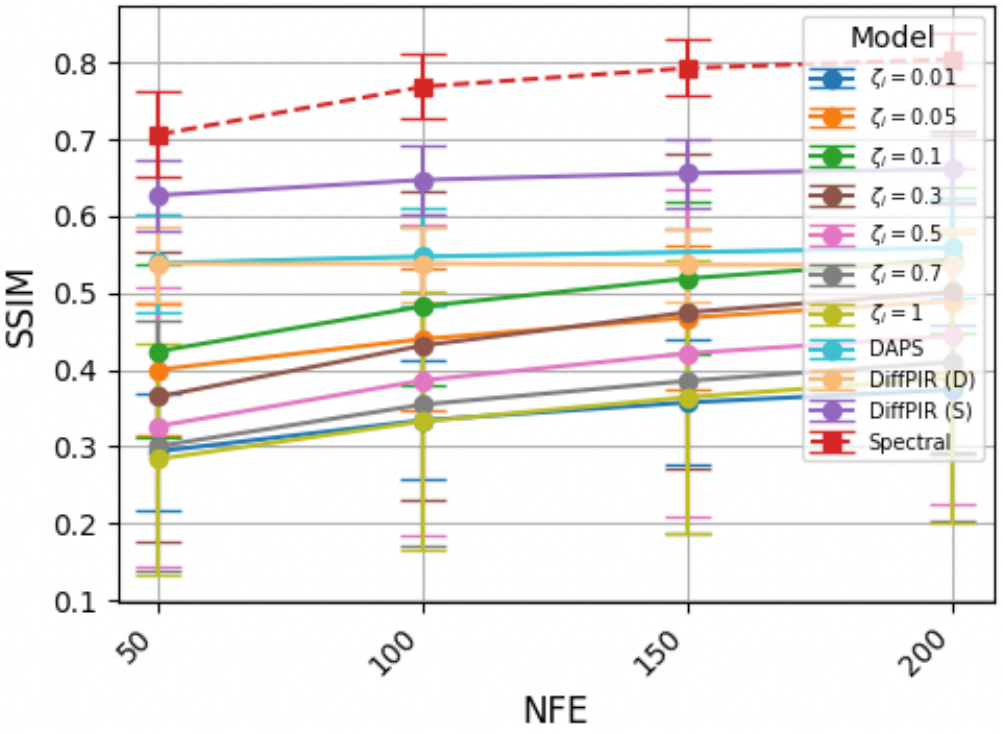}
        \caption{SSIM $\uparrow$}
        \label{fig:inpainting_box_ssim}
    \end{subfigure}
    \hfill
    \begin{subfigure}[b]{0.48\textwidth}
        \centering
        \includegraphics[width=\textwidth]{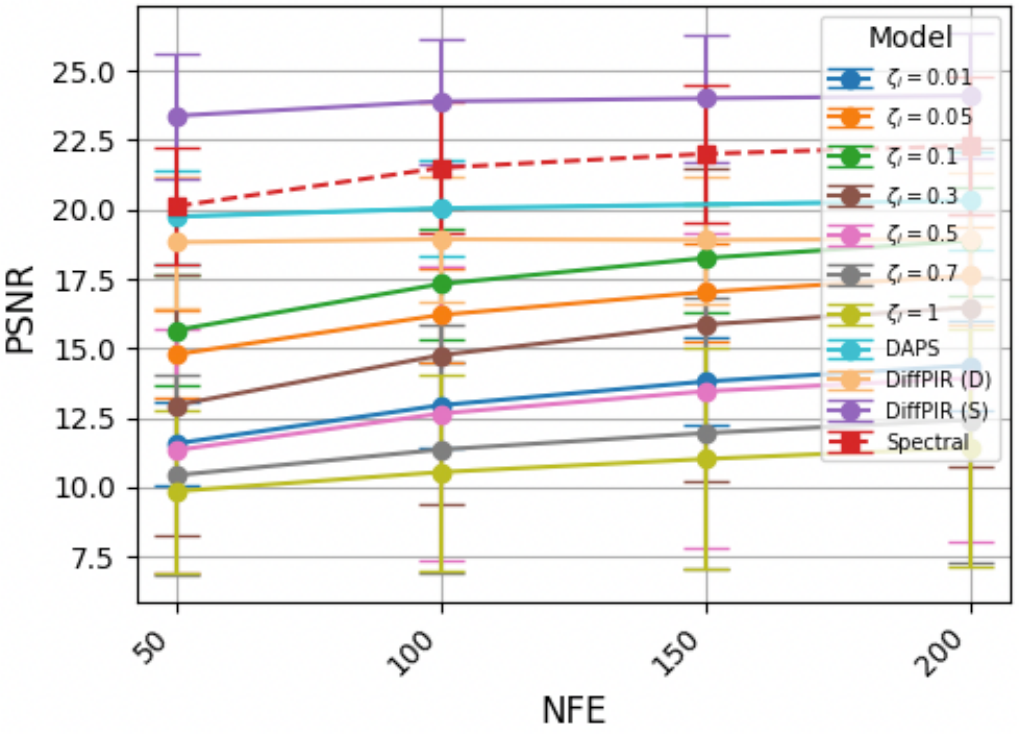}
        \caption{PSNR $\uparrow$}
        \label{fig:inpainting_box_psnr}
    \end{subfigure}
    \caption{\textbf{Box inpainting degradation.}
Comparison between DPS, DiffPIR, and DAPS heuristic configurations and the proposed DPS spectral recommendations under box inpainting with a $128\times128$ mask and additive Gaussian noise $\sigma_y=0.1$, evaluated across the number of function evaluations (NFE) on the FFHQ dataset.
}
    \label{fig:inpainting_box_comparison}
\end{figure}

Figure~\ref{fig:inpainting_random_comparison} presents the results for the random inpainting task. The proposed spectral recommendations consistently outperform all DPS heuristic configurations in terms of measurement fidelity, achieving improved PSNR values across diffusion steps. In terms of perceptual quality, the spectral recommendations also obtain lower FID values than the heuristic alternatives. However, for small guidance heuristic weights, such as $\zeta'=\{0.01,0.05\}$, slightly better FID values are observed for a small number of diffusion steps (e.g., $50$ steps). As discussed in Section~\ref{sec:Empirical_Distribution}, perceptual quality alone does not fully characterize performance in inverse problems, where accurate reconstruction of the underlying signal is also required. Figure~\ref{fig:inpainting_box_comparison} presents the results for the box inpainting task, showing the same conclusions in terms of both reconstruction fidelity and perceptual quality.

Figures~\ref{fig:inpainting_random_comparison} and~\ref{fig:inpainting_box_comparison} further compare the spectral recommendations derived for the DPS framework with DAPS and DiffPIR. In both cases, we use the heuristic configurations proposed for the FFHQ inpainting task, including both the deterministic and stochastic DiffPIR variants. Overall, the spectral recommendations outperform both methods across most settings. In the random inpainting task, however, the stochastic DiffPIR process achieves improved FID values, which is consistent with the stochastic sampling process. We emphasize that this comparison is intended to position the proposed framework relative to existing posterior sampling approaches. Since the spectral recommendations are optimized separately for each posterior sampling method, improved performance for one algorithm does not necessarily imply superior performance relative to alternative methods across all metrics and experimental settings. Rather, the results demonstrate clear gains in key aspects. For degradations closer to a circulant LPF (e.g., blur with non-circular padding), generalization is more natural, and the optimization can be performed using a circulant approximation.




\subsubsection{Additional visual results}
\begin{figure*}[h]

    \centering
    \setlength{\tabcolsep}{2pt}
    \renewcommand{\arraystretch}{1}
    \begin{tabular}{cc|c|c}
        \toprule
        Reference & Measurement
        & \multicolumn{1}{c}{DPS Heuristic}
        & \multicolumn{1}{c}{Spectral Weights} \\

        \midrule

        \includegraphics[width=0.15\textwidth]{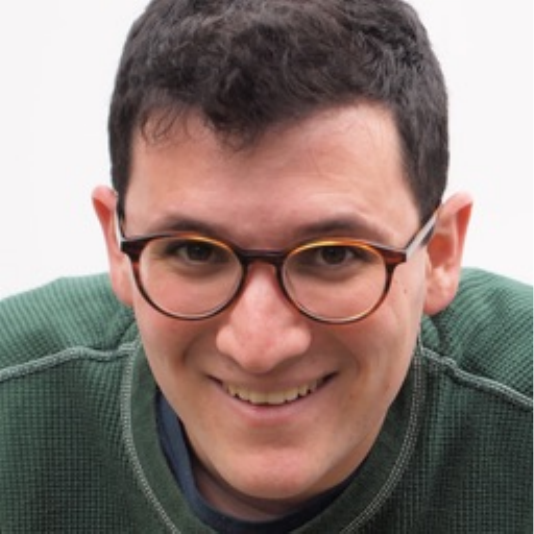}
        & \includegraphics[width=0.15\textwidth]{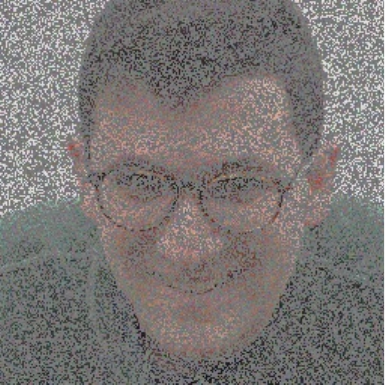}
        & \includegraphics[width=0.15\textwidth]{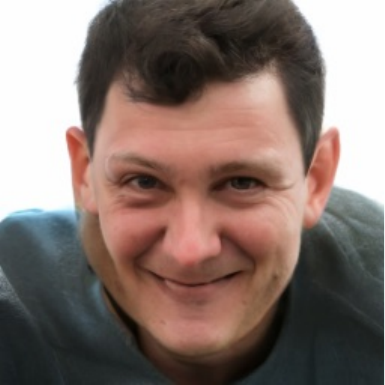}
        & \includegraphics[width=0.15\textwidth]{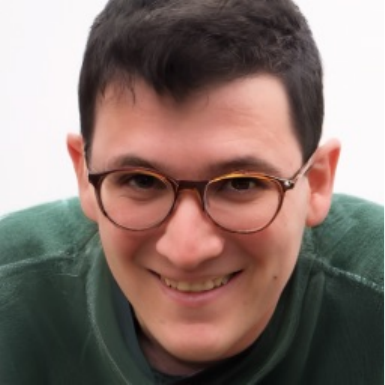}\\

        \includegraphics[width=0.15\textwidth]{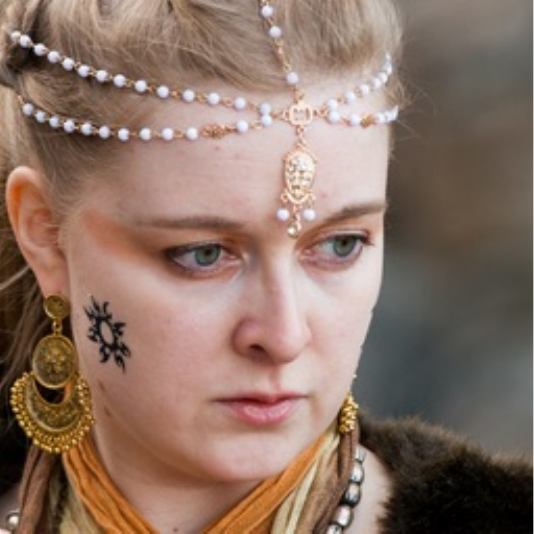}
        & \includegraphics[width=0.15\textwidth]{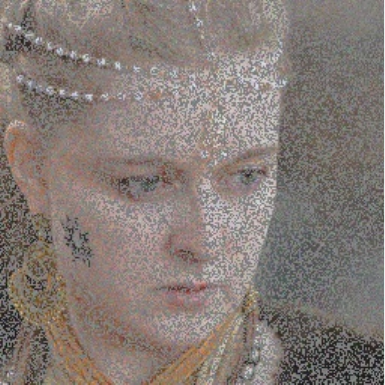}
        & \includegraphics[width=0.15\textwidth]{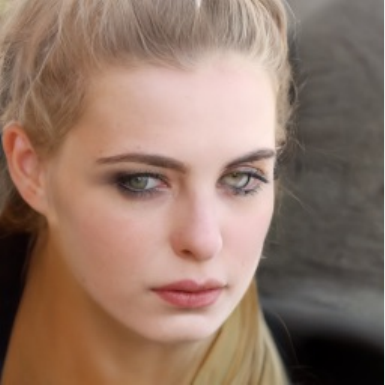}
        & \includegraphics[width=0.15\textwidth]{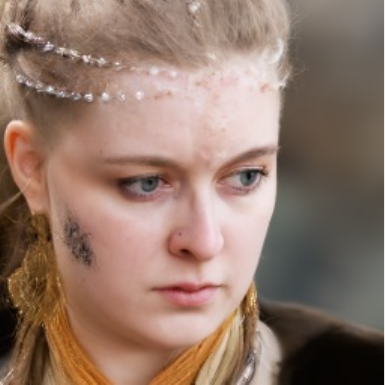}\\

        \includegraphics[width=0.15\textwidth]{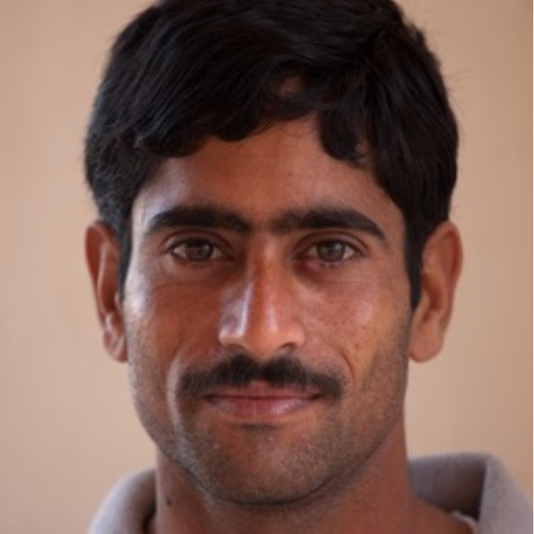}
        & \includegraphics[width=0.15\textwidth]{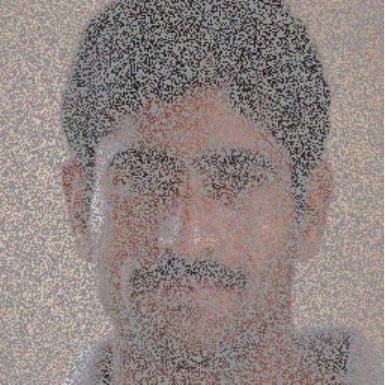}
        & \includegraphics[width=0.15\textwidth]{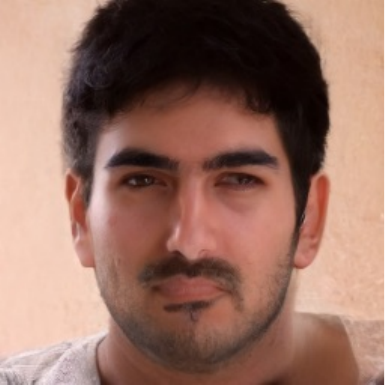}
        & \includegraphics[width=0.15\textwidth]{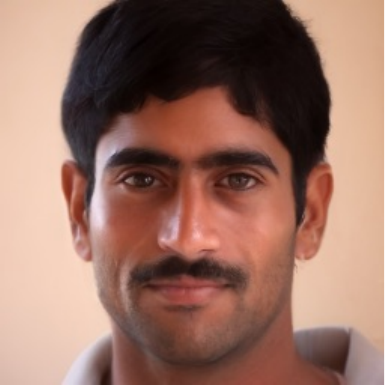}\\

    \end{tabular}

\caption{Qualitative comparison of reconstructions on the FFHQ dataset under random inpainting with $70\%$ missing pixels, comparing the DPS heuristic ($\zeta'=0.1$) and the proposed spectral recommendations across different diffusion step counts.}
        \label{fig:FFHQ_inpainting_dps_random_compare}
\end{figure*}

\begin{figure*}[h]

    \centering
    \setlength{\tabcolsep}{2pt}
    \renewcommand{\arraystretch}{1}
    \begin{tabular}{cc|c|c}
        \toprule
        Reference & Measurement
        & \multicolumn{1}{c}{DPS Heuristic}
        & \multicolumn{1}{c}{Spectral Weights} \\

        \midrule

        \includegraphics[width=0.15\textwidth]{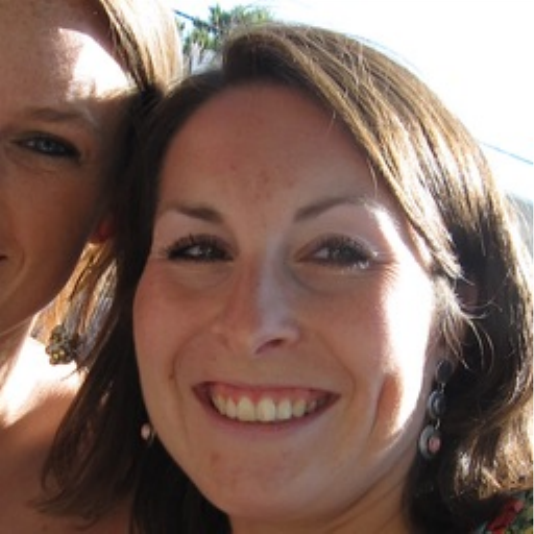}
        & \includegraphics[width=0.15\textwidth]{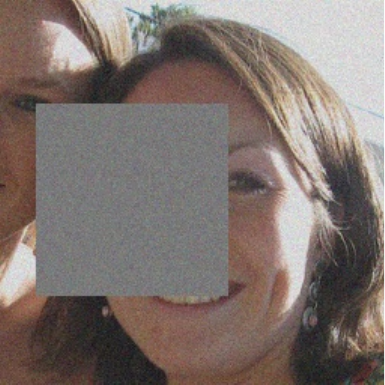}
        & \includegraphics[width=0.15\textwidth]{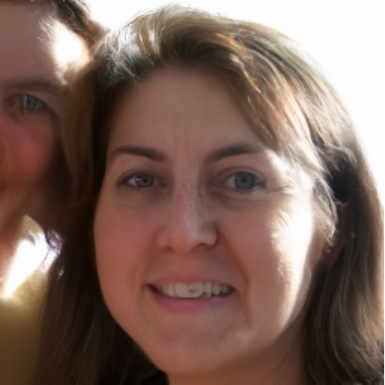}
        & \includegraphics[width=0.15\textwidth]{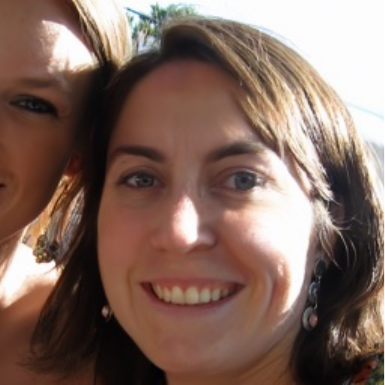}\\

        \includegraphics[width=0.15\textwidth]{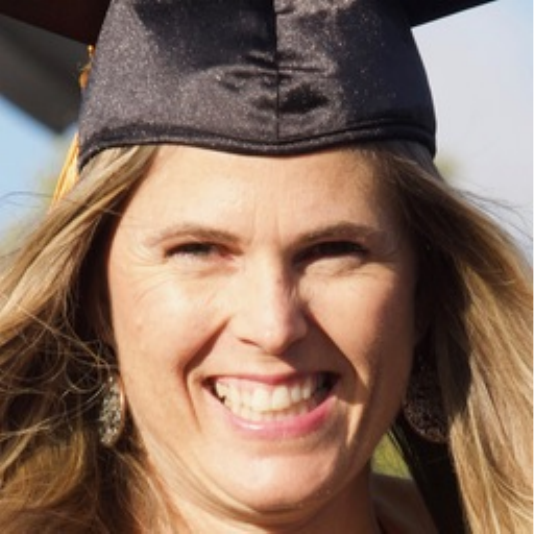}
        & \includegraphics[width=0.15\textwidth]{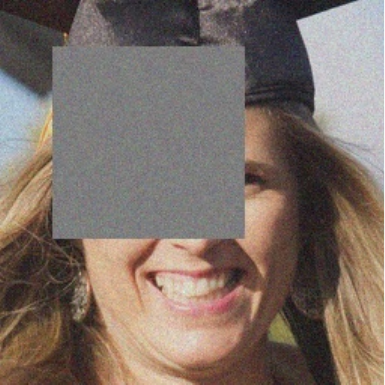}
        & \includegraphics[width=0.15\textwidth]{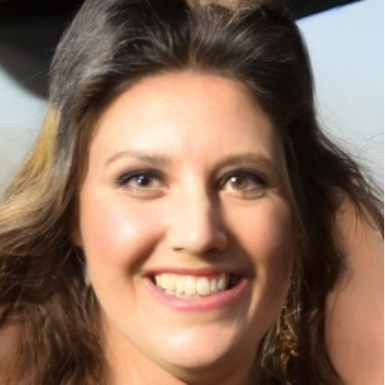}
        & \includegraphics[width=0.15\textwidth]{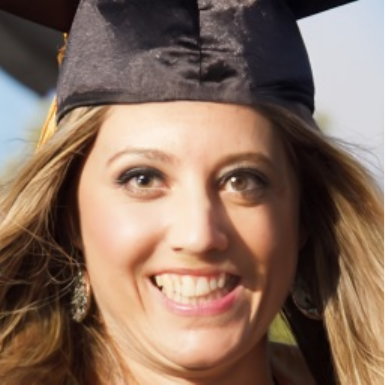}\\
        
        \includegraphics[width=0.15\textwidth]{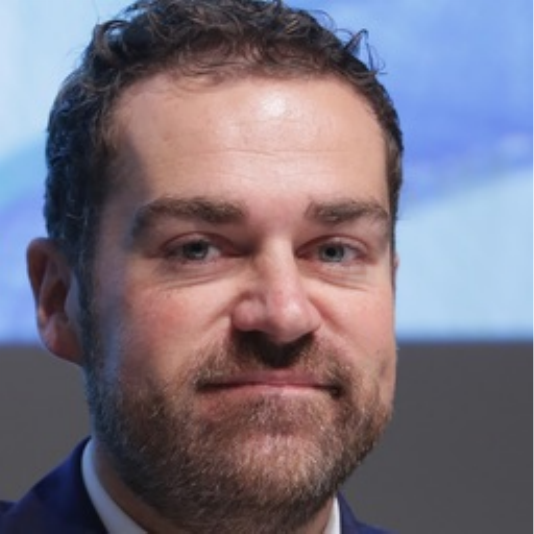}
        & \includegraphics[width=0.15\textwidth]{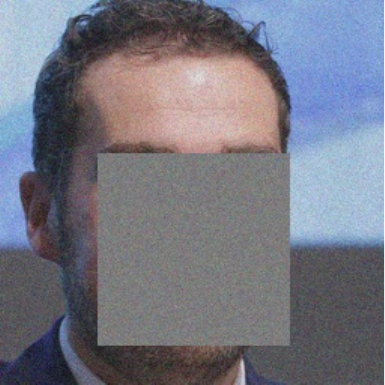}
        & \includegraphics[width=0.15\textwidth]{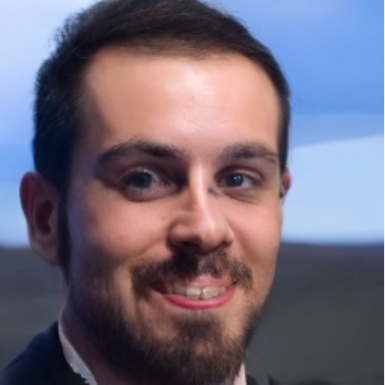}
        & \includegraphics[width=0.15\textwidth]{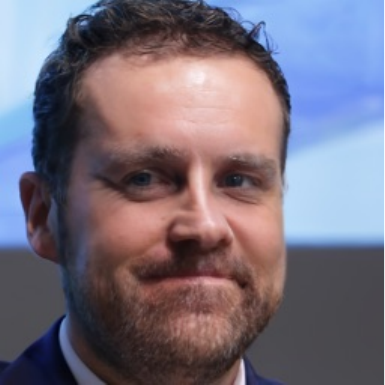}\\

    \end{tabular}

\caption{Qualitative comparison of reconstructions on the FFHQ dataset under box inpainting with a $128 \times 128$ missing region, comparing the DPS heuristic ($\zeta'=0.1$) and the proposed spectral recommendations across different diffusion step counts.}
        \label{fig:FFHQ_inpainting_dps_box_compare}
\end{figure*}

Figures~\ref{fig:FFHQ_inpainting_dps_random_compare} and \ref{fig:FFHQ_inpainting_dps_box_compare} present representative examples of the DPS algorithm on the FFHQ dataset under random and box inpainting, respectively. In both figures, the first column shows the original image, and the second column shows the degraded input. The next column corresponds to DPS reconstructions using the predefined heuristic for $200$ diffusion steps, while the final column presents reconstructions obtained using the spectral recommendations at the same number of steps.

For the DPS heuristic, we show results with $\zeta'=0.1$, as it provides favorable performance among the tested values of $\zeta'$, as observed in Figures~\ref{fig:inpainting_random_comparison} and \ref{fig:inpainting_box_comparison}. For both degradation settings, the spectral recommendations produce reconstructions that are more consistent with the original image while preserving a natural visual appearance. In contrast, the DPS heuristic is less faithful to the original image and tends to introduce new information. For example, in the box inpainting setting, it may also alter regions outside the inpainted area and introduce new content, even when using relatively large guidance weights.




\clearpage
\newpage
\subsection{fastMRI dataset}
\label{sec:fastMRI_DPS}
\subsubsection{Dataset preparation}
\label{sec:Dataset_Preparation}


In our work, we consider the single-coil knee MRI scans from the fastMRI dataset \cite{zbontar2018fastmri,knoll2020fastmri}. For training, we sample 15 slices at uniformly spaced indices between the one-third and two-third positions of each volume, since slices near the edges typically contain less anatomical information. This results in a training set of $12,546$ images. For validating our spectral recommendations, we use $200$ samples extracted from the middle slice of each volume in the validation set. The selected slices are taken from the provided magnitude reconstructions. We center-crop the images to a spatial resolution of $320\times320$, following \cite{jalal2021robust}, and normalize the pixel intensities to the range $[-1,1]$ before diffusion model training and evaluation.

\subsubsection{Degradation operators}
Magnetic Resonance Imaging (MRI) is a widely used non-invasive imaging technique that provides clinically important diagnostic information. In MRI systems, measurements are acquired in the Fourier domain, commonly referred to as k-space, through magnetic field encoding. Acquiring the full k-space, however, is often associated with long scan times and increased acquisition costs. As a result, reconstructing high-quality images from incomplete, noisy, or undersampled k-space measurements has become a central inverse problem in accelerated MRI, attracting significant attention in recent years \cite{weiss2019pilot, wang2022one}. 

For the fastMRI single-coil knee MRI slices, we follow the degradation protocol introduced by \cite{jalal2021robust}. In particular, we consider a horizontal frequency undersampling operator in k-space. Given an acceleration factor $R$, the operator preserves the central $\frac{120}{R}$ low-frequency components and additionally samples $\frac{200}{R}$ frequencies uniformly across the remaining spectrum. Overall, this results in retaining $\frac{320}{R}$ frequency measurements out of the original $320$ acquired frequencies.

\begin{figure}[h]
    \centering

    \begin{subfigure}{0.32\linewidth}
        \centering
        \includegraphics[width=\linewidth]{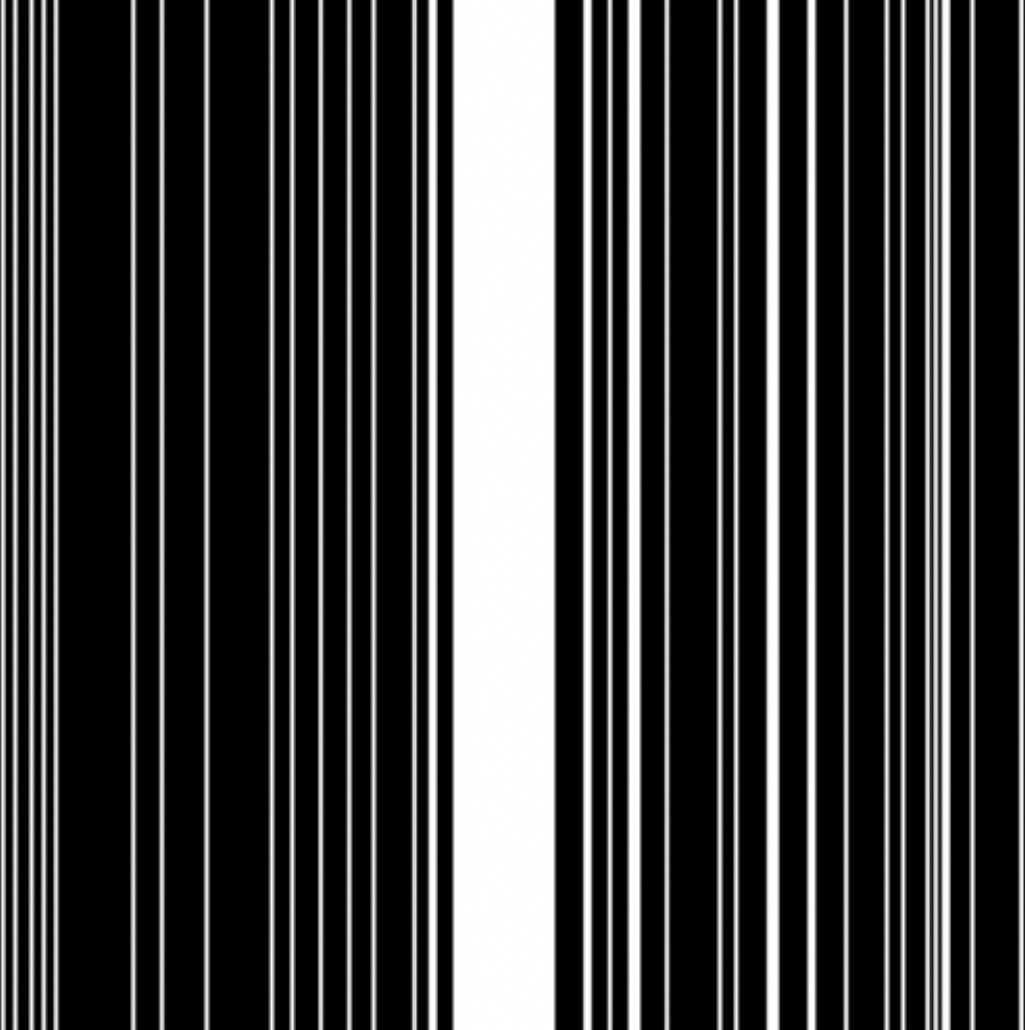}
        \caption{$R=4$}
    \end{subfigure}
    \hfill
    \begin{subfigure}{0.32\linewidth}
        \centering
        \includegraphics[width=\linewidth]{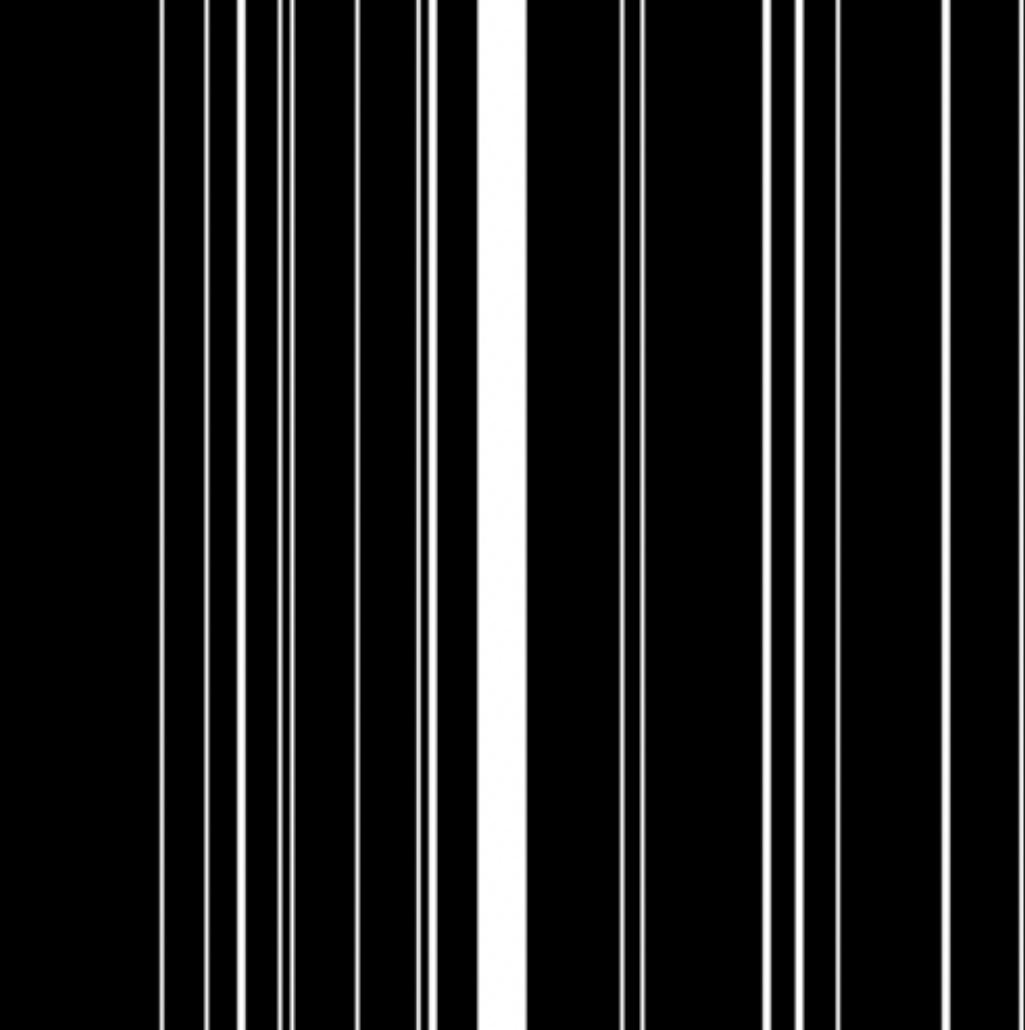}
        \caption{$R=8$}
    \end{subfigure}
    \hfill
    \begin{subfigure}{0.32\linewidth}
        \centering
        \includegraphics[width=\linewidth]{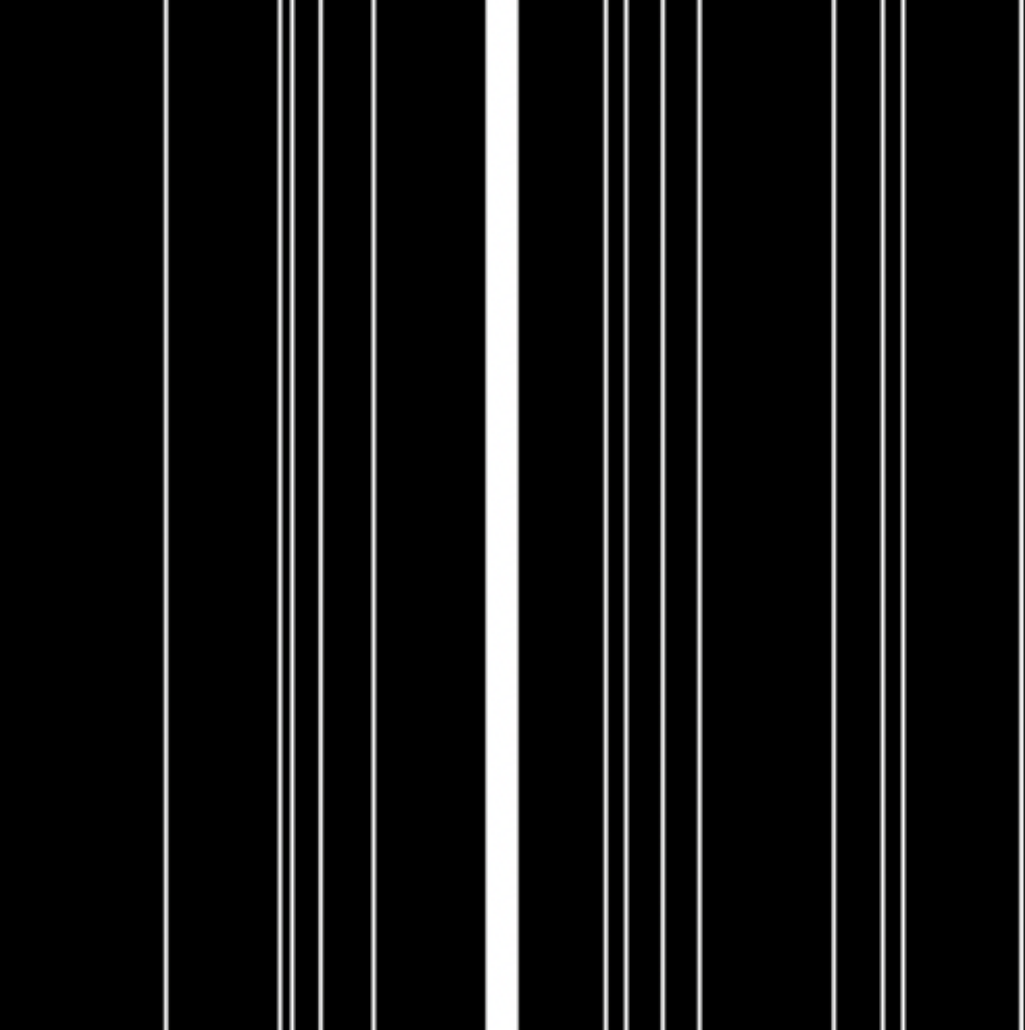}
        \caption{$R=12$}
    \end{subfigure}

    \caption{Frequency subsampling masks used for accelerated MRI acquisition with acceleration factors $R\in\{4,8,12\}$. As the acceleration factor increases, fewer frequency components are retained, leading to more severe undersampling in k-space.}
    \label{fig:mri_masks_R}
\end{figure}

Notably, the accelerated MRI degradation operator satisfies the shift-invariant property considered in our analysis. In the Fourier domain, its spectral representation $\boldsymbol{\Lambda}_{h}$ corresponds to a diagonal sampling mask that selects the acquired k-space frequencies, naturally yielding a diagonal representation of the degradation process. This provides a practically relevant setting in which our framework applies directly to a widely studied real-world inverse problem. Examples of the masks used for acceleration factors $R\in\{4,8,12\}$ are shown in Figure~\ref{fig:mri_masks_R}.

\subsubsection{Training details}
\label{sec:Training_Details}



Since no pretrained prior model is available for the fastMRI dataset, we used the same U-Net architecture employed in \cite{chung2022diffusion} for the FFHQ and ImageNet datasets. We adapted the input size to the fastMRI dataset and trained the model using the dataset constructed as described in Section~\ref{sec:Dataset_Preparation}. The model was trained for $80{,}000$ steps using eight NVIDIA A40 GPUs with a batch size of $16$, using the AdamW optimizer with a learning rate of $10^{-4}$ and weight decay $0$, together with an exponential moving average (EMA) of the model parameters.

\subsubsection{Comprehensive comparison - accelerated sampling}

\begin{figure}[h]
    \centering

    \begin{subfigure}[b]{0.48\textwidth}
        \centering
        \includegraphics[width=\textwidth]{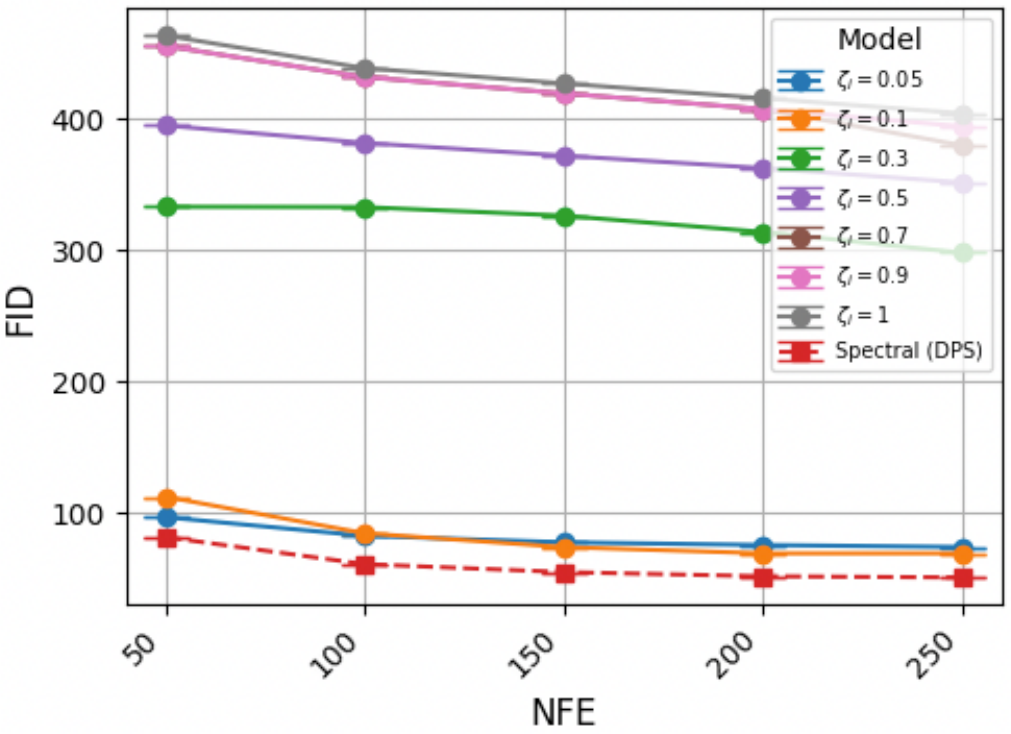}
        \caption{FID $\downarrow$}
        \label{fig:fmri_lps_fid}
    \end{subfigure}
    \hfill
    \begin{subfigure}[b]{0.48\textwidth}
        \centering
        \includegraphics[width=\textwidth]{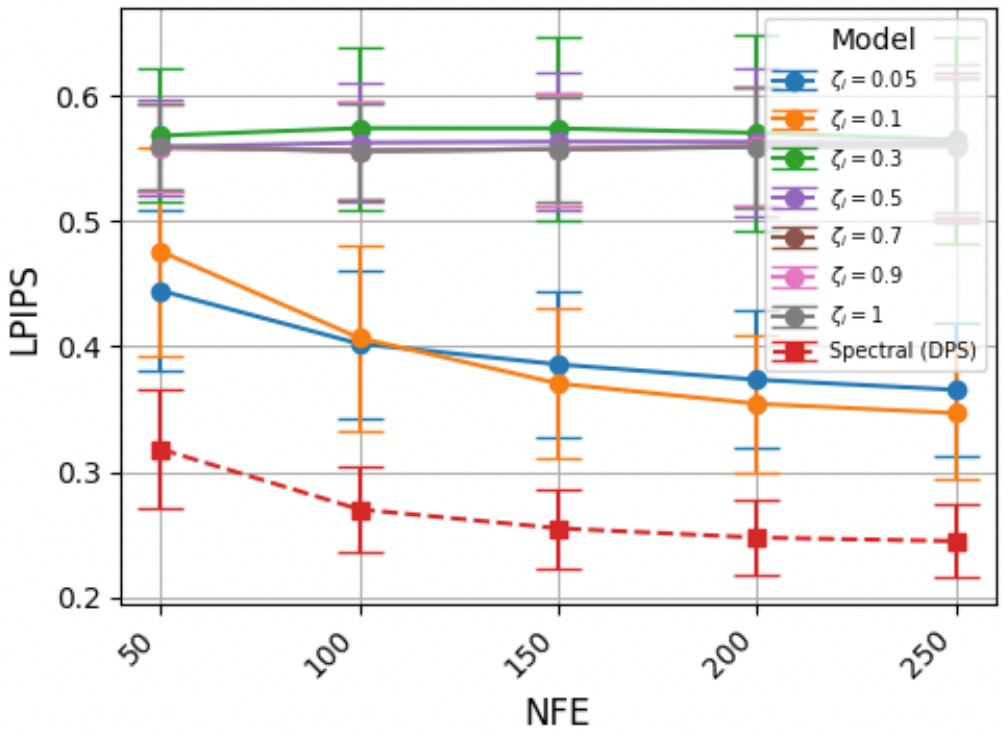}
        \caption{LPIPS $\downarrow$}
        \label{fig:fmri_lps_lpips}
    \end{subfigure}

    \vspace{0.5em}

    \begin{subfigure}[b]{0.48\textwidth}
        \centering
        \includegraphics[width=\textwidth]{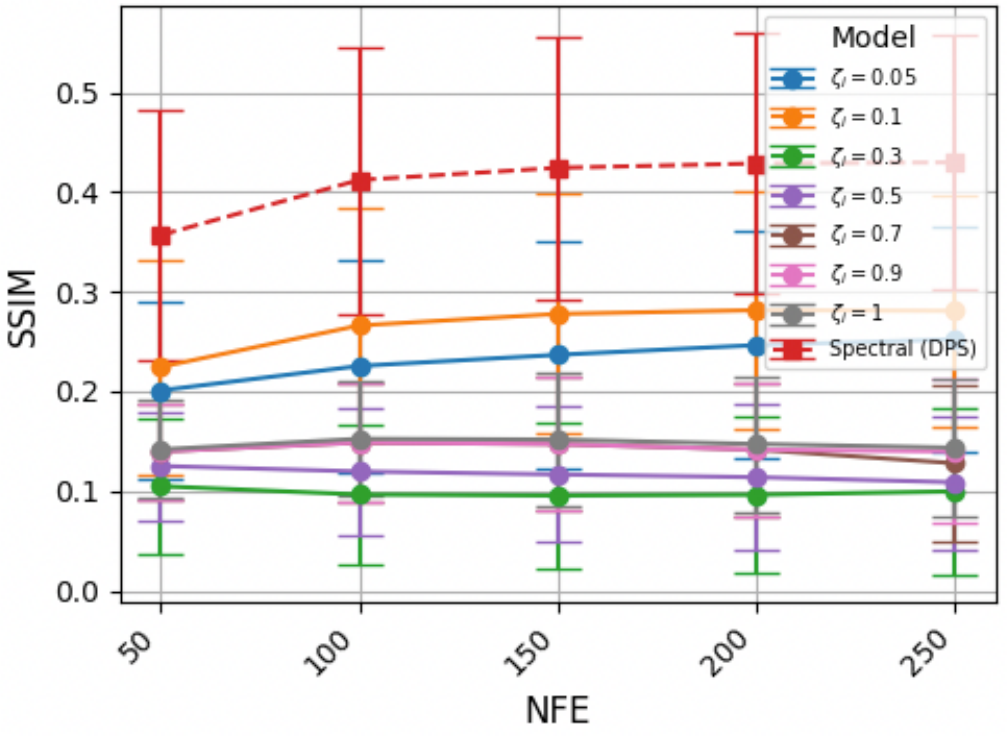}
        \caption{SSIM $\uparrow$}
        \label{fig:fmri_lps_ssim}
    \end{subfigure}
    \hfill
    \begin{subfigure}[b]{0.48\textwidth}
        \centering
        \includegraphics[width=\textwidth]{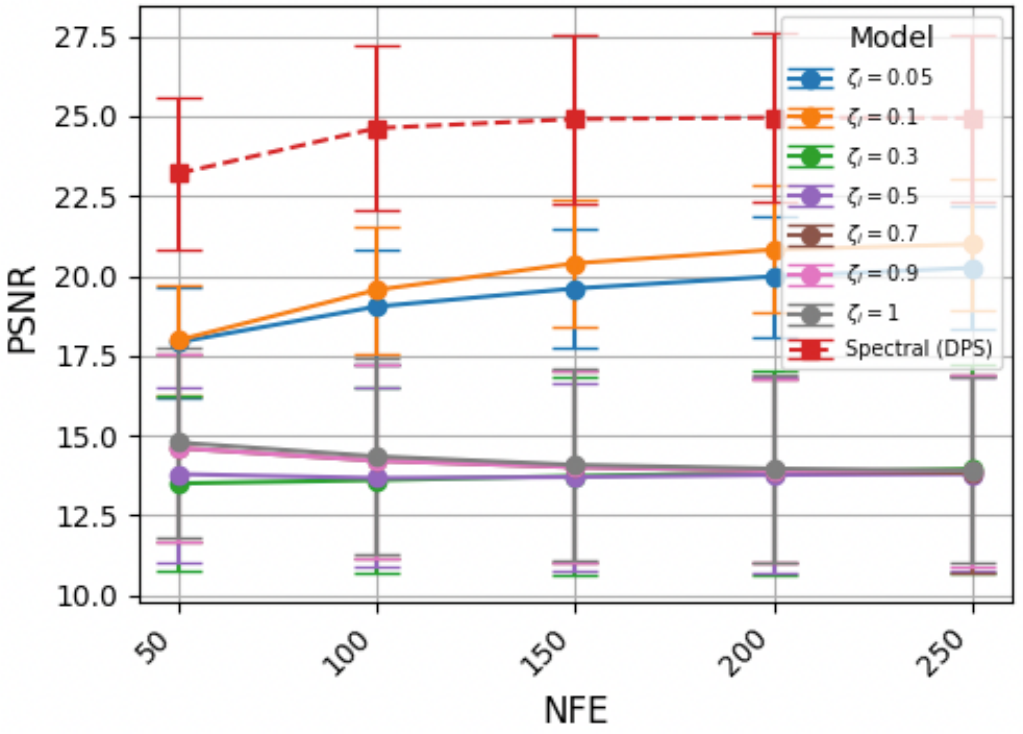}
        \caption{PSNR $\uparrow$}
        \label{fig:fmri_lps_psnr}
    \end{subfigure}

\caption{\textbf{LPF degradation.} Comparison between DPS heuristic configurations with $\zeta'=\{0.05,0.1,0.3,0.5,0.7,0.9,1\}$ and the proposed spectral recommendations under LPF degradation with $\mathcal{V}=0.1$ and additive Gaussian noise $\sigma_y=0.1$, following the setting in Section~\ref{sec:Empirical_Distribution}. Results are evaluated across the number of function evaluations (NFE) on the fastMRI dataset.}
    \label{fig:fmri_lps_comprehenssive_comparison}
\end{figure}

Figure~\ref{fig:fmri_lps_comprehenssive_comparison} compares the proposed spectral recommendations with multiple DPS heuristic configurations, $\zeta'\in\{0.05,0.1,0.3,0.5,0.7,0.9,1\}$, on the fastMRI validation set described in Section~\ref{sec:Dataset_Preparation}, under the LPF degradation setting introduced in Section~\ref{sec:Empirical_Distribution}. The spectral recommendations consistently outperform the heuristic configurations across both perceptual and distortion metrics. In particular, noticeable improvements are observed in PSNR, reflecting stronger adherence to the underlying clean scans.


We now evaluate additional degradation operators. Table~\ref{tab:fastmri_results_R} presents results on the fastMRI validation set, described in Section~\ref{sec:Dataset_Preparation}, under horizontal frequency subsampling degradation with acceleration factors $R\in\{4,8,12\}$. For brevity, for each acceleration factor we report only the best-performing DPS heuristic configuration, corresponding to $\zeta'=0.1$, selected from the range $\zeta'\in\{0.05,0.1,0.3,0.5,0.7,0.9,1\}$.

It can be seen that across all acceleration factors, the spectral recommendations consistently outperform the heuristic configurations in terms of adherence to the original signal $\x_0$. This property is particularly relevant in medical imaging, where, alongside perceptual quality, accurate reconstruction of the underlying scan is required in order to avoid introducing artificial structures that may lead to incorrect interpretation. In terms of perceptual quality, the spectral recommendations also perform well for lower acceleration factors. For the more challenging setting of $R=12$, however, the selected heuristic configuration achieves slightly improved FID values for a small number of diffusion steps. This observation further highlights the trade-off between perceptual naturalness and reconstruction fidelity. In particular, under severe degradations, improved PSNR values may be preferable to lower FID scores, since visually plausible scans that deviate from the underlying signal are undesirable in medical reconstruction tasks.

\begin{table*}[h]
\caption{
Quantitative comparison between the proposed spectral recommendations and the DPS heuristic on the fastMRI dataset under accelerated MRI acquisition with acceleration factors $R\in\{4,8,12\}$, using PSNR, SSIM, LPIPS, and FID.
}
\label{tab:fastmri_results_R}
\centering
\begin{small}
\begin{sc}
\setlength{\tabcolsep}{4.5pt}
\begin{tabular}{c c c cccc}
\toprule
$R$ & Method & Steps &
PSNR$\uparrow$ & SSIM$\uparrow$ & LPIPS$\downarrow$ & FID$\downarrow$ \\
\midrule

\multirow{8}{*}{4}
& \multirow{4}{*}{Spectral}
& $50$  & $\mathbf{21.66}$ & $\mathbf{0.31}$ & $\mathbf{0.37}$ & $\mathbf{107.17}$ \\
& & $100$ & $\mathbf{23.44}$ & $\mathbf{0.40}$ & $\mathbf{0.31}$ & $\mathbf{75.05}$ \\
& & $150$ & $\mathbf{24.08}$ & $\mathbf{0.44}$ & $\mathbf{0.28}$ & $\mathbf{65.66}$ \\
& & $200$ & $\mathbf{24.44}$ & $\mathbf{0.46}$ & $\mathbf{0.27}$ & $\mathbf{61.591}$ \\
\cmidrule(lr){2-7}
& \multirow{4}{*}{DPS}
& $50$  & $18.32$ & $0.26$ & $0.46$ & $110.15$ \\
& & $100$ & $19.78$ & $0.27$ & $0.38$ & $75.06$ \\
& & $150$ & $20.29$ & $0.27$ & $0.35$ & $68.59$ \\
& & $200$ & $20.57$ & $0.27$ & $0.34$ & $67.56$ \\

\midrule

\multirow{8}{*}{8}
& \multirow{4}{*}{Spectral}
& $50$  & $\mathbf{19.21}$ & $\mathbf{0.23}$ & $\mathbf{0.44}$ & ${143.15}$ \\
& & $100$ & $\mathbf{20.61}$ & $\mathbf{0.29}$ & $\mathbf{0.38}$ & ${103.48}$ \\
& & $150$ & $\mathbf{21.25}$ & $\mathbf{0.32}$ & $\mathbf{0.36}$ & $\mathbf{86.19}$ \\
& & $200$ & $\mathbf{21.75}$ & $\mathbf{0.34}$ & $\mathbf{0.34}$ & $\mathbf{81.991}$ \\
\cmidrule(lr){2-7}
& \multirow{4}{*}{DPS}
& $50$  & $17.62$ & $0.23$ & $0.49$ & $\mathbf{120.27}$ \\
& & $100$ & $19.06$ & $0.27$ & $0.42$ & $\mathbf{102.31}$ \\
& & $150$ & $19.79$ & $0.27$ & $0.38$ & $88.74$ \\
& & $200$ & $19.99$ & $0.26$ & $0.36$ & $83.13$ \\

\midrule

\multirow{8}{*}{12}
& \multirow{4}{*}{Spectral}
& $50$  & $\mathbf{17.57}$ & $\mathbf{0.21}$ & $0.47$ & $142.00$ \\
& & $100$ & $\mathbf{18.82}$ & $\mathbf{0.25}$ & $\mathbf{0.42}$ & $119.50$ \\
& & $150$ & $\mathbf{19.31}$ & $\underline{0.26}$ & $\mathbf{0.40}$ & $\mathbf{95.15}$ \\
& & $200$ & $\mathbf{19.61}$ & $\underline{0.27}$ & $\mathbf{0.39}$ & $\mathbf{90.55}$ \\
\cmidrule(lr){2-7}
& \multirow{4}{*}{DPS}
& $50$  & $17.10$ & $0.20$ & $\mathbf{0.50}$ & $\mathbf{129.92}$ \\
& & $100$ & $18.26$ & $0.24$ & $0.46$ & $\mathbf{114.59}$ \\
& & $150$ & $18.96$ & $\underline{0.26}$ & $0.42$ & ${96.63}$ \\
& & $200$ & $19.27$ & $\underline{0.27}$ & $0.40$ & ${92.89}$ \\

\bottomrule
\end{tabular}
\end{sc}
\end{small}
\end{table*}

\clearpage
\subsection{Additional visual results}



\begin{figure*}[h]

    \centering
    \setlength{\tabcolsep}{2pt}
    \renewcommand{\arraystretch}{1}
    \begin{tabular}{cc|ccc|ccc}
        \toprule
        Reference & Measurement
        & \multicolumn{3}{c}{DPS Heuristic}
        & \multicolumn{3}{c}{Spectral Weights} \\

        \midrule


        \includegraphics[width=0.1\textwidth]{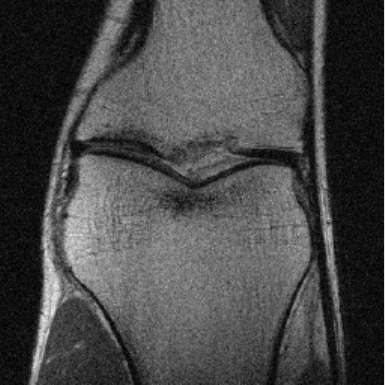}
        & \includegraphics[width=0.1\textwidth]{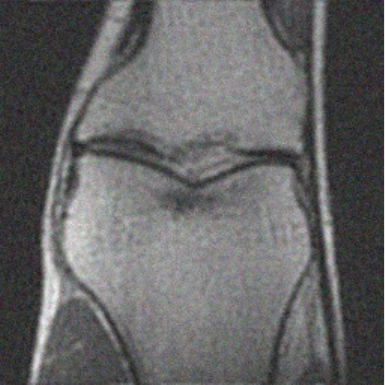}
        & \includegraphics[width=0.1\textwidth]{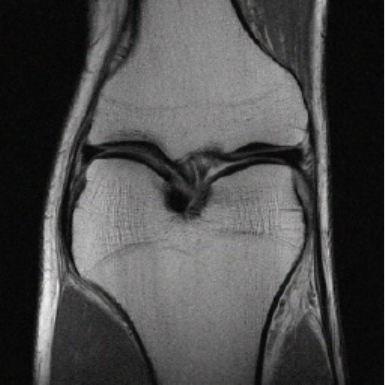}
        & \includegraphics[width=0.1\textwidth]{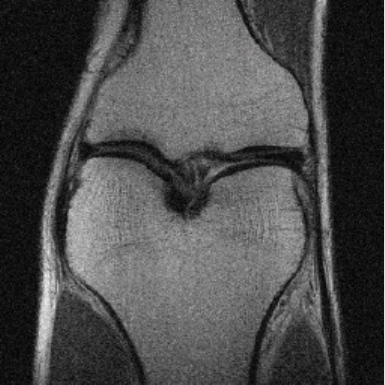}
        & \includegraphics[width=0.1\textwidth]{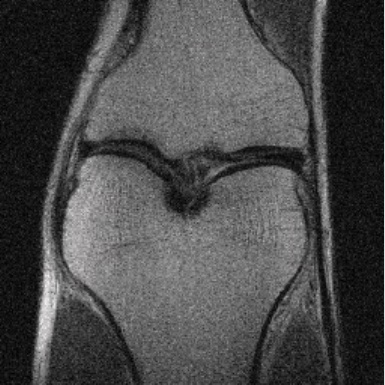}
        & \includegraphics[width=0.1\textwidth]{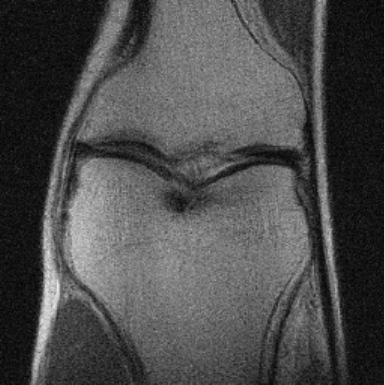}
        & \includegraphics[width=0.1\textwidth]{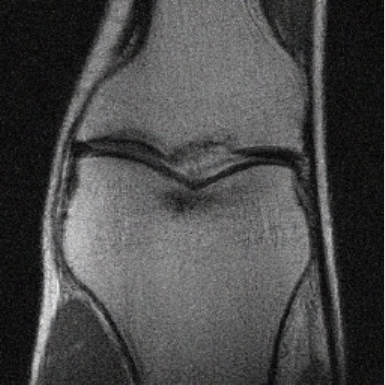}
        & \includegraphics[width=0.1\textwidth]{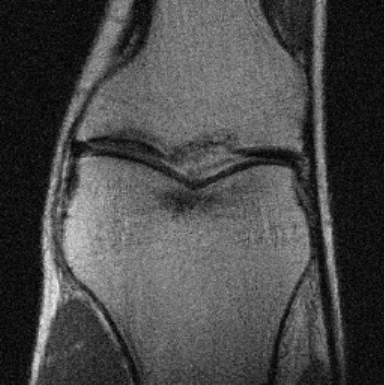}\\

        \includegraphics[width=0.1\textwidth]{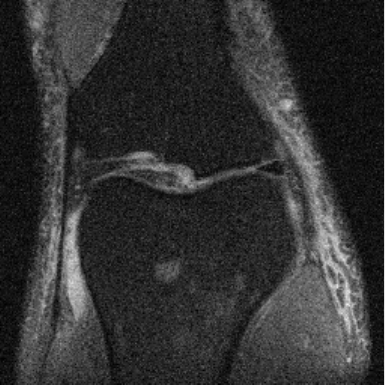}
        & \includegraphics[width=0.1\textwidth]{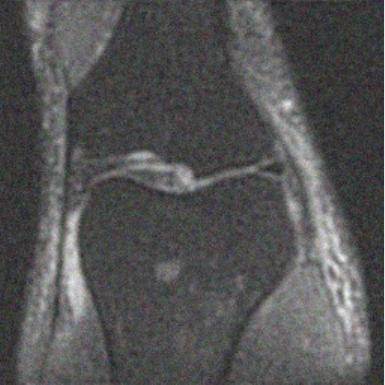}
        & \includegraphics[width=0.1\textwidth]{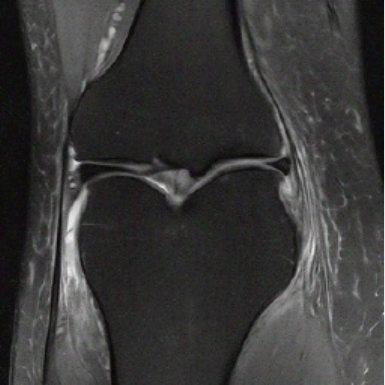}
        & \includegraphics[width=0.1\textwidth]{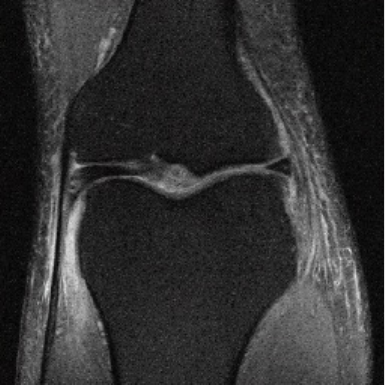}
        & \includegraphics[width=0.1\textwidth]{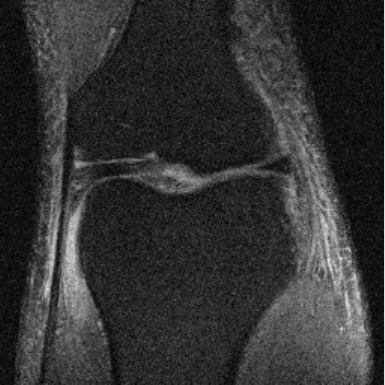}
        & \includegraphics[width=0.1\textwidth]{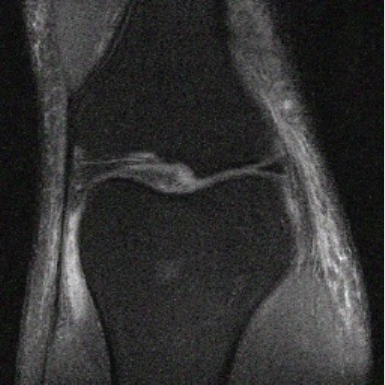}
        & \includegraphics[width=0.1\textwidth]{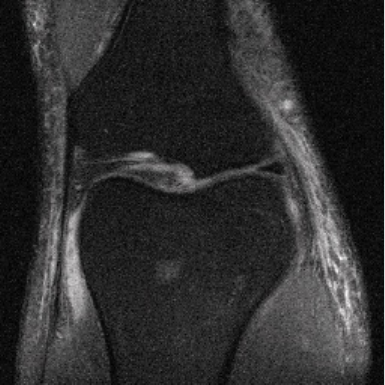}
        & \includegraphics[width=0.1\textwidth]{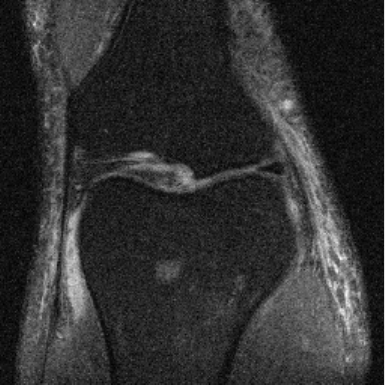} \\

        \includegraphics[width=0.1\textwidth]{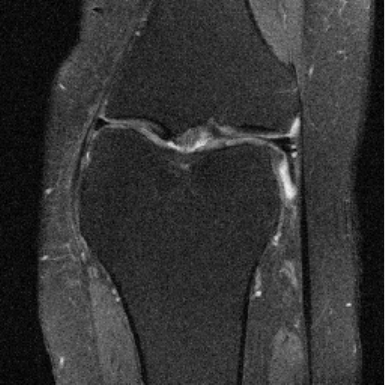}
        & \includegraphics[width=0.1\textwidth]{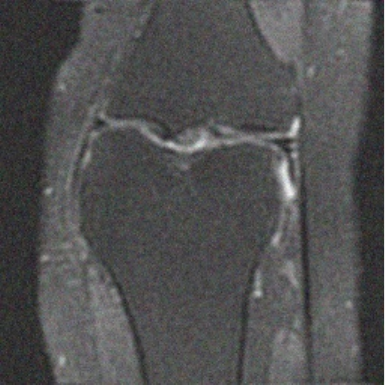}
        & \includegraphics[width=0.1\textwidth]{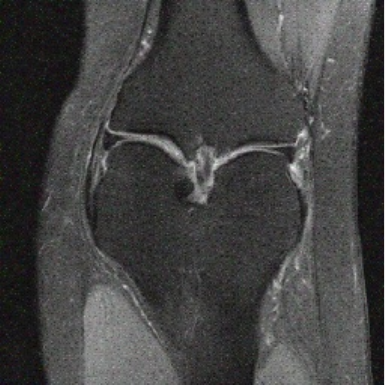}
        & \includegraphics[width=0.1\textwidth]{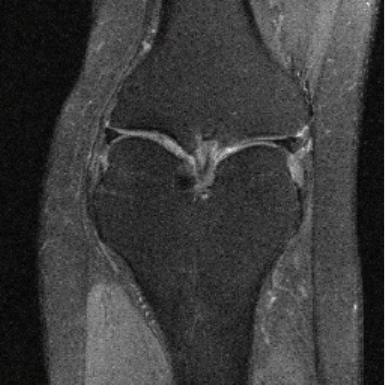}
        & \includegraphics[width=0.1\textwidth]{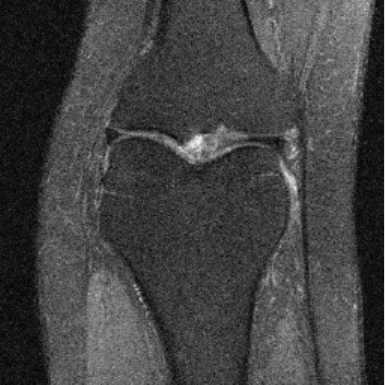}
         & \includegraphics[width=0.1\textwidth]{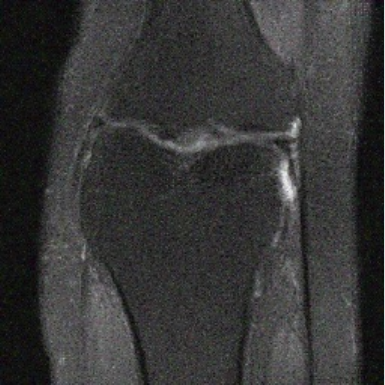}
        & \includegraphics[width=0.1\textwidth]{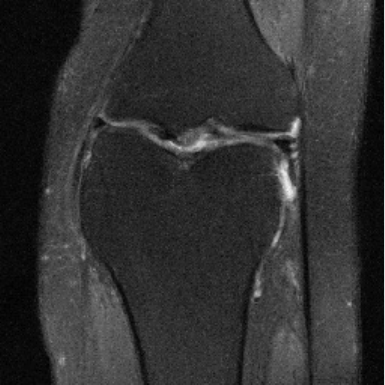}
        & \includegraphics[width=0.1\textwidth]{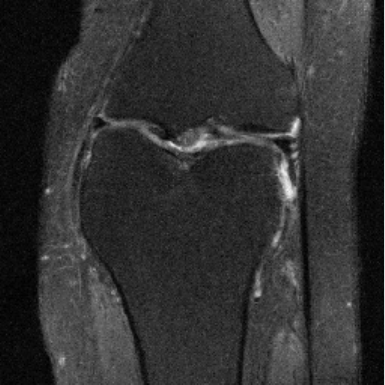} \\

        \includegraphics[width=0.1\textwidth]{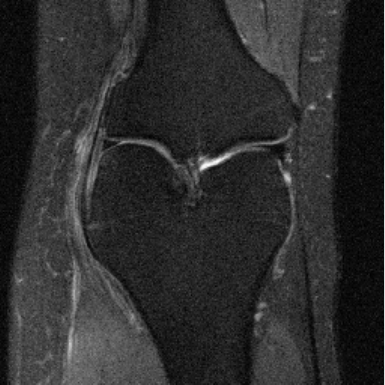}
        & \includegraphics[width=0.1\textwidth]{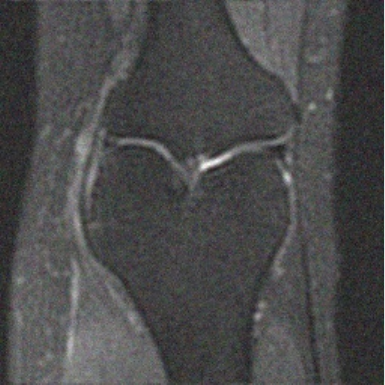}
       & \includegraphics[width=0.1\textwidth]{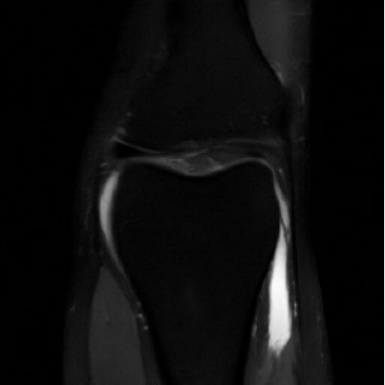}
        & \includegraphics[width=0.1\textwidth]{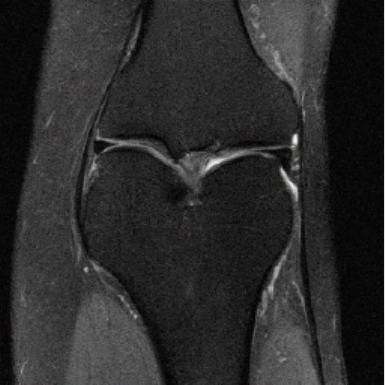}
        & \includegraphics[width=0.1\textwidth]{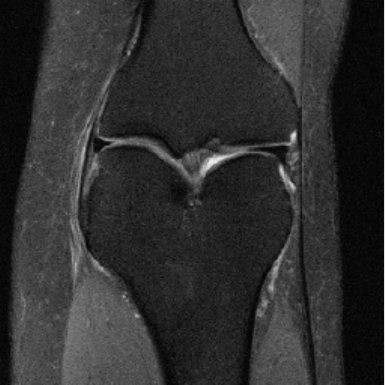}
         & \includegraphics[width=0.1\textwidth]{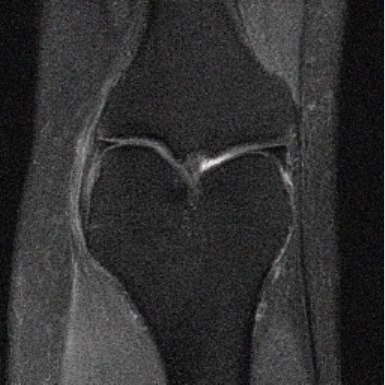}
        & \includegraphics[width=0.1\textwidth]{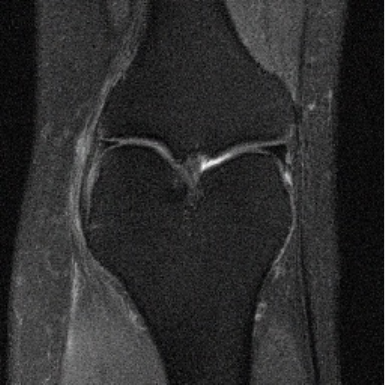}
        & \includegraphics[width=0.1\textwidth]{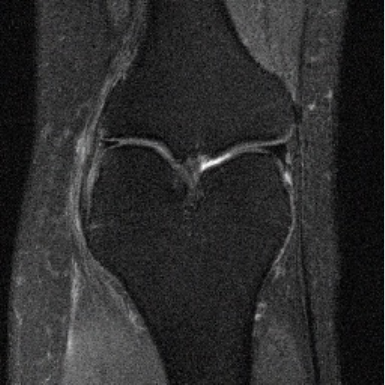} \\

    \end{tabular}

\caption{Qualitative comparison of reconstructions on the fastMRI dataset under LPF degradation, comparing the DPS heuristic ($\zeta'=0.1$) and the proposed spectral recommendations.}
        \label{fig:fastMRI_LPF}
\end{figure*}

\begin{figure*}[h]

    \centering
    \setlength{\tabcolsep}{2pt}
    \renewcommand{\arraystretch}{1}
    \begin{tabular}{cc|ccc|ccc}
        \toprule
        Reference & Measurement
        & \multicolumn{3}{c}{DPS Heuristic}
        & \multicolumn{3}{c}{Spectral Weights} \\

        \midrule


        \includegraphics[width=0.1\textwidth]{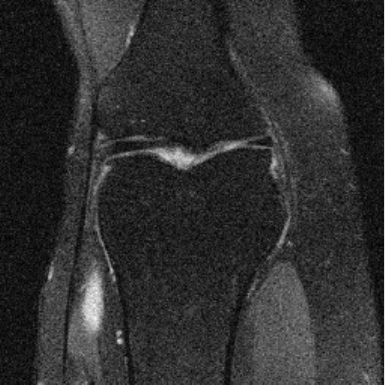}
        & \includegraphics[width=0.1\textwidth]{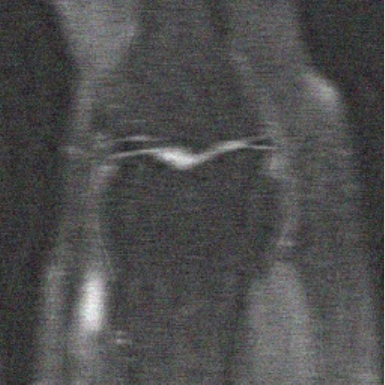}
        & \includegraphics[width=0.1\textwidth]{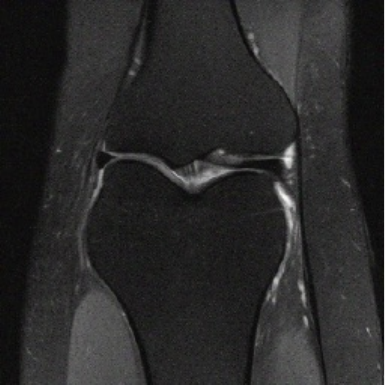}
        & \includegraphics[width=0.1\textwidth]{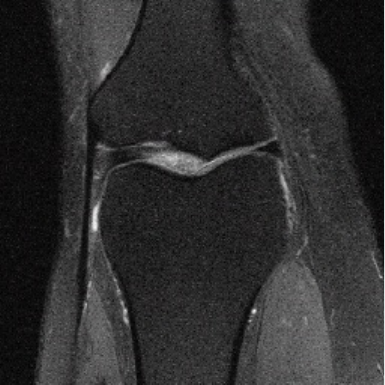}
        & \includegraphics[width=0.1\textwidth]{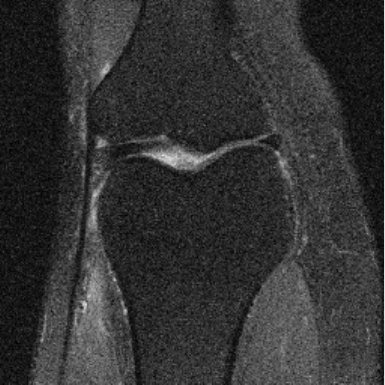}
        & \includegraphics[width=0.1\textwidth]{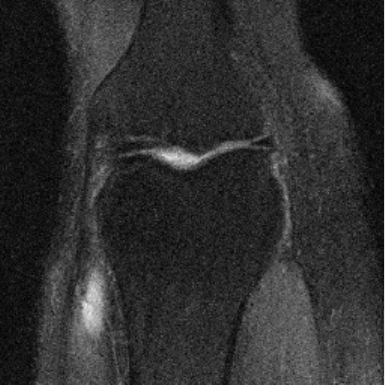}
        & \includegraphics[width=0.1\textwidth]{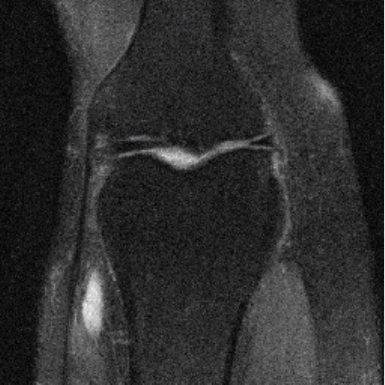}
        & \includegraphics[width=0.1\textwidth]{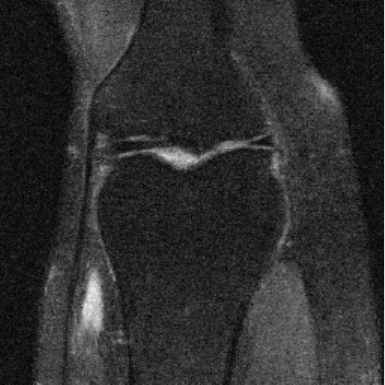}\\

        \includegraphics[width=0.1\textwidth]{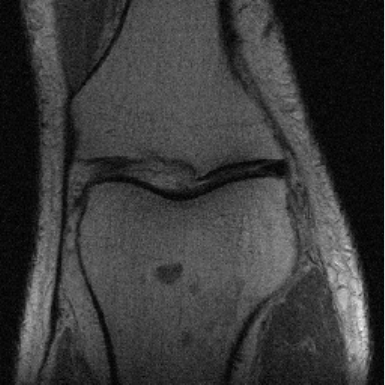}
        & \includegraphics[width=0.1\textwidth]{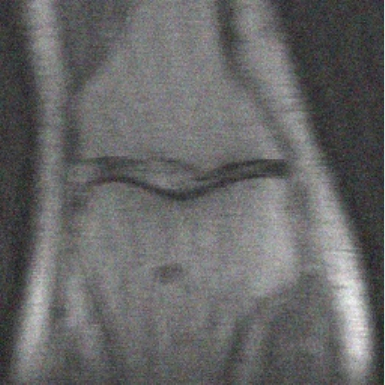}
        & \includegraphics[width=0.1\textwidth]{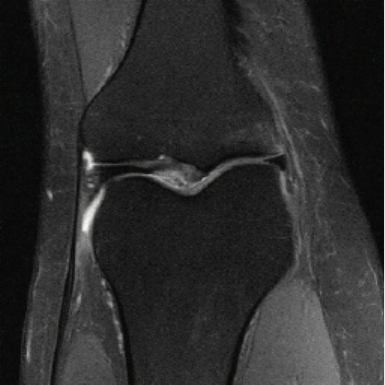}
        & \includegraphics[width=0.1\textwidth]{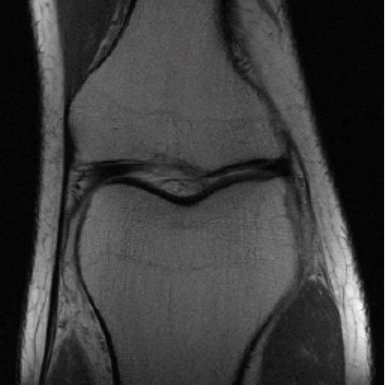}
        & \includegraphics[width=0.1\textwidth]{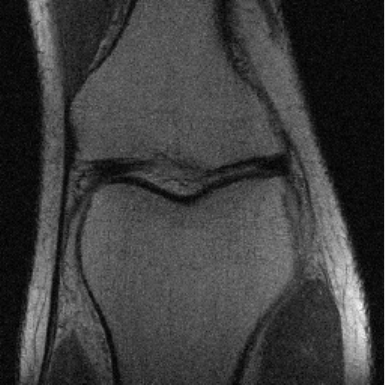}
        & \includegraphics[width=0.1\textwidth]{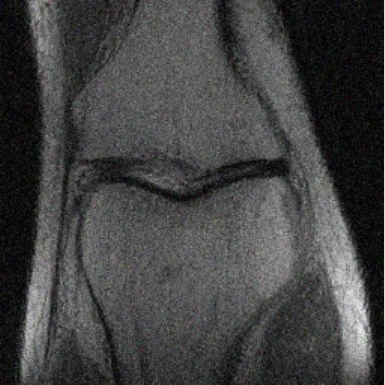}
        & \includegraphics[width=0.1\textwidth]{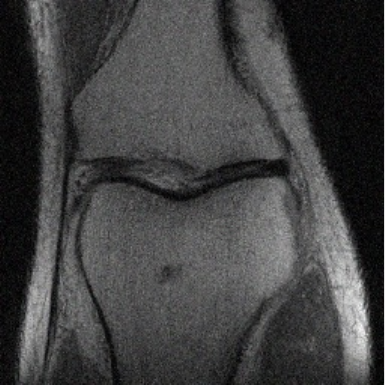}
        & \includegraphics[width=0.1\textwidth]{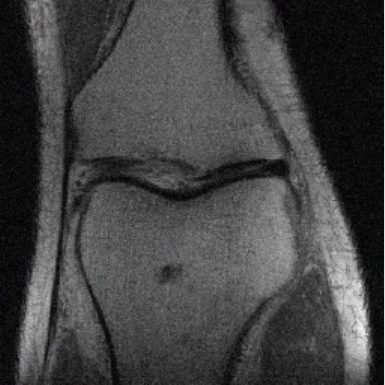}\\

        \includegraphics[width=0.1\textwidth]{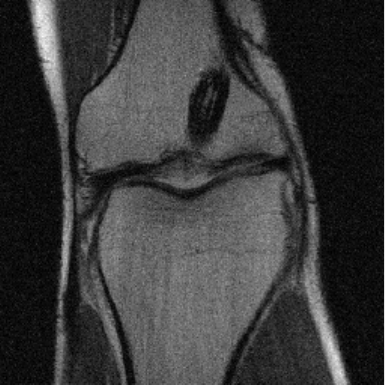}
        & \includegraphics[width=0.1\textwidth]{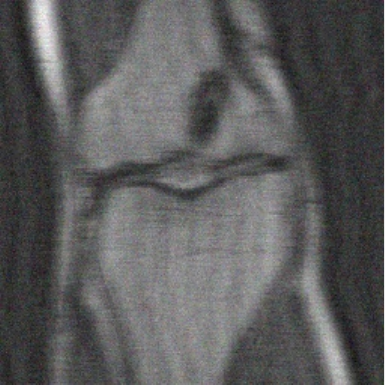}
        & \includegraphics[width=0.1\textwidth]{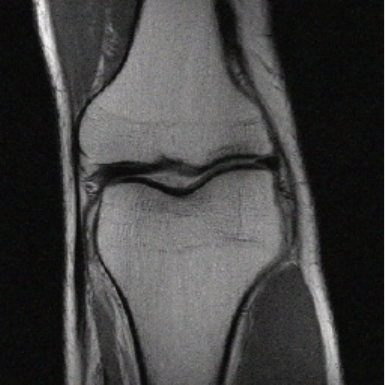}
        & \includegraphics[width=0.1\textwidth]{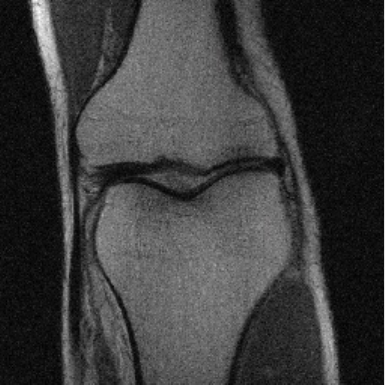}
        & \includegraphics[width=0.1\textwidth]{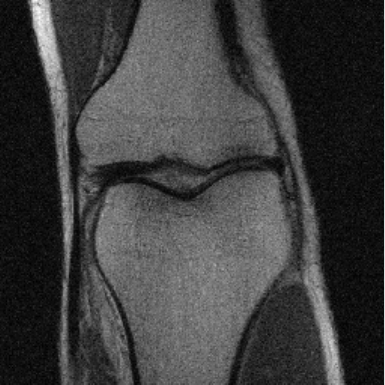}
        & \includegraphics[width=0.1\textwidth]{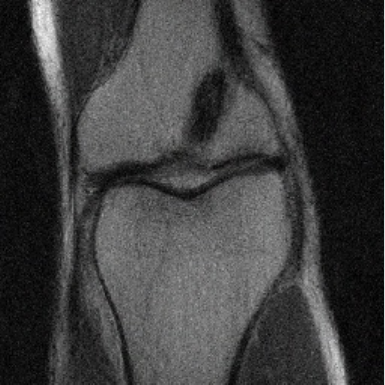}
        & \includegraphics[width=0.1\textwidth]{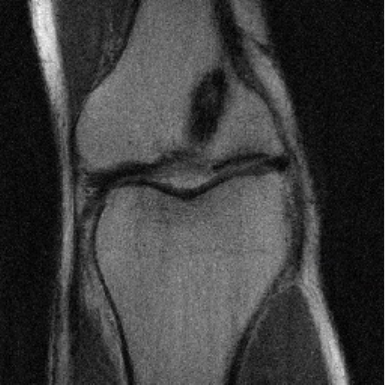}
        & \includegraphics[width=0.1\textwidth]{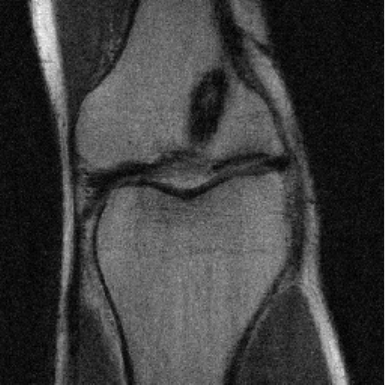}\\

         \includegraphics[width=0.1\textwidth]{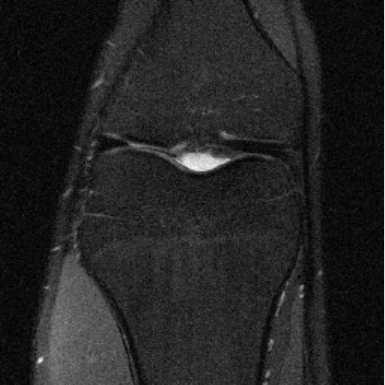}
        & \includegraphics[width=0.1\textwidth]{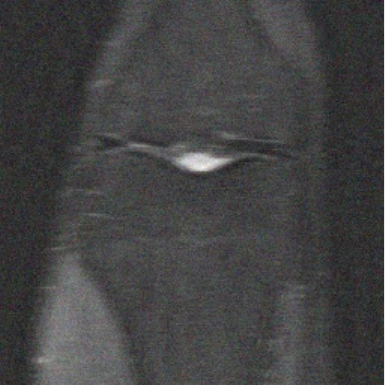}
        & \includegraphics[width=0.1\textwidth]{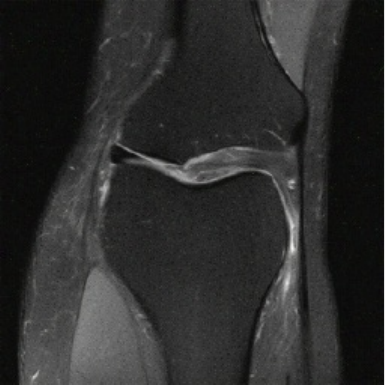}
        & \includegraphics[width=0.1\textwidth]{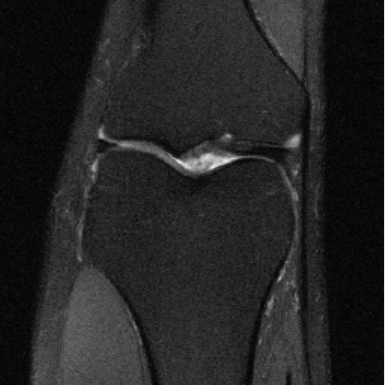}
        & \includegraphics[width=0.1\textwidth]{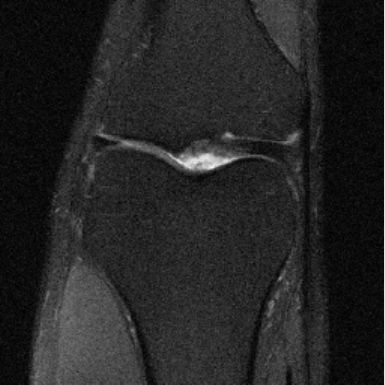}
        & \includegraphics[width=0.1\textwidth]{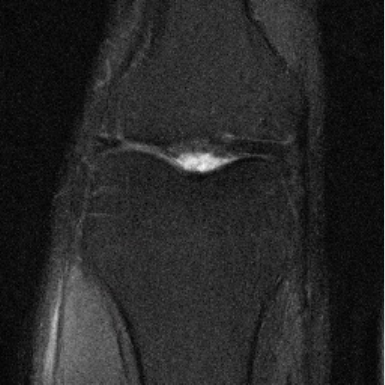}
        & \includegraphics[width=0.1\textwidth]{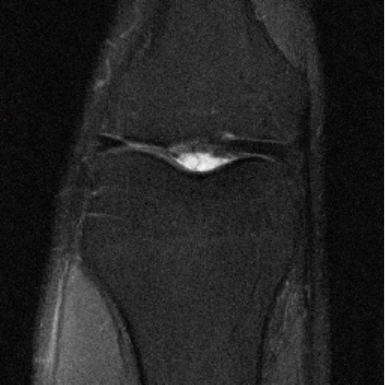}
        & \includegraphics[width=0.1\textwidth]{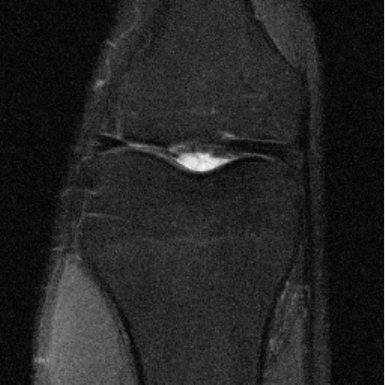}\\

    \end{tabular}

\caption{Qualitative comparison of reconstructions on the fastMRI dataset under accelerated MRI acquisition with $R=4$, comparing the DPS heuristic ($\zeta'=0.1$) and the proposed spectral recommendations.}
        \label{fig:fastMRI_R_4}
\end{figure*}

\begin{figure*}[h]

    \centering
    \setlength{\tabcolsep}{2pt}
    \renewcommand{\arraystretch}{1}
    \begin{tabular}{cc|ccc|ccc}
        \toprule
        Reference & Measurement
        & \multicolumn{3}{c}{DPS Heuristic}
        & \multicolumn{3}{c}{Spectral Weights} \\

        \midrule

        \includegraphics[width=0.1\textwidth]{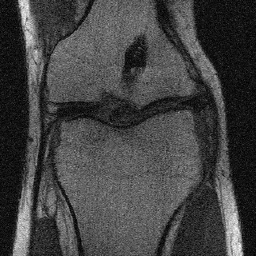}
        & \includegraphics[width=0.1\textwidth]{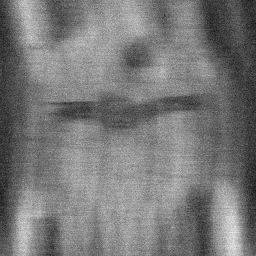}
        & \includegraphics[width=0.1\textwidth]{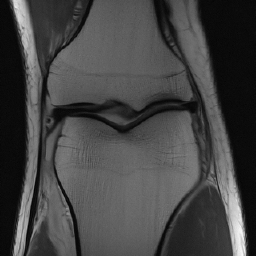}
        & \includegraphics[width=0.1\textwidth]{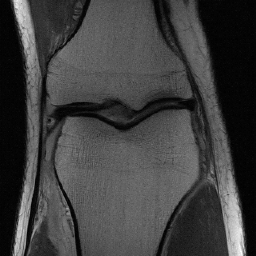}
        & \includegraphics[width=0.1\textwidth]{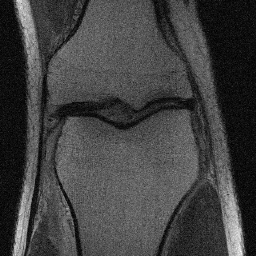}
        & \includegraphics[width=0.1\textwidth]{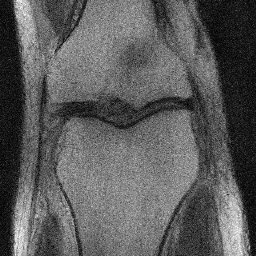}
        & \includegraphics[width=0.1\textwidth]{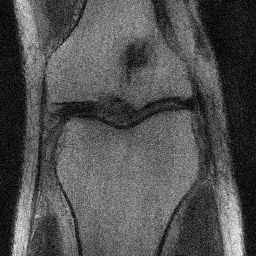}
        & \includegraphics[width=0.1\textwidth]{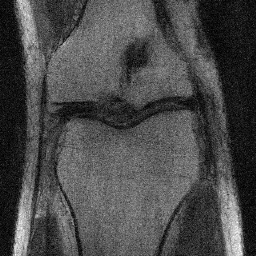}\\

        \includegraphics[width=0.1\textwidth]{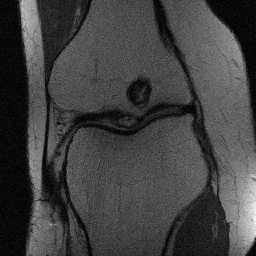}
        & \includegraphics[width=0.1\textwidth]{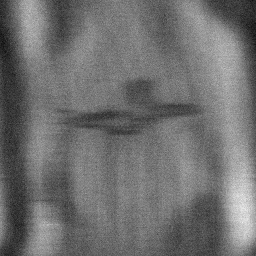}
        & \includegraphics[width=0.1\textwidth]{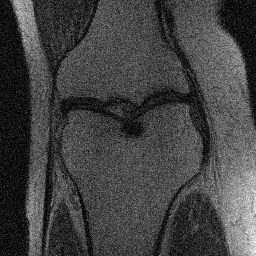}
        & \includegraphics[width=0.1\textwidth]{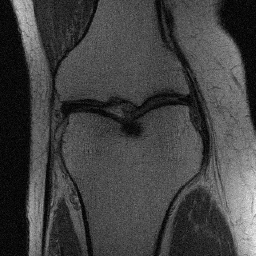}
        & \includegraphics[width=0.1\textwidth]{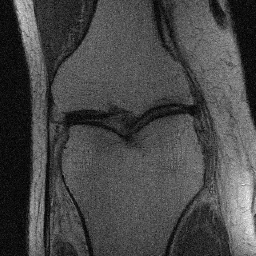}
        & \includegraphics[width=0.1\textwidth]{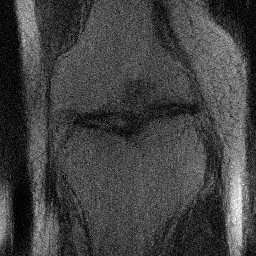}
        & \includegraphics[width=0.1\textwidth]{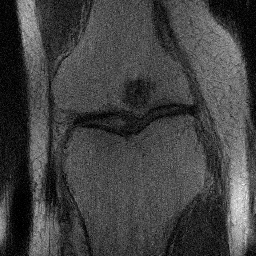}
        & \includegraphics[width=0.1\textwidth]{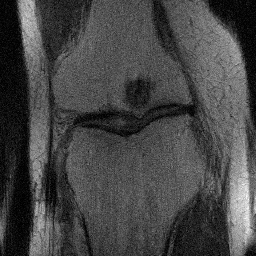}\\

        \includegraphics[width=0.1\textwidth]{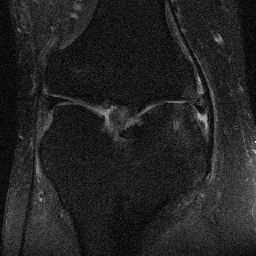}
        & \includegraphics[width=0.1\textwidth]{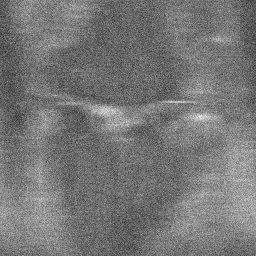}
        & \includegraphics[width=0.1\textwidth]{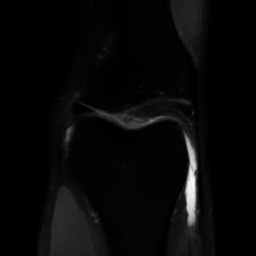}
        & \includegraphics[width=0.1\textwidth]{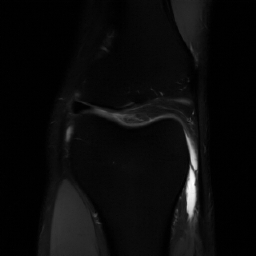}
        & \includegraphics[width=0.1\textwidth]{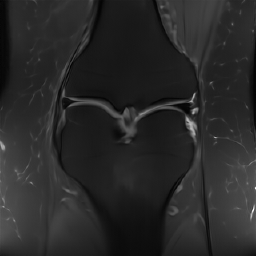}
        & \includegraphics[width=0.1\textwidth]{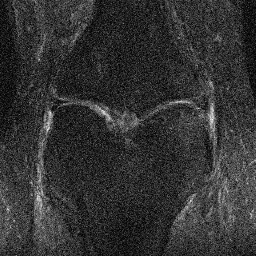}
        & \includegraphics[width=0.1\textwidth]{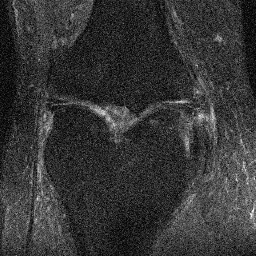}
        & \includegraphics[width=0.1\textwidth]{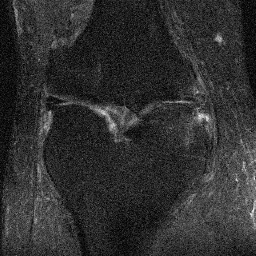}\\

    \end{tabular}

\caption{Qualitative comparison of reconstructions on the fastMRI dataset under accelerated MRI acquisition with $R=8$, comparing the DPS heuristic ($\zeta'=0.1$) and the proposed spectral recommendations.}
        \label{fig:fastMRI_R_8}
\end{figure*}

\begin{figure*}[h]

    \centering
    \setlength{\tabcolsep}{2pt}
    \renewcommand{\arraystretch}{1}
    \begin{tabular}{cc|ccc|ccc}
        \toprule
        Reference & Measurement
        & \multicolumn{3}{c}{DPS Heuristic}
        & \multicolumn{3}{c}{Spectral Weights} \\

        \midrule


        \includegraphics[width=0.1\textwidth]{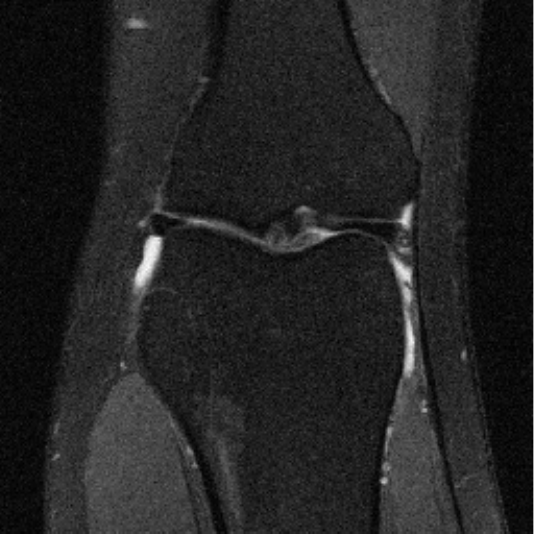}
        & \includegraphics[width=0.1\textwidth]{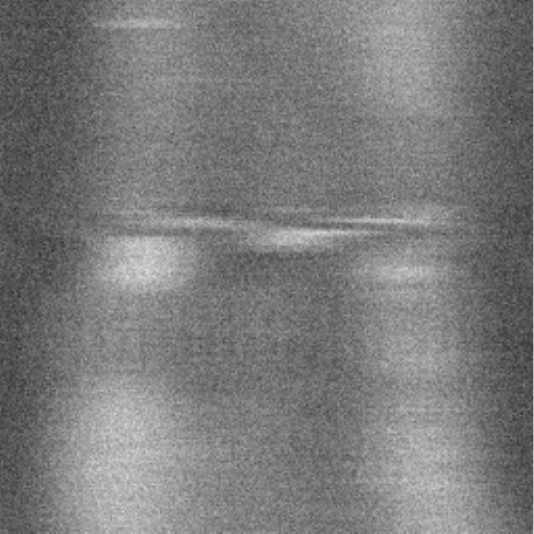}
        & \includegraphics[width=0.1\textwidth]{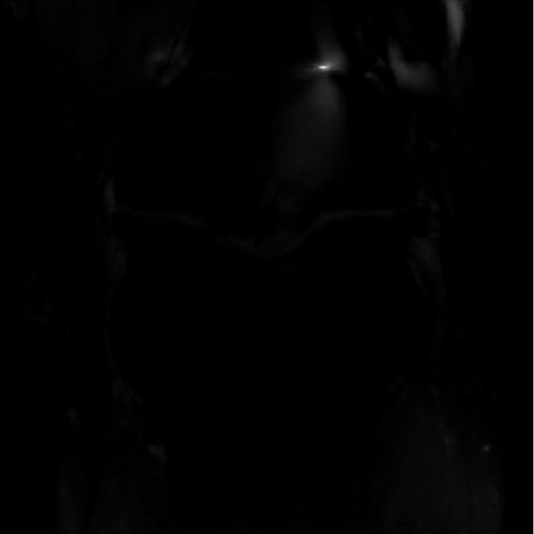}
        & \includegraphics[width=0.1\textwidth]{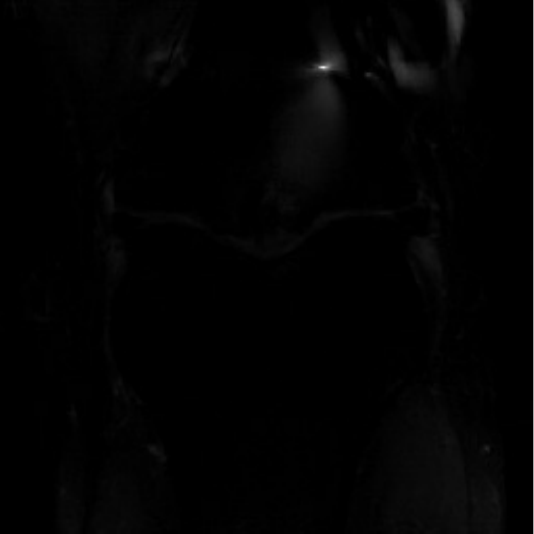}
        & \includegraphics[width=0.1\textwidth]{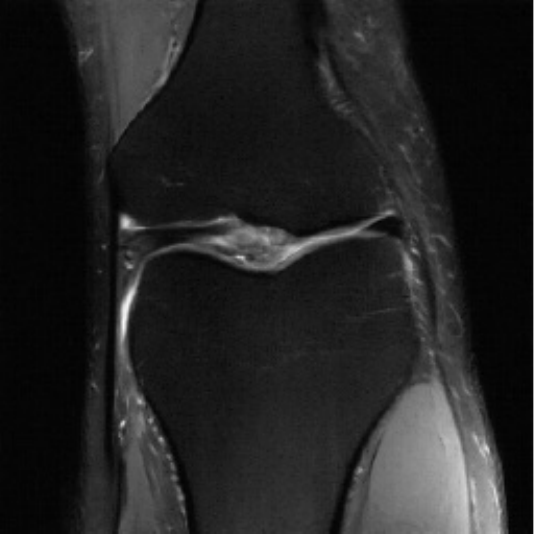}
        & \includegraphics[width=0.1\textwidth]{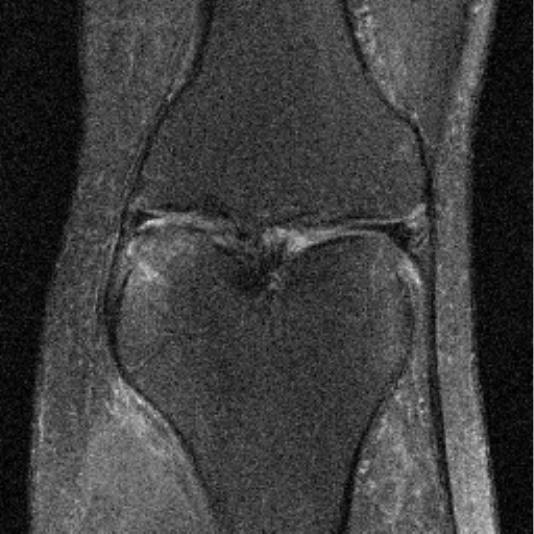}
        & \includegraphics[width=0.1\textwidth]{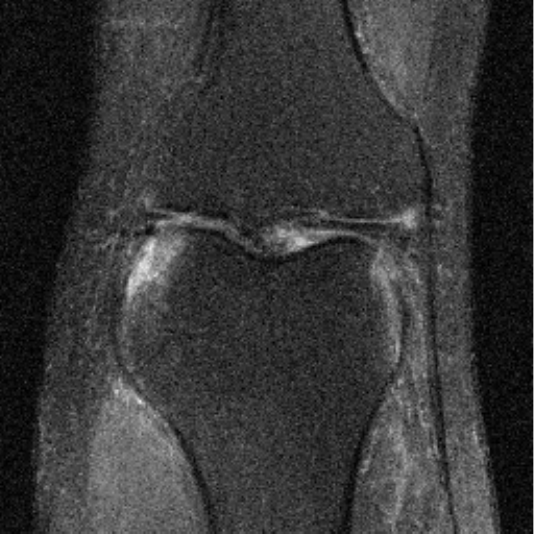}
        & \includegraphics[width=0.1\textwidth]{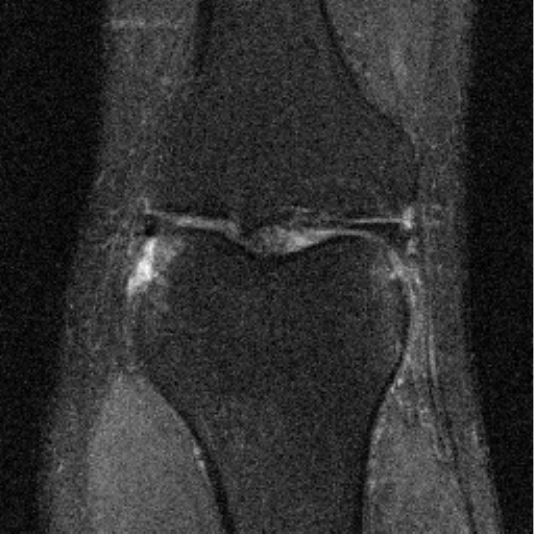}\\

          \includegraphics[width=0.1\textwidth]{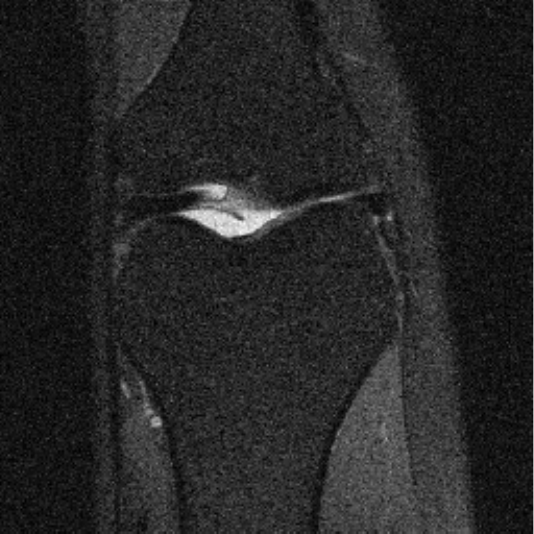}
        & \includegraphics[width=0.1\textwidth]{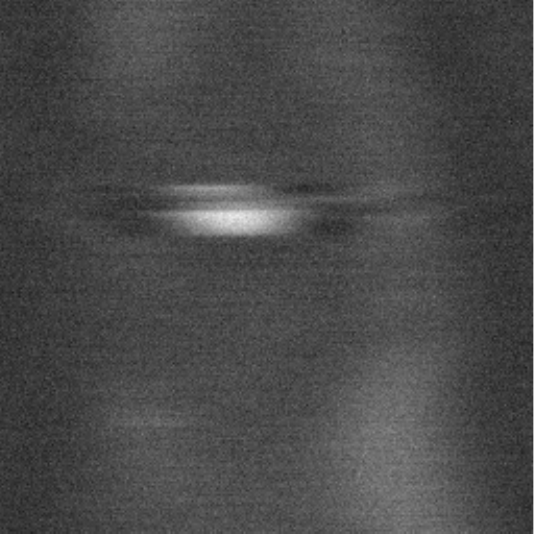}
        & \includegraphics[width=0.1\textwidth]{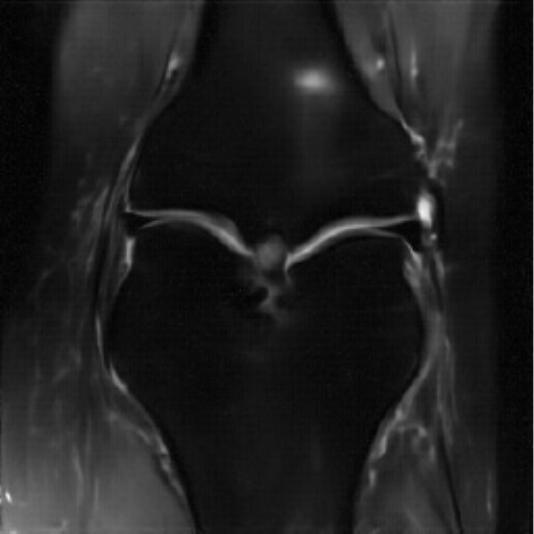}
        & \includegraphics[width=0.1\textwidth]{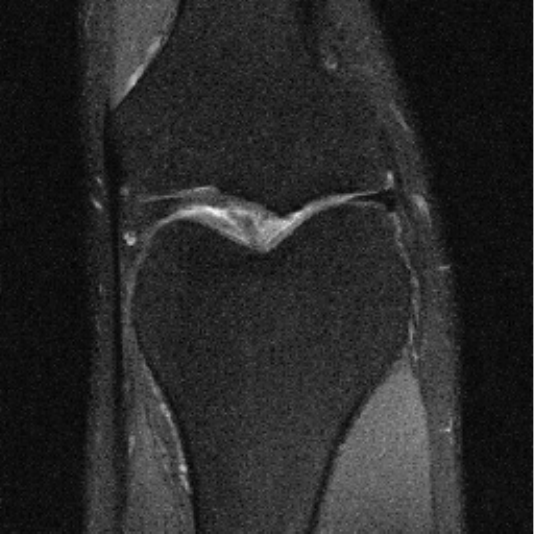}
        & \includegraphics[width=0.1\textwidth]{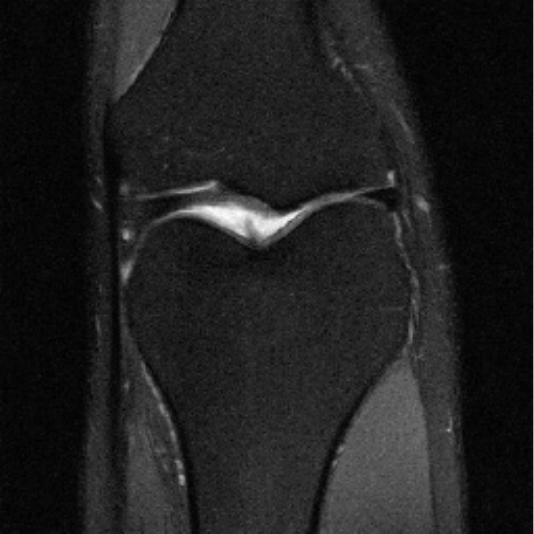}
        & \includegraphics[width=0.1\textwidth]{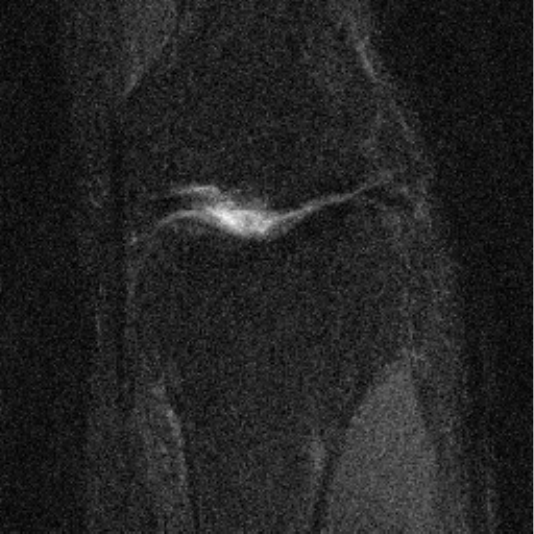}
        & \includegraphics[width=0.1\textwidth]{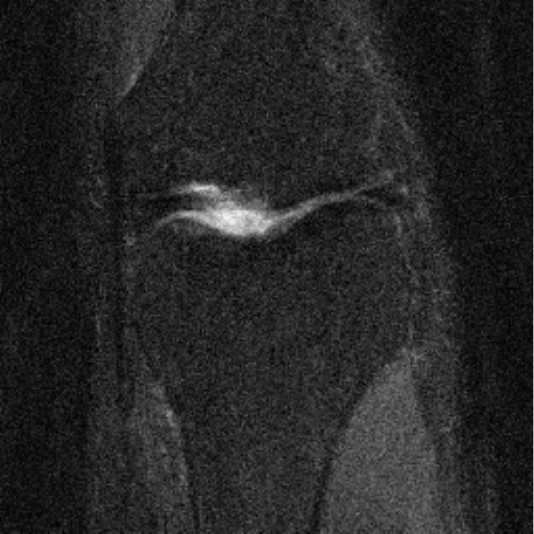}
        & \includegraphics[width=0.1\textwidth]{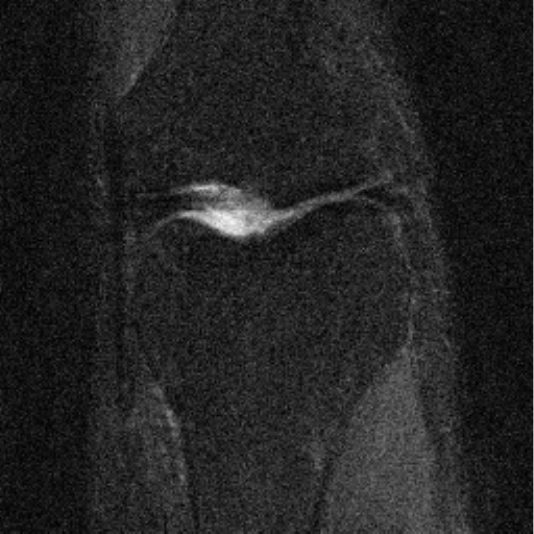}\\

        \includegraphics[width=0.1\textwidth]{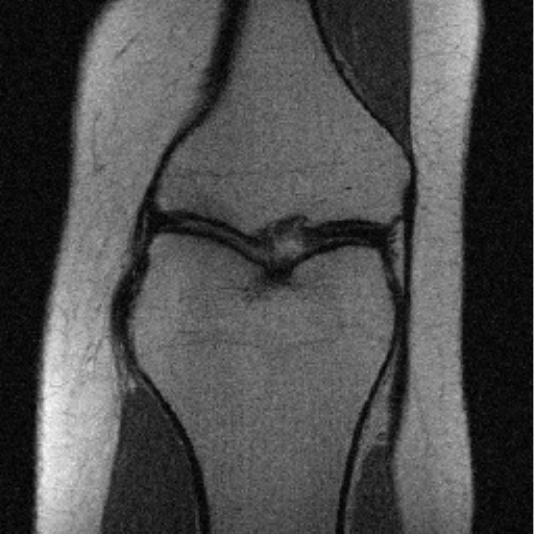}
        & \includegraphics[width=0.1\textwidth]{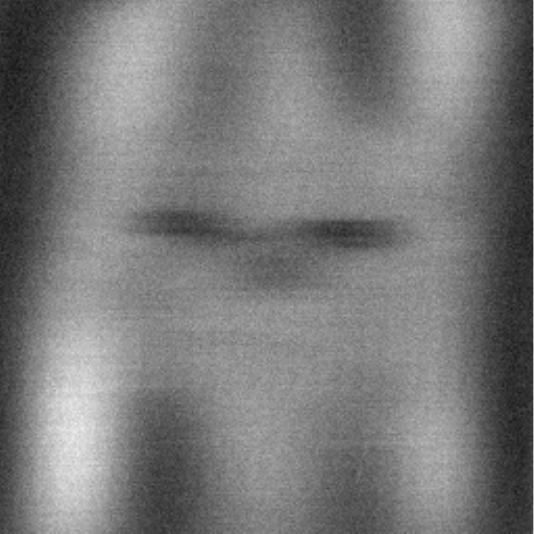}
        & \includegraphics[width=0.1\textwidth]{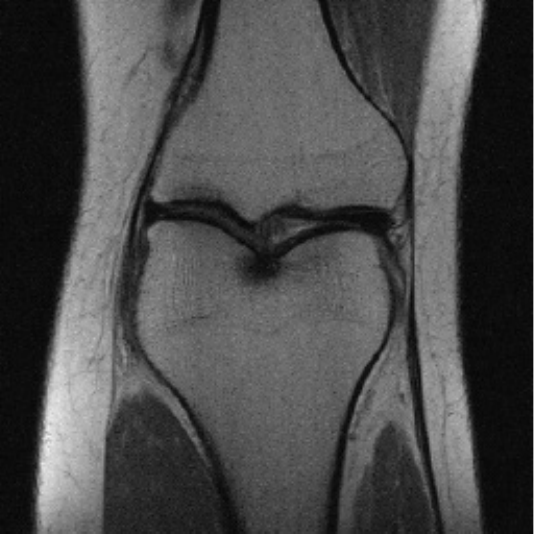}
        & \includegraphics[width=0.1\textwidth]{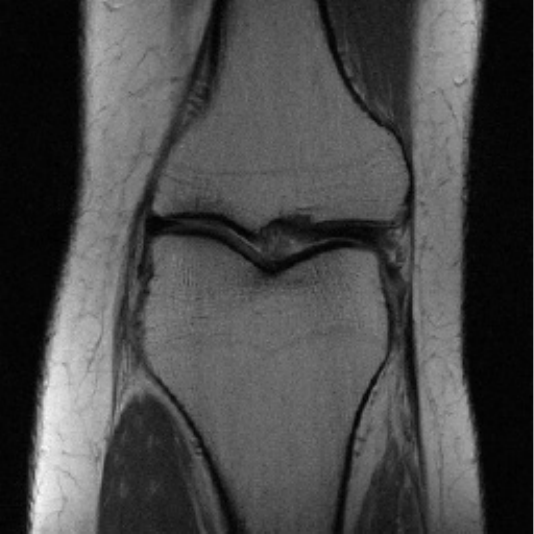}
        & \includegraphics[width=0.1\textwidth]{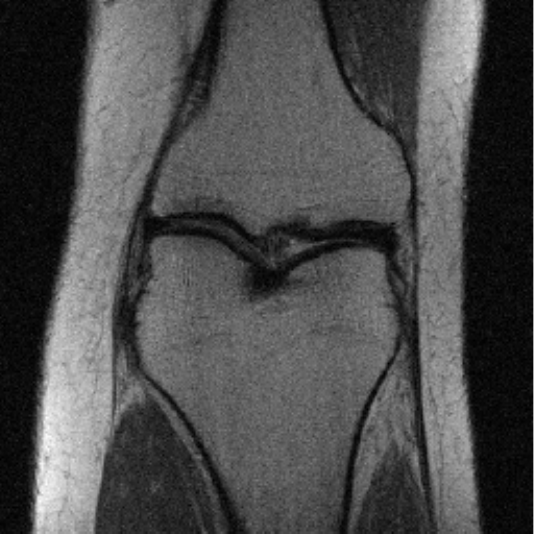}
        & \includegraphics[width=0.1\textwidth]{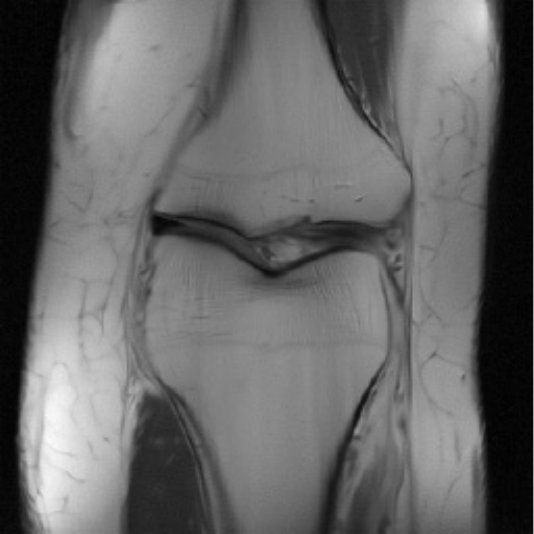}
        & \includegraphics[width=0.1\textwidth]{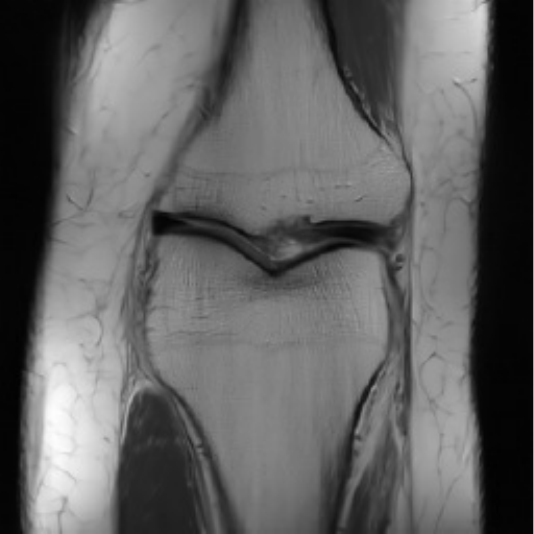}
        & \includegraphics[width=0.1\textwidth]{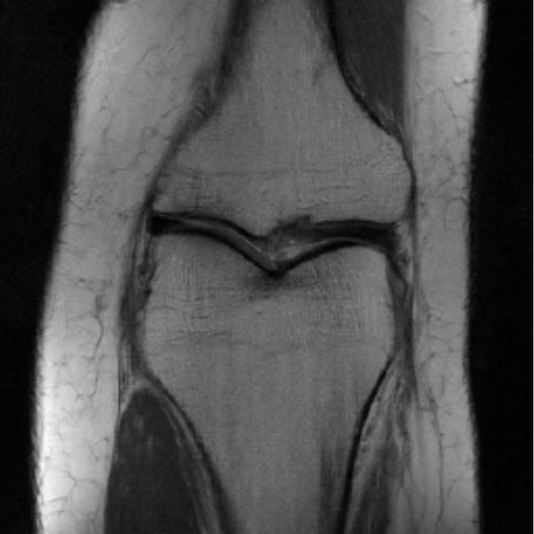}\\

        \includegraphics[width=0.1\textwidth]{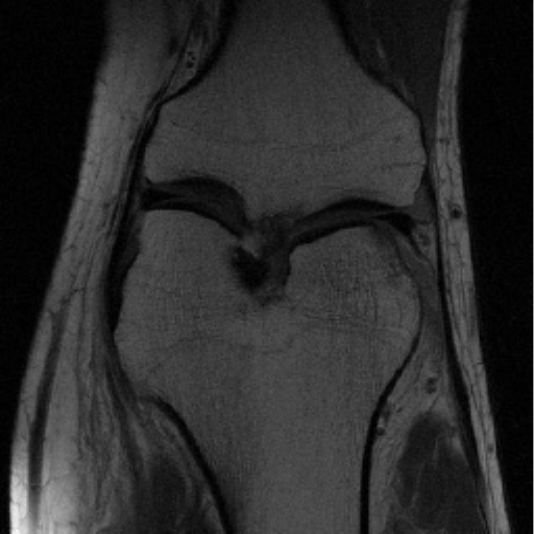}
        & \includegraphics[width=0.1\textwidth]{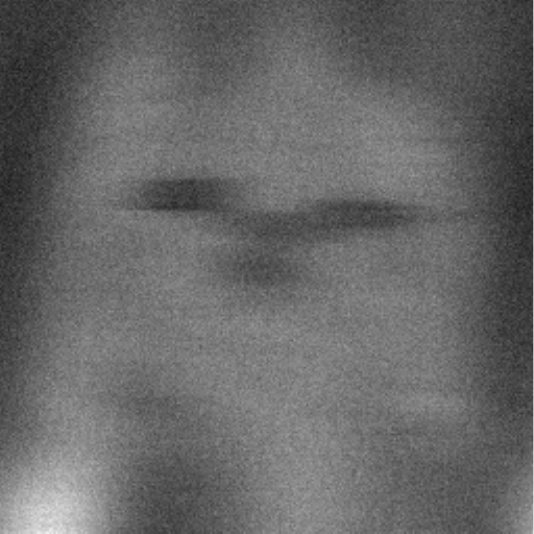}
        & \includegraphics[width=0.1\textwidth]{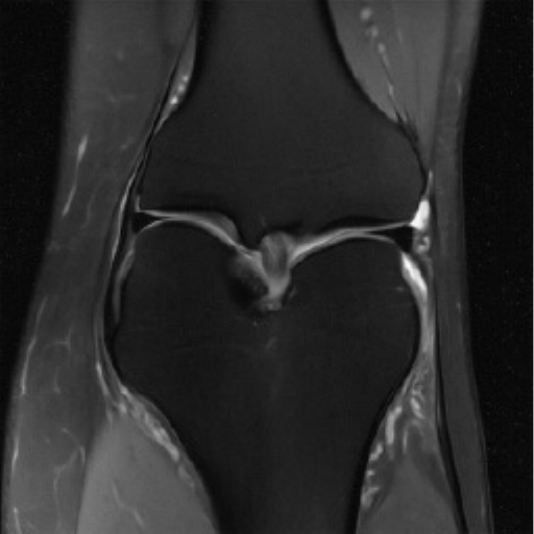}
        & \includegraphics[width=0.1\textwidth]{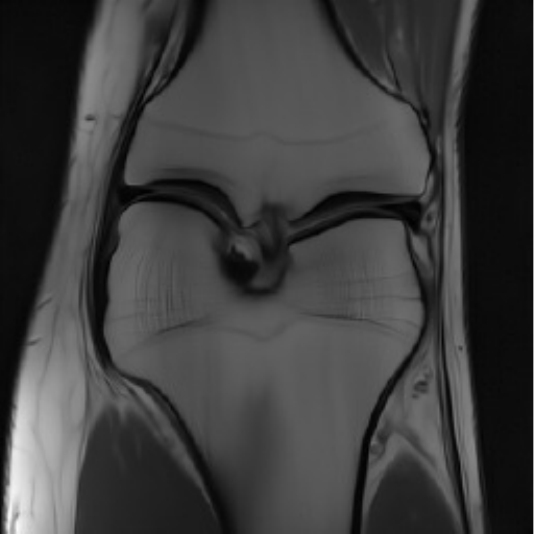}
        & \includegraphics[width=0.1\textwidth]{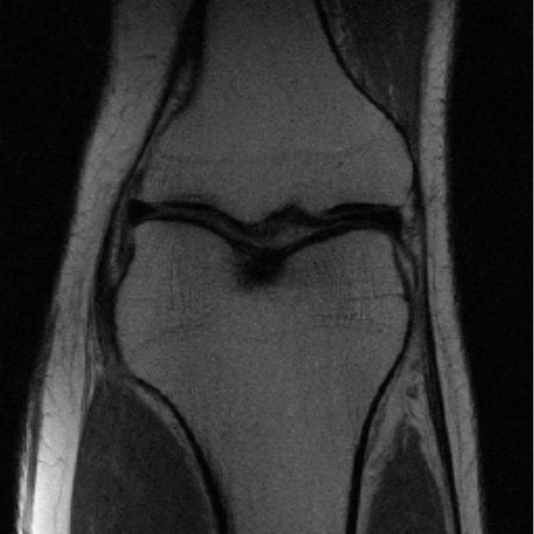}
        & \includegraphics[width=0.1\textwidth]{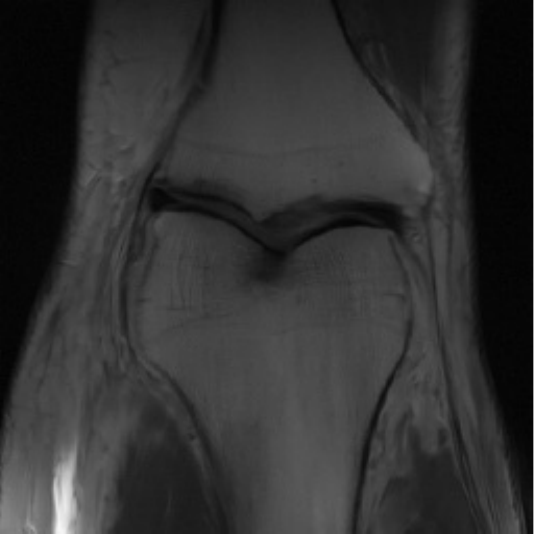}
        & \includegraphics[width=0.1\textwidth]{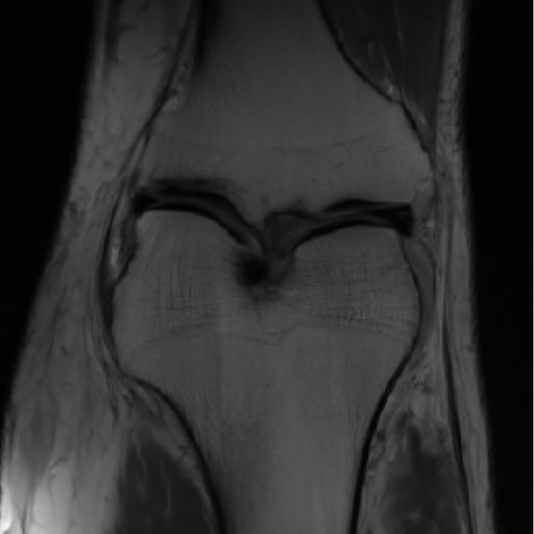}
        & \includegraphics[width=0.1\textwidth]{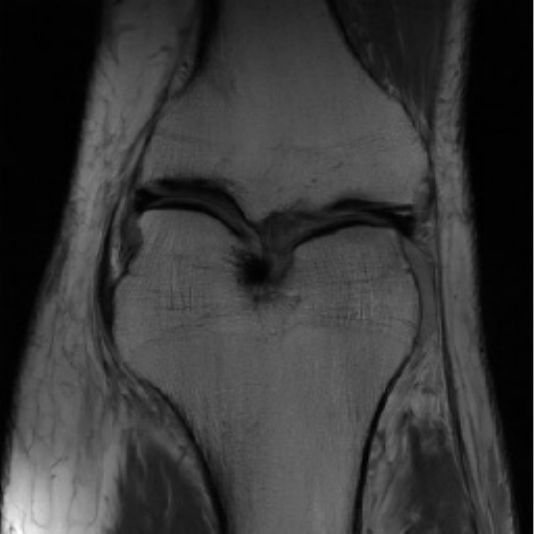}\\

    \end{tabular}

\caption{Qualitative comparison of reconstructions on the fastMRI dataset under accelerated MRI acquisition with $R=12$, comparing the DPS heuristic ($\zeta'=0.1$) and the proposed spectral recommendations.}
        \label{fig:fastMRI_R_12}
\end{figure*}

In the following section, we present qualitative comparisons between our spectral guidance recommendations and the DPS heuristic on the fastMRI dataset. Figure~\ref{fig:fastMRI_LPF} considers LPF degradation with $\mathcal{V}=0.1$ and $\sigma_y=0.1$, while Figures~\ref{fig:fastMRI_R_4}–\ref{fig:fastMRI_R_12} show results for accelerated MRI with $R\in\{4,8,12\}$. 

In each figure, the first column corresponds to the original image, $\x_0$, and the second to the degraded measurement, $\y$. The following three columns show samples obtained using the DPS heuristic at $50$, $100$, and $200$ diffusion steps (from left to right), while the remaining columns present results obtained using the proposed spectral recommendations at the same step counts. 

Since the fastMRI dataset is not considered in the original DPS work \cite{chung2022diffusion}, no predefined heuristic is available for selecting the guidance parameter. Therefore, for each degradation, we evaluate a wide range of candidate values $\zeta' \in \{0.05, 0.1, 0.3, 0.5, 0.7, 0.9, 1\}$, and select $\zeta' = 0.1$, which consistently provides the best trade-off between perceptual quality and distortion metrics.
This setting highlights the practical advantage of our approach. Rather than tuning multiple heuristics, the spectral recommendations are computed once for each degradation and naturally adapt to the prior characteristics, the degradation model, and the diffusion step count.


In Figure~\ref{fig:fastMRI_LPF}, the spectral recommendations better preserve the global structure of the original image and the measurement, while the DPS heuristic shows noticeable deviations, especially at low diffusion step counts. As the number of steps increases, the spectral approach progressively refines fine details. For instance, in the second row, a small white spot present in the measurement is recovered using spectral weights, whereas the DPS heuristic fails to reconstruct it.

Although the DPS heuristic often produces visually natural images even at low step counts, as also reflected in the quantitative results, visual quality alone is insufficient in inverse problems, where accurate reconstruction of the underlying signal is required.


Figure~\ref{fig:fastMRI_R_4} demonstrates a similar trend under accelerated sampling using $R=4$. Across the rows, subtle details such as small white or black spots, which may correspond to clinically relevant structures, are better recovered using the spectral weights, whereas the DPS heuristic typically fails to reconstruct them accurately.

Figures~\ref{fig:fastMRI_R_8} and \ref{fig:fastMRI_R_12} present comparisons under more severe degradations, using acceleration factors of $R=8$ and $R=12$. The same trends persist, with the spectral recovering finer details more accurately. In this regime, the visual naturalness of the DPS heuristic becomes, in some cases, comparable to that of the spectral approach, a behavior also reflected in the FID results in Table~\ref{tab:fastmri_results_R}. This is expected, as relatively small guidance weights such as $\zeta'=0.1$ tend to favor sampling from the prior, but often struggle to enforce consistency with the measurements. In certain instances, this leads to the introduction of spurious structures. For example, in the second row of the $R=8$ results, the DPS heuristic produces an artificial dark spot for $50$ and $100$ diffusion steps that is not present in the measurement.

\clearpage
\subsection{DiffPIR Algorithm}
\subsubsection{spectral recommendations}
\begin{figure}[h]
    \centering
    \includegraphics[width=0.6\textwidth]{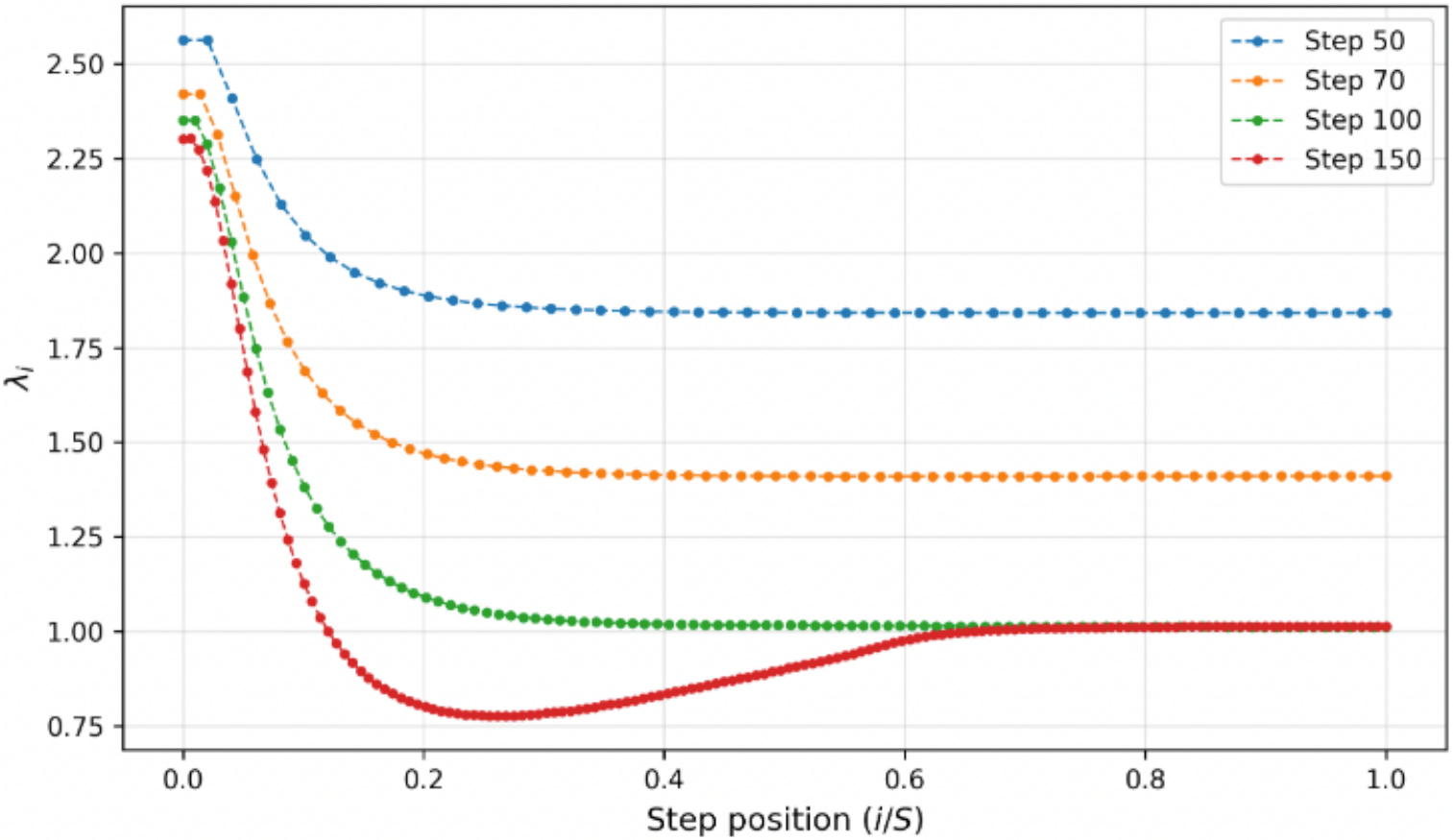}
    \caption{Comparison of the spectral recommendations for the DiffPIR regularization parameter $\lambda_i$ on the FFHQ $256 \times 256$ dataset across selected diffusion step counts $S \in \{50,70,100,150\}$.}
    \vspace{-0.35cm}
    \label{fig:Comparison_DiffPIR_and_Spectral_steps_FFHQ}
\end{figure}

In this section, we evaluate our framework using the DiffPIR algorithm. As an initial step, Figure~\ref{fig:Comparison_DiffPIR_and_Spectral_steps_FFHQ} presents the spectral recommendations for the guidance parameters ${\lambda_i}_{i=1}^S$, derived from the deterministic DiffPIR process described in Section~\ref{sec:diffpir_time_domain}, corresponding to the setting $\zeta=0$. The recommendations are obtained on the FFHQ dataset under the LPF degradation setting described in Section~\ref{sec:Empirical_Distribution}. Unlike the original DiffPIR heuristic, which employs a constant guidance value independently of the number of diffusion steps, the proposed spectral recommendations vary both across different diffusion step counts and throughout the diffusion process itself.

\subsubsection{Comprehensive comparison}
Table~\ref{tab:DiffPIR_results_FFHQ} compares the heuristic configurations proposed in DiffPIR with the proposed spectral recommendations. We first evaluate both approaches under the LPF degradation setting, and then assess the effectiveness of the spectral recommendations for degradation operators beyond the shift-invariant setting. Specifically, we consider two inpainting tasks: random masking with $70\%$ randomly removed pixels and box inpainting with mask size $128\times128$. For the inpainting tasks, we report results with $\lambda=7$ and $\lambda=6$ for the random and box inpainting settings, respectively, corresponding to the best-performing configurations selected from the heuristic values $\lambda\in{3,6,7}$, as also recommended in~\cite{zhu2023denoising}.
 The spectral recommendations consistently outperform the heuristic configurations under LPF degradation, while a similar trend is observed for the inpainting tasks. In the random inpainting setting, however, the relative performance between the two methods varies across diffusion steps, with a slight advantage shifting between the two methods across diffusion steps.


\begin{table*}[h]
\caption{
Quantitative comparison between the proposed spectral recommendations and DiffPIR on the FFHQ dataset under LPF degradation, random inpainting with $70\%$ masking, and box inpainting with mask size $128\times128$, using PSNR, SSIM, LPIPS, and FID.
}
\label{tab:DiffPIR_results_FFHQ}
\centering
\begin{small}
\begin{sc}
\setlength{\tabcolsep}{4.5pt}
\begin{tabular}{c c c cccc}
\toprule
Degradation & Method & Steps &
PSNR$\uparrow$ & SSIM$\uparrow$ & LPIPS$\downarrow$ & FID$\downarrow$ \\
\midrule

\multirow{8}{*}{LPF}
& \multirow{4}{*}{Spectral}
& $50$  & $\mathbf{19.45}$ & $\mathbf{0.60}$ & $\mathbf{0.32}$ & $\mathbf{40.22}$  \\
& & $100$ & $\mathbf{19.60}$ & $\mathbf{0.61}$ & $\mathbf{0.32}$ &  $\mathbf{39.69}$ \\
& & $150$ & $\mathbf{19.75}$ & $\mathbf{0.61}$ & $\mathbf{0.31}$ &    $\mathbf{39.07}$\\
& & $200$ & $\mathbf{19.86}$ & $\mathbf{0.62}$ & $\mathbf{0.31}$ &    $\mathbf{38.60}$\\
\cmidrule(lr){2-7}
& \multirow{4}{*}{DiffPIR}
& $50$  & $18.65$ & $0.56$ & $0.36$ & $41.03$\\
& & $100$ & $18.16$ & $0.55$ & $0.36$ & $39.74$  \\
& & $150$ & $18.20$ & $0.56$ & $0.35$ &$39.41$  \\
& & $200$ & $18.20$ & $0.56$ & $0.35$ &  $39.58$  \\

\midrule

\multirow{8}{*}{Inpainting (random)}
& \multirow{4}{*}{Spectral}
& $50$  & $\mathbf{29.95}$ & $\mathbf{0.82}$ & $\mathbf{0.21}$ & $\mathbf{32.74}$ \\
& & $100$ & $\mathbf{30.65}$ & $\mathbf{0.86}$ & $\mathbf{0.19}$ & ${29.03}$ \\
& & $150$ & $\mathbf{30.87}$ & $\mathbf{0.87}$ & $\underline{0.18}$ & $25.81$ \\
& & $200$ & $\mathbf{30.98}$ & $\mathbf{0.87}$ & $\underline{0.18}$ & $\mathbf{23.07}$ \\
\cmidrule(lr){2-7}
& \multirow{4}{*}{DiffPIR}
& $50$  & $29.40$ & $0.78$ & $0.23$ & ${35.62}$ \\
& & $100$ & $30.27$ & $0.82$ & $0.20$ & $\mathbf{28.06}$ \\
& & $150$ & $30.65$ & $0.84$ & $\underline{0.18}$ & $\mathbf{24.71}$ \\
& & $200$ & $30.83$ & $0.85$ & $\underline{0.18}$ & $24.25$ \\

\midrule

\multirow{8}{*}{Inpainting (box)}
& \multirow{4}{*}{Spectral}
& $50$ & $\mathbf{24.01}$ & $\mathbf{0.70}$ & $\mathbf{0.27}$ & $\mathbf{46.52}$ \\
& & $100$ & $\mathbf{24.56}$ & $\mathbf{0.73}$ & $\mathbf{0.26}$ & $\mathbf{46.39}$ \\
& & $150$ & $\mathbf{24.71}$ & $\mathbf{0.75}$ & $\mathbf{0.26}$ & $\mathbf{46.33}$ \\
&& $200$  & $\mathbf{24.76}$ & $\mathbf{0.76}$ & $\mathbf{0.26}$ & $\mathbf{46.25}$ \\
\cmidrule(lr){2-7}
& \multirow{4}{*}{DiffPIR}
& $50$  & $23.37$ & $0.62$ & ${0.32}$ & ${57.73}$ \\
& & $100$ & $23.89$ & $0.64$ & $0.32$ & ${57.11}$ \\
& & $150$ & $24.00$ & ${0.65}$ & $0.32$ & ${56.58}$ \\
& & $200$ & $24.09$ & ${0.66}$ & $0.32$ & ${55.81}$ \\

\bottomrule
\end{tabular}
\end{sc}
\end{small}
\end{table*}

\clearpage
\subsubsection{Additional visual results}

\begin{figure*}[h]

    \centering
    \setlength{\tabcolsep}{2pt}
    \renewcommand{\arraystretch}{1}
    \begin{tabular}{cc|cc|cc}
        \toprule
        Reference & Measurement
        & \multicolumn{2}{c}{DiffPIR Heuristic}
        & \multicolumn{2}{c}{Spectral Weights} \\

        \midrule


        \includegraphics[width=0.15\textwidth]{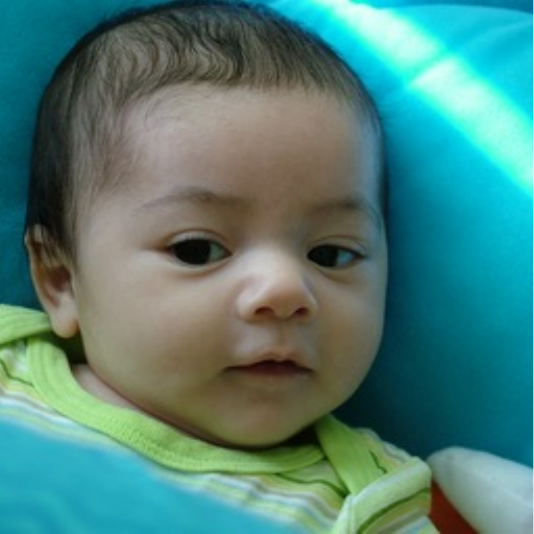}
        & \includegraphics[width=0.15\textwidth]{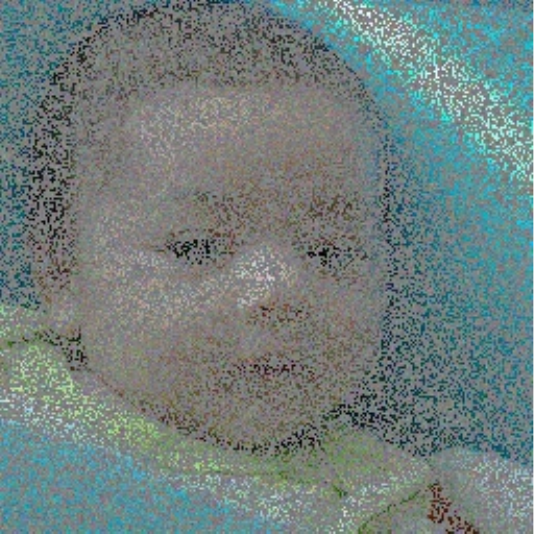}
        & \includegraphics[width=0.15\textwidth]{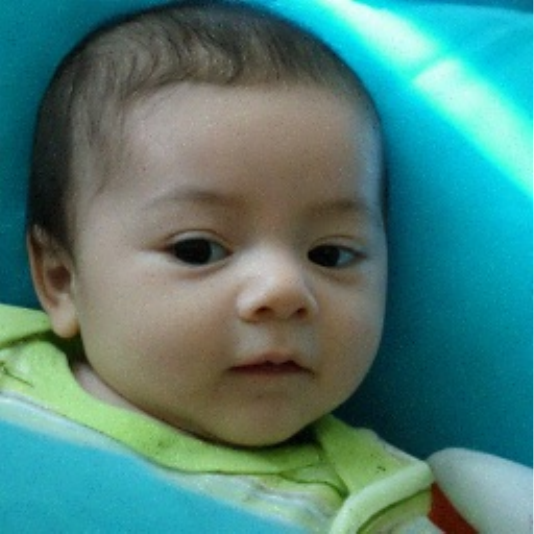}
        & \includegraphics[width=0.15\textwidth]{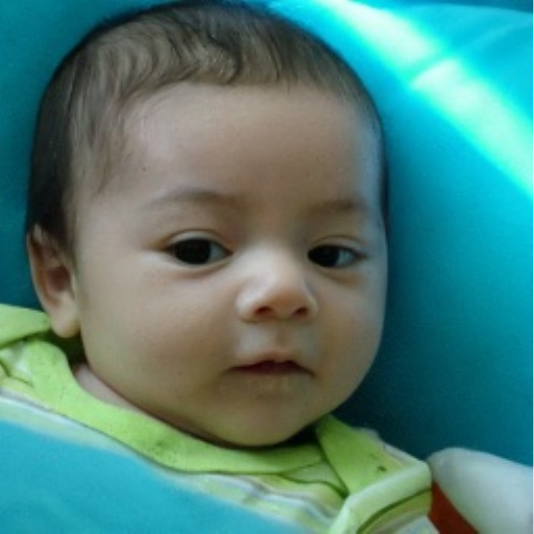}
        & \includegraphics[width=0.15\textwidth]{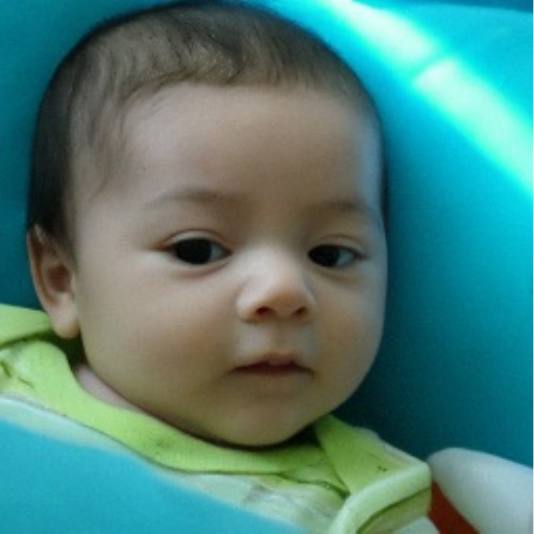}
        & \includegraphics[width=0.15\textwidth]{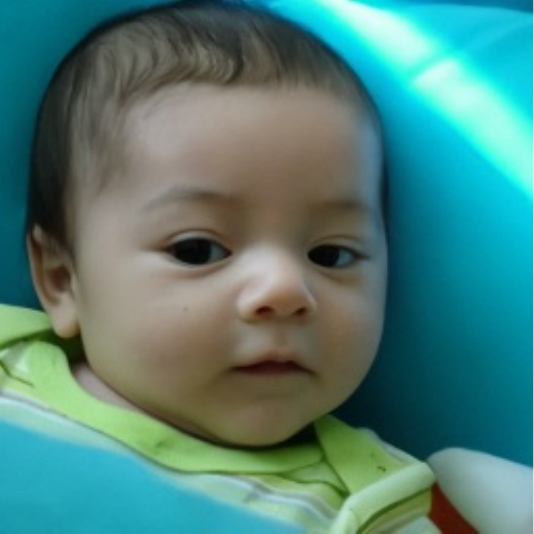}\\



        \includegraphics[width=0.15\textwidth]{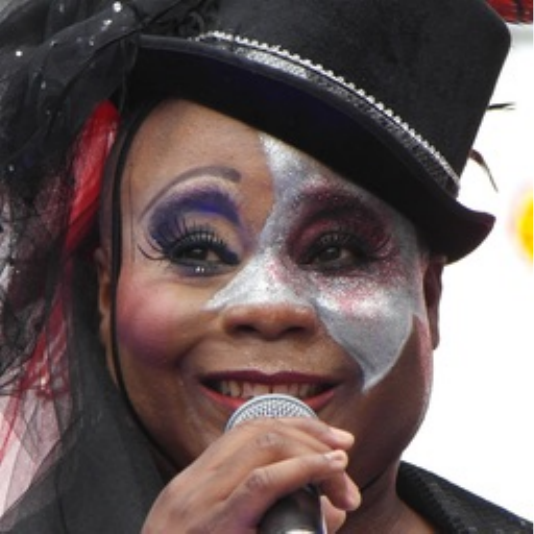}
        & \includegraphics[width=0.15\textwidth]{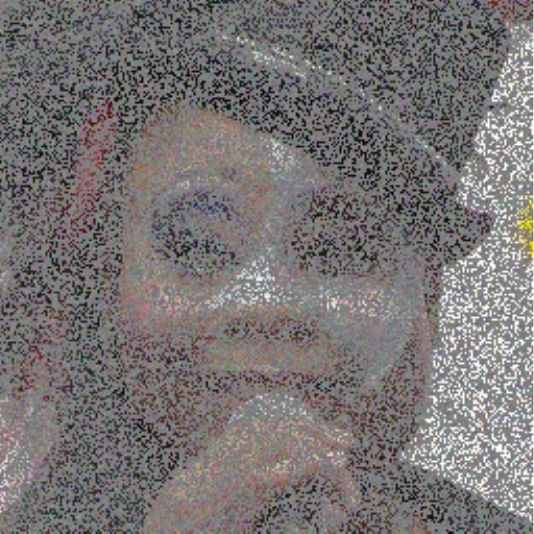}
        & \includegraphics[width=0.15\textwidth]{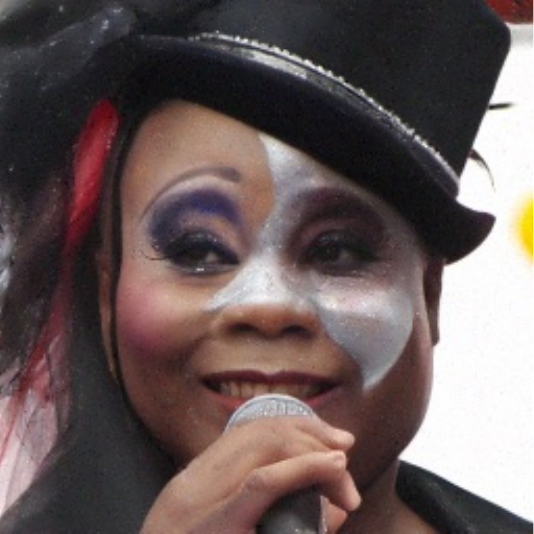}
        & \includegraphics[width=0.15\textwidth]{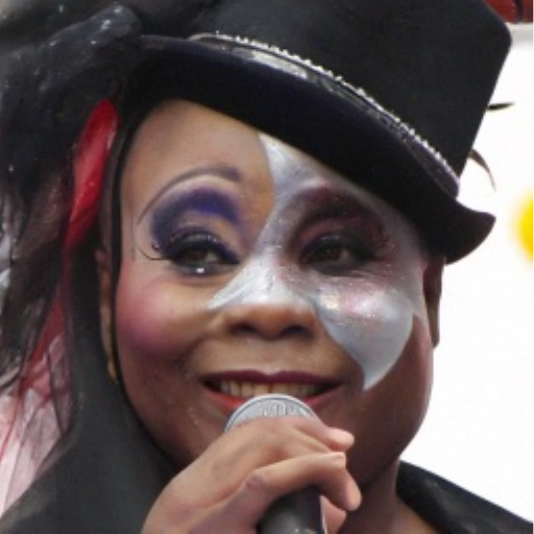}
        & \includegraphics[width=0.15\textwidth]{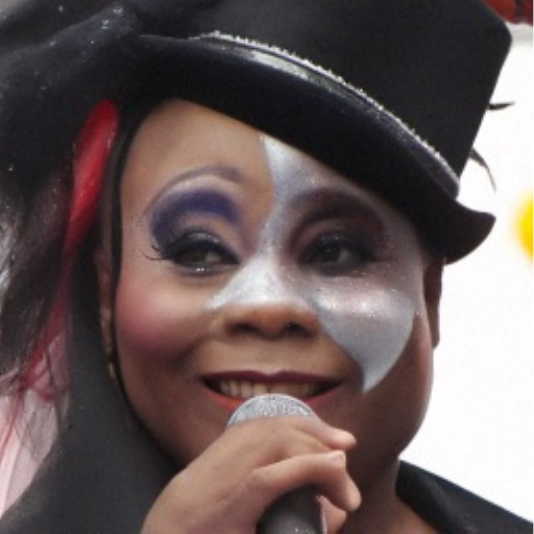}
        & \includegraphics[width=0.15\textwidth]{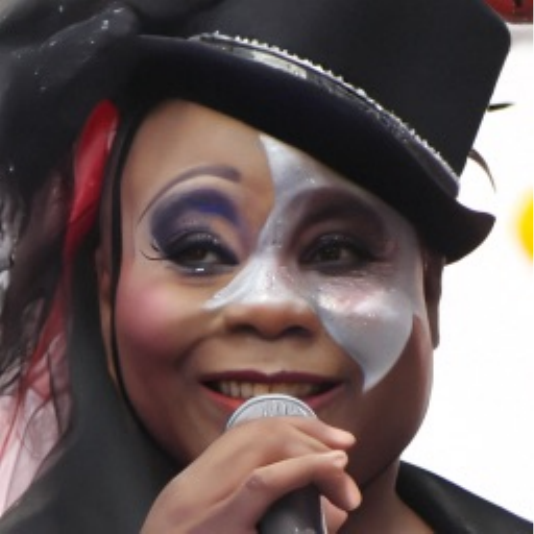}\\

    \end{tabular}

\caption{Qualitative comparison of reconstructions on the FFHQ dataset under random inpainting with $70\%$ missing pixels, comparing DiffPIR with predefined heuristics ($\lambda=7$) and the proposed spectral recommendations across different diffusion step counts.}
        \label{fig:FFHQ_inpainting_diffpir_random}
\end{figure*}

\begin{figure*}[h]

    \centering
    \setlength{\tabcolsep}{2pt}
    \renewcommand{\arraystretch}{1}
    \begin{tabular}{cc|cc|cc}
        \toprule
        Reference & Measurement
        & \multicolumn{2}{c}{DiffPIR Heuristic}
        & \multicolumn{2}{c}{Spectral Weights} \\

        \midrule


        \includegraphics[width=0.15\textwidth]{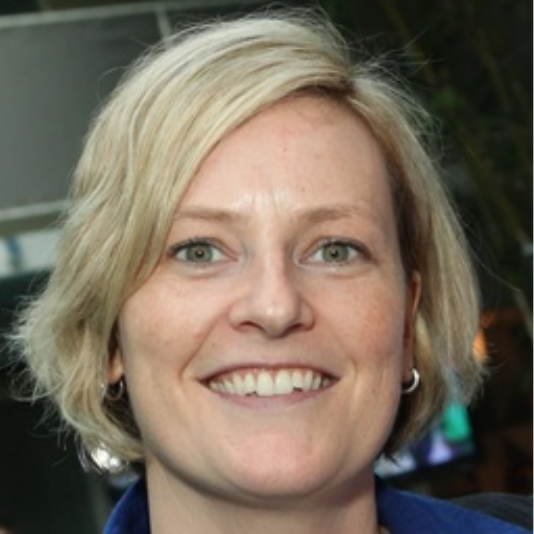}
        & \includegraphics[width=0.15\textwidth]{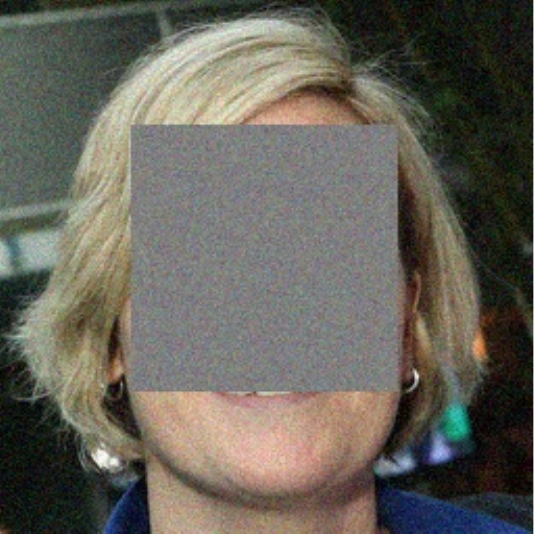}
        & \includegraphics[width=0.15\textwidth]{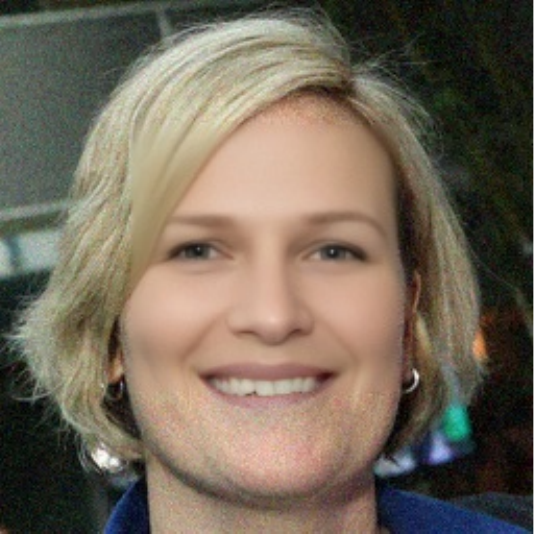}
        & \includegraphics[width=0.15\textwidth]{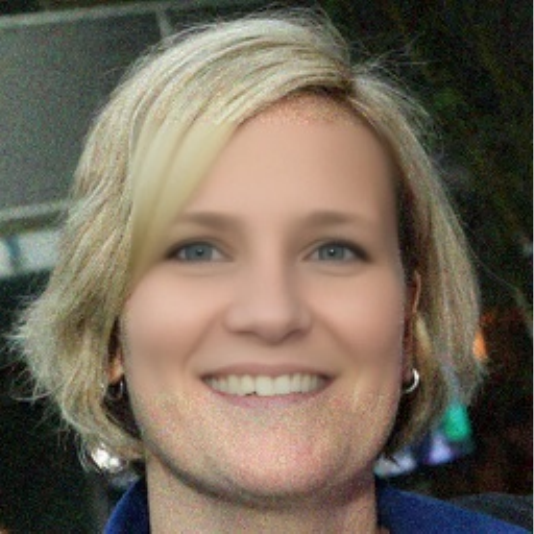}
        & \includegraphics[width=0.15\textwidth]{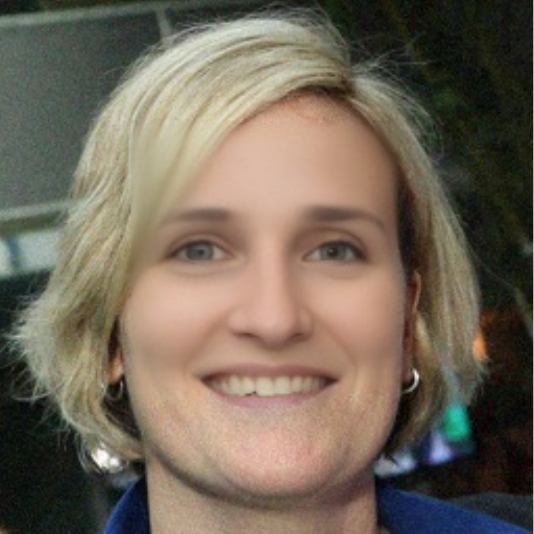}
        & \includegraphics[width=0.15\textwidth]{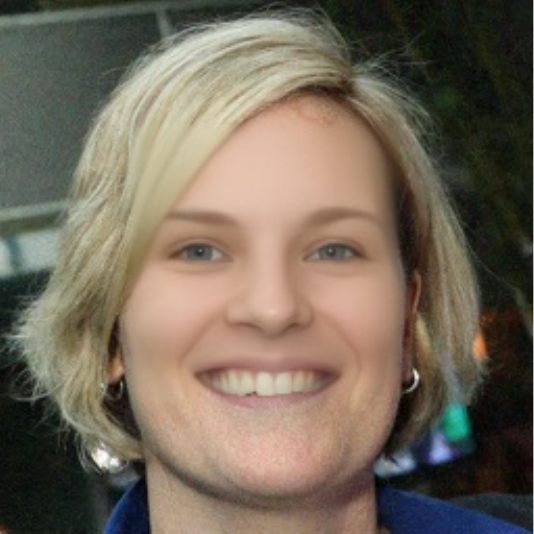}\\



        \includegraphics[width=0.15\textwidth]{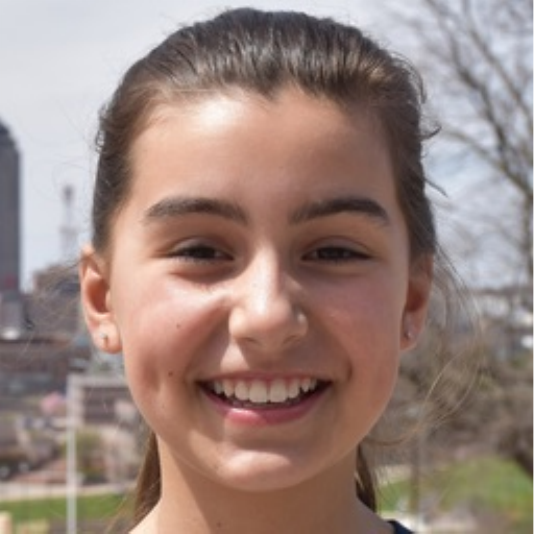}
        & \includegraphics[width=0.15\textwidth]{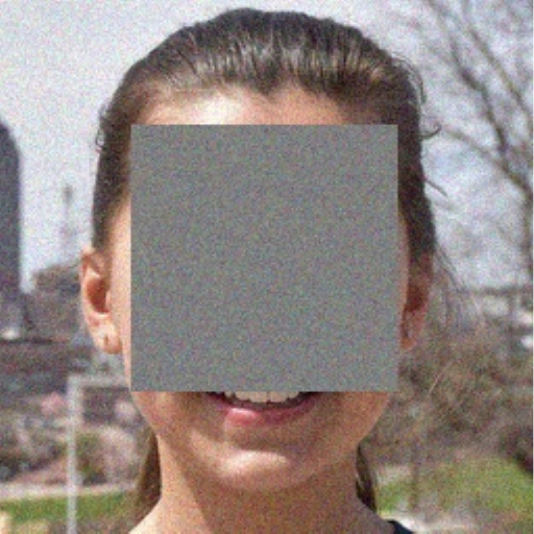}
        & \includegraphics[width=0.15\textwidth]{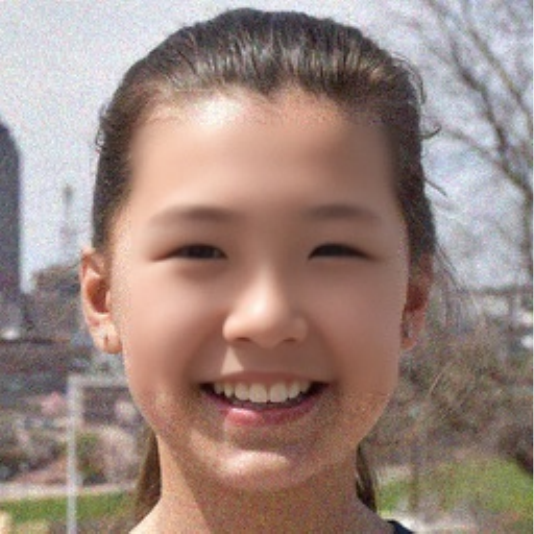}
        & \includegraphics[width=0.15\textwidth]{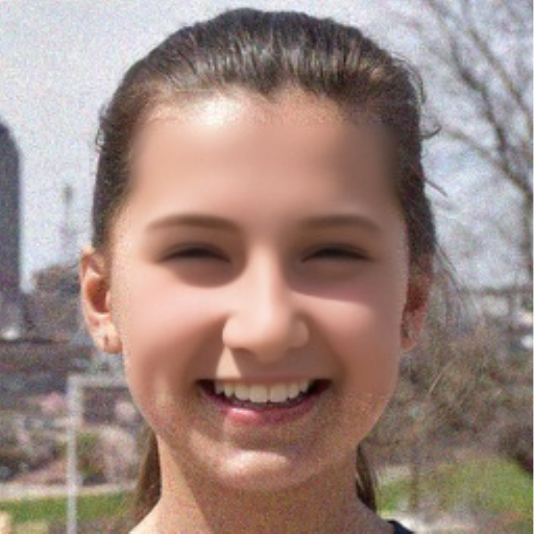}
        & \includegraphics[width=0.15\textwidth]{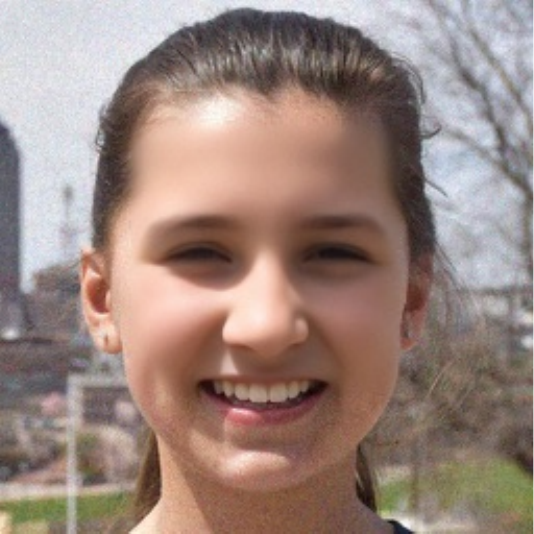}
        & \includegraphics[width=0.15\textwidth]{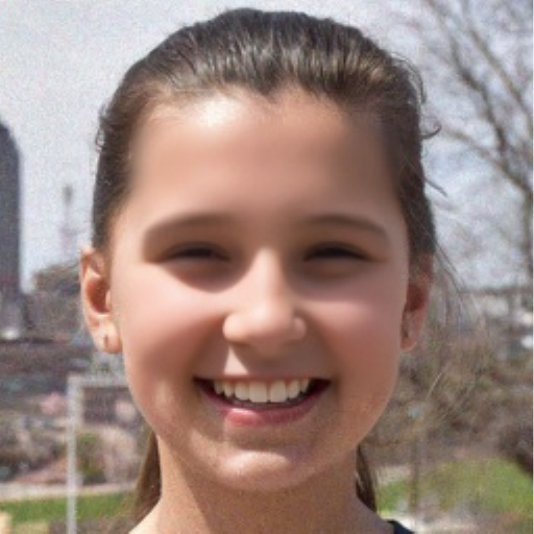}\\


    \end{tabular}


\caption{Qualitative comparison of reconstructions on the FFHQ dataset under box inpainting with a $128 \times 128$ missing region, comparing DiffPIR with predefined heuristics ($\lambda=6$) and the proposed spectral recommendations across different diffusion step counts.}
        \label{fig:FFHQ_inpainting_diffpir_box}
\end{figure*}

Figures~\ref{fig:FFHQ_inpainting_diffpir_random} and \ref{fig:FFHQ_inpainting_diffpir_box} present representative examples of the DiffPIR algorithm on the FFHQ dataset under random and box inpainting, respectively. In both figures, the first column shows the original image, and the second column shows the degraded input. The next two columns correspond to DiffPIR reconstructions using the predefined heuristics, for $50$ and $200$ diffusion steps (from left to right). The final two columns present the reconstructions obtained using the spectral recommendations, also for $50$ and $200$ diffusion steps.


For random inpainting (Figure~\ref{fig:FFHQ_inpainting_diffpir_random}), the spectral recommendations produce less noisy reconstructions than the DiffPIR heuristics at lower diffusion step counts. As the number of steps increases, this difference becomes less pronounced, and the reconstructions appear increasingly similar. This behavior is also reflected in Table~\ref{tab:DiffPIR_results_FFHQ}, where the gap between the scores becomes gradually less pronounced across all metrics. For box inpainting (Figure\ref{fig:FFHQ_inpainting_diffpir_box}), a similar behavior is observed. Under both guidance strategies, the reconstructions appear natural, with a slightly larger gap at lower diffusion step counts in favor of the spectral recommendations. In addition, Compared to the random inpainting setting, the reconstructions appear to be slightly noisier.



\clearpage
\section{Posterior Sampling Results}
\label{sec:Posterior_comparison_results}

This section includes further analysis of posterior behavior under fixed observations.
As described in Section~\ref{sec:Empirical_Distribution}, for three realizations $\x_i$ we generate corresponding degraded observations $\y_i$ and produce $1{,}000$ reconstructions for each case. Evaluation metrics are computed separately for each realization and then averaged.

Table~\ref{tab:ffhq_lpf_10_sigma_01_1} reports the resulting metrics.
Here, \emph{Spectral} denotes results obtained using spectral recommendations derived by solving the optimization problem~\ref{eq:optimization_problem_many_realizations}, which captures expected posterior behavior across degraded measurements and is solved once for a fixed degradation operator.In contrast, \emph{Spectral $K{=}1$} corresponds to solving the optimization problem independently for each realization. Overall, both spectral strategies tend to achieve more favorable performance in terms of measurement fidelity and perceptual similarity compared to the DPS heuristic. In addition, solving the optimization problem separately for each realization yields slightly improved results, reflecting the benefit of adapting the weighting to the specific observation.

\begin{table}[H]
  \caption{Quantitative results for posterior sampling under fixed observations.}
  \label{tab:ffhq_lpf_10_sigma_01_1}
  \centering
  \begin{small}
  \setlength{\tabcolsep}{6pt}
  \begin{tabular}{c c ccc}
    \toprule
    Method & Steps &
    PSNR$\uparrow$ & SSIM$\uparrow$ & LPIPS$\downarrow$ \\
    \midrule

    \multirow{6}{*}{DPS}
      & 50  & $12.52$ & $0.27$ & $0.69$ \\
      & 100 & $14.99$ & $0.35$ & $0.61$ \\
      & 200 & $18.15$ & $0.44$ & $0.52$ \\
      & 300 & $19.88$ & $0.49$ & $0.48$ \\
      & 400 & $21.04$ & $0.53$ & $0.45$ \\
      & 500 & $21.89$ & $0.55$ & $\mathbf{0.43}$ \\
    \midrule

    \multirow{6}{*}{Spectral}
      & 50  & $\mathbf{16.63}$ & $\mathbf{0.48}$ & $\mathbf{0.60}$ \\
      & 100 & $\underline{17.55}$ & $\underline{0.51}$ & $\mathbf{0.56}$ \\
      & 200 & $\underline{19.01}$ & $\underline{0.53}$ & $\mathbf{0.52}$ \\
      & 300 & $\underline{23.05}$ & $\underline{0.62}$ & $\mathbf{0.43}$ \\
      & 400 & $\underline{23.47}$ & $\underline{0.61}$ & $\underline{0.42}$ \\
      & 500 & $\underline{22.07}$ & $\underline{0.61}$ & $\underline{0.45}$ \\
    \midrule

    \multirow{6}{*}{Spectral $K{=}1$}
      & 50  & $\underline{12.95}$ & $\underline{0.38}$ & $\underline{0.66}$ \\
      & 100 & $\mathbf{18.48}$ & $\mathbf{0.53}$ & $\underline{0.53}$ \\
      & 200 & $\mathbf{22.24}$ & $\mathbf{0.59}$ & $\underline{0.44}$ \\
      & 300 & $\mathbf{23.08}$ & $\underline{0.62}$ & $\mathbf{0.41}$ \\
      & 400 & $\mathbf{24.76}$ & $\mathbf{0.63}$ & $\mathbf{0.39}$ \\
      & 500 & $\mathbf{22.72}$ & $\underline{0.61}$ & $0.44$ \\
    \bottomrule
  \end{tabular}
  \end{small}
  
\end{table}

\begin{figure}[h]
    \centering
       \includegraphics[width=0.18\linewidth]{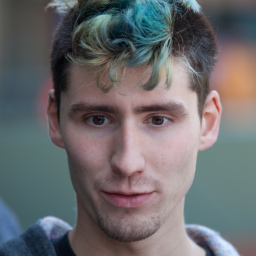}%
    \hspace{0.05\linewidth} 
     \includegraphics[width=0.18\linewidth]{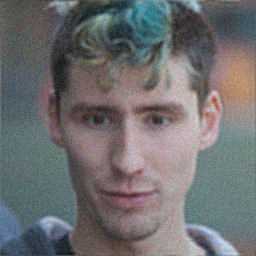}
    
  \setlength{\tabcolsep}{4pt}
  \renewcommand{\arraystretch}{1.0}

  \begin{tabular}{c c c}
        \multicolumn{3}{c}{} \\[2pt]

    \includegraphics[width=0.18\linewidth]{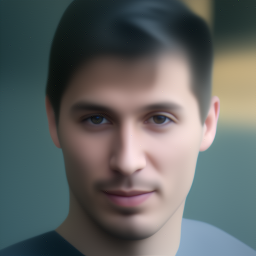} &
    \includegraphics[width=0.18\linewidth]{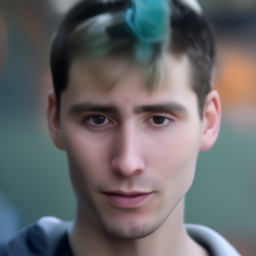} &
    \includegraphics[width=0.18\linewidth]{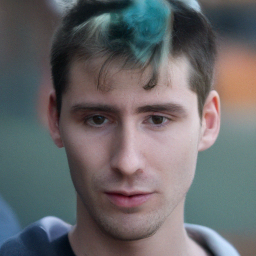} \\

    \includegraphics[width=0.18\linewidth]{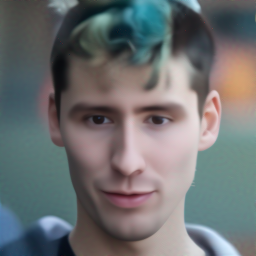} &
    \includegraphics[width=0.18\linewidth]{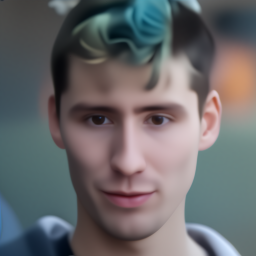} &
    \includegraphics[width=0.18\linewidth]{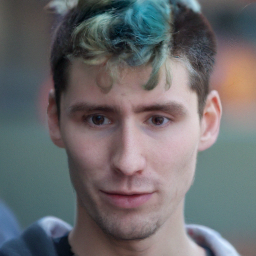} \\

    \includegraphics[width=0.18\linewidth]{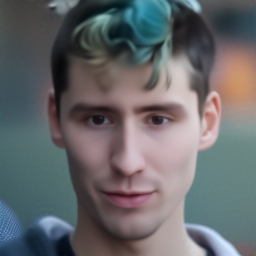} &
    \includegraphics[width=0.18\linewidth]{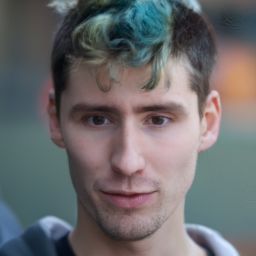} &
    \includegraphics[width=0.18\linewidth]{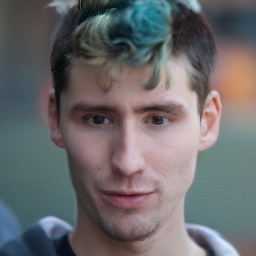} \\

  \end{tabular}
  \caption{Qualitative comparison of posterior samples under a fixed observation. The top row shows the reference image (left) and the corresponding degraded observation (right). Subsequent rows present reconstructions using different guidance schemes: the DPS heuristic (top), spectral recommendations (middle), and the realization-specific spectral variant with $k=1$ (bottom). Samples corresponding to $100$, $200$, and $400$ diffusion steps are shown from left to right.}
  \label{fig:posterior_comparison_00015}
\end{figure}

\begin{figure}[h]
    \centering
       \includegraphics[width=0.18\linewidth]{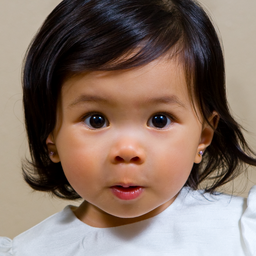}%
    \hspace{0.05\linewidth} 
     \includegraphics[width=0.18\linewidth]{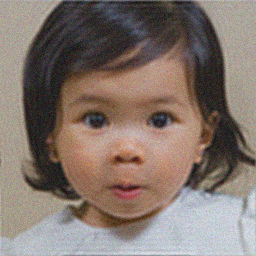}
    
  \setlength{\tabcolsep}{4pt}
  \renewcommand{\arraystretch}{1.0}

  \begin{tabular}{c c c}
        \multicolumn{3}{c}{} \\[2pt]

    \includegraphics[width=0.18\linewidth]{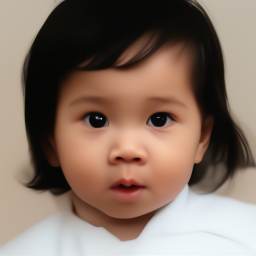} &
    \includegraphics[width=0.18\linewidth]{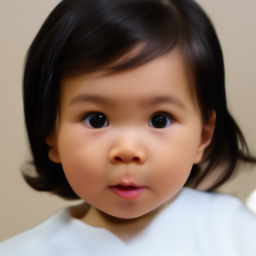} &
    \includegraphics[width=0.18\linewidth]{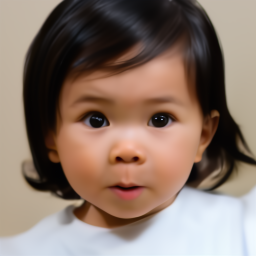} \\

    \includegraphics[width=0.18\linewidth]{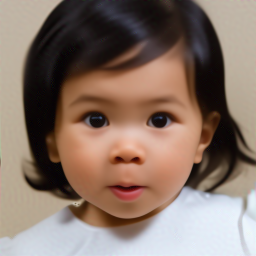} &
    \includegraphics[width=0.18\linewidth]{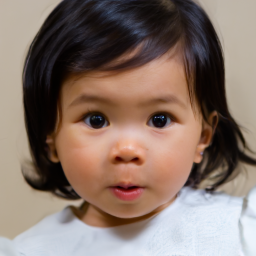} &
    \includegraphics[width=0.18\linewidth]{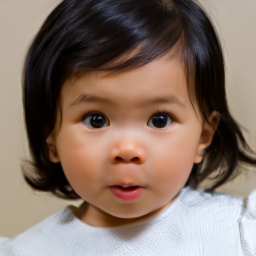} \\

    \includegraphics[width=0.18\linewidth]{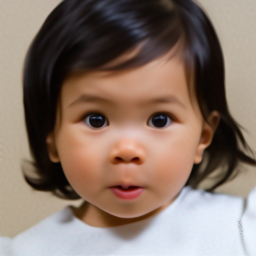} &
    \includegraphics[width=0.18\linewidth]{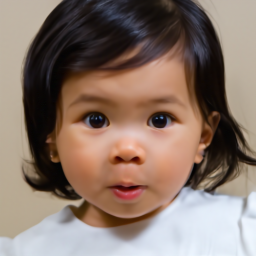} &
    \includegraphics[width=0.18\linewidth]{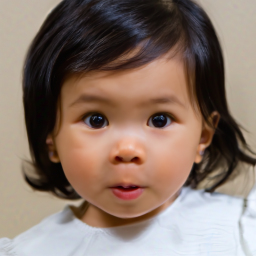} \\

  \end{tabular}

  \caption{Qualitative comparison of posterior samples under a fixed observation. The top row shows the reference image (left) and the corresponding degraded observation (right). Subsequent rows present reconstructions using different guidance schemes: the DPS heuristic (top), spectral recommendations (middle), and the realization-specific spectral variant with $k=1$ (bottom). Samples corresponding to $100$, $200$, and $400$ diffusion steps are shown from left to right.}

  \label{fig:posterior_comparison_00003}
\end{figure}

Figures~\ref{fig:posterior_comparison_00015} and~\ref{fig:posterior_comparison_00003} present qualitative comparisons for two different images.
In each figure, the top row shows the reference image on the left and the corresponding degraded observation on the right.
Each subsequent row corresponds to a different weighting scheme: DPS (top), the proposed spectral recommendations (middle), and the realization-specific spectral variant with $k{=}1$ (bottom).
Within each row, posterior samples obtained with $100$, $200$, and $400$ diffusion steps are shown from left to right. For visualization purposes, each example was selected from the $1,000$ generated samples by choosing the instance that best preserves visual naturalness while remaining consistent with the reference image.

As illustrated in Figures~\ref{fig:posterior_comparison_00015} and~\ref{fig:posterior_comparison_00003}, For a small number of diffusion steps, the DPS heuristic produces samples that reflect the global structure of the reference image, while remaining less constrained by the measurements. In contrast, the spectral recommendations yield samples that adhere more closely to the observed data. We also observe variability across samples produced by all weighting schemes, indicating that the methods do not converge to a single estimate such as the posterior mean.

\newpage
\clearpage

\section{Covariance Matrix Estimation}
\label{sec:appendix_estimating_covarince}

Under stationarity, the covariance matrix of a color image
\( \x_0 \in \mathbb{R}^{3N} \) has block-circulant with circulant blocks (BCCB) structure:
\[
\bSigmaZ =
\begin{pmatrix}
\Sigma_{RR} & \Sigma_{RG} & \Sigma_{RB} \\
\Sigma_{GR} & \Sigma_{GG} & \Sigma_{GB} \\
\Sigma_{BR} & \Sigma_{BG} & \Sigma_{BB}
\end{pmatrix}.
\]
Consequently, each block is diagonalized by the 2D discrete Fourier transform. Let \( \F_N \) denote the unitary 2D DFT matrix; then
\[
\Sigma_{ij} = \F_N^{*} \Lambda_{ij} \F_N,
\]
where \( \Lambda_{ij} \) are diagonal matrices containing the auto- and cross-channel power spectra.

This yields the Fourier-domain representation of the full covariance
\[
\bSigmaZ =
\left( \I_3 \otimes \F_N^{*} \right)
\bLambda
\left( \I_3 \otimes \F_N \right),
\]
where \( \Lambda \) is block-diagonal across spatial frequencies.

This spectral representation allows the optimization problem from Section~\ref{sec:guided_zero_shot_posterior} to be applied, and solved using the acceleration approaches described in Appendix~\ref{sec:appendix_optimization_time_analysis}.
\section{Optimization Time Analysis}
\label{sec:appendix_optimization_time_analysis}


The computational cost of solving the proposed optimization problem is influenced by several factors, including the signal resolution and the number of diffusion steps. An additional consideration is whether the optimization is performed separately for each measurement realization $\y_i$ (Equation \eqref{eq:optimization_problem_specific_realization}), yielding realization-specific improvements across metrics, or whether a single optimization is solved once for a given degraded system (Equation \eqref{eq:optimization_problem_specific_realization}), and then reused across realizations.

To accelerate the optimization procedure, we adopt two principles introduced in~\cite{benita2025spectral}. The first is \emph{iterative optimization}, in which the problem is solved progressively: the process starts with a small number of diffusion steps, allowing fast convergence with relatively low computational requirements. The resulting solution is then interpolated to the appropriate length and used to initialize the optimization with a larger number of steps.

The second principle is \emph{dimensionality reduction}. For high-resolution signals, computational requirements increase significantly; therefore, a low-rank covariance approximation is assumed. Specifically, only the $d$ most significant eigen-directions are retained using principal component analysis (PCA), while components associated with near-zero eigenvalues are discarded. Together, these methods substantially reduce optimization time without compromising performance.

Table\ref{tab:optimization_time} reports the solution times of the optimization problem on the FFHQ dataset at a resolution of $256 \times 256 \times 3$, using iterative initialization and dimensionality reduction. All experiments are conducted on a standard CPU using the L-BFGS-B via SciPy, with the maximum number of optimization iterations set to $2500$ and a function tolerance of $10^{-6}$.


\begin{table}[H]
  \centering
  \caption{Optimization time as a function of the number of diffusion steps for the FFHQ dataset
  ($256 \times 256 \times 3$). Results are obtained using iterative initialization and
  dimensionality reduction}
  \label{tab:optimization_time}
  \begin{small}
    \setlength{\tabcolsep}{10pt}
    \begin{tabular}{c c}
      \toprule
      Number of diffusion steps & Optimization time (seconds) \\
      \midrule
      $10$   &  $1.77$\\
      $50$   &  $13.73$\\
      $100$  &  $32.40$\\
      $200$  &  $393.30$\\
      $300$  &  $189.71$\\
      $400$  &  $333.50$\\
      $600$  &  $997.22$\\
      $800$  &  $1406.56$\\
      $1000$  &  $1902.48$\\
      \bottomrule
    \end{tabular}
  \end{small}
\end{table}

Table~\ref{tab:optimization_time} shows that the time required to solve the optimization problem increases with the number of diffusion steps.
However, when the optimization is performed once per degradation setting, the resulting computation time remains within a practical range, as the same spectral recommendations can be reused across multiple inference runs for that setting.





\clearpage
\newpage

\end{document}